\pdfoutput=1
\documentclass[12pt]{report}
\usepackage[numbers]{natbib}

\usepackage{amsmath,amsfonts,amssymb,amsthm, color}
\usepackage[section]{algorithm}
\usepackage{algorithmicx}
\usepackage{epsfig,wrapfig,epsf}
\usepackage[outercaption]{sidecap} 
\usepackage{times}
\usepackage{fancyvrb}
\usepackage{enumerate}
\usepackage{authblk}
\usepackage{tikz}
\usepackage{bm}
\usetikzlibrary{arrows,shapes}
\usetikzlibrary{patterns}
\usepackage{graphicx}
\usepackage{fullpage}
\usepackage{ltxtable}
\usepackage{nicefrac}
\usepackage{multirow}
\usepackage{setspace}
\usepackage[colorlinks=true, linkcolor=blue, bookmarks=true]{hyperref}
\usepackage[noend]{algpseudocode}
\usepackage[utf8]{inputenc} 
\usepackage[T1]{fontenc}    
\usepackage{url}            
\usepackage{booktabs}       
\usepackage{microtype}      
\usepackage{pbox}
\usepackage{natbib}
\renewcommand{\ref}{\hyperref}

\newcommand{\figref}[1]{Figure~\ref{#1}}
\newcommand{\secref}[1]{Section~\ref{#1}}

\newcommand{\algoref}[1]{Algorithm~\ref{#1}}

\renewcommand{\eqref}[1]{Eq.~(\ref{#1})}

\makeatletter
\newtheorem*{rep@theorem}{\rep@title}
\newcommand{\newreptheorem}[2]{%
\newenvironment{rep#1}[1]{%
 \def\rep@title{#2 \ref{##1}}%
 \begin{rep@theorem}}%
 {\end{rep@theorem}}}
\makeatother

\newreptheorem{theorem}{Theorem}
\newreptheorem{lemma}{Lemma}
\newreptheorem{definition}{Definition}

\theoremstyle{definition}

\newcounter{saveenumi}

\algrenewcommand\algorithmicrequire{\textbf{Parameters:}}
\algrenewcommand\algorithmicensure{\textbf{Initialization:}}

\usepackage[left=1.65in, right=1.65in, top=1in, bottom=1in]{geometry}
\usepackage{makeidx}

\makeindex

\usepackage{mysty}
\usepackage{enumitem}
\usepackage{xspace}

\usepackage{times}
\usepackage{epsfig}
\usepackage{graphicx}

\usepackage{bm}
\usepackage{makecell}
\usepackage{multirow}

\usepackage{color}
\usepackage{rotating}
\usepackage{overpic}
\usepackage{xcolor}
\usepackage{colortbl}
\usepackage{hhline}

\usepackage{booktabs}

\usepackage{caption}
\usepackage{subcaption}

\usepackage{amsmath,amssymb,amsfonts,gensymb}

\usepackage{tikz}
\usetikzlibrary{decorations.pathmorphing}
\usetikzlibrary{intersections}
\usetikzlibrary{positioning}
\usetikzlibrary{calc}
\usetikzlibrary{fit}
\usepackage[skins]{tcolorbox}

\usepackage{geometry}

\usepackage{adjustbox}
\usepackage{array}

\usepackage{tabu}

\usepackage{upgreek}
\usepackage{bm}
\usepackage{comment}

\usepackage{algorithm}

\usepackage[nottoc,notlot,notlof]{tocbibind}

\makeatletter
\newcommand{\bfgreek}[1]{\bm{\@nameuse{#1}}}
\newcommand{\bfgreekco}[2]{\bm{\textcolor{#2}{\@nameuse{#1}}}}
\makeatother

\newif\ifshowcomment
\newcommand{\xxcom}[2]{\ifshowcomment\textcolor{#1}{#2}\fi}
\newcommand{\gscom}[1]{\xxcom{red}{greg: #1}} 
\newcommand{\rtcom}[1]{\xxcom{blue}{rt: #1}} 

\newcommand{\eg}[1]{e.g.}
\newcommand{\val}[1]{\texttt{val}}
\newcommand{\test}[1]{\texttt{test}}
\newcommand{\train}[1]{\texttt{train}}
\newcommand{\aug}[1]{\texttt{train-aug}}

\numberwithin{equation}{section}

   \def\url#1{}

\newcommand{\thesisTitle}{GOAL-DRIVEN TEXT DESCRIPTIONS FOR IMAGES}
\newcommand{\thesisAuthor}{Ruotian Luo}
\newcommand{\thesisInstitute}{Toyota Technological Institute at Chicago}
\newcommand{\thesisDate}{August, 2021}
\newcommand{\thesisAdvisor}{Dr. Gregory Shakhnarovich}
\newcommand{\committeeFirst}{Dr. Kevin Gimpel}
\newcommand{\committeeSecond}{Dr. David McAllester}
\newcommand{\committeeThird}{Dr. Gal Chechik}

\newcommand{\thesisAbstract}{

\rtcom{Mostly from proposal}
A big part of achieving Artificial General Intelligence(AGI) is to build a machine that can see and listen like humans. Much work has focused on designing models for image classification, video classification, object detection, pose estimation, speech recognition, etc., and has achieved significant progress in recent years thanks to deep learning.
However, understanding the world is not enough. An AI agent also needs to know how to talk, especially how to communicate with a human. While perception (vision, for example) is more common across animal species, the use of complicated language is unique to humans and is one of the most important aspects of intelligence.

In this thesis, we focus on generating textual output given visual input. This involves both visual perception and language generation. Nevertheless, we will use existing visual perception models, and focus primarily on how to generate more meaningful texts, by studying different goals of language.
In Chapter \ref{chap:ref_exp}, we focus on generating the referring expression, a text description for an object in the image so that a receiver can infer which object is being described. We use a comprehension machine to directly guide the generated referring expressions to be more discriminative.
In Chapter \ref{chap:disc_cap}, we introduce a method that encourages \textbf{discriminability} in image caption generation. We show that more discriminative captioning models generate more descriptive captions.
In Chapter \ref{chap:diversity}, we study how training objectives and sampling methods affect the models' ability to generate \textbf{diverse} captions. We want diverse captions  because one caption may not be sufficient to describe an image. We find that a popular captioning training strategy will be detrimental to the diversity of generated captions.
In Chapter \ref{chap:length}, we propose a model that can control the \textbf{length} of generated captions. By changing the desired length, one can influence the style and descriptiveness of the captions.
Finally, in Chapter \ref{chap:tagging}, we show how we rank/generate \textbf{informative} image tags according to their information utility. The proposed method better matches what humans think are the most important tags for the images.

}

\usepackage{float}

\newcommand{\spicen}{AllSPICE\xspace}

\newcommand{\tagcap}{\textsc{tag2cap}\xspace}

\newcommand{\tagtag}{\textsc{tag2tag}\xspace}
\newcommand{\tagfeat}{\textsc{tag2feat}\xspace}

\newcommand{\tagc}{\textsc{t2c}}
\newcommand{\tagf}{\textsc{t2f}}

\begin{document}

\pagenumbering{roman}
\newgeometry{left=1.65in, right=1.65in, top=1.65in, bottom=1.65in}

\begin{center}

\Large{\bf \thesisTitle} \\[1cm]
\large{\MakeUppercase{by}} \\
\large{\MakeUppercase \thesisAuthor} \\[2cm]
\large{A thesis submitted\\ in partial fulfillment of the requirements for\\the degree of}\\[0.8cm]
\large{Doctor of Philosophy in Computer Science}\\[0.8cm]
\large{at the}\\[0.8cm]
\large{{\MakeUppercase \thesisInstitute}\\Chicago, Illinois}\\[0.8cm]
\large{\thesisDate}\\[2cm]
\large{Thesis Committee: \\ {\thesisAdvisor \;(Thesis advisor)}\\ {\committeeFirst}\\ {\committeeSecond} \\{\committeeThird}}	

\end{center}

\thispagestyle{empty}
\newgeometry{left=1.2in, right=1.2in, top=1in, bottom=1.25in}

\begin{center}

\Large{\bf \thesisTitle} \\[0.2cm]
\large{by} \\[0.2cm]
\Large{\thesisAuthor}

\end{center}

\section*{Abstract}
\thesisAbstract\par

\noindent  Thesis Advisor: Gregory Shakhnarovich

\newpage





\clearpage           
\vspace*{\stretch{1}}
{\itshape             
\begin{center}
To my family
\end{center}
}
\par 
\vspace{\stretch{3}} 
\clearpage           

\section*{Acknowledgments}

I would like to thank my advisor Greg Shakhnarovich. I feel very fortunate to have worked with him. He is a great supporter and a great mentor. He gave us the freedom to work on our own interests at our own pace. He is one of the smartest people I have ever seen. I would also like to thank my other faculty collaborators Allyson Ettinger, Michael Maire, Bohyung Han, Rebecca Willett. I am especially grateful to David McAllester, Kevin Gimpel, and Gal Chechik for being part of the committee.  They all provided support and assistance at some stage of my Ph.D. studies.

I would also like to thank the faculty and staff of TTIC for making TTIC such a great place for research. I would especially like to thank Matt Walter, Avrim Blum, Madhur Tulsiani, Karen Livescu, Yury Makarychev, Chrissy Coleman, Adam Bohlander, Mary Marre, Erica Cocom, Amy Minick, Matthew Turk.

I want to thank all my internship mentors, Scott Cohen, Brian Price, Linjie Yang, Ning Zhang, Zhe Gan, Jingjing Liu.

I also thank my student collaborators and TTIC fellows, Davis Gilton, Lior Bracha, Haochen Wang, Falcon Dai, Andrea Daniele, Jiading Fang, Nick Kolkin, Chip Schaff, Shane Settle, Bowen Shi, Freda Shi, Shubham Toshniwal, Igor Vasiljevic, Takuma Yoneda, Minda Chen, Zewei Chu, Siqi Sun, Lifu Tu, Lorraine Li, Xiaoming Zhao, Chenxi Yang.

I would also like to thank all the fellow students or researchers I met during internships or conferences.
They are
Mengtian Li, Licheng Yu, Chongruo Wu, Yuxin Chen, Hengshuang Zhao, Shuai Li, Yingcong Chen, Yinan Zhao, Jiahui Yu, Yilin Wang, Huiwen Chang, Shanghang Zhang, Xiang Cheng, Kaichun Mo,
Zhou Ren, Ziyu Zhang, Jianfei Yu, Zhizhong Li, Anhong Guo, Xuecheng Nie, Shibi He, Liuhao Ge, Yiqun Cao, Han Wang, Zhenglin Geng, Yongxi Lu, Yifan Sun, Yuchong Xiang, Jianwei Yang, Tianye Li, Zhengyuan Yang, 
Luowei Zhou, Jie Lei, Yu Cheng, Yen-Chun Chen, 
Jiasen Lu, Zhiyuan Fang, Tianlu Wang, Chenyun Wu, Xin Eric Wang, Jiuxiang Gu, Xi Yin, Hexiang Hu, Xinlei Chen, Yuxin Wu, Xu Yang, Huaizu Jiang, Chenxi Liu, Yin Cui,
Kaihua Tang, Qianyu Feng, Yu Wu, Xin Wang.

I would like to express my deepest gratitude to my parents and family in China. They have provided me with a warm home and a good educational environment. I am grateful that they have always supported my doctoral studies and were willing to let go of me to pursue my studies in a foreign country.

To my wife, Jinhua Zhang, I express my highest respect and gratitude. I am so blessed to have her by my side. She is the star of my life.

\newgeometry{left=1.2in, right=1.2in, top=1in, bottom=1.25in}

{\hypersetup{linkcolor=black} \tableofcontents }
{\hypersetup{linkcolor=black} \listoffigures}
{\hypersetup{linkcolor=black}  \listoftables}
%

\doublespacing 

\clearpage
\pagenumbering{arabic}

\chapter{Introduction}  \label{chap:intro}

\section{Motivation}
Most of the artificial intelligence (AI) research only focuses on a specific field. For example, computer vision researchers study vision perception: image classification, object detection, pixel-level segmentation. NLP researchers study how to comprehend text: translation, parsing, generation, etc. However, the ultimate AI should be multi-modal. Not only does the AI need to understand the world, but also to speak and communicate with human users. Think of an example: in a smart home, you ask your AI assistant: Can you find my keys? The assistant needs first to understand the language, search through the home with vision models, and then, more importantly, tell you where it is through language. The ability to talk is an essential step towards building intelligent machines.

\begin{figure}[!ht]
  {\small
  \centering
  \begin{minipage}[b]{0.3\textwidth}
    \includegraphics[width=\linewidth]{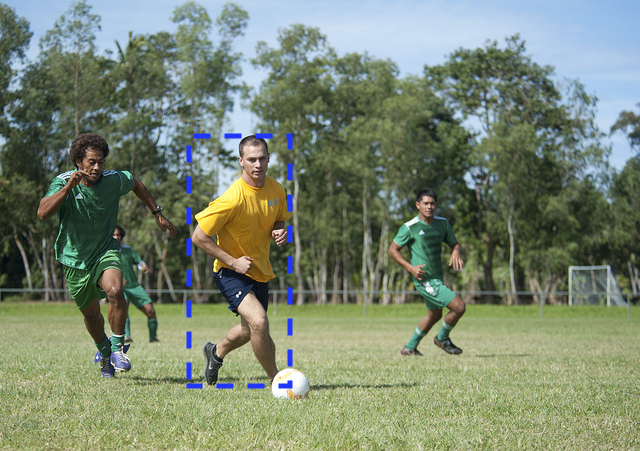}
    Referring expression: person in yellow
  \end{minipage}
  \hspace{1em}
  \begin{minipage}[b]{0.3\textwidth}
    \includegraphics[width=\linewidth]{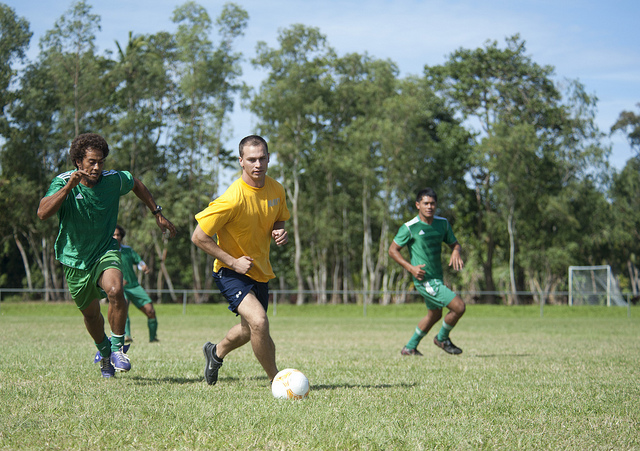}
    Captioning: a group of young men playing a game of soccer.
  \end{minipage}
  \hspace{1em}
  \begin{minipage}[b]{0.3\textwidth}
    \includegraphics[width=\linewidth]{figures/fig_pic_from_mattnetweb.png}
    Tagging: soccer, person, tree, run, goal
  \end{minipage}
  }

  \caption{Examples of the three tasks we consider in this thesis. Referring expression generation: generate a discriminative/unambiguous description for an object in the image. Image captioning: describe the scene in a sentence. Image tagging: label the scene with a set of tags.}
  \label{fig:thesis_fig1}
\end{figure}

In this thesis, we focus on how to translate perceptual output into textual output: empowering machines to speak. We will study the task of referring expression generation (generate a discriminative/unambiguous description for an object in the image), image captioning (describe the scene in a sentence), image tagging (label the scene with a set of tags). (See examples in \figref{fig:thesis_fig1}).
More specifically, we want to achieve different goals in the generation process. One type of goal is based on the cooperative principle proposed by Linguist Paul Grice in pragmatics theory~\cite{grice1975logic}. As a speaker, the model needs to take into account that the generated text will be consumed by a \textit{listener}, which in most cases is a person. The cooperative principle includes four maxims, namely quality (truthful), quantity (right amount of information), relevance (relevant and pertinent), and manner (clear). These maxims suggest that the text description should be informative, concise, and unambiguous. We will examine such goals in all three tasks (Chapter \ref{chap:ref_exp}, \ref{chap:disc_cap}, \ref{chap:tagging}). In addition, in the end, AI is a tool, and that tool needs to be embedded with some specific functionality. These functionalities can also be the goals we pursue. In this direction, specifically, we study diversity and length control in image captioning task (Chapter \ref{chap:diversity}, \ref{chap:length}).

\section{Thesis Goals and Contributions}
This thesis presents a series of goal-driven vision-conditioned text generation methods/models.
We begin with referring expression generation task in Chapter \ref{chap:ref_exp}.
Unlike generic ``image captioning''
which lacks natural standard evaluation criteria, quality of a
referring expression may be measured by the receiver's ability to
correctly infer which object is being described. Following this
intuition, we propose two approaches to utilize models trained for
referring expression comprehension task to generate better expressions. First, we use a
comprehension module trained on human-generated expressions, as a
``critic" of referring expression generator. The
comprehension module serves as a differentiable proxy of human
evaluation, providing training signal to the generation
module. Second, we use the comprehension module in a
generate-and-rerank pipeline, which chooses from candidate expressions
generated by a model according to their performance on the
comprehension task. We show that both approaches lead to improved
referring expression generation on multiple benchmark
datasets.

We further extend the concept of discriminability idea into image captioning task in Chapter \ref{chap:disc_cap}: 
being able to tell two images apart given the caption for one of them. We propose a way to improve this aspect of caption generation. By incorporating into the captioning training objective a component directly related to ability (by a machine) to disambiguate image/caption matches, we obtain systems that produce much more discriminative caption, according to human evaluation. Remarkably, our approach leads to improvement in other aspects of generated captions, reflected by a battery of standard scores such as BLEU, SPICE etc. Our approach is modular and can be applied to a variety of model/loss combinations commonly proposed for image captioning.

In Chapter \ref{chap:diversity}, we study diversity in image captioning task.
We investigate the effect of different model
architectures, training objectives, hyperparameter
settings, and decoding procedures on the diversity of automatically generated
image captions. Our results show that 1) simple decoding by naive sampling, coupled with
low temperature is a competitive and fast method to produce diverse and accurate
caption sets; 2) training with CIDEr-based reward using Reinforcement learning 
harms the diversity properties of the resulting
generator, which cannot be mitigated by manipulating decoding parameters. In addition,
we propose a new metric \spicen for evaluating both accuracy and
diversity of a set of captions by a single value.

We further study how to control length for image captioning models in Chapter \ref{chap:length}.
Modern image captioning models lack controllability. For example, the length of the generated caption is determined by when the $\langle$eos$\rangle$ (End Of Sentence) token is spit out by the model. To enable length control, we propose two length conditional models based on existing work in abstract summarization domain. We show that our models can faithfully generate captions of the designated length. Compared to baseline method based on manipulating decoding algorithm, our models can generate better long captions and more fluent captions. Qualitatively, we show our methods can generate captions of different style and descriptiveness.

In Chapter \ref{chap:tagging}, we study how to rank/generate informative image tags, focusing on selecting tags that convey the most information about an image. Image tagging systems can be highly accurate in producing tags that are ``technically correct,'' but are often irrelevant, non-useful or not interesting and may differ substantially from tags provided by people for the same image.
Building on the observation that only a small set of words is necessary to recover meaning, we define the \textit{information utility} of a tag: how much information a tag conveys about the image. To quantify utility, we design and train two models,  \tagcap and \tagfeat, which can rank tags by their ability to reconstruct image captions or image features. Our method outperforms multiple baselines in predicting the most important tags, evaluated on a new dataset of importance-focused image tags collected from human raters.


\chapter{Background}
\section{Text generation with visual input}
\subsection{Image Captioning}


Image captioning has been a long-standing topic in the computer vision domain.
%
%
%
%
It was thought to be the most ambitious image understanding test, a natural evolution of other tasks like image classification and object detection. It requires the model to understand images and also have the ability of syntactic and semantic understanding of the language. At the same time, automatically generated image descriptions can be used for image indexing/search, help visually impaired people (applied in IOS magnifier, Facebook etc.), and facilitate office work (used in Microsoft Powerpoint).

Before the deep learning era, the most popular approaches for image captioning were template-based methods and retrieval-based methods. Template-based methods treat captioning as a fill-in-the-blanks problem. Objects, attributes, actions are detected and then filled into the blanks in the pre-defined templates~\cite{farhadi2010every,li2011composing,kulkarni2013babytalk}. Template-based methods are useful but can not generate arbitrary captions; the space of captions is limited to the templates.

Retrieval-based methods are also simple yet effective. Given a target image to caption, retrieval-based methods will first find the similar images in the image pool and then select the final caption from the caption pool constructed by the captions of these similar images~\cite{gong2014improving,hodosh2013framing,ordonez2011im2text,sun2015automatic}. However, these methods cannot generate novel captions either, and the performance is bounded by the size of image pool size. While template-based and retrieval-based methods are not as popular as before, the insight still inspired recent work like Neural baby talk~\cite{lu2018neural}.

Following the success of neural models in machine translation~\cite{bahdanau2014neural,cho2014learning}, many works started to explore deep learning models in image captioning~\cite{karpathy2015deep,donahue2015long,Vinyals2014,Mao2015}. The general approach is first to encode an
image using a deep visual backbone (most likely some convolutional neural network (CNN) like ResNet), and feed this as input to a neural language model (e.g., a recurrent network like LSTM). The neural language model can generate an arbitrary-length sequence of words. Within this generic framework, many efforts~\cite{Vinyals2014,Mao2015, Xu2015,Karpathy2015,lu2016knowing,yang2016review,anderson2018bottom,wang2017skeleton,lu2017knowing,gu2018stack,cornia2020meshed,yang2019auto,huang2019attention,Liu2016,pan2020x} explored different encoder-decoder structures, including attention-based models.

Image features play an important role in caption generation. While early deep learning captioning models used the image features from CNNs like ResNet~\cite{he2016deep}, InceptionNet~\cite{szegedy2015going}, \cite{anderson2018bottom} found that using object features extracted from a pre-trained object-attribute detection network can achieve much stronger performance. Specifically, \cite{anderson2018bottom} train a Faster-RCNN that detects objects and identifies object attributes/category. The model is trained on Visual Genome dataset~\cite{krishna2017visual}. \cite{zhang2021vinvl} further explored different pre-training detection datasets for vision language tasks and found that using more datasets can further boost the captioning performance as well as other vision language tasks.

The training objective for image captioning has also developed in recent years. Following the standard sequence to sequence model, most deep learning based captioning models are trained with Maximum Likelihood Estimation (MLE) combined with some tricks like scheduled sampling.
The limitation of MLE training is its mismatch with the text generation quality. To alleviate this problem ,many reinforcement learning based methods~\cite{Ranzato2016,zhang2017actor,liu2017improved,rennie2017self} were proposed to optimize metrics like BLEU directly. Among these, Self-Critical training~\cite{rennie2017self}, due to its good performance and simplicity, is used as a standard training method for most later captioning papers, as well as some other text generation tasks.

While most image captioning models are based on pre-trained vision models like ResNet or pre-trained Faster-RCNN, the text decoder (language model) was usually trained from scratch, following common practice in NLP field at that time. After the boost of the pre-trained language models like ELMO~\cite{peters2018deep}, BERT~\cite{devlin2018bert}, GPT~\cite{radford2018improving}, vision-language pre-training models also became popular and have shown advantages in image captioning~\cite{zhou2020unified,li2020oscar,zhang2021vinvl}, achieving the new state-of-the-art in image captioning task.

To get more thorough knowledge about image captioning techniques, please refer to the following survey papers~\cite{stefanini2021show,hossain2019comprehensive,liu2019survey,bai2018survey}.

\rtcom{where should I put these.}
\noindent\textbf{Captioning with different setups.}
\cite{sadovnik2012image} proposed the task of discriminative image captioning, with the goal of distinguishing one image from a set of images.
\cite{park2019robust,jhamtani2018learning} proposed the change captioning task, which is to describe only the change between two similar images.
Different from discriminative captioning, change captioning only needs to describe the change/difference.
\cite{gilton2020detection} extended change captioning of two images to that of a visual stream, and proposed CLEVR-Sequence and ``Street Change'' datasets.

Researchers have also focused on novel object captioning: describing images containing objects that are not present in the training caption corpora. \cite{mao2015learning} addressed the task of learning novel visual concepts from only a few captioned images of novel concepts. \cite{hendricks2016deep} benchmarked this task with a simple proof-of-concept dataset where 8 novel object classes are held out from the COCO dataset. The captions of the held out images are not seen during training. \cite{agrawal2019nocaps} later presented a large-scale benchmark, using the images and annotations from the Open Images dataset~\cite{krasin2017openimages}.

Although a big reason for studying image captioning is to help blind people to ``see'' the world, not much work focused on a real-world application for blind people. Recently, \cite{gurari2020captioning} proposed VizWiz-Captions dataset where the images are taken by blind people. Compared to other datasets, VizWiz-Captions suffers more image quality like improper lighting, focus, etc. In addition, during collections, the mechanical turkers were explicitly asked to describe the images in a way that can help a blind person.

To study how to comprehend texts in describing images, \cite{sidorov2020textcaps} proposed image captioning with reading comprehension task and collected a new dataset, TextCaps. This task requires machines to recognize the text in the image, relate to the visual context, and decide how to compose a caption by copying or paraphrasing the detected text.

Other captioning tasks include styled captioning~\cite{gan2017stylenet,shuster2019engaging,you2018image,mathews2018semstyle}, video captioning~\cite{chen2011collecting,xu2016msr,wang2019vatex}, dense captioning~\cite{johnson2016densecap}, dense video captioning~\cite{krishna2017dense,zhou2018towards,lei2020tvr}, visual story telling~\cite{huang2016visual}, news captioning~\cite{tariq2016context,feng2010topic,biten2019good,liu2020visualnews}, vision language navigation instruction generation~\cite{tan2019learning,fried2018speaker} etc.


%

\subsubsection{Image captioning datasets}

While the methods evolve, the datasets people train and evaluate also change through time.
Much early work used Pascal-1K~\cite{rashtchian2010collecting} which includes 1000 images from Pascal-VOC 2008 dataset~\cite{hoiem2009pascal} and 5 captions for each image. Each caption is provided by an Amazon Mechanical Turker who is asked to ``Please describe the image in one complete but simple sentence.''
To increase the size and complexity of the dataset, the same group published back-to-back another two datasets Flickr8k~\cite{hodosh2013framing} and Flickr30k~\cite{young-etal-2014-image}. FLickr8k~\cite{hodosh2013framing} collected 8,092 images from Flickr and 5 captions for each image, focusing on mainly people and animals performing some actions. Flickr30k~\cite{young-etal-2014-image} contains a total of 31,783 images (including original images from Flickr8k), 158,915 captions, and also a denotation graph derived from the captions.

In 2015, COCO-Captions~\cite{chen2015microsoft} was released, extending the COCO dataset with caption annotations. It has 123,287 images and 616,767 captions. The annotations were collected in a similar fashion as the datasets above. In addition to the original simple prompt, COCO-Captions also adds a few quality control criteria, for example, a mandatory minimum length. It is the most popular image caption dataset and is the standard benchmark for image text retrieval and image captioning task. A commonly used train/val/test split for COCO-Captions is from \cite{karpathy2015deep}, with 113,287 training images, 5000 validation images, and 5000 for test.

The datasets above all focus on generic conceptual descriptions~\cite{hodosh2013framing}. \textit{Conceptual} image descriptions focus only on the information that can be obtained from the image alone, in contrast to {\it non-visual} descriptions where captions may include background information that is not contained from the image alone. \textit{Generic} descriptions don't identify the people or locations by their names, e.g., ``describe a person as a woman or a skateboarder, and the scene as a city street or a room.''\footnote{Quoted from \cite{hodosh2013framing}.} That is the reason why the above datasets focus mostly on collecting data from mechanical turkers instead of harvesting the web.

However, it has always been an interest to use the ``free'' data on the internet. BBC captions~\cite{feng2010topic} collected an image caption dataset by downloading 3,361 articles from BBC News website and extracted the news images and their corresponding captions. However, the texts have little connection to images or include information not in the images, describing the news instead of the image.
\cite{ordonez2011im2text} collected SBU Captions containing 1 million images and captions from Flickr. However, the captions 
often describe the situation where the images were taken, rather than what is actually in the image.
Most of the captions are non-visual, and many of the rest only describe a small portion of image detail.

More recently, \cite{mao2016training} collected Pinterest40M, a dataset of 40 million image caption pairs from Pinterest. \cite{sharma2018conceptual} collected Conceptual Captions (CC), a dataset of 3M+ images and their alt-text from the web, which are filtered from a larger set and post-processed to be generic captions. Later, after the wild use of CC in vision-language pre-training literature~\cite{lu2019vilbert,lu202012,chen2019uniter,gan2020large}, a larger version of CC, Conceptual Captions 12M~\cite{changpinyo2021conceptual}, was released. CC12M applies a relaxed filter, thus larger and noisier, and is designed for vision-language pre-training. Recent image-text pre-training models~\cite{radford2021learning,jia2021scaling} also followed this strategy of collecting a large amount of image text pairs from the web to train a powerful model.

While most datasets focus on descriptions of only the salient aspects of the photograph and usually in one short sentence, many datasets also include lengthy captions. IAPR TC-12 dataset (also known as ImageClef)~\cite{Grubinger2006}, proposed in 2006, consists of 20,000 images and 
free-text descriptions. These descriptions typically consist of several sentences or sentence fragments and tend to be long (average length: 23.1 words) and excessively detailed.
%
\cite{krause2017hierarchical} released a dataset of image paragraphs -- telling a story about an image, as part of the Visual Genome~\cite{krishna2017visual} dataset. The dataset contains 19,561 images, and each image contains one paragraph. Localized narrative~\cite{pont2020connecting} was released as part of Open Images V6, providing similar lengthy descriptions, however at the same time, each word in the description is grounded with a trace segment over the image.




Synthetic datasets are also popular benchmarks because it is harder to control the image distribution and the caption distribution when using real images, and the cost of annotating (if there is) is linear to the size of the data. For example, \cite{zitnick2014adopting} used clip art images. They can explicitly generate sets of semantically similar scenes, which is not nearly impossible with real images. CLEVR~\cite{johnson2017clevr} rendered 3d shapes in a scene and enabled automatic synthesis of captions with programs~\cite{kottur2019clevr,park2019robust}. This can help researchers to diagnose different abilities of existing models.


\subsubsection{Caption generation evaluation}
Evaluation protocols can generally be divided into human and non-human/automatic evaluation.

\noindent\textbf{Automatic metrics}

Automatic metrics are popular because they do not involve any human labor and are good for benchmarking and prototyping models. We can further divide automatic metrics into reference-based, reference+image, reference-free methods~\cite{hessel2021clipscore}.

The most commonly used metrics are BLEU~\cite{papineni2002bleu}, METEOR~\cite{denkowski2014meteor}, ROUGE~\cite{lin2004rouge}, CIDEr~\cite{vedantam2015cider}, and SPICE~\cite{anderson2016spice}, where BLEU, METEOR, ROUGE are borrowed from text generation domain, and CIDEr and SPICE are designed for caption generation.

BLEU is a popular metric for machine translation. It computes the n-gram precision between generated text and reference text(s). ROUGE is designed for evaluating text summarization and computes the F-measure of shared common subsequence.
METEOR calculates word-level alignment considering synonym, stemming and paraphrasing. CIDEr measures consensus in image captions by performing a TFIDF weighting for each n-gram. SPICE converts captions into scene graphs and computes the graph matching F1 score.

These metrics are all reference-based methods. During evaluation, for each test image, we will have ground truth captions as references, and we evaluate the generated caption by comparing the generated caption to the ground truth captions.

Other reference-based methods include perplexity, a variant of BERTScore~\cite{yi2020improving}, SMURF~\cite{feinglass2021smurf}, WMD~\cite{kusner2015word,kilickaya2016re}, TER~\cite{snover2005study,elliott2014comparing} and WEmbSim~\cite{sharif2020wembsim}.

Reference+image methods include REO~\cite{jiang2019reo}, LEIC~\cite{cui2018learning}, TIGEr~\cite{jiang2019tiger}, ViLBERTScore~\cite{lee2020vilbertscore}, and FAIEr~\cite{wang2021faier} which use images in addition to textual references. Reference-free methods only take the generated caption and the image as input, by looking at their ``compatibility.'' This category includes CLIPScore~\cite{hessel2021clipscore}, VIFIDEL~\cite{madhyastha2019vifidel}, UMIC~\cite{lee2021umic}.

Some other metrics are focusing on different aspects of caption generation. For example, discriminability metrics, instead of directly measuring the caption quality, measure the ability of the generated caption to correctly retrieve the image from which it is generated~\cite{dai2017contrastive,luo2018discriminability}. In Chapter \ref{chap:disc_cap}, we use a pre-trained image-text retrieval model to calculate such a metric. CHAIR~\cite{rohrbach2018object} evaluates object hallucination level of image captioning models. There are also caption diversity metrics which we will look into in Chapter \ref{chap:diversity}.



\noindent\textbf{Human evaluation}

\cite{Vinyals2014} evaluate the generated captions by asking humans to select from ``Describes without errors,'' ``Describes with minor errors,'' ``Somewhat related to the image,'' and ``Unrelated to the image.''

COCO Image Captioning Challenge~\cite{chen2015microsoft} designed five human judgment metrics:
\begin{itemize}
    \item \textbf{M1}	Percentage of captions that are evaluated as better or equal to human caption.
    \item \textbf{M2}	Percentage of captions that pass the Turing Test.
    \item \textbf{M3}	Average correctness of the captions on a scale 1-5 (incorrect - correct).
    \item \textbf{M4}	Average amount of detail of the captions on a scale 1-5 (lack of details - very detailed).
    \item \textbf{M5}	Percentage of captions that are similar to human description.
\end{itemize}

Conceptual Captions Challenge~\cite{levinboim2019quality} conducted a human evaluation by asking, ``Is this a good caption for the image?'' To measure if this simple question is enough, the authors also collected fine-grained human evaluations on two dimensions: helpfulness and correctness. Helpfulness is divided into three levels (no useful info - info very useful), and correctness is divided into three levels (not correct - correct) and an ``I cannot tell.''

In \cite{vedantam2017context}, human experiments are conducted to measure the discriminability of the image captions. We will describe and use it in Chapter \ref{chap:disc_cap}.

\subsection{Referring expression generation}
The referring expression generation (REG) task has been studied since 1972~\cite{winograd1972understanding}.
Much early work in this area focused on relatively limited datasets,
using synthesized objects in artificial scenes
or simple objects in controlled environments
~\cite{mitchell2013generating,golland2010game,krahmer2012computational,fitzgerald2013learning}.

Recently, more real-world and large-scale datasets were released for this task, resulting in more interest among researchers. \cite{kazemzadeh2014referitgame,yu2016modeling} collected the datasets in the same way by using an interactive game. A speaker describes a specific object in the image, and the listener tries to guess which object is being referred to. This game is cooperative, and only when the listener correctly identifies the object can both people get a reward; the reward will be higher if the time spent is short. RefClef~\cite{kazemzadeh2014referitgame} uses the 20k natural images from ImageClef~\cite{Grubinger2006}, while RefCOCO(+)~\cite{yu2016modeling} uses the images from COCO dataset~\cite{lin2014microsoft}. Concurrent to \cite{yu2016modeling}, \cite{Mao2016} also collected a referring expression dataset RefCOCOg on COCO images. However, \cite{Mao2016} collected it in a non-interactive setting, similar to image caption collection, 
thus resulting in more complex and lengthy expressions.

Synthetic datasets are also built for REG. The benefits of using synthetic datasets are
full control over the scene, and the ability to minimize dataset bias. \cite{liu2019clevr} proposed CLEVR-Ref+ based on CLEVR~\cite{johnson2017clevr}. The ground truth referring expressions can be automatically generated by a program.
\cite{tanaka2019generating} proposed RefGTA, a person-only referring expression dataset where the images are rendered from GTA V. The complexity of the scene can be controlled. Unlike CLEVR-Ref+, RefGTA requires humans to annotate the ground truth.


Following image captioning, the standard evaluation metrics include BLEU, METEOR, CIDEr. As found in \cite{Mao2016}, these automatic evaluation metrics do not correlate to actual referring accuracy. Therefore, human evaluations (human comprehension accuracy) are always preferred. In \cite{tanaka2019generating}, the authors evaluate not only the human comprehension accuracy but also the comprehension time to understand the easy-to-understandness of the expressions.

In general, the architecture for referring expression generation is very similar to image captioning. Instead of encoding an image, the model encodes an object and generates a text description with the object features. However, unlike image captioning which does not have a subjective criterion, referring expressions have a specific goal -- correctly refer to the object discriminatively. Thus, some techniques are specifically designed for this task, as well as our work in Chapter \ref{chap:ref_exp}. We will discuss more in \secref{sec:re_related}.

\subsection{Goal-driven generation}
Much work has considered generating texts that achieve a certain goal.
\cite{krishna2019information} learned to generate questions that maximize
mutual information with an expected answer and the image, to tackle the problem of too generic question generation.
%
\cite{hendricks2016generating} generated visual explanations that can justify the image classification result.
\cite{chen2015mind} aimed to generate captions from image features that can also reconstruct the image features.
\cite{andreas2016reasoning,luo2018discriminability, cohn2018pragmatically,liu2018show} tried to learn a model that can produce captions that are not merely correct but also distinct so as to distinguish the input image from other images. (We will describe our work~\cite{luo2018discriminability} in Chapter \ref{chap:disc_cap}.)
\cite{das2017learning} considered the task of playing a dialog game to guess one image from a set of images. 
%
\cite{holtzman2018learning} incorporated discriminators that focus on different principles to guide the language generation.
\cite{bracha2019informative} proposed to generate tags that minimize the uncertainty in the remaining tags.
\cite{sun2020summarize} proposed SOE, which selects a sentence from a passage as its summary by looking at how well the sentence can reconstruct the whole passage.
\cite{fisch2020capwap} developed a captioning system that generates captions with intended information. To evaluate if a caption contains the intended information, they fed the caption and questions about the image to a Question-Answering machine to see if the caption can answer the questions. This can be helpful for generating captions for blind people.


\section{Autoregressive sequence generation model}
\subsection{Brief introduction}
The autoregressive model is a common way to solve sequence generation problems. It follows humans' intuition to generate sequences in a left-to-right manner. All the models we mentioned in this thesis follow this model.

\noindent\textbf{Math formulation.}
Consider a conditional model for image captioning, $P(\bc|\I)$, where $\bc = \{w_i\}$ is the caption and $\I$ is the image\footnote{We take the image captioning model as an example. It is similar for other conditional generation models.}. By applying the Bayesian rule, we can factorize the probability to:
\begin{align}
    P(\bc|\I) = \prod_{t} P(w_t|\I, w_{<t}). \label{eq:factorization}
\end{align}
Note that, there is no assumption required for this factorization.

Based on this factorization, we can easily sample/decode from the distribution to get the caption, if we have a module that can model $P(w_t|\I, w_{<t})$. A model that models $P(w_t|\I, w_{<t})$ is called autoregressive.

\rtcom{should I put it or not}
Alternatively, you can get $P(\bc|\I)$, using masked language model (modeling fill-in-the-blanks distribution $P(\bc|\bc^\prime, \I)$), or an energy model $P(\bc|\I) \propto \exp(f(\bc, \I))$. Although mathematically we can connect these different formulations, these models would require different model architectures, training strategies, and decoding methods.

\noindent\textbf{Commonly used models} include Long Short Term Memory networks (LSTM)~\cite{hochreiter1997long}, Gated Recurrent Unit (GRU)~\cite{cho2014learning}, Transformer~\cite{vaswani2017attention} and their variants~\cite{rennie2017self,anderson2018bottom,cornia2020meshed,li2019entangled,herdade2019image}. \cite{aneja2018convolutional,gu2017empirical} also attempted to use convolutional networks.

\rtcom{Do we need to introduce LSTM? Att2in???}

\subsection{Training methods}

\subsubsection{MLE and its variants}\label{sec:bg_mle}
The standard way to train such a model is to train with Maximum Likelihood Estimation. After the factorization of the probability in ~\eqref{eq:factorization}, the loss is also equivalent to calculate the cross entropy loss on each word at each time step. This is also known as teacher forcing, because during training, you always feed ground truth tokens in the condition.

\begin{align}
    L_{\text{MLE}} & = -\sum_{i} \log P(\bc^i|\I^i) \\
    & = -\sum_{i}\sum_{t} - \log P(w^i_t|\I^i, w^i_{<t}),
\end{align}
where $(\I^i, \bc^i)$ are image caption pairs in the training set.

One problem of MLE is exposure bias. During training, the model always takes in ground truth caption tokens. However, during generation, the model will take in the tokens generated by the previous time steps. Once the model generates a bad token, the error will propagate because the model has never seen such inputs.



To avoid this, one can instead feed predicted output as next time-step input ($P(w^i_t|\I^i, \widehat w^i_{<t})$). However, such a method does not work well in practice. To get the best of both worlds, \cite{Bengio2015} propose a mixture of two strategies, called scheduled sampling. Instead of always feeding ground truth tokens or predicted tokens, the authors propose to alter the choice according to a curriculum schedule. At the beginning of the training, the model always takes in the ground truth tokens. As learning proceeds, the input at each time step will have an increasingly higher probability (up to a cap) to be a predicted token. Scheduled sampling is shown to be beneficial to generate better sequences.

Another modification to the standard MLE is label smoothing~\cite{szegedy2016rethinking,muller2019does}. As mentioned, MLE can be treated as cross entropy loss between the estimated word posterior from the model and the one-hot ground truth probability. Instead, label smoothing replaces one-hot to a smoothed probability, where the value is high at the ground truth (e.g., 0.9) and the rest density is evenly distributed to the other words. Label smoothing can improve the generalization ability of models.


\subsubsection{Reinforcement learning algorithm}\label{sec:bg_rl}
In addition to train test discrepancy on generation process, there is also a discrepancy between training objective and test time evaluation. Lower MLE loss does not necessarily correspond to higher caption quality. Thus, reinforcement learning methods are borrowed to directly optimize the caption metrics, because the metrics are non-differentiable.
MIXER~\cite{Ranzato2016} use REINFORCE with baseline algorithm for sequence generation training. The results show better performance but the method requires training an additional parameterized function as the baseline.

Self-Critical Sequence Training (SCST)~\cite{rennie2017self} inherits the REINFORCE algorithm from MIXER but discards the learned baseline function. Instead, SCST uses the reward of the greedy decoding result as the baseline, achieving better captioning performance and lower gradient variance. While originally proposed for image captioning task, SCST not only has become the new standard for training captioning models~\cite{yao2017boosting,dognin2019adversarial,luo2018discriminability,jiang2018recurrent,liu2018show,ling2017teaching,liu2018context,chen2019improving,yang2019auto,zhao2018multi}, but also has been applied to many other tasks, like video captioning~\cite{li2018jointly,chen2018less,li2019end}, reading comprehension~\cite{hu2017reinforced}, summarization~\cite{celikyilmaz2018deep,paulus2017deep,wang2018reinforced,pasunuru2018multi}, image paragraph generation~\cite{melas2018training}, speech recognition~\cite{zhou2018improving}.
Many variants of SCST have also been proposed~\cite{anderson2018bottom,gao2019self,luo2020better,cornia2020meshed}. Many methods based on other reinforcement learning methods have also been applied to captioning~\cite{zhang2017actor, ding2017cold, seo2020reinforcing}.

\noindent\textbf{Formal formulation}

The goal of SCST, for example in captioning, is to maximize the expected score of generated captions.
\begin{align}
\max_{\bfgreek{theta}} \bbE_{{\widehat \bc}\sim P(\bc|\I; \bfgreek{theta})}[R({\widehat \bc})]
\end{align}
where ${\widehat \bc}$ is a sampled caption; $\I$ is the image; $P(\bc|\I;\bfgreek{theta})$ is the captioning model parameterized by $\bfgreek{theta}$, and $R(\cdot)$ is the score (e.g., CIDEr score).

Since this objective is non-differentiable with respect to $\bfgreek{theta}$, back propagation is not feasible. To optimize it, a policy gradient method, specifically REINFORCE with baseline~\cite{williams1992simple} is used.
\begin{align}
\nabla_{\bfgreek{theta}} \bbE[R] \approx (R(\widehat \bc)-b) \nabla_{\bfgreek{theta}} \log P(\widehat \bc| \I;\bfgreek{theta})
\end{align}

The policy gradient method allows estimating the gradient of the expected reward from individual samples (the right-hand side) and applying gradient ascent. To reduce the variance of the estimation, a baseline $b$ is needed, and $b$ has to be independent of $\widehat \bc$.

In SCST, the baseline is set to be the score of the greedy decoding output $\bc^\ast=(w^\ast_1,\ldots,w^\ast_T)$, where
\begin{align}
w^\ast_t = \argmax_w P(w_t|\I, w^\ast_{<t}).
\end{align}
Thus, we have
\begin{align}
\nabla_{\bfgreek{theta}} \bbE[R] \approx \left (R(\widehat \bc) - R(\bc^\ast)\right ) \nabla_{\bfgreek{theta}} \log P(\widehat \bc| \I; \bfgreek{theta}).
\end{align}


\subsection{Other methods}
There has also been an exploration of different training objectives. For example,~\cite{yao2016boosting} added some auxiliary tasks like word appearance prediction. \cite{edunov2017classical} used classical structured prediction losses to optimize text generation models. \cite{welleck2019neural} proposed unlikelihood training to avoid the model generating repeated words/phrases. \cite{zhao2018multi} trained image captioning models along with two other related tasks: multi-object classification and syntax generation. \cite{dai2017contrastive} trained the captioning model with a loss inspired by noise contrastive estimation. \cite{chen2015mind} proposed a cycle consistency type of loss where the caption features shall also be able to reconstruct the image. We will describe our training objectives in Chapter \ref{chap:ref_exp} and \ref{chap:disc_cap}.




\subsection{Non-autoregressive models}
There is an interest in recent years to build non-autoregressive models. One drawback for autoregressive models is their sequential generation. The generation time is linear to the sequence length and can not benefit from parallelization. Much work has looked at model architecture, training objectives, decoding methods for non-autoregressive models \cite{gu2017non,wang2019bert,ghazvininejad2019mask,ma2019flowseq,tu2020engine,lee2018deterministic}.
%
In the image captioning domain, \cite{deng2020length} proposed a model that can generate a caption of a designated length (the same goal as Chapter \ref{chap:length}). To do so, they proposed a non-autoregressive model LaBERT based on UniVLP~\cite{zhou2020unified}.

\section{Decoding methods for autoregressive models}\label{sec:bg_decoding}
\subsection{Optimal inference}
The optimal inference is very common for conditional language generation, for example, abstract summarization, image captioning, etc. Take image captioning as an example. The goal is to find $\argmax_{\bc} P(\bc|\I)$, the most likely sentence given the image as the condition. (It is also the general philosophy for any structured prediction problems, where the objective could also be some energy function or potentials.)

Generally speaking, for such a structured prediction problem, the search space for the solution is exponentially large ($O(|V|^l)$, where $V$ is the vocabulary and $l$ is the length), so brute force search is not applicable. When the model has a strong assumption, for example, Hidden Markov Model\rtcom{what to cite for hmm and forward-backward}, there exists a fast exact inference algorithm (forward-backward algorithm) that is of time complexity $O(|V|\cdot l)$. However, for the autoregressive models we consider in this thesis, there exists no polynomial exact inference algorithm. So, there are many approximate inference/decoding algorithms people commonly use.

One common way to get an approximate optimal is to apply \textbf{greedy decoding}. At each time step, we just take the most likely token as the prediction, $w^\ast_t = \argmax_w P(w_t|\I, w^\ast_{<t})$.
Another common algorithm is \textbf{beam search}~\cite{medress1977speech}. It is a heuristic search algorithm that explores the search tree by expanding the most promising node in a limited set. The size of the set is called beam size. When beam size is 1, beam search is equivalent to greedy decoding. When beam size is $|V|^l$, then beam search is exact inference. The larger the beam size, the result is closer to the optimum. However, in practice, people find larger beam size does not always perform well in terms of actual text generation quality.
 Thus, the beam size is usually a hyperparameter that people tune according to the performance of the actual text generation metrics like BLEU, etc.

Another approximate method is to first randomly sample multiple candidates from the distribution $P(\bc|\I)$, and then pick the one with the highest $P(\bc|\I)$ or based on some other criteria. We used this approach in Chapter \ref{chap:ref_exp}. It is also proved to be extremely useful for code completion~\cite{chen2021evaluating}.

\subsection{Sampling}\label{sec:bg_sampling}
Sampling methods are more common in unconditioned language models, for example, to generate a sentence from a pre-trained language model like GPTs~\cite{radford2018improving,radford2019language,brown2020language}; it is also useful for many conditional generation tasks when one wants to get multiple outputs, like diverse image captioning.

Naive sampling is to directly sample from the distribution $P(\bc|\I)$. Because the model is autoregressive, one can sample each word sequentially: $\widehat w_t \sim P(w_t|\I, \widehat w_{<t})$.

Although naive sampling is an exact sampling from the distribution, people have found that the probability distributions of the best current language models have an unreliable tail~\cite{Ari2019nucleus}. Thus sampling from it does not always give us good results. Especially when a bad prediction is sampled, the error propagation will lead to catastrophic failure in the generation.

To alleviate this problem, biased sampling methods are proposed. The intuition behind biased sampling is to avoid selecting ``bad'' tails. Top-K sampling is proposed in \cite{fan2018hierarchical}, and was used in GPT-2~\cite{radford2019language}. At each time step, the model will only sample from the top K most likely tokens according to their relative probabilities.
Formally, given $P(w_t|\I, \widehat w_{<t})$, the Top-K vocabulary $V^{(k)} \subset V$ as the set of size k which maximizes $\sum_{w_t\in V^{(k)}} P(w_t|\I,\widehat w_{<t})$. Then, $\widehat w_t$ is sampled from $P^\prime(w_t|\I, \widehat w_{<t})$, which is defined as follows:
\begin{align}
    P^\prime(w_t|\I, \widehat w_{<t}) = \left\{
        \begin{aligned}
            &\frac{P(w_t|\I, \widehat w_{<t})}{\sum_{w\in V^{(k)}} P(w|\I, \widehat w_{<t})} & & & w_t \in V^{(k)}\\
            &0 & & & w_t \notin V^{(k)}
        \end{aligned}
        \right.
\end{align}
The assumption is whatever is out of Top-K choices are bad predictions.

Another biased sampling is nucleus sampling (Top-p sampling), proposed in \cite{Ari2019nucleus}. The intuition is that at different time steps, the word posterior distributions vary a lot, and the shape of the probability distribution should be used to determine the candidate token set. Given $P(w_t|\I, \widehat w_{<t})$, the top-p vocabulary $V^{(p)} \subset V$ as the smallest set such that $\sum_{w_t\in V^{(p)}} P(w_t|\I,\widehat w_{<t})\geq p$. Then, $w_t$ is sampled from $P^\prime(w_t|\I, \widehat w_{<t})$, which is defined as follows:

\begin{align}
    P^\prime(w_t|\I, \widehat w_{<t}) = \left\{
        \begin{aligned}
            &\frac{P(w_t|\I, \widehat w_{<t})}{\sum_{w\in V^{(p)}} P(w|\I, \widehat w_{<t})} & & & w_t \in V^{(p)}\\
            &0 & & & w_t \notin V^{(p)}
        \end{aligned}
        \right.
\end{align}

For the same time step, nucleus sampling and Top-K sampling are equivalent when $|V^{(p)}| = k$. When $p=0$ and $k=1$, nucleus sampling and Top-K sampling are equivalent to greedy decoding.

\chapter{Discriminability for referring expression generation}\label{chap:ref_exp}

%


%





\section{Problem statement}\label{sec:re_intro}

\begin{figure}[t]
\centering
\begin{minipage}[c]{0.23\linewidth}
\includegraphics[width=\textwidth]{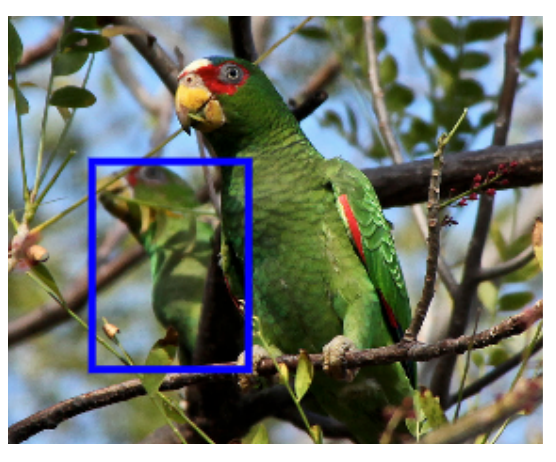}
\end{minipage}
\begin{minipage}[c]{0.24\linewidth}
green bird {\color{red}$\times$}\\
blurry bird $\checkmark$\\
left bird $\checkmark$
\end{minipage}
\begin{minipage}[c]{0.25\linewidth}
\includegraphics[width=\textwidth]{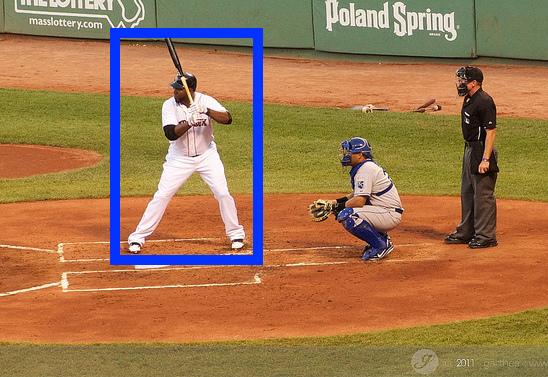}
\end{minipage}
\begin{minipage}[c]{0.24\linewidth}
baseball player {\color{red}$\times$}\\
batter $\checkmark$\\
player in white $\checkmark$
\end{minipage}
\caption{Good and bad referring expressions of objects in the bounding box. On the left, ``green bird'' cannot distinguish between two birds, while the lower two can. On the right, ``baseball player'' cannot distinguish two players while the lower two can.}
\label{fig:re_ref_exp_example}
\end{figure}

Referring expressions~\cite{winograd1972understanding,kazemzadeh2014referitgame} are a special case of image captions. Such
expressions describe an object or region in the image, with the goal
of identifying it uniquely to a listener.
Such expressions are used by humans in daily life to refer things in the scene, for example, when one wants to ask a store worker to retrieve an item from the shelf. A method that can generate such expressions is likely to be important for building multi-modal conversational AI agents, to facilitate the human-machine interaction.
Due to the definition, in contrast to generic
captioning, referring expression generation has a natural
evaluation metric: a human should easily comprehend the description and identify the object being described. Two examples are shown in  \figref{fig:re_ref_exp_example}. Each example contains an object within an image, and a few referring expressions, some flawed and some successful in uniquely identifying the object.

In this chapter, we will consider two tasks related to referring expressions. One is the \emph{generation} task: generating a
discriminative referring expression for an object in an image. The other is the \emph{comprehension}
task:
localizing an object in an image given a referring
expression.  Most
prior work~\cite{Mao2016,Hu2015} address both tasks by building a sequence generation model. 
Such a model can
be used discriminatively for the comprehension task, by inferring the region which maximizes the expression posterior.

In contrast, we draw inspiration from the generator-discriminator structure
in Generative Adversarial
Networks~\cite{Goodfellow2014,Radford2015}. In GANs, the generator
module tries
to generate a signal (e.g., natural image), and the discriminator module
tries to tell real images apart from the generated ones. For our task,
the generator produces referring expressions. We would like these
expressions to be both intelligible/fluent and unambiguous to
humans. Fluency can be encouraged by using the standard
cross entropy loss with respect to human-generated
expressions. On the other hand, we use a comprehension model as the ``discriminator'' which tells if the expression can be correctly referred back to. Note that we can also regard the comprehension model as a ``critic'' of the ``action'' made by the generator where the ``action'' is each generated word~\cite{konda2000actor}. 

Instead of an adversarial relationship between the two modules in GANs,
our architecture is collaborative -- the comprehension module
``tells'' the generator how to improve the expressions it produces. \rtcom{*}In practice, our methods are simpler than GANs as we avoid the alternating
optimization strategy -- the comprehension model is separately trained
on ground truth data and then fixed. We use it as a proxy for human judgment of discriminability, providing an additional signal for referring expression generation.


Specifically there are two ways that we utilize the comprehension
model. The {\bf generate-and-rerank} method uses comprehension at inference time, similar to~\cite{Andreas2016}, where they
tried to produce unambiguous captions for clip-art images. The generation model generates some candidate expressions and passes them through the comprehension model. The final output expression is the one with highest generation-comprehension score which we will describe later.

The {\bf training-by-proxy} method is closer in spirit to GANs. The generation and
comprehension model are connected and the generation model is
optimized to lower discriminative comprehension loss (in addition to
the cross entropy loss). We investigate several training strategies for this method and a trick to make the proxy model trainable by standard back-propagation.  Compared to generate-and-rerank method, the training-by-proxy method doesn't require additional region proposals during test time.\footnote{In our paper, we always fix the comprehension model, however such a cooperative framework can also be trained jointly as shown in later work~\cite{vered2019joint}.}

\section{Related work}\label{sec:re_related}
Referring expressions have
attracted wider interest after the release of the standard large-scale
datasets~\cite{kazemzadeh2014referitgame,yu2016modeling,Mao2016}. In~\cite{Hu2015} an autoregressive sequence
generation model is appropriated for the generation task; comprehension can be done by evaluating
the probability of a sentence $\s$ given an image $P(\s|\I, \r)$ as the matching
score. \cite{Mao2016} proposed a joint
model, in which comprehension and generation aspects are trained using
max-margin Maximum Mutual Information (MMI) training. Both papers used
whole image, region and location/size features. Based on the model
in~\cite{Mao2016}, both~\cite{Nagaraja2016} and~\cite{Yu2016a}
tried to model context regions in their frameworks. 

Our method is trying to combine simple models and replace the max margin
loss, which is orthogonal to modeling context, with a surrogate closer
to the eventual goal -- human comprehension. This requires a
comprehension model, which, given a referring expression, infers the appropriate region in the image.

Among comprehension models proposed in literature,~\cite{Rohrbach2015}
used multi-modal embedding and sets up the comprehension task as multi-class classification. Later,~\cite{Fukui2016} achieved a slight improvement by replacing the concatenation layer with a compact bilinear pooling layer. The comprehension model used in this paper belongs to this multi-modal embedding category.

The ``speaker-listener'' model in~\cite{Andreas2016} attempted to produce discriminative captions that
can tell images apart. The speaker is trained to generate captions, and a
listener to prefer the correct image over a wrong one, given the
caption. At test time, the listener reranks the captions sampled
from the speaker. Our generate-and-rerank method is based on translating this idea to referring expression generation.

Concurrent to us, \cite{yu2017joint} proposed a unified model, Speaker-Listener-Reinforcer, that jointly learns to solve both comprehension and generation. To encourage the discriminability of generated expressions, they add a discriminative reward-based reinforcer to guide the generation of more discriminative expressions, with the help of a reinforcement learning algorithm.

\section{Generation and comprehension models}\label{sec:re_model}
We start by defining the two modules used in the collaborative
architecture we propose. Each of these can be trained as a standalone
machine for the task it solves, given a dataset with ground truth
regions/referring expressions. 

\begin{figure}[t]
\centering
\begin{minipage}[c]{0.48\textwidth}
\centering
\includegraphics[width=\textwidth]{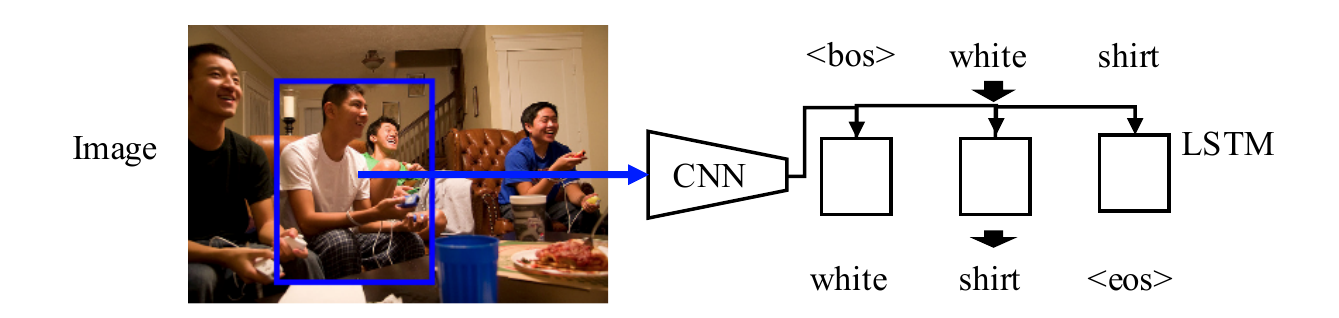}
\caption{Illustration of how the generation model describes the region inside the blue bounding box. $\langle$bos$\rangle$ and $\langle$eos$\rangle$ stand for beginning and end of sentence.}
\label{fig:re_generation} 
\end{minipage}
\hspace{1em}
\begin{minipage}[c]{0.48\textwidth}
\centering
\includegraphics[width=\textwidth]{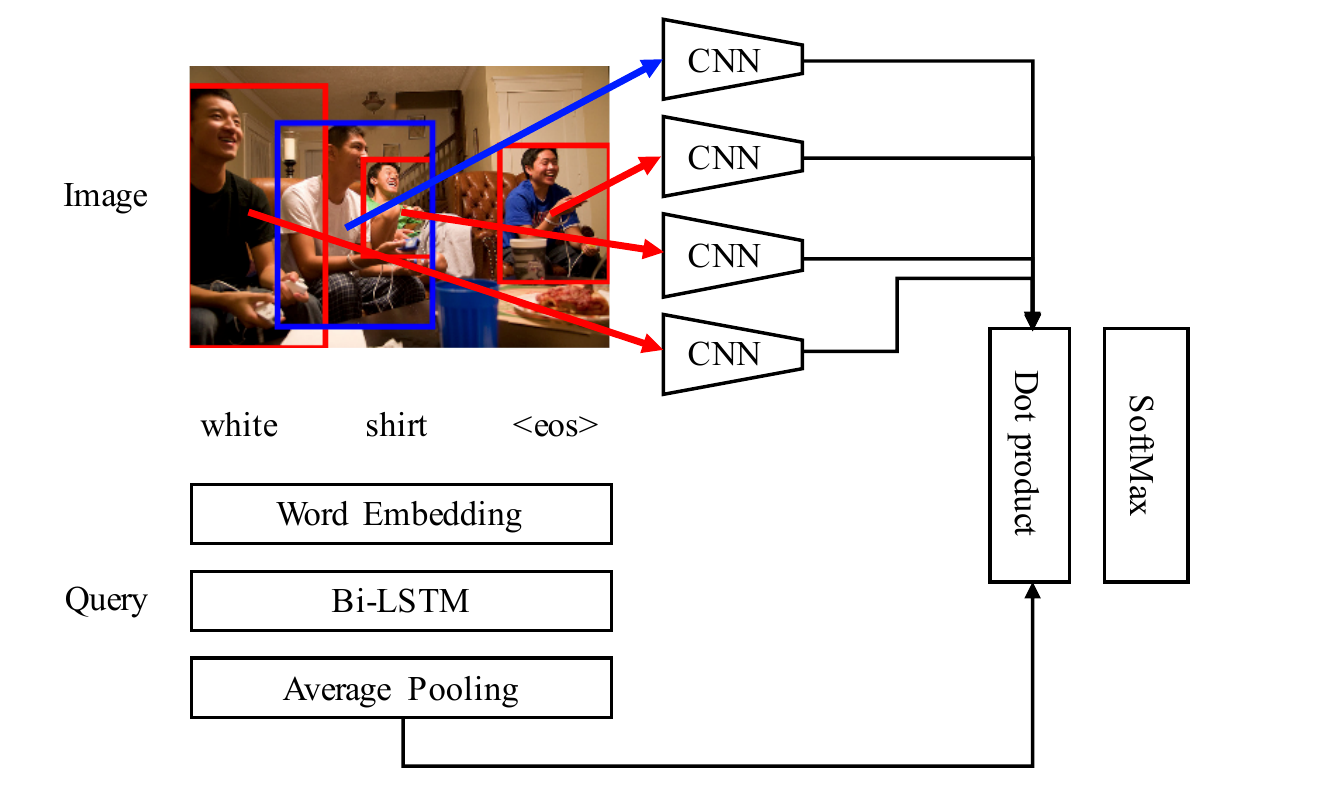}
\caption{Illustration of comprehension model. The blue bounding box is the target region, and the red ones are incorrect regions. The CNNs share the weights.}
\label{fig:re_comprehension} 
\end{minipage}
\end{figure}

\subsection{Expression generation model}\label{sec:re_generation}
We use a simple expression generation model introduced
in~\cite{yu2016modeling,Mao2016}. The generation task takes inputs of an
image $\I$ and an internal region $\r$, and outputs an expression $\s$,
$\G:\I\times \r \rightarrow \s$. To solve this task, we consider a conditional model $P_\G(\s|\I,\r)$.

Specifically, the generation model is an encoder-decoder network, where the encoder
encodes the visual information from $\r$ and $\I$ and the decoder generates the expression $\s$. \figref{fig:re_generation} shows the structure of the
generation model.

\noindent\textbf{Encoder.} Following~\cite{Hu2015,yu2016modeling,Nagaraja2016}, we use encoding that includes: target
object representation $\o$, global context feature $\g$, and
location/size feature $\l$. In our experiments, $\o$ is the
activation on the cropped region $\r$ of the last fully connected layer
{\tt fc7} of VGG-16~\cite{simonyan2014very}; $\g$ is the {\tt fc7} activation on the
whole image $\I$; $\l$ is a 5D vector encoding
 the relative location of top left corner and bottom right corner of the bounding box $\r$, as well as the bounding box
size relative to the image size, i.e. $\l=[\frac{x_{tl}}{W}, \frac{y_{tl}}{H}, \frac{x_{br}}{W}, \frac{y_{br}}{H}, \frac{w\cdot h}{W\cdot H}]$, 
where $W$, $H$, $w$, $h$ are the width and height of the image and the bounding box; $x_{tl/br}$ and $y_{tl/br}$ are the coordinates of the corners.

The final visual feature vector $\bv$ of the
region is a linear transformation (plus bias terms) of the concatenation of three
features $[\o,\g,\l]$:
\begin{align}
\bv &= f_\bv(\I, \r)=\W_\bv([\o, \g, \l])+\b_\bv.
\end{align}

\noindent\textbf{Decoder.} We
use a uni-directional LSTM~\cite{hochreiter1997long} as the decoder. At each time step the LSTM takes in the visual feature and the previous word and outputs the distribution of next
word:
\begin{align}
    \h_t = \text{LSTM}([\bv, \s_{t-1}], \h_{t-1}) \\
    P_\G(\s_t|\s_{<t}, \I, \r) = \text{Softmax}(\W_\G \h_t+\b_\G),
\end{align}
where $\s_{t}$ is the t-th word of  
ground truth expression $\s$.

The model is trained to minimize the cross entropy loss, equivalent to maximum likelihood estimation (MLE). The per-sample loss is defined as follows:
\begin{align}
  \label{eq:re_loss_gen}
L_{\G} &=  - \sum_{t = 1}^{T} \log P_\G(\s_{t}|\s_{<t}, \I, \r),\\
\end{align}
where $T$ is the length of $\s$.

In practice, to sample an expression $\s$ from $P_\G(\s|\I,\r)$, one can use beam search, greedy decoding, or sampling.

\subsection{Expression comprehension model}
The comprehension task is to select a region (bounding box) $\widehat \r$ from a
set of regions  $\calR = \{\r_i\}$ given a query expression $\q$ and the image $\I$.
\begin{align}
\C: \I\times \q\times \calR \rightarrow \r,\  \r\in \calR
\end{align}

We also define the comprehension model as a conditional model $P_\C(\r|\I, \q, \calR)$.
The estimated region given a comprehension model is: $\widehat \r = \argmax_\r P_\C(\r|\I, \q, \calR)$. As shown latter, $P_\C$ can be formulated as either a multi-class or multi-label probability. 

We design the comprehension model similar to~\cite{Rohrbach2015}. To build the model, we first define a similarity function $f_{\text{sim}}$ that computes the similarity between a region $\r_i$ and a query $\q$.
We use the same visual feature encoder as in generation model. For the query expression, we use a one-layer bi-directional LSTM~\cite{Graves2013} to encode it. We take the averaging over the hidden vectors of each time step so that we can get a fixed-length representation for an arbitrary-length query: 
\begin{align}
\h_\q = \text{LSTM}(\E \Q),
\end{align}
where $\E$ is the word embedding matrix
and $\Q$ is a one-hot representation of the query expression, i.e. $\Q_{t,j} = \1(\q_t = j)$.

Unlike \cite{Rohrbach2015}, which uses concatenation + MLP to
calculate the similarity, we use a simple dot product as in \cite{Dhingra2016}.
\begin{align}
f_{\text{sim}}(\I, \r_i, \q) = \bv_i^\intercal \h_\q.
\end{align}

We consider two formulations of the comprehension
task as classification. The
per-region logistic loss 
\begin{align}
  \label{eq:re_com_bin}
  P_\C(\r_i|\I, \q)=\sigma(f_{\text{sim}}(\I, \r_i, \q)),\\
  L_{\C_{\text{bin}}}=-\log P_\C(\r_{i^\ast}|\I, \q) - \sum_{i \neq i^\ast}\log (1- P_\C(\r_i|\I, \q)),
\end{align}
where $\r_{i^\ast}$ is ground truth region. This loss corresponds to a per-region
classification: is this region the right match for the expression or not. 

The softmax loss
\begin{align}
  \label{eq:re_com_multi}
  P_\C(\r_i|\I, \q, \calR)&=\frac{\exp(f_{\text{sim}}(\I, \r_i, \q))}{\sum_j \exp(f_{\text{sim}}(\I, \r_j, \q))},\\
  L_{\C_{\text{multi}}}&=-\log P_\C(\r_{i^\ast}|\I, \q, \calR),
\end{align}
frames the task as a multi-class
classification: which region in the set should be matched to the
expression. 

The model is trained to minimize the comprehension loss $L_\C$, where $L_\C$ is either $L_{\C_{\text{bin}}}$ or $L_{\C_{\text{multi}}}$.

\figref{fig:re_comprehension} shows the structure of our generation model under multi-class classification formulation.

\section{Methods}\label{sec:method}

\newcommand{\propo}{\mathcal{R}}

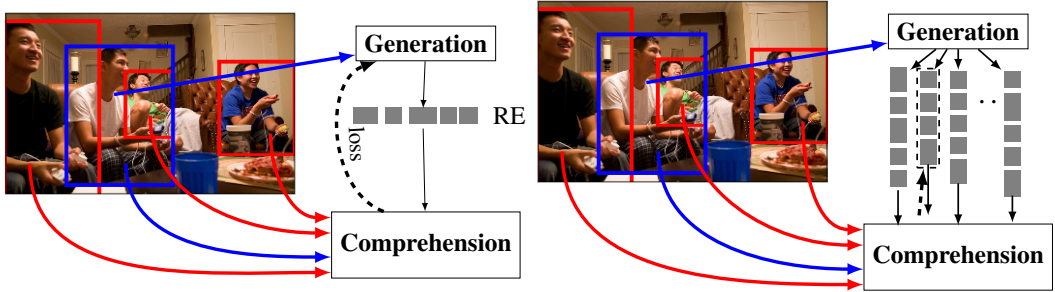
\begin{figure}[!thb]
\begin{center}
\begin{minipage}[b]{0.45\textwidth}
\scalebox{0.8}{
  \begin{tikzpicture}
\node[rectangle,draw=black,thick,fill overzoom
    image=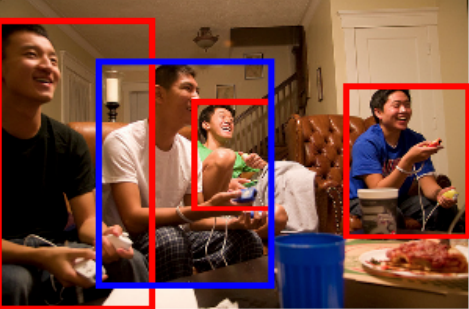,minimum width=4.8cm,minimum height=3cm,opacity=.5] (image) {};    
    \node[rectangle,minimum width=2cm,draw=black,right=of image,yshift=1cm] (g) {\textbf{Generation}};
    \draw[ultra thick,blue,-latex] (-.6,.15)--(g);
    \node[black!50,rectangle,fill=black!50,minimum width=.4cm,minimum height=.2cm] at
    ($ (g)+(-1,-1.2) $) (t1) {};
    \node[right=of t1,right=3pt,black!50,rectangle,fill=black!50,minimum
    width=.24cm,minimum height=.2cm] (t2) {};
    \node[right=of t2,right=3pt,black!50,rectangle,fill=black!50,minimum
    width=.46cm,minimum height=.2cm] (t3) {};
    \node[right=of t3,right=1pt,black!50,rectangle,fill=black!50,minimum
    width=.18cm,minimum height=.2cm] (t4) {};
    \node[right=of t4,right=1pt,black!50,rectangle,fill=black!50,minimum
    width=.3cm,minimum height=.2cm] (t5) {};
    \node[right=of t5,right=3pt] {RE};

    \node[below=of g,below=2.5cm,rectangle,minimum width=2cm,minimum
    height=1.1cm,draw=black] (c) {\textbf{Comprehension}};

    \draw[black,-latex] (g) -- ($ (g)!.9!(t3) $);
    \draw[black,-latex] ($ (t3)!.1!(c) $) -- (c);
    \draw[ultra thick,blue,-latex] (-.4,-1) to[out=270,in=180] ($
    (c.west) + (0,-.2)$);
    \draw[ultra thick,red,-latex] (-2,-1) to[out=270,in=180] ($
    (c.south west) + (0,.1) $);
    \draw[ultra thick,red,-latex] (2,-.5) to[out=270,in=180] ($
    (c.north west) + (0,-.1) $);
    \draw[ultra thick,red,-latex] (0,-.2) to[out=270,in=180] ($
    (c.west) + (0,.2) $);
    \draw[ultra thick,dashed,-latex] ($ (c.north) + (-.7,0)$)
    to[out=160,in=-160] node[midway,above,sloped] {\small \ \ \ loss} (g);
  \end{tikzpicture}
}
\end{minipage}
%
\begin{minipage}[b]{0.45\textwidth}
\scalebox{0.8}{
 \begin{tikzpicture}
\node[rectangle,draw=black,thick,fill overzoom
    image=figures/refexp/image.pdf,minimum width=4.8cm,minimum height=3cm,opacity=.5] (image) {};    
    \node[rectangle,minimum width=2cm,draw=black,right=of image,yshift=1cm] (g) {\textbf{Generation}};
    \draw[ultra thick,blue,-latex] (-.6,.15)--(g);
    \node[black!50,rectangle,fill=black!50,minimum width=.2cm,minimum height=.4cm] at
    ($ (g)+(-1,-.8) $) (t11) {};
    \node[below=of t11,below=2pt,black!50,rectangle,fill=black!50,minimum
    width=.2cm,minimum height=.24cm] (t12) {};
    \node[below=of t12,below=2pt,black!50,rectangle,fill=black!50,minimum
    width=.2cm,minimum height=.4cm] (t13) {};
    \node[below=of t13,below=2pt,black!50,rectangle,fill=black!50,minimum
    width=.2cm,minimum height=.15cm] (t14) {};
    \node[below=of t14,below=2pt,black!50,rectangle,fill=black!50,minimum
    width=.2cm,minimum height=.22cm] (t15) {};

    \node[black!50,rectangle,fill=black!50,minimum width=.2cm,minimum height=.1cm] at
    ($ (g)+(-.5,-.8) $) (t31) {};
    \node[below=of t31,below=2pt,black!50,rectangle,fill=black!50,minimum
    width=.2cm,minimum height=.3cm] (t32) {};
    \node[below=of t32,below=2pt,black!50,rectangle,fill=black!50,minimum
    width=.2cm,minimum height=.3cm] (t33) {};
    \node[below=of t33,below=2pt,black!50,rectangle,fill=black!50,minimum
    width=.2cm,minimum height=.43cm] (t34) {};

    \node[black!50,rectangle,fill=black!50,minimum width=.2cm,minimum height=.29cm] at
    ($ (g)+(0,-.8) $) (t21) {};
    \node[below=of t21,below=2pt,black!50,rectangle,fill=black!50,minimum
    width=.2cm,minimum height=.24cm] (t22) {};
    \node[below=of t22,below=2pt,black!50,rectangle,fill=black!50,minimum
    width=.2cm,minimum height=.1cm] (t23) {};
    \node[below=of t23,below=2pt,black!50,rectangle,fill=black!50,minimum
    width=.2cm,minimum height=.22cm] (t24) {};
    \node[below=of t24,below=2pt,black!50,rectangle,fill=black!50,minimum
    width=.2cm,minimum height=.4cm] (t25) {};
   
     \node[right=of t22,right=1pt,draw=none] {\large \ldots};

     \node[black!50,rectangle,fill=black!50,minimum width=.2cm,minimum height=.1cm] at
    ($ (g)+(0.9,-.8) $) (t41) {};
    \node[below=of t41,below=2pt,black!50,rectangle,fill=black!50,minimum
    width=.2cm,minimum height=.45cm] (t42) {};
    \node[below=of t42,below=2pt,black!50,rectangle,fill=black!50,minimum
    width=.2cm,minimum height=.1cm] (t43) {};
    \node[below=of t43,below=2pt,black!50,rectangle,fill=black!50,minimum
    width=.2cm,minimum height=.3cm] (t44) {};
    \node[below=of t44,below=2pt,black!50,rectangle,fill=black!50,minimum
    width=.2cm,minimum height=.44cm] (t45) {};

    \draw[thick,black,-latex] (g) -- ($ (g)!.8!(t11)$);
    \draw[thick,black,-latex] (g) -- ($ (g)!.8!(t21)$);
    \draw[thick,black,-latex] (g) -- ($ (g)!.8!(t31)$);
    \draw[thick,black,-latex] (g) -- ($ (g)!.8!(t41)$);

    \node[below=of g,below=2.9cm,rectangle,minimum width=2cm,minimum
    height=1.1cm,draw=black] (c) {\textbf{Comprehension}};

    \draw[thick,black,-latex] ($ (t15)+(0,-.2) $) -- ++(0,-.6);
    \draw[thick,black,-latex] ($ (t25)+(0,-.2)$) -- ++(0,-.7);
    \draw[thick,black,-latex] ($ (t34)+(0,-.2) $) -- ++(0,-.85)
    coordinate (p1);
    \draw[thick,black,-latex] ($ (t45)+(0,-.2) $) -- ++(0,-.45);

    \node[draw,thick,dashed,inner sep=1pt,fit=(t31) (t34)] (selbox) {};
    \draw[dashed,ultra thick,-latex] ($ (p1) + (-.15,0) $) -- (selbox);

    \draw[ultra thick,blue,-latex] (-.4,-1) to[out=270,in=180] ($
    (c.west) + (0,-.2)$);
    \draw[ultra thick,red,-latex] (-2,-1) to[out=270,in=180] ($
    (c.south west) + (0,.1) $);
    \draw[ultra thick,red,-latex] (2,-.5) to[out=270,in=180] ($
    (c.north west) + (0,-.1) $);
    \draw[ultra thick,red,-latex] (0,-.2) to[out=270,in=180] ($
    (c.west) + (0,.2) $);
  \end{tikzpicture}
}
\end{minipage}

\end{center}
\caption{Left: Training-by-proxy. The comprehension model must
  correctly identify the target (blue) region based on a referring expression (RE);
  comprehension loss (dashed) is propagated to the generator. Right:
Generate-and-rerank. Generator produces multiple REs; comprehension
model evaluates them based on its ability to identify the true (blue)
region from them, and selects (dashed) the best RE.}
\end{figure}

Once we have trained the comprehension model, we can start using it as
a proxy for human comprehension, to guide the expression generator. 
\subsection{Training-by-proxy}\label{sec:re_proxy}
Consider a referring expression $\widehat \s$ generated by $\G$ for a given training example
of an image/region pair $(\I,\r)$. The generation loss $L_{\G}$ tries to maximize the probability of the
ground truth expression $\s$. The comprehension model $\C$ can provide an
alternative, complementary signal: how to modify $\G$ to maximize
the discriminability of the generated expression, so that $\C$ selects
the correct region $\r$ among the proposal set
$\propo$. Intuitively, this signal should push down on probability of
a word if it is unhelpful for comprehension, and pull that probability
up if it is helpful. 
%
%
%
%
Ideally, we hope to minimize the comprehension loss
of the output of the generation model $L_{\C}(\r|\I, \calR, \widehat \Q)$,
where $\widehat \Q$ is the 1-hot encoding of $\widehat \s = \G(\I, \r)$,
with $K$ rows (vocabulary size) and $T$ columns (sequence length).

We hope to update the generation model according to the gradient of loss with respect to the model parameter $\theta_\G$. By chain rule, 
\begin{align}
\frac{\partial L_{\C}}{\partial \theta_\G} = 
\frac{\partial L_{\C}}{\partial \tilde \Q}\frac{\partial \tilde \Q}{\partial \theta_\G}
\end{align}
However, $\tilde \Q$ is inferred by sampling from the word posterior which is not differentiable. To address this issue, \cite{Ranzato2016,Bahdanau2016,Yu2016a} applied reinforcement
learning methods. Instead, we use the continuous relaxation as in~\cite{tu2018learning,Xu2015, DzmitryBahdana2014}.

We define a matrix $\BP$ which has the same size as $\tilde \Q$. The t-th column of $\BP$ is -- instead of the one-hot vector of the generated t-th word -- the distribution of the t-th word produced by $P_\G$, i.e. 
\begin{align}
\BP_{t,j} = P_\G(\widehat \s_t = j).
\end{align}
$\BP$ has several good properties. First, $\BP$ has the same size as
$\tilde \Q$, so that the we can still compute the query feature by
replacing the $\tilde \Q$ by $\BP$, i.e. $h_{\widehat \s} =
\text{LSTM}(\E\BP)$. Secondly, the sum of each column in $\BP$ is 1, just
like $\tilde \Q$. Thirdly, $\BP$ is differentiable with respect to
generator's parameters $\theta_\G$.

Now, the gradient of $\theta_\G$ is approximated by:
\begin{align}\label{eq:re_diffapprox}
\frac{\partial L_{\C}}{\partial \theta_\G} = 
\frac{\partial L_{\C}}{\partial \BP}\frac{\partial \BP}{\partial \theta_\G}
\end{align}

We will use this approximate gradient in the following three algorithms. The three algorithms share the same loss form but differ in the input token of the LSTM.

\noindent\textbf{Compound loss.}\label{sec:re_CL}
The cross entropy loss~\eqref{eq:re_loss_gen}
encourages fluency of the generated expression, but disregards its
discriminability. Thus, the first way is to use the comprehension model as
a source of an additional loss signal. 

Technically, we define a
compound loss 
\begin{equation}\label{eq:re_lcompound}
L = L_{\G} + \lambda L_{\C} 
\end{equation}
where the comprehension loss $L_\C$ is either the
logistic~\eqref{eq:re_com_bin} or the softmax~\eqref{eq:re_com_multi} loss; the
balance term $\lambda$ determines the relative importance of fluency
vs. discriminability in $L$.

Both $L_{\G}$ and $L_{\C}$ take as input $\G$'s distribution over the
t-th word $P_\G(\s_t|\s_{<t}, \I,\r)$, where
the preceding words $\s_{<t}$ are from the \textbf{ground truth} expression $\s$.



\noindent\textbf{Modified Scheduled sampling training.}
Our final goal is to generate comprehensible expression during test time. However, in compound loss, the loss is calculated given the ground truth input while during test time each token is sampled from the model, thus yielding a discrepancy between how the model is used during training and at test time. Inspired by similar motivation,~\cite{Bengio2015} proposed scheduled sampling which allows the model to be trained with a mixture of ground truth data and predicted data. Here, we propose modified schedule sampling training to train our model.

During training, at each iteration $i$,
we draw a random variable $\alpha$ from a Bernoulli
distribution with probability $\epsilon_i$. If $\alpha = 1$, we feed
the ground truth expression to the LSTM and minimize cross entropy loss $L_\G$. If
$\alpha = 0$, we sample the whole sequence from $P_\G$,
and the input of comprehension model is $P_\G(\s_i|\widehat \s_{<t}, \I, \r)$,
where $\widehat \s_{<t}$
are the \emph{sampled} words. Different from original scheduled sampling where the input at each time step is randomly selected from the ground truth or prediction, our modified version operates on the sequence level so the input sequence is either the ground truth sequence or the predicted sequence.
We update
the model by minimizing the comprehension loss $L_{\C}$. Therefore, $\alpha$
serves as a dispatch mechanism, randomly alternating between the
sources of data for the LSTMs and the components of the compound loss.

We start the modified scheduled sampling training from a pre-trained generation model trained on cross entropy loss using the ground truth sequences. As the training progresses, we linearly decay $\epsilon_i$ until a preset minimum value $\epsilon$. The minimum probability prevents the model from degeneration. If we don't set the minimum, when $\epsilon_i$ goes to 0, the model will lose all the ground truth information, and will be purely guided by the comprehension model. 
This would lead the generation model to discover those pathological optima that exist in neural models~\cite{goodfellow2014explaining,lewis2017deal}.
In this case, the generated expressions would do ``well'' on comprehension model, but no longer be intelligible to humans. See \algoref{alg:re_mss} for the pseudo-code.

\makeatletter
\def\BState{\State\hskip-\ALG@thistlm}
\makeatother

\begin{algorithm}
\caption{Modified scheduled sampling training}\label{alg:re_mss}
\begin{algorithmic}[1]
\State Train the generation model $\G$.
\State Set the offset $k$ ($0\leq k\leq 1$), the slope of decay $c$, minimum probability $\epsilon$, number of iterations $N$.
\For{$i=1, N$}
\State $\epsilon_i \gets \max(\epsilon, k - ci)$
\State Get a sample from training data $(\I, \r, \s)$,
\State Sample the $\alpha$ from Bernoulli distribution, where $P(\alpha = 1) = \epsilon_i$
\If {$\alpha = 1$}
\State Minimize $L_{\G}$ with the ground truth input $\s$.
\Else
\State Sample a sequence $\widehat \s$ from $P_\G(\s|\I, \r)$
\State Minimize $L_{\C}$ with the input $P_\G(\s_t|\widehat \s_{<t}, \I, \r)$, $t \in [1, T]$.
\EndIf
\EndFor

\end{algorithmic}
\end{algorithm}

\noindent\textbf{Stochastic mixed sampling.}
Since modified scheduled sampling training samples a whole sentence at a time, it would be hard to get a useful signal due to error propagation through time. We hope to find a method that can slowly deviate from the original model and explore.

Here we borrow the idea from mixed incremental cross entropy reinforce (MIXER)~\cite{Ranzato2016}. Again, we start the model from a pre-trained generator. Then we introduce model predictions during training with an annealing schedule so as to gradually teach the model to produce stable sequences. For each iteration $i$, we feed the input for the first $m_i$ steps, and sample the rest $T - m_i$ words, where $0\leq m_i \leq T$, and $T$ is the maximum length of expressions. We define $m_i = m + \Delta m$, where $m$ is a base step size which gradually decreases during training, and $\Delta m$ is a random variable which follows geometric distribution: $P(\Delta m = k) = (1-p)^{k+1}p$. This $\Delta m$ is the difference between our method and MIXER.

By introducing this term $\Delta m$, we can control how much supervision we want to get from ground truth by tuning the value $p$. This is also for preventing the model from producing pathological optima. Note that, when $p$ is 0, $\Delta m$ will always be large enough so that it's just cross entropy loss training. When $p$ is 1, $\Delta m$ will always equal to 0, which is equivalent to MIXER annealing schedule. See \algoref{alg:re_smixec} for the pseudo-code.

\begin{algorithm}
\caption{Stochastic mixed sampling}\label{alg:re_smixec}
\begin{algorithmic}[1]
\State Train the generation model $\G$.
\State Set the geometric distribution parameter $p$, maximum sequence length $T$, period of decay $d$, number of iterations $N$.
\For{$i=1, N$}
\State $m \gets \max(0, T - \lceil i / d\rceil)$
\State Sample $\Delta m$ from geometric distribution with success probability $p$
\State $m_i \gets \min(T, m + \Delta m)$
\State Get a sample from training data, $(\I, \r, \s)$
\State Run the $\G$ with ground truth input in the first $m_i$ steps, and sampled $\widehat \s_{m_i+1 \ldots T}$ from $\G$ for the remaining steps.
\State Get $L_{\G}$ on first $m_i$ steps, and $L_{\C}$ on the mixed sampled sentence $\{\s_{1\ldots m_i}, \widehat \s_{m_i+1 \ldots T}\}$.
\State Minimize $L_{\G} + \lambda L_{\C}$. 
\EndFor

\end{algorithmic}
\end{algorithm}

\subsection{Generate-and-rerank}
Here we propose a different strategy to generate better expressions. Instead of using comprehension model for fine-tuning a generation model, we compose the comprehension model during test time. The pipeline is similar to \cite{Andreas2016}.

Unlike in \secref{sec:re_generation} where we only need image $\I$ and region $\r$ as input, we will optionally need the region set $\calR$ too during generation when $L_\C$ is chosen to be $L_{\C_{\text{multi}}}$.

Suppose we have a generation model and a comprehension model which are pre-trained. The steps are as follows:
\begin{enumerate}
\item Generate candidate expressions $\{\widehat \s^1, \ldots, \widehat \s^n \}$ according to $P_\G(\s|\I,\r)$.
\item Select $\widehat \s^k$ with $k = \argmax_i \text{Score}(\widehat \s^i)$.
\end{enumerate}
Here, we use sampling because we want the candidate set to be more diverse. We define the score function as a weighted combination of the log perplexity and comprehension loss (we use $L_{\C_{\text{multi}}}$ as an example here).
\begin{align}
	\text{Score}(\widehat\s) = \frac{1}{T}\log P_\G(\widehat\s|\I, \r) + \gamma \log P_\C(\r|\I, \calR, \widehat\s),
\end{align}
where $T$ is the length of $\widehat\s$.

This can be viewed as a weighted joint log probability of an expression to be both natural and unambiguous. The log perplexity term ensures the fluency, and the comprehension loss ensures the chosen expression to be discriminative.

\section{Experiments and results}\label{sec:re_experiment}

We evaluate our methods on the following referring expression datasets.

{\bf RefClef(ReferIt)}~\cite{kazemzadeh2014referitgame} contains 20,000 images from IAPR TC-12 dataset~\cite{Grubinger2006}, together with segmented image regions from SAIAPR-12 dataset~\cite{Escalante2010}. The dataset is split into 10,000 for training/validation and 10,000 for test. There are 59,976 (image, bounding box, description) tuples in the trainval set and 60,105 in the test set.

{\bf RefCOCO(UNC RefExp)}~\cite{yu2016modeling} consists of 142,209 referring expressions for
50,000 objects in 19,994 images from COCO~\cite{lin2014microsoft}, collected using the ReferitGame~\cite{kazemzadeh2014referitgame}.

{\bf RefCOCO+}~\cite{yu2016modeling} has 141,564 expressions for 49,856 objects in
19,992 images from COCO. ``Location words'' are disallowed, focusing
the dataset more on appearance based descriptions.

{\bf RefCOCOg(Google RefExp)}~\cite{Mao2016} consists of 85,474
referring expressions for 54,822 objects in 26,711 images from COCO; it contains
longer and more flowery expressions than RefCOCO and RefCOCO+.

\subsection{Comprehension}
We first evaluate our comprehension model on human expressions,
to assess its ability to provide useful signals. We consider two comprehension settings as in
~\cite{Mao2016,yu2016modeling,Nagaraja2016}. First, the input region set
$\calR$ contains only ground truth bounding boxes for objects, and a
hit is defined by the model choosing the correct region the expression
refers to. In the second setting, $\calR$ contains proposal regions
generated by Fast-RCNN detector~\cite{girshick2015fast}, or by other
proposal generation
methods~\cite{edge-boxes-locating-object-proposals-from-edges}. Here a
hit occurs when the model chooses a proposal with intersection over
union (IoU) with the ground truth of 0.5 or higher. We used pre-computed proposals from~\cite{yu2016modeling,Mao2016,Hu2015} for all four datasets.

In RefCOCO and RefCOCO+, there are two test sets: testA contains people and testB contains all other objects. For RefCOCOg, we evaluate on the validation set. For RefClef, we evaluate on the test set.

We train the model using Adam optimizer~\cite{kingma2014adam}. The word embedding size is 300, and the hidden size of bi-LSTM is 512. The length of visual feature is 1024. For RefCOCO, RefCOCO+ and RefCOCOg, we train the model using softmax loss, with ground truth regions as training data. For RefClef dataset, we use the logistic loss. The training regions are composed of ground truth regions and all the proposals from Edge Boxes~\cite{edge-boxes-locating-object-proposals-from-edges}. The binary classification is to tell if the proposal is a hit or not.

\begin{table}[ht]
  \setlength{\tabcolsep}{4pt}
  \centering
    \begin{tabular}{l|c|c|c|c|c|c|c|c|c|c}
    \multirow{3}[6]{*}{} & \multicolumn{4}{c|}{RefCOCO}  & \multicolumn{4}{c|}{RefCOCO+} & \multicolumn{2}{c}{RefCOCOg}\\
\cline{2-11}          & \multicolumn{2}{c|}{testA} & \multicolumn{2}{c|}{testB} & \multicolumn{2}{c|}{testA} & \multicolumn{2}{c|}{testB} & \multicolumn{2}{c}{Val}\\
\cline{2-11}          & GT    & DET   & GT    & DET   & GT    & DET   & GT    & DET   & GT    & DET\\
    \toprule
    MLE~\cite{yu2016modeling} & 63.15 & 58.32 & 64.21 & 48.48 & 48.73 & 46.86 & 42.13 & 34.04 & 55.16 & 40.75\\
    MMI~\cite{yu2016modeling}   & 71.72 & 64.90 & 71.09 & 54.51 & 52.44 & 54.03 & 47.51 & 42.81 & 62.14 & 45.85\\
    visdif+MMI~\cite{yu2016modeling} & 73.98 & 67.64 & 76.59 & 55.16 & 59.17 & 55.81 & \textbf{55.62} & 43.43 & 64.02 & 46.86\\
    Neg Bag~\cite{Nagaraja2016} & \textbf{75.6} & 58.6 & \textbf{78.0} & \textbf{56.4} & -     & -     & -     & -     & \textbf{68.4} & 39.5\\
    \midrule
    Ours  & 74.14 & \textbf{68.11} & 71.46 & 54.65 & 59.87 & 56.61 & 54.35 & \textbf{43.74} & 63.39 & 47.60\\
    Ours(w2v) & 74.04 & 67.94 & 73.43 & 55.18 & \textbf{60.26} & \textbf{57.05} & 55.03 & 43.33 & 65.36 & \textbf{49.07}\\
    \bottomrule
    \end{tabular}%
  \caption{Comprehension results on RefCOCO, RefCOCO+, RefCOCOg
    datasets. Numbers are accuracies in percentage. GT: the region set contains ground truth bounding boxes; DET: region set contains proposals generated from detectors. w2v means initializing the embedding layer using pretrained word2vec.}
   \label{tab:re_comprehension}

\end{table}%

\begin{table}[!bth]
  \centering
    \begin{tabular}{l|c}
          & RefCLEF Test\\
    \toprule
    SCRC~\cite{Hu2015}  & 17.93\\
    GroundR~\cite{Rohrbach2015} & 26.93\\
    MCB~\cite{Fukui2016}   & 28.91\\
    \midrule
    Ours  & 31.25\\
    Ours(w2v) & \textbf{31.85}\\
    \bottomrule
    \end{tabular}%
    \caption{Comprehension on RefClef (Edge Boxes proposals). Numbers in percentage.}\label{tab:re_compre_clef}
\end{table}%

Table~\ref{tab:re_comprehension} shows our results on RefCOCO, RefCOCO+
and RefCOCOg compared to recent algorithms. Among these, MMI
represents Maximum Mutual Information which uses max-margin loss to
help the generation model better comprehend. With the same visual
feature encoder, our model can get a better result compared to MMI in
\cite{yu2016modeling}. Our model is also competitive with recent, more complex
state-of-the-art models~\cite{yu2016modeling,Nagaraja2016}. Table~\ref{tab:re_compre_clef} shows our results on RefClef where we only test
in the second setting to compare to existing results; our model, which
is a modest modification of~\cite{Rohrbach2015}, obtains state-of-the-art
accuracy in this experiment.

\subsection{Generation}

\begin{table}[thbp!]
  \setlength{\tabcolsep}{3.5pt}
  \centering
    \begin{tabular}{l|ccccc|ccccc}
    \multicolumn{11}{c}{RefCOCO}\\
    \toprule
    \multirow{2}[4]{*}{} & \multicolumn{5}{c|}{testA}           & \multicolumn{5}{c}{testB}\\
\cline{2-11}          & Acc   & B1 & B2 & R & M & Acc   & B1 & B2 & R & M \\
    \midrule
    MLE~\cite{yu2016modeling} & 74.80\% & 47.7 & 29.0 & 41.3 & 17.3 & 72.81\% & 55.3 & \textbf{34.3} & 49.9 & 22.8\\
    MMI~\cite{yu2016modeling}   & 78.78\% & 47.8 & \textbf{29.5} & 41.8 & 17.5 & 74.01\% & 54.7 & 34.1 & 49.7 & 22.8\\
    CL    & \textbf{80.14\%} & 45.86 & 25.52 & 40.96 & 17.8 & 75.44\% & 54.34 & 32.66 & \textbf{50.56} & \textbf{23.26}\\
    MSS   & 79.94\% & 45.74 & 25.32 & 41.26 & 17.59 & \textbf{75.93\%} & 54.03 & 32.32 & 50.10 & 22.97\\
    SMIXEC & 79.99\% & \textbf{48.55} & 28.00 & \textbf{42.12} & \textbf{18.48} & 75.60\% & \textbf{55.36} & 34.26 & 50.12 & 23.20\\
    \midrule
    \midrule
    MLE+sample & 78.38\% & 52.01 & \textbf{33.91} & 44.84 & 19.74 & 73.08\% & 58.42 & 36.86 & 51.61 & 24.25\\
    Rerank & \textbf{97.23\%} & \textbf{52.09} & \textbf{33.91} & \textbf{45.82} & \textbf{20.49} & \textbf{94.96\%} & \textbf{59.35} & \textbf{37.63} & \textbf{52.59} & \textbf{25.05}\\
    \bottomrule
    \multicolumn{11}{c}{}\\ 
    \multicolumn{11}{c}{RefCOCO+}\\
    \toprule
    \multirow{2}[4]{*}{} & \multicolumn{5}{c|}{testA}           & \multicolumn{5}{c}{testB}\\
\cline{2-11}          & Acc   & B1 & B2 & R & M & Acc   & B1 & B2 & R & M \\
    \midrule
    MLE~\cite{yu2016modeling} & 62.10\% & \textbf{39.1} & \textbf{21.8} & \textbf{35.6} & 14.0 & 46.21\% & 33.1 & 17.4 & 32.2 & 13.5\\
    MMI~\cite{yu2016modeling}   & 67.79\% & 37.0 & 20.3 & 34.6 & 13.6 & 55.21\% & 32.4 & 16.7 & 32.0 & 13.3\\
    CL    & 68.54\% & 36.83 & 20.41 & 33.86 & 13.75 & \textbf{55.87\%} & \textbf{34.09} & \textbf{18.29} & \textbf{34.32} & \textbf{14.55}\\
    MSS   & \textbf{69.41\%} & 37.63 & 21.26 & 34.25 & 14.01 & 55.59\% & 33.86 & 18.23 & 33.65 & 14.24\\
    SMIXEC & 69.05\% & 38.47 & 21.25 & 35.07 & \textbf{14.36} & 54.71\% & 32.75 & 17.16 & 31.94 & 13.54\\
    \midrule
    \midrule
    MLE+sample & 62.45\% & 39.25 & 22.56 & 35.81 & 14.56 & 47.86\% & 33.54 & 18.19 & 33.70 & 14.70\\
    Rerank & \textbf{77.32\%} & \textbf{39.56} & \textbf{22.84} & \textbf{36.36} & \textbf{14.84} & \textbf{67.65\%} & \textbf{33.68} & \textbf{18.43} & \textbf{34.41} & \textbf{15.09}\\
    \bottomrule
    \end{tabular}
  \caption{Expression generation evaluated by automated metrics.
  Numbers (except Acc) are scaled by 100x.
  Acc: accuracy of the trained comprehension model on generated
    expressions. B1: BLEU-1. B2: BLEU-2. R: ROUGE. M: METEOR. We separately mark in bold the best results for
    single-output methods (top) and sample-based methods (bottom) that
  generate multiple expressions and select one.}
  \label{tab:re_generation}
\end{table}%

\begin{table}[thbp!]
  \centering
    \begin{tabular}{lccccc}
    \multicolumn{6}{c}{RefCOCOg (val)}\\
    \toprule
          & Acc   & BLEU-1 & BLEU 2 & ROUGE & METEOR\\
  \midrule
    Max Lik. & 61.96\% & 43.7 & 27.3 & 36.3 & 14.9\\
    MMI   & 70.38\% & 42.8 & 26.3 & 35.4 & 14.4\\
    CL    & 70.74\% & \textbf{44.39} & \textbf{27.51} & \textbf{36.95} & 15.52\\
    MSS   & \textbf{70.80\%} & 43.77 & 26.97 & 36.33 & 15.24\\
    SMIXEC & 70.02\% & 43.38 & 26.83 & 36.50 & \textbf{15.75}\\
    \midrule
    \midrule
    MLE+sample & 66.72\% & 44.06 & 27.55 & 37.48 & 15.26\\
    Rerank & \textbf{76.65\%} & \textbf{44.10} & \textbf{27.72} & \textbf{37.82} & \textbf{15.36}\\
    \bottomrule
    \end{tabular}%
  \caption{Expression generation result on RefCOCOg val. Numbers (except Acc) are scaled by 100x.}
  \label{tab:re_generation_cocog}
\end{table}%

Table~\ref{tab:re_generation},~\ref{tab:re_generation_cocog} shows our evaluation of different methods based on automatic caption generation metrics (BLEU, ROUGE, METEOR). We also add an `Acc' column, which is the ``comprehension accuracy'' of the generated expressions according to our comprehension model: how well our comprehension model can comprehend the generated expressions.

The two baseline models are maximum likelihood estimation (MLE) and maximum mutual information (MMI) from~\cite{yu2016modeling}. Our methods include compound loss (CL), modified scheduled sampling (MSS), stochastic mixed incremental cross entropy comprehension (SMIXEC) and also generate-and-rerank (Rerank). We also include a baseline MLE+sample to better analyze generate-and-rerank method. 

For the two baseline models and our three strategies for training-by-proxy
 method, we use greedy decoding to generate an expression. The MLE+sample
and Rerank methods generate an expression by choosing a best one from 100 sampled expressions.

Our generate-and-rerank (Rerank in Table~\ref{tab:re_generation},~\ref{tab:re_generation_cocog}) model gets consistently better results on automatic comprehension accuracy and on fluency-based metrics like
BLEU. To see if the improvement is from sampling or reranking, we also
sample 100 expressions on MLE model and choose the one with
the lowest perplexity (MLE+sample in Table~\ref{tab:re_generation},~\ref{tab:re_generation_cocog}). The
generate-and-rerank method still has better results, showing benefit
from comprehension-guided reranking.

Compared to two baselines, our three variants of training-by-proxy methods can all get higher
accuracy under the comprehension model (Acc), indicating the effectiveness of our training pipeline.

Among the three training schedules of training-by-proxy, there is no
clear winner. In RefCOCO, our SMIXEC method outperforms basic MMI
method with higher comprehension accuracy and higher caption
generation metrics. The compound loss and modified scheduled sampling
seem to suffer from optimizing over the accuracy. However, in RefCOCO+
and RefCOCOg, our three models seem to perform very differently. The
compound loss works better on RefCOCO+ testB and RefCOCOg; the SMIXEC
works best on RefCOCO+ testA. The source of this disparity is unclear
to us.
\paragraph{Human evaluations.} We also evaluated
human comprehension of the generated expressions, since this is known
to not be perfectly correlated with automatic metrics~\cite{yu2016modeling}. For 100 images randomly chosen from each
split of RefCOCO and RefCOCO+, subjects had to click on the object which
they thought was the best match for a generated expression. Each
image/expression example was presented to two subjects, with a hit
recorded only when both subjects clicked inside the correct region.

\begin{table}[tbph]
  \centering
    \begin{tabular}{l|cc|cc}
    \multirow{2}[4]{*}{} & \multicolumn{2}{c|}{RefCOCO} & \multicolumn{2}{c}{RefCOCO+}\\
          & testA & testB & testA & testB\\
    \toprule
    MMI~\cite{yu2016modeling}   & 53\%  & 61\%  & 39\%  & 35\%\\
    SMIXEC & 62\%  & 68\%  & \textbf{46\%} & 25\%\\
    Rerank & \textbf{66\%} & \textbf{75\%} & 43\%  & \textbf{47\%}\\
    \bottomrule
    \end{tabular}%
  \caption{Human evaluation results.}
  \label{tab:re_generation_human}%
\end{table}%

\begin{figure*}[hbt]
\centering
\includegraphics[width=\textwidth]{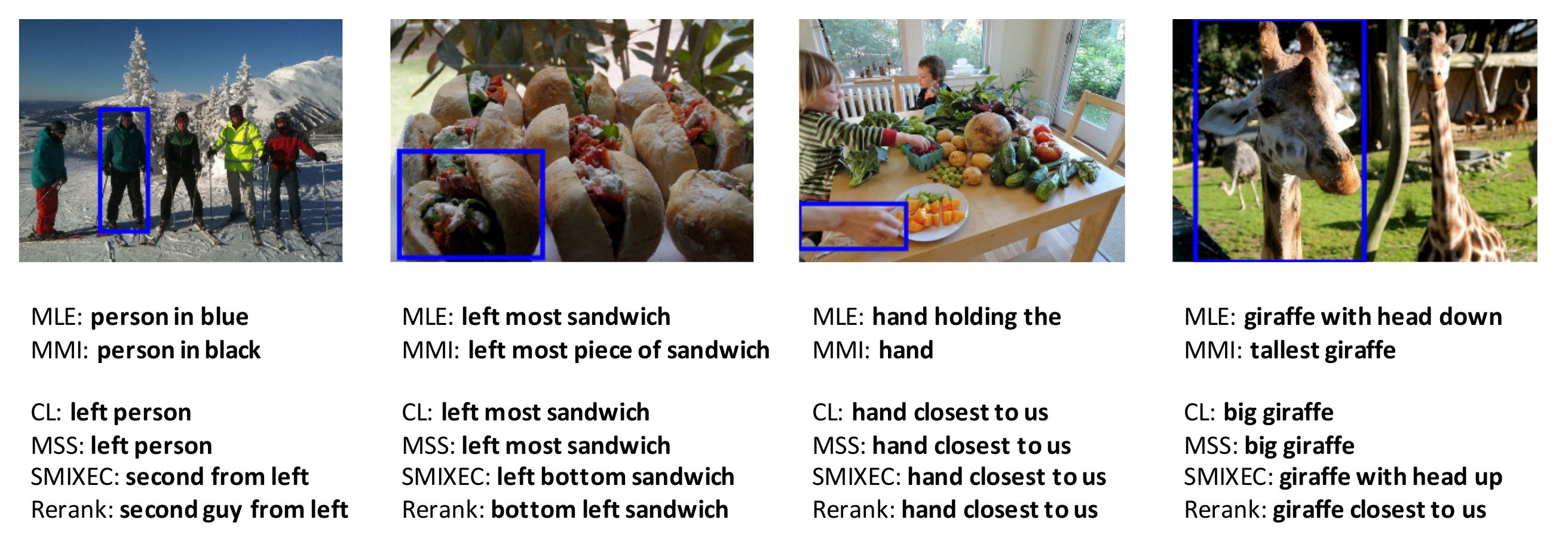}
\caption{Generation results on (left to right)RefCOCO testA, RefCOCO testB, RefCOCO+ testA and RefCOCO+ testB.}
\label{fig:re_gen_result}
\end{figure*}

The results from human evaluations with MMI, SMIXEC and our
generate-and-rerank method are in
Table~\ref{tab:re_generation_human}. On RefCOCO, both of our
comprehension-guided methods appear to generate better (more
informative) referring expressions. 
On RefCOCO+, the result are similar to those on RefCOCO on testA, but our training-by-proxy methods
performs less well on testB.
\figref{fig:re_gen_result} shows some example generation results on test images.

\section{Conclusion}\label{sec:re_end}
In this work, we propose to use learned comprehension models to guide
generating better referring expressions. Comprehension guidance can be
incorporated at training time, with a training-by-proxy method, where
the discriminative comprehension loss (region retrieval based on
generated referring expressions) is included in training the
expression generator. Alternatively comprehension guidance can be used
at
test time, with a generate-and-rerank method which uses model
comprehension score to select among
multiple proposed expressions. Empirical evaluation
shows both to be promising, with the generate-and-rerank method
obtaining particularly good results across datasets.

In the next chapter, we will apply ``training-by-proxy'' to image captioning task, to increase the discriminability of generated image captions.

\chapter{Discriminability in image captioning}\label{chap:disc_cap}





\section{Problem Statement}\label{sec:disccap_intro}

Image captioning is a task of mapping images to text for human consumption. Broadly speaking, in order for a caption to be good it must satisfy two requirements: it should be a \emph{fluent}, well-formed phrase or sentence in the target language; and it should be \emph{informative}, or \emph{descriptive}, conveying meaningful non-trivial information about the visual scene it describes. Our goal in the work presented here is to improve captioning on both of these fronts. 

Because these properties are somewhat vaguely defined, objective evaluation of caption quality is a challenge.
However, a number of metrics have emerged as preferred, if imperfect, evaluation measures. Comparison to human (``gold standard'') captions collected for test images is done by means of metrics, such as BLEU, CIDEr and SPICE. 

In contrast, to assess how informative a caption is, we may design an explicitly discriminative task the success of which would depend on how accurately the caption describes the visual input. One approach to this is to consider \emph{referring expressions}~\cite{kazemzadeh2014referitgame} (as in Chapter \ref{chap:ref_exp}): captions for an image region, produced with the goal of unambiguously identifying the region within the image to the recipient. We can also consider the ability of a recipient to identify an entire image that matches the caption, out of two (or more) images~\cite{vedantam2017context}. In this chapter, we want to extend our idea in Chapter \ref{chap:ref_exp} to image captioning: focusing on the \emph{discriminability} of generated captions.

\begin{figure}[t]
\begin{minipage}[t]{.47\linewidth}
\begin{centering}
\includegraphics[height=4.4cm]{figures/disccap/COCO_train2014_000000489297.jpg}\\
\end{centering}
{\bf Human}: a large jetliner taking off from an airport runway
\\
{\bf Att2in+CIDER}: a large airplane is flying in the sky\\
{\bf Ours}: a large airplane taking off from \textcolor{green!50!black}{runway}
\end{minipage}%
\hspace{1em}
\begin{minipage}[t]{.47\linewidth}
\begin{centering}
\includegraphics[height=4.4cm]{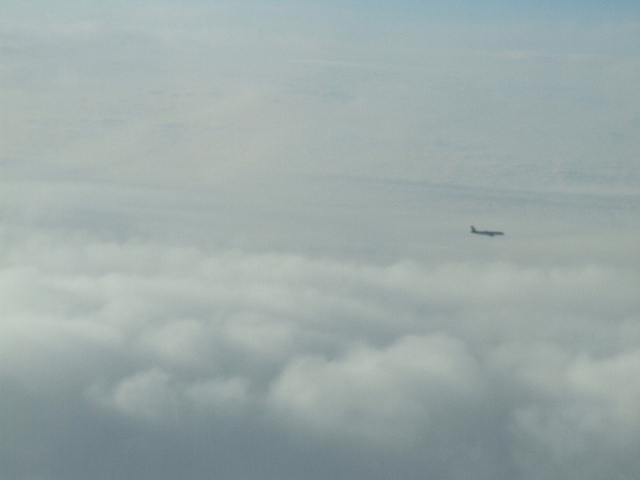}\\
\end{centering}
{\bf Human}: a jet airplane flying above the clouds in the distance\\
{\bf Att2in+CIDER}: a large airplane is flying in the sky\\
{\bf Ours}: a plane flying in the sky with a \textcolor{green!50!black}{cloudy sky}
\end{minipage}
\caption{Example captions generated by humans, an existing automatic system (Att2in+CIDER~\cite{rennie2017self}), and a model trained with our proposed method (Att2in+CIDER+DISC(1), see \secref{sec:disccap_exp}).}
\label{fig:disccap_human_vs_machine_vs_baseline} 
\end{figure}

Traditionally used training objectives, such as maximum likelihood estimation (MLE) or CIDEr (with Self-Critical Training),
tend to encourage the model to ``play it safe,'' often yielding overly general captions as illustrated in \figref{fig:disccap_human_vs_machine_vs_baseline}. Despite the visual differences in the image, a top captioning system~\cite{rennie2017self} produces the same caption for both images. In contrast, humans appear to notice ``interesting'' details that are likely to distinguish the image from other potentially similar images, even without explicitly being requested to do so. (We confirm this assertion empirically in \secref{sec:disccap_exp}.)

To reduce this gap, we propose to incorporate an explicit measure for discriminability into learning a caption generator, as part of the training loss. Our discriminability objective is derived from the ability of a (pre-trained) \emph{retrieval model} to match the correct image to its caption significantly stronger than any other image in a set, and vice versa (caption to correct image above other captions).
The role of retrieval model is similar to the referring expression comprehension model in Chapter \ref{chap:ref_exp}.

Language-based measures like CIDEr reward machine captions for mimicking human captions, and so since, as we state above, human captions are discriminative, one could expect these measures to be correlated with descriptiveness. However, in practice, given an imperfect caption generator, there may be a tradeoff between fluency and descriptiveness; our training regime allows us to negotiate this tradeoff and ultimately improve both aspects of a generator. 

Our discriminability objective can be added to any gradient-based learning procedure for caption generators. We show in \secref{sec:disccap_exp} that it can improve existing models for all metrics evaluated. In particular, to our knowledge, we established new state of the art in discriminative captioning at the time.

\section{Related work}\label{sec:disccap_related}

\noindent\textbf{Visual Semantic Embedding methods.}
Image-Caption retrieval has been considered as a task relying on image captioning~\cite{Vinyals2014,Karpathy2015,Ma2016,Xu2015}. However, it can also be regarded as a multi-modal embedding task. In previous work~\cite{frome2013devise,Wang2016a,WSABIE,wang2017learning,nam2016dual,Vendrov2015} visual and textual embeddings are trained with the objective to minimize matching loss, e.g., ranking loss on cosine distance, or to enforce partial order on captions and images. 


\noindent\textbf{Discriminative captioning} was proposed in~\cite{sadovnik2012image,vedantam2017context,Andreas2016}: given a set of other images, called distractors, the generated captions of each image have to distinguish one from others. In the ``speaker-listener'' model~\cite{Andreas2016}, 
the speaker is trained to generate captions, and a listener to prefer the correct image over a wrong one, given the caption. At test time, the listener reranks the captions sampled from the speaker.
\cite{vedantam2017context} proposed a decoding mechanism which can suppress the caption elements that are common for both target image and distractor image. In contrast to our work, both~\cite{vedantam2017context} and~\cite{Andreas2016}  require the distractor to be presented prior to caption production. We aim to generate distinctive captions a-priori, without a specific distractor at hand, like humans appear to do.

Referring expressions is another flavor of discriminative captioning task. \cite{yu2016joint} and \cite{luo2017comprehension}~(Chapter \ref{chap:ref_exp}) learned to generate more discriminative referring expressions guided by a referring expression comprehension model. The techniques there are tied to the task of describing a region within an image, while our goal here is to describe natural scenes in their entirety.

\noindent\textbf{Visual Dialog} 
has recently attracted interests in the field~\cite{geman2015visual,das2016visual}.
While it's hard to evaluate generic `chat',
\cite{das2017learning,de2016guesswhat} propose goal-driven visual dialog tasks and datasets. \cite{das2017learning} propose the `image guessing' game where two agents -- Q-BOT and A-BOT -- who communicate in natural language dialog so that Q-BOT can select an unseen image from a lineup of images.  GuessWhat Game~\cite{de2016guesswhat} is similar, but guess an object in an image during a dialog. In another related effort~\cite{li2017learning} the machine must show understanding the difference between two images by asking a question that has different answers for two images. Our work shares the ultimate purpose (producing text that allows image identification) with these efforts, but in contrast to those, our aim is to generate a caption in a single ``shot''. This is somewhat similar to round 0 of the dialog in~\cite{das2017learning}, where the agent is given a caption generated by~\cite{Karpathy2015} (without regard to any discrimination task) and chooses an image from a set. Since our captions are shown in \secref{sec:disccap_exp} to be both fluent and discriminative, switching to using them may improve/shorten visual dialog.



\noindent\textbf{Similar work.} Finally, some work is similar to ours in its goals (learning to produce discriminative captions) and, to a degree, in techniques.
The motivation in~\cite{dai2017contrastive} is similar, but the focus is on caption (rather than image) retrieval. The objective is contrastive: pushing the negative captions from different images to have lower probability than positive captions using noise contrastive learning.
In \cite{lu2017best}, more meaningful visual dialog responses are generated by distilling knowledge from a discriminative model trained to rank different dialog responses given the previous dialog context.
\cite{dai2017towards,shetty2017speaking} proposed using Conditional Generative Adversarial Network to train image captioning. They both learn a discriminator to distinguish human captions from machine captions. For more detailed discussion of \cite{dai2017contrastive,dai2017towards,shetty2017speaking}, see Appendix \ref{app:disccap_comparison}. \rtcom{We add the discussion in our original appendix. I wanted to merge, but I feel like that would break the flow, so I still put it in the appendix. What do you think?}

Despite being motivated by a desire to improve caption discriminability, all these methods are fundamentally remain tied to the objective of matching the surface form of human captions, and do not include an explicitly discriminative objective in training. 
Ours was the first work to incorporate both image retrieval and caption retrieval into caption generation training. We can easily ``plug'' our method into existing models, for instance combine it with CIDEr optimization, leading to improvements in metrics across the board: both the discriminative metrics (image identification) and traditional metrics such as ROUGE end METEOR (Tables~\ref{tab:disccap_val_result_lang},~\ref{tab:disccap_val_result_disc},~\ref{tab:disccap_disc_result_on_test}).

Concurrent to us, \cite{liu2018show} proposed a similar speaker-listener model for image captioning, but also leverage unlabeled images for semi-supervised training. Later, \cite{vered2019joint} extended our work and enabled joint training between the captioning and the retrieval model, achieving better performance. With the same goal, \cite{wang2020compare} proposed a simple yet effective method: reweighing the importance of ground truth captions to encourage the discriminability of generated captions.
\section{Models}\label{sec:disccap_models}
Our model involves two main ingredients: a \emph{retrieval} model that scores images caption pairs, and a caption generator that maps an image to a caption. We describe the models used in our experiments below; however we note that our approach is very modular, and can be applied to different retrieval models and/or different caption generators.
Then we describe the key element of our approach: combining these two ingredients in a collaborative framework. We use the retrieval score derived from the retrieval model to help guide training of the generator.
The intuition is similar to the idea of training-by-proxy (Section \ref{sec:re_proxy}) in Chapter \ref{chap:ref_exp} (but the technical detail is different).

\subsection{Retrieval model}
The retrieval model we use is taken from~\cite{faghri2017vse++}. It is a visual semantic embedding network which embeds both text and image into a shared semantic space in which a similarity/compatibility score can be calculated between a caption and an image. We outline the model below, for details see~\cite{faghri2017vse++}.

We start with an image $\I$ and caption $\bc$. First, domain-specific encoders compute an image feature vector $\bfgreek{phi}(\I)$, e.g., using a CNN, and a caption feature vector $\bfgreek{psi}(\bc)$, e.g. using an RNN-based text encoder. These feature vectors are then projected into a joint space by $\W_\I$ and $\W_\bc$.
\begin{align}
f(\I) = \mathbf{W}_\I^T\bfgreek{phi}(\I)\\
g(\bc) = \mathbf{W}_\bc^T\bfgreek{psi}(\bc)
\end{align}
The similarity score between $\I$ and $\bc$ is now computed as the cosine similarity in the embedding space:
\begin{align}
s(\I,\bc) = \frac{f(\I)\cdot g(\bc)}
{\|f(\I)\|\|g(\bc)\|}
\end{align}
The parameters of the caption embedding $\bfgreek{psi}$, as well as the maps $\mathbf{W}_\I$ and $\mathbf{W}_\bc$, are learned jointly, end-to-end, by minimizing the contrastive loss defined below. In our case, the image embedding network $\bfgreek{phi}$ is a pre-trained CNN and the parameters are fixed during training. 


\noindent\textbf{Contrastive loss} is a sum of two hinge losses:
\begin{align}
L_{\textsc{CON}}(\bc,\I) &=
\max_{\bc^\prime}[\alpha+s(\I,\bc^\prime)-s(\I,\bc)]_+ + \max_{\I^\prime}[\alpha+s(\I^\prime,\bc)-s(\I,\bc)]_+
\label{eq:disccap_con_loss}
\end{align}
where $[x]_+ \equiv \max(x,0)$. The max in~\eqref{eq:disccap_con_loss} is taken, in practice, over a batch of $B$ images and corresponding captions. The (image,caption) pairs $(\I,\bc)$ are  correct matches, while $(\I^\prime,\bc)$ and $(\I,\bc^\prime)$ are incorrect (e.g., $\bc^\prime$ is a caption that does not describe $\I$).
Intuitively, this loss ``wants'' the model to assign the matching pair $(\I,\bc)$ the score higher (by at least $\alpha$) than the score of any mismatching pair, either $(\I^\prime,\bc)$ or $(\I,\bc^\prime)$ that can be formed from the batch. This objective can be viewed as a hard negative mining version of triplet loss~\cite{schroff2015facenet}.





\subsection{Discriminability objective}
The ideal way to measure discriminability is to pass it to a human and get feedback from them, like in~\cite{LingNIPS2017}. However it is rather costly and very slow to collect. Here, we propose instead to use a pre-trained retrieval model to work as a proxy for human perception. Specifically, we define the discriminability objective as follows.

Suppose we have a captioning system, parameterized by a set of parameters $\bfgreek{theta}$, that can output conditional distribution over captions for an image, $P(\bc|\I;\bfgreek{theta})$. Then, the objective of maximizing the discriminability objective is 
\begin{align}
\max_{\bfgreek{theta}}\mathbb{E}_{\widehat{\bc}\;\sim\;P(\bc|\I;\bfgreek{theta})} \left[-L_{\textsc{CON}}(\widehat{\bc}, \I)\right]
\end{align}
In other words, the objective involves the same contrastive loss used to train the retrieval model. However, when training the retrieval model, the loss relies on ground truth image-caption pairs (with human-produced captions), and is back-propagated to update parameters of the retrieval model. 
Now, when using the loss to train caption generators, an input batch (over which the max in~\eqref{eq:disccap_con_loss} is computed) will include pairs of images with captions that are sampled from the posterior distribution produced by a caption generator; the signal derived from the discriminability objective will be used to update parameters $\bfgreek{theta}$ of the generator, while holding the retrieval model fixed.

\subsection{Caption generation models}
We now briefly describe two caption generation models used in our experiments; both are introduced in~\cite{rennie2017self} where further details can be found. Discussion on training these models with the discriminability objective is deferred until \secref{sec:disccap_training}.

\noindent\textbf{FC Model.}
The first model is a simple sequence encoder initialized with visual features. Words are represented with an embedding matrix (a vector per word). Visual features are extracted from an image using a CNN. 

The caption sequence is generated by a form of LSTM model. Its output at time $t$ depends on the previously generated word and on the context/hidden state (evolving as per LSTM update rules). At training time the word fed to the state $t$ is the ground truth word $w_{t-1}$; at test time, it is the predicted word $\widehat{w}_{t-1}$. The first word is a special $\langle$bos$\rangle$ (beginning of sentence) token. The sequence production is terminated when the special $\langle$eos$\rangle$ token is output. The image features (mapped to the dimensions of word embeddings) serve at the initial ``word'' $w_{-1}$, fed to the state at $t=0$.

\noindent\textbf{Att2in model} is an attention-based LSTM. Each image is now encoded into a set of spatial features: each encodes a sub-region of the image. At each word $w_t$, the context (and thus the output) depends not only on the previous output and the internal state of the LSTM, but also a weighted average of all the spatial features (attention mechanism). The attention weights are computed by a parametric function.


Both models provide us with a posterior distribution over sequence of words $\bc=(w_0,\ldots,w_T)$, factorized as
\begin{equation}
P(\bc|\I;\bfgreek{theta})=\prod_{t=1}^T P(w_t|\I, w_{0,\ldots,t-1};\bfgreek{theta})
\label{eq:disccap_p_w}
\end{equation}
where $w_0$ is the $\langle$bos$\rangle$ token. (This is the same as \eqref{eq:factorization}.)

\section{Learning to reward discriminability}\label{sec:disccap_training}
Given a caption generation model, we may want to train it to maximize the discriminability objective~\eqref{eq:disccap_con_loss}. A natural approach would be to use gradient descent. Unfortunately, the objective is non-differentiable since it involves sampling captions for input images in a batch. 

One way to tackle this is by the Gumbel-softmax reparametrization trick~\cite{jang2016categorical,maddison2016concrete} which has been used in image captioning and visual dialog~\cite{shetty2017speaking,lu2017best}. Another way is to use the continuous relaxation (as in \secref{sec:re_proxy}). Instead, in this paper, we follow the philosophy of~\cite{Ranzato2016, hendricks2016generating,yu2016joint,rennie2017self} and treat captioning as a reinforcement learning problem. Specifically we use the ``Self-Critical'' training~\cite{rennie2017self}, as outlined below (more details in \secref{sec:bg_rl}).
Empirically we find ``Self-Critical'' works better than the other two ways.




The objective is to learn parameters $\bfgreek{theta}$ of the policy (here defining a mapping from $\I$ to $\bc$, i.e., $P$) that would maximize the reward computed by function $R(\bc,\I)$. The algorithm computes an update to approximate the gradient of the expected reward:
\begin{equation}
\nabla_{\bfgreek{theta}} \bbE_{\widehat{\bc}\sim P(\bc|\I;\bfgreek{theta})}[R(\widehat{\bc},\I)] \approx (R(\widehat{\bc},\I) - R(\bc^\ast, \I)) \nabla_{\bfgreek{theta}} \log p(\widehat{\bc}|\I;\bfgreek{theta})
\label{eq:disccap_policy-gradient}
\end{equation}
where $\bc^\ast$ is greedy decoding output.

We could apply this to maximizing the reward defined simply as the discriminability objective $-L_{\textsc{CON}}(\widehat{\bc}, \I)$. However, similar to the observation in Chapter \ref{chap:ref_exp}, this does not yield human-friendly captions since discriminability objective will not directly hurt from influency. So we will combine the discriminability objective with other traditional objectives in defining the reward, as described below.

\subsection{Training with maximum likelihood}
The standard objective in training a sequence prediction model is maximum likelihood estimation:
\begin{align}
R_{\textsc{LL}}(\bc,\I)=\log P(\bc|\I;\bfgreek{theta})
\end{align}
The parameters $\bfgreek{theta}$ here include word embedding matrix and LSTM weights which are updated as part of training. The CNN weights for visual representation are held fixed after pre-training on a vision task such as ImageNet classification.

Combining the MLE objective with discriminability objective in the REINFORCE framework corresponds to defining the objective as
\begin{equation}
R(\bc,\I)=R_{\textsc{LL}}(\bc,\I)\,-\,\lambda L_{\textsc{CON}}(\widehat \bc,\I),
\label{eq:disccap_R-ll}
\end{equation}
yielding gradient:
\begin{equation}
\begin{split}
\nabla_{\bfgreek{theta}}&\bbE[R(\bc,\I)]\approx\nabla_{\bfgreek{theta}}R_{\textsc{LL}}(\bc,\I)\\
&-\,\lambda
\left[L_{\textsc{CON}}(\widehat{\bc},\I)-L_{\textsc{CON}}(\bc^\ast,\I)\right]
\nabla_{\bfgreek{theta}}\log P(\widehat{\bc}|\I;\bfgreek{theta})
\end{split}
\label{eq:disccap_mle-policy}
\end{equation}
The coefficient $\lambda$ determines the tradeoff between matching human captions (expressed by the cross entropy) and discriminative properties expressed by $L_{\textup{CON}}$. 

\subsection{Training with CIDEr optimization}
In our experiments, it was hard to train with the combined objective in~\eqref{eq:disccap_mle-policy}. For small $\lambda$, the solutions seemed stuck in a local minimum; but increasing $\lambda$ would abruptly make output less fluent.

An alternative to MLE is to train the model to maximize some other reward/score, such as CIDEr following~\cite{rennie2017self}.
Optimizing over CIDEr can also benefit other metrics~\cite{rennie2017self}.
We found that in practice, the discriminability objective appears to ``cooperate'' better with CIDEr than with log-likelihood; we also observed better performance, across many metrics, on validation set as described in Section~\ref{sec:disccap_exp}. 


Compared to \cite{rennie2017self}, which uses CIDEr as the reward function, the difference here is we use a weighted sum of CIDEr score and discriminability objective.
\begin{equation}
\nabla_\theta \bbE[R(\widehat{\bc},\I)] \approx \left (R(\widehat{\bc},\I) - R(\bc^\ast, \I)\right ) \nabla_\theta \log P(\widehat \bc|\I;\bfgreek{theta}),
\end{equation}
where the reward is the combination
\begin{equation}
R(\widehat{\bc},\I) = \mbox{CIDEr}(\widehat{\bc}) - \lambda L_{\textsc{CON}}(\widehat{\bc}, \I),
\end{equation}
with $\lambda$ again representing the relative weight of discriminability objective vs. CIDEr.





\section{Experiments and results}\label{sec:disccap_exp}
The main goal of our experiments is to evaluate the utility of the proposed discriminability objective in training image captions. Recall that our motivation for introducing this objective is two-fold: to make the captions more discriminative, and to improve caption quality in general (with the implied assumption that expected discriminability is part of the unobservable human ``objective'' in describing images).

\noindent\textbf{Dataset.} We train and evaluate our model on COCO-Captions dataset \cite{chen2015microsoft}. To enable direct comparisons on discriminative captioning, we use the data split from \cite{vedantam2017context} instead of the commonly used Karpathy split, which includes 113,287 images for training, 5,000 images for validation, and another 5,000 held out for test.

\subsection{Implementation details}\label{sec:disccap_details}

As the basis for caption generators, we used two models described in Section~\ref{sec:disccap_models}, FC and Att2in, with relevant implementation details as follows.

For image encoder in retrieval and FC captioning model, we used a pretrained Resnet-101~\cite{renNIPS15fasterrcnn}. For each image, we take the global average pooling of the final convolutional layer output, which results in a vector of dimension 2048.

For Att2in model, we use the bottom-up feature from~\cite{anderson2018bottom}. Specifically, the spatial features are extracted from output of a Faster R-CNN~\cite{renNIPS15fasterrcnn,anderson2018bottom} with ResNet-101~\cite{he2016deep}, trained by object and attribute annotations from Visual Genome~\cite{krishna2017visual}. The number of spatial features varies from image to image. Each feature encodes a region in the image which is proposed by region proposal network. Both the FC features and Spatial features are pre-extracted, and no finetuning is applied on image encoders.
For captioning models, the dimension of LSTM hidden state, image feature embedding, and word embedding are all set to 512. 

The retrieval model uses GRU-RNN to encode text, and the FC features above as the image feature. The word embedding has 300 dimensions and the GRU hidden state size and joint embedding size are 1024. The margin $\alpha$ is set to 0.2, as suggested by~\cite{faghri2017vse++}.


\noindent\textbf{Training.}
All of our captioning models are trained according to the following scheme. We first pre-train the captioning model using MLE, with Adam~\cite{kingma2014adam}. After 40 epochs, the model is switched to self-critical training with appropriate reward (CIDEr alone or CIDEr combined with discriminability)  and continues for another 20 epochs. For fair comparison, we also train another 20 epochs for MLE-only models.

For both retrieval and captioning models, the batch size is set to 128 images. The learning rate is initialized to be 5e-4 and decayed by a factor of 0.8 for every three epochs.\footnotetext[1]{We find that this hyperparameter setting may not be optimal. The absolute performance would change with different setups.}

During test time, we apply beam search to sample captions from captioning model. The beam size is set to 2.

\subsection{Experiment design}
We consider a variety of possible combination of captioning objective (MLE/CIDEr), captioning model (FC/Att2in), and inclusion/exclusion of discriminability, abbreviating the model references for brevity, so, e.g., Att2in+CIDER+DISC(5) corresponds to fine-tuning the attention model with a combination of CIDEr and discriminability objective, with $\lambda=5$.

\noindent\textbf{Evaluation metrics.} Our experiments consider two families of metrics. The first family of standard metrics that have been proposed for caption evaluation, mostly based on comparing generated captions to human captions, includes BLEU, METEOR, ROUGE, CIDEr and SPICE.

The second set of metrics directly assesses how discriminative the captions are. This includes automatic assessment, by measuring accuracy of the trained retrieval model on generated captions.

We also assess how discriminative the generated captions are when presented to humans. To measure this, we conducted an image discrimination task on Amazon Mechanical Turk (AMT), following the protocol in~\cite{vedantam2017context}.
A single task (HIT) involves displaying, along with a caption, a pair of images (in randomized order). One image is the \emph{target} for which the caption was generated, and the second is a \emph{distractor} image. The worker is asked to select which image is more likely to match the caption. Each target/distractor pair is presented to five distinct workers; we report the fraction of HITs with correct selection by at least $k$ out of five workers, with $k=3,4,5$. Note that $k=3$ suffers from highest variance since the forced choice nature of the task would produce a non-trivial chance of 3/5 correct selections when the caption is random. In our opinion, $k=4$ is the most reliable indicator of human ability to discriminate based on the caption.

The test set used for this evaluation is the set from~\cite{vedantam2017context}, constructed as follows. For each image in the original test set, its nearest neighbor is found based on visual similarity, estimated as Euclidean distance between the {\tt fc7} feature vectors computed by VGG-16 pre-trained on ImageNet~\cite{simonyan2014very}. Then a captioning model is run on the nearest neighbor images, and the word-level overlap (intersection over union) of the generated captions is used to and pick (out of 5000 pairs) the top (highest overlap) 1000 pairs.

For preliminary evaluation, we followed a similar protocol to construct our own validation set of target/distractor pairs; both the target images and distractor images were taken from the caption validation set (and so were never seen by any training procedure, nor included in the test set).

\subsection{Retrieval model quality}

Before proceeding with the main experiments, we report in Table~\ref{tab:disccap_retrieval-result} the accuracy of the retrieval model on validation set, with human-generated captions. This is relevant since we rely on this model as a proxy for discriminability in our training procedure. While this model does not achieve state of the art for image caption retrieval, it is good enough for providing training signals to improve caption results.

\begin{table}[!tbh]
  \centering
    \begin{tabular}{l|rrrrr}
      \toprule
          & \multicolumn{1}{l}{R@ 1} & \multicolumn{1}{l}{R@ 5} & \multicolumn{1}{l}{R@ 10} & \multicolumn{1}{l}{Med r} & \multicolumn{1}{l}{Mean r} \\
    \midrule
          & \multicolumn{5}{c}{Caption Retrieval} \\
    \midrule
    1k val & 63.9  & 90.4  & 95.9  & 1.0   & 2.9 \\
    5k val & 38.0  & 68.9  & 81.1  & 2.0   & 10.4 \\
    \midrule
          & \multicolumn{5}{c}{Image Retrieval} \\
    \midrule
    1k val & 47.9  & 80.7  & 89.9  & 2.0   & 7.7 \\
    5k val & 26.1  & 54.7  & 67.5  & 4.0   & 34.6 \\
    \bottomrule
    \end{tabular}
  \caption{Retrieval model performance on validation set.}\label{tab:disccap_retrieval-result}
\end{table}

\subsection{Captioning performance}

\begin{table}[!th]
  \centering
    \begin{tabular}{lrrrrrr}
      \toprule
          & BLEU-4 & METEOR & ROUGE & CIDEr & SPICE\\
    \midrule
    FC+MLE~\cite{rennie2017self} & \textbf{33.08} & 25.66 & 54.07 & 100.05 & 18.55\\
    FC+CIDER~\cite{rennie2017self} & 32.49 & 25.50 & 54.28 & 101.54 & 18.99\\
    FC+MSE+DISC(100) &	29.02 & 25.23 & 52.61 & 91.90 & 18.81\\
    FC+CIDER+DISC(1) & 32.74 & \textbf{25.74} & \textbf{54.57} & \textbf{102.31} & \textbf{19.39}\\
    FC+CIDER+DISC(5) & 30.72 & 25.34 & 53.82 & 96.78 & 19.04\\
    FC+CIDER+DISC(10) & 27.27 & 24.73 & 52.24 & 87.95 & 18.07\\
    \midrule
    Att2in+MLE~\cite{rennie2017self} & 35.82 & 27.19 & 56.49 & 110.78 & 20.19\\
    Att2in+CIDER~\cite{rennie2017self} & 35.92 & 26.95 & 56.78 & 113.32 & 20.83\\
    Att2in+MLE+DISC(100) & 32.66 & 26.97 & 55.42 & 104.48 & 20.57\\
    Att2in+CIDER+DISC(1) & \textbf{36.27} & \textbf{27.28} & \textbf{57.06} & \textbf{114.06} & \textbf{21.13}\\
    Att2in+CIDER+DISC(5) & 35.04 & 27.04 & 56.36 & 110.26 & 20.97\\
    Att2in+CIDER+DISC(10) & 32.61 & 26.73 & 55.49 & 105.52 & 20.70\\
    \bottomrule
    \end{tabular}%
  \caption{Standard captioning metrics on the validation set. The numbers in the parenthesis are discriminability objective weight $\lambda$. All scores are scaled by 100x. The top half is the performance with FC model and the bottom Att2in model. The bold indicates the best score within the same generation model.}\label{tab:disccap_val_result_lang}
\end{table}%

\begin{table}[!th]
  \centering
    \begin{tabular}{lrrrrrr}
      \toprule
          & Acc  & Acc'  & 3 in 5 & 4 in 5 & 5 in 5 \\
    \midrule
    FC+MLE~\cite{rennie2017self} & 77.23 & 77.23 & 71.78 & 50.28 & 18.79 \\
    FC+CIDER~\cite{rennie2017self} & 74.00 & 74.32 & 73.04 & 50.58 & 24.83 \\
    FC+MSE+DISC(100) & 87.42	& 87.42 & 76.91 & 54.62 & 23.20 \\
    FC+CIDER+DISC(1) & 79.26 & 79.49 & 74.26 & 55.53 & 24.13 \\
    FC+CIDER+DISC(5) & 85.90 & 85.68 & 78.63 & 58.03 & 32.64 \\
    FC+CIDER+DISC(10) & \textbf{88.69} & \textbf{88.01} & \textbf{80.01} & \textbf{62.71} & \textbf{37.15} \\
    \midrule
    Att2in+MLE~\cite{rennie2017self} & 72.40 & 73.12 & 69.90 & 54.60 & 28.07 \\
    Att2in+CIDER~\cite{rennie2017self} & 71.05 & 71.13 & 69.97 & 51.34 & 27.34 \\
    Att2in+MLE+DISC(100) & 82.64 & 83.03 & 78.18 & 55.63 & 21.71 \\
    Att2in+CIDER+DISC(1) & 75.74 & 76.60 & 72.70 & 53.23 & 34.33 \\
    Att2in+CIDER+DISC(5) & 80.98 & 81.43 & 76.69 & 60.94 & 33.49 \\
    Att2in+CIDER+DISC(10) & \textbf{83.69} & \textbf{83.50} & \textbf{81.93} & \textbf{65.12} & \textbf{35.41} \\
    \bottomrule
    \end{tabular}%
  \caption{Discriminability metrics (machine and human) on the validation set. All scores are scaled by 100x.}\label{tab:disccap_val_result_disc}
\end{table}%

In Table \ref{tab:disccap_val_result_lang} and \ref{tab:disccap_val_result_disc}, we show the results on validation set with a variety of model/loss settings. Note that all the FC*/Att2in* model with different settings are finetuned from the same model pre-trained with MLE. 
The results in the tables for the machine scores(Acc, Acc', BLEU, METEOR, ROUGE, CIDEr and SPICE) are based on all the 5k images. For ``k in 5'' metrics, we randomly select a subset of 300 image pairs from validation set.
We can draw a number of conclusions from these results.

\noindent\textbf{Effectiveness of reinforcement learning.} In the first column of Table \ref{tab:disccap_val_result_disc}, we report the retrieval accuracy (Acc, \% of pairs in which the model correctly selects the target vs. distractor) on pairs given the output of the captioning model. Training with the discriminability objective produces higher values here, meaning that our captions are more discriminative \emph{to the retrieval model}\footnote{It is the same model we use to calculate the discriminability objective during training.}, as intended. As a control experiment, we also report the accuracy (Acc') obtained by a same architecture but separately trained retrieval model, not used in training caption generators. Acc and Acc' are very similar for all models, showing that our model does not overfit to the retrieval model it uses during training time.

\noindent\textbf{Human discrimination.} More importantly, we observe that incorporating discriminability in training yields captions that are more discriminative to humans, with higher $\lambda$ leading to better human accuracy.

\noindent\textbf{Improved caption quality.} We also see that, as hoped, incorporating discriminability indeed improves caption quality as measured by a range of metrics that are not tied to discriminability, such as BLEU, etc. Even the CIDEr scores are improved when adding discriminability to the CIDEr optimization objective with moderate $\lambda$. This is somewhat surprising since the addition of $L_{\textsc{CON}}$ could be expected to detract from the original objective of maximizing CIDEr; we presume that the improvement is due to the additional objective ``nudging'' the RL process and helping it escape less optimal solutions.

\noindent\textbf{Model/loss selection.} While discriminability objective works for both Att2in model and FC model, and with both MLE and CIDEr learning, to make captions more discriminative, and with mild $\lambda$ to improve other metrics, the overall performance analysis favors Att2in+CIDEr combination. We also note that Att2in is better than FC on discriminability metrics even when trained without $L_{\textup{CON}}$, but the gains are less significant than in automatic metrics.

\noindent\textbf{Effect of $\lambda$.} As stated above, mild $\lambda=1$, combined with Att2in+CIDER, appear to yield the optimal tradeoff, improving measures of discriminative \emph{and} descriptive quality across the board. Higher values of $\lambda$ do make resulting captions more discriminative to both humans and machines, but at the cost of reduction in other metrics, and in our observations (see Section~\ref{sec:disccap_qual}) in perceived fluency. This analysis is applicable across model/loss combinations.
We also notice a relatively large range of $\lambda$(0.5-1.2) can yield similar improvements on automatic metrics.


\begin{table*}[ht]
  \centering
  \setlength{\tabcolsep}{3pt}
    \begin{tabular}{lrrrrrrrrrr}
      \toprule
          & Acc  & Acc' & B4 & M & R & C & S & 3 in 5 & 4 in 5 & 5 in 5 \\
    \midrule
    Human & 74.30 & 74.14 & -     & -     & -     & -     & -     & 91.14 & 82.38 & 57.08 \\
    \midrule
    Att2in+MLE~\cite{rennie2017self} & 68.60 & 66.90 & 39.07 & 29.13 & 59.56 & 121.98 & 21.32 & 72.06 & 59.06 & 44.25 \\
    Att2in+CIDER~\cite{rennie2017self} & 68.19 & 65.12 & 38.71 & 29.08 & 59.71 & 126.04 & 22.60 & 70.07 & 55.95 & 35.95 \\
    \midrule
    CACA~\cite{vedantam2017context} & 75.80 & 76.00 & 23.57 & 21.86 & 47.19 & 76.56 & 15.26 & 74.1\footnotemark[1] & 56.88\footnotemark[1] & 35.19\footnotemark[1] \\
    Att2in+C+D(1) & 72.63 & 70.68 & \textbf{39.71} & \textbf{29.31} & \textbf{60.43} & \textbf{127.70} & \textbf{23.02} & 76.91 & 61.67 & 40.09 \\
    Att2in+C+D(10) & \textbf{79.75} & \textbf{79.14} & 35.38 & 28.21 & 58.11 & 114.29 & 22.04 & \textbf{77.70} & \textbf{64.63} & \textbf{44.63} \\
    \bottomrule
    \end{tabular}%
  \caption{Experiment results on 1k test set. Acc, Acc' and k in 5 are in percentage. +C+D=+CIDER+DISC. B4=BLEU-4, M=METEOR, R=ROUGE, C=CIDEr, S=SPICE.}\label{tab:disccap_disc_result_on_test}%
\end{table*}%
\footnotetext[1]{3 in 5 is quoted from~\cite{vedantam2017context}; 4 in 5 and 5 in 5 computed by us on the set of captions provided by the authors.}

Following the observations above, we select a subset of methods to evaluate on the (previously untouched) test set, with results shown in Table~\ref{tab:disccap_disc_result_on_test}.

Here, we add two more results for comparison. The first involves presenting AMT workers with \emph{human} captions for the target images. Recall that these captions are collected for each image independently, without explicit instructions related to discriminability, and without showing potential distractors. However, human captions prove to be highly discriminative. This, not surprisingly, indicates that humans may be incorporating an implicit objective of describing elements in an image that are surprising, notable or otherwise may help in distinguishing the scene from other scenes. While this performance is not perfect (4/5 accuracy of 82\%) it is much higher than for any automatic caption model.

The second additional set of results is for the model in~\cite{vedantam2017context}, evaluated on captions provided by the authors. Note that in contrast to our model (and to human captions), this method has the benefit of seeing the distractor prior to generating the caption; nonetheless, its performance is dominated across metrics by our attention models trained with CIDEr optimization combined with discriminability objective. It also appears that this models'  gains on discrimination are offset by a significant deterioration under other metrics.

Unlike~\cite{vedantam2017context}, our Att2in+CIDER+DISC(10) model achieves the most discriminative image captioning result without major degradation under other metrics; the Att2in+CIDER+DISC(1) again shows the best discriminability/descriptiveness tradeoff among the evaluated models.

{
\begin{table}[ht]
  \centering
    \begin{tabular}{lcccccc}
      \toprule
          & Color & Attribute & Cardinality & Object & Relation & Size \\
    \midrule
    FC+MLE &                9.32  &                8.74  &                1.73  &              34.04  &                4.81  & \textbf{              2.74 } \\
    FC+MLE+DISC(100) & \textbf{             15.85 } & \textbf{             10.31 } &               5.33  &              34.57  &                4.43  &                2.62  \\
    FC+CIDER  &                5.77  &                7.01  &                1.80  &              35.70  &                5.17  &                1.70  \\
    FC+CIDER+DISC(1) &                8.28  &                7.81  &                3.45  & \textbf{            36.37 } & \textbf{               5.25 } &               2.10  \\
    FC+CIDER+DISC(5) &              10.87  &                9.11  &                6.72  &              35.58  &                4.75  &                2.08  \\
    FC+CIDER+DISC(10) &              12.80  &                9.90  & \textbf{              8.50 } &             34.60  &                4.40  &                1.70  \\
    \midrule
    Att2in+MLE &              11.78  &              10.13  &                3.00  &              36.42  &                5.52  &                3.67  \\
    Att2in+MLE+DISC(100) & \textbf{            15.80 } & \textbf{             11.83 } &             14.30  &              37.16  &                5.13  & \textbf{              3.97 } \\
    Att2in+CIDER &               7.24  &                8.77  &                8.93  &              38.38  & \textbf{              6.21 } &               2.39  \\
    Att2in+CIDER+DISC(1) &                9.25  &                9.49  &              10.51  & \textbf{            38.96 } &               5.91  &                2.58  \\
    Att2in+CIDER+DISC(5) &              11.99  &              10.40  &              15.23  &              38.57  &                5.59  &                2.53  \\
    Att2in+CIDER+DISC(10) &              12.88  &              10.88  & \textbf{            15.72 } &             38.09  &                5.35  &                2.53  \\
    \bottomrule
    \end{tabular}%
  \caption{SPICE subclass scores on 5k validation set. All the scores here are scaled up by 100.}  \label{tab:disccap_model_spice}%
\end{table}%
}

\noindent\textbf{Effect on SPICE score}
To further break down how our discriminabilty objective does, we analyze the effect of different models on the SPICE score~\cite{vedantam2015cider}. It estimates caption quality by transforming both candidate and reference (human) captions into a scene graph
and computing the matching between the graphs.
SPICE is known to have higher correlation with human ratings than other conventional metrics like BLEU, CIDEr. Furthermore, it provides subclass scores on Color, Attributes, Cardinality, Object, Relation, Size. We report the results on SPICE subclass scores (on validation set) in detail for different models in Table~\ref{tab:disccap_model_spice}.

By adding the discriminability objective, we improve scores on Color, Attribute, and Cardinality. With the latter, qualitative results suggest that the improvement may be due to a refined ability to distinguish ``one'' or ``two'' from ``group of'' or ``many''. With small $\lambda$, we can also get the best score on Object. Since the object score is dominant in SPICE, $\lambda=1$ also obtains the highest SPICE score overall in Tables~\ref{tab:disccap_val_result_lang},~\ref{tab:disccap_disc_result_on_test}.

\begin{table}[htbp]
  \centering
    \begin{tabular}{lcc}
    \toprule
          & \# distinct captions & Avg. length \\
    \midrule
    FC+MLE~\cite{rennie2017self} & 2700  & 8.99 \\
    FC+CIDER~\cite{rennie2017self} & 2242  & 9.30 \\
    FC+CIDER+DISC(1) & 3204  & 9.32 \\
    FC+CIDER+DISC(5) & 4379  & 9.45 \\
    FC+CIDER+DISC(10) & 4634  & 9.78 \\
    \midrule
    Att2in+MLE~\cite{rennie2017self} & 2982  & 9.01 \\
    Att2in+CIDER~\cite{rennie2017self} & 2640  & 9.20 \\
    Att2in+CIDER+DISC(1) & 3235  & 9.28 \\
    Att2in+CIDER+DISC(5) & 4089  & 9.54 \\
    Att2in+CIDER+DISC(10) & 4471  & 9.84 \\
    \bottomrule
    \end{tabular}%
  \caption{Distinct caption number and average sentence length on validation set.}
  \label{tab:disccap_num_distinct}%
\end{table}

\noindent\textbf{Improved diversity.}
Finally, we can evaluate the diversity in captions generated by different models, shown in Table \ref{tab:disccap_num_distinct}. We find that including discriminability objective, and using higher $\lambda$, are correlated with captions that are more diverse (4471 distinct captions with Att2in+CIDER+DISC(10) for the 5000 images in validation set, compared to 2640 with Att2in+CIDER) and slightly longer (average length 9.84 with Att2in+CIDER+DISC(10) vs. 9.20 with Att2in+CIDER). We also observe pure CIDEr optimization will harm the diversity of output captions.

\begin{figure}[!th]
  \begin{minipage}[t]{.45\linewidth}
  \includegraphics[height=4cm]{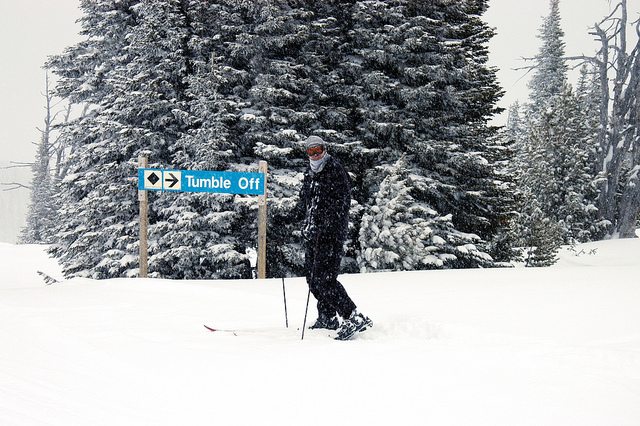}\\
  {\footnotesize
  {\bf Human}: a man riding skis next to a blue sign near a forest
  \\
  {\bf Att2in+CIDER}: a man standing on skis in the snow\\
  {\bf Ours}: a man standing in the snow with \textcolor{green!50!black}{a sign}
  }
  \end{minipage}%
  \hspace{1em}
  \begin{minipage}[t]{.45\linewidth}
  \includegraphics[height=4cm]{figures/disccap/COCO_val2014_000000464144.jpg}\\
  {\footnotesize
  {\bf Human}: the man is skiing down the hill with his goggles up\\
  {\bf Att2in+CIDER}: a man standing on skis in the snow\\
  {\bf Ours}: a man riding skis on a snow covered slope
  }
  \end{minipage}\\
  \begin{minipage}[t]{.45\linewidth}
  \includegraphics[height=4cm]{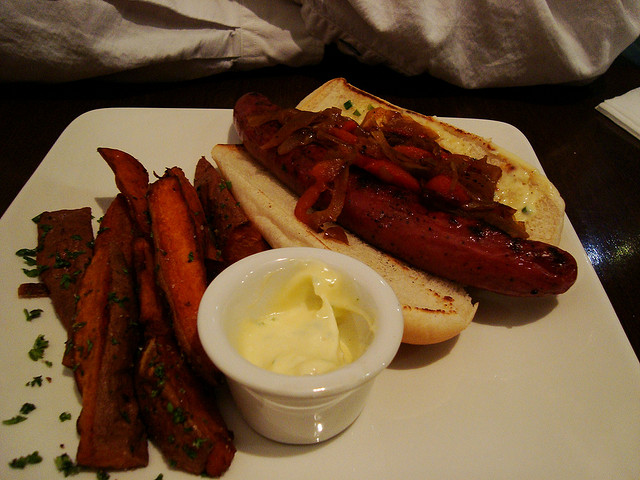}\\
  {\footnotesize
  {\bf Human}: a hot dog serves with fries and dip on the side
  \\
  {\bf Att2in+CIDER}: a plate of food with meat and vegetables on a table\\
  {\bf Ours}: \textcolor{green!50!black}{a hot dog and french fries} on a plate
  }
  \end{minipage}%
  \hspace{1em}
  \begin{minipage}[t]{.45\linewidth}
  \includegraphics[height=4cm]{figures/disccap/COCO_train2014_000000337365.jpg}\\
  {\footnotesize
  {\bf Human}: a plate topped with meat and vegetables and sauce\\
  {\bf Att2in+CIDER}: a plate of food with meat and vegetables on a table\\
  {\bf Ours}: a plate of food with \textcolor{green!50!black}{carrots and vegetables} on a plate
  }
  \end{minipage}
  \\
  \begin{minipage}[t]{.45\linewidth}
  \includegraphics[height=4cm]{figures/disccap/COCO_train2014_000000551470.jpg}\\
  {\footnotesize
  {\bf Human}: a train on an overpass with people under it
  \\
  {\bf Att2in+CIDER}: a train is on the tracks at a train station\\
  {\bf Ours}: a \textcolor{green!50!black}{red} train parked on the side of a building
  }
  \end{minipage}%
  \hspace{1em}
  \begin{minipage}[t]{.45\linewidth}
  \includegraphics[height=4cm]{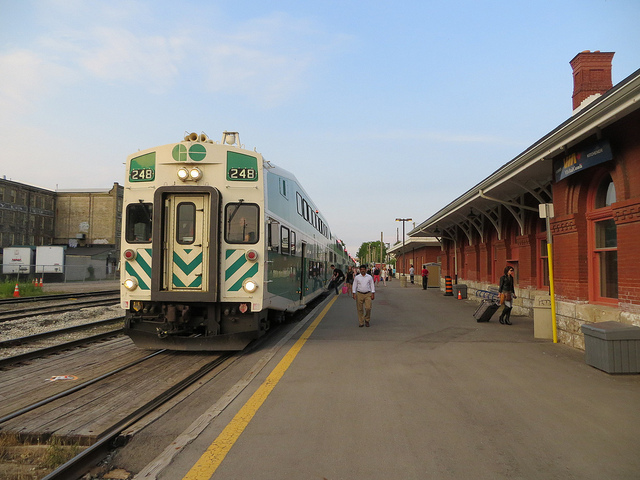}\\
  {\footnotesize
  {\bf Human}: a train coming into the train station\\
  {\bf Att2in+CIDER}: a train is on the tracks at a train station\\
  {\bf Ours}: a \textcolor{green!50!black}{green} train traveling down a train station
  }
  \end{minipage}
  \caption{Examples of image captions; Ours refers to Att2in+CIDER+DISC(1).}\label{fig:disccap_qual}
  \end{figure}
  
  \begin{figure}
  
  \begin{minipage}[t]{.30\linewidth}
  \includegraphics[height=4cm]{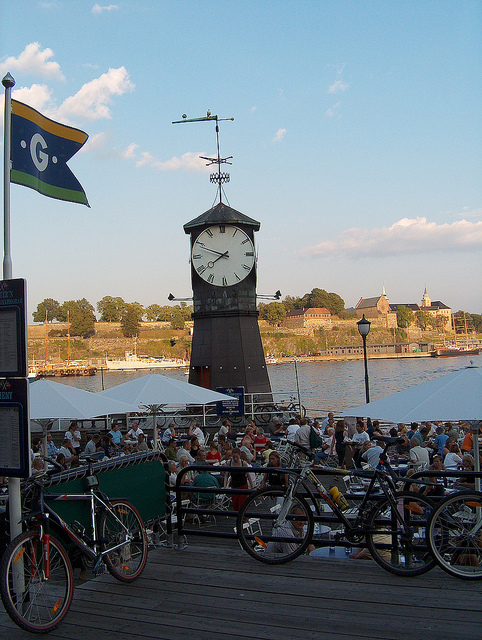}
  \end{minipage}%
  \hspace{1em}
  \begin{minipage}[t]{.60\linewidth}
  \includegraphics[height=4cm]{figures/disccap/COCO_val2014_000000527248.jpg}
  \end{minipage}\\
  {\footnotesize
  {\bf Att2in+MLE}: a large clock tower with a clock on it
  \\
  {\bf Att2in+CIDER}: a clock tower with a clock on the side of it\\
  {\bf Att2in+CIDER+DISC(1)}: a clock tower with \textcolor{green!50!black}{bikes} on the side of \textcolor{green!50!black}{a river}\\
  {\bf Att2in+CIDER+DISC(10)}: a clock tower with \textcolor{green!50!black}{bicycles} on the \textcolor{green!50!black}{boardwalk} near \textcolor{green!50!black}{a harbor}
  }
  \begin{minipage}[t]{\linewidth}
  \begin{minipage}[t]{.35\linewidth}
  \includegraphics[height=4cm]{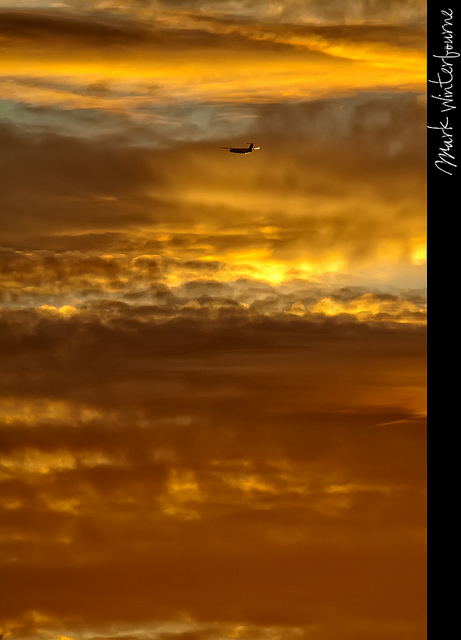}
  \end{minipage}%
  \hspace{1em}
  \begin{minipage}[t]{.55\linewidth}
  \includegraphics[height=4cm]{figures/disccap/COCO_val2014_000000345329.jpg}
  \end{minipage}
  \end{minipage}
  {\footnotesize
  {\bf Att2in+MLE}: a view of an airplane flying through the sky
  \\
  {\bf Att2in+CIDER}: a plane is flying in the sky\\
  {\bf Att2in+CIDER+DISC(1)}: a plane flying in the sky with \textcolor{green!50!black}{a sunset}\\
  {\bf Att2in+CIDER+DISC(10)}: \textcolor{green!50!black}{a sunset} of \textcolor{red}{a sunset with a sunset in the sunset}
  }
  %
  %
  \begin{minipage}[t]{\linewidth}
  \begin{minipage}[t]{.65\linewidth}
  \includegraphics[height=4cm]{figures/disccap/COCO_train2014_000000334400.jpg}
  \end{minipage}%
  \begin{minipage}[t]{.30\linewidth}
  \includegraphics[height=4cm]{figures/disccap/COCO_val2014_000000293782.jpg}
  \end{minipage}
  \end{minipage}
  {\footnotesize
  {\bf Att2in+MLE}: a couple of people standing next to a stop sign
  \\
  {\bf Att2in+CIDER}: a stop sign on the side of a street\\
  {\bf Att2in+CIDER+DISC(1)}: a stop sign in front of \textcolor{green!50!black}{a store with umbrellas}\\
  {\bf Att2in+CIDER+DISC(10)}: a stop sign sitting in front of \textcolor{green!50!black}{a store} with \textcolor{red}{shops}
  }
  %
  \caption{Captions from different models describing the target images(left). Right images are the corresponding distractors selected in val/test set; these pairs were included in AMT experiments.}
  \label{fig:disccap_amt}
  \end{figure}

\subsection{Qualitative result}\label{sec:disccap_qual}
In addition to \figref{fig:disccap_human_vs_machine_vs_baseline}, we further show, in \figref{fig:disccap_qual}, a sample of validation set images and for each include a human caption, the caption generated by Att2in+CIDER, and our captions produced by Att2in+CIDER+DISC(1). To emphasize the discriminability gap, the images are organized in pairs\footnote{Note that these pairs are formed for the purpose of this figure; these are not pairs shown to AMT workers for human evaluation.}, where both images have the same Att2in+CIDER caption; although these are mostly correct, compared to our and human captions, they tend to lack discriminative specificity. 


To illustrate the task on which we base our evaluation of discriminability to humans, we show in Figure~\ref{fig:disccap_amt} a sample of image pairs and associated captions. In each case, the target is on the left (in AMT experiments the order was randomized), and we show captions produced by four automatic systems, two without added discriminability objective in training, and two with (with low and high $\lambda$, respectively). Again, we can see that discriminability objective encourages learning to produce more discriminative captions, and that with higher $\lambda$ this may be associated with reduced fluency. We highlight in \textcolor{green!50!black}{green} caption elements that (subjectively) seem to aid discriminability, and in \textcolor{red}{red} the portions that seem incorrect or jarringly non-fluent. For additional qualitative results, see Appendix \ref{app:disccap_qual}.

\subsection{Additional results on Karpathy split and COCO test server}
Above we report results on the split from~\cite{vedantam2017context}, here we also report results (with automatic metrics) on the commonly used Karpathy split~\cite{Karpathy2015} and on COCO test server, in Table~\ref{tab:disccap_ka_scores},~\ref{tab:disccap_ka_spice},~\ref{tab:disccap_mscoco}. Our observations and conclusions are further confirmed by this evaluation.

\begin{table}[htbp]
  \centering
    \begin{tabular}{lccccc}
    \toprule
          & BLEU-4 & METEOR & ROUGE & CIDER & SPICE \\
    \midrule
          & \multicolumn{5}{c}{val} \\
    \midrule
    Att2in+CIDER & 35.11 & 27.00 & 56.76 & 112.25 & 20.43 \\
    Att2in+CIDER+DISC(1) & \textbf{35.83} & \textbf{27.33} & \textbf{57.20} & \textbf{113.81} & \textbf{20.94} \\
    Att2in+CIDER+DISC(5) & 34.16 & 26.98 & 56.33 & 109.02 & 20.54 \\
    Att2in+CIDER+DISC(10) & 31.96 & 26.64 & 55.20 & 104.02 & 20.24 \\
    \midrule
          & \multicolumn{5}{c}{test} \\
    \midrule
    Att2in+CIDER & 35.66 &27.10 & 56.88 & 113.45 & 20.58 \\
    Att2in+CIDER+DISC(1) & \textbf{36.14} & \textbf{27.38} & \textbf{57.29} & \textbf{114.25} & \textbf{21.05} \\
    Att2in+CIDER+DISC(5) & 34.39 & 26.96 & 56.40 & 110.18 & 20.82 \\
    Att2in+CIDER+DISC(10) & 32.03 & 26.71 & 55.31 & 105.30 & 20.50 \\
    \bottomrule
    \end{tabular}%
  \caption{Standard metrics on Karpathy split.}
  \label{tab:disccap_ka_scores}%
\end{table}

\begin{table}[htbp]
  \centering
    \begin{tabular}{lcccccc}
      \toprule
          & \multicolumn{1}{c}{Color} & \multicolumn{1}{c}{Attribute} & \multicolumn{1}{c}{Cardinality} & \multicolumn{1}{c}{Object} & \multicolumn{1}{c}{Relation} & \multicolumn{1}{c}{Size} \\
    \midrule
          & \multicolumn{6}{c}{val} \\
    \midrule
    Att2in+CIDER & 6.27 & 8.19 & 9.07 & 38.18 & 5.67 & \textbf{2.76} \\
    Att2in+CIDER+DISC(1) & 8.17 & 8.89 & 10.94 & \textbf{38.97} & \textbf{5.70} & 2.41 \\
    Att2in+CIDER+DISC(5) & 10.83 & 9.70 & 13.98 & 38.27 & 5.19 & 2.68 \\
    Att2in+CIDER+DISC(10) & \textbf{12.33} & \textbf{10.25} & \textbf{14.47} & 37.72 & 4.87 & 2.25 \\
    \midrule
          & \multicolumn{6}{c}{test} \\
    \midrule
    Att2in+CIDER & 5.54 & 7.82 & 7.83 & 38.43 & 6.04 & 2.21 \\
    Att2in+CIDER+DISC(1) & 7.55 & 8.72 & 9.75 & \textbf{39.08} & \textbf{6.05} & 2.24 \\
    Att2in+CIDER+DISC(5) & 10.46 & 9.89 & 13.05 & 38.46 & 5.66 & \textbf{2.59} \\
    Att2in+CIDER+DISC(10) & \textbf{12.21} & \textbf{10.42} & \textbf{14.84} & 38.04 & 5.19 & 2.54 \\
    \bottomrule
    \end{tabular}%
  \caption{SPICE subclass scores on Karpathy split. All the scores here are scaled up by 100.}
  \label{tab:disccap_ka_spice}%
\end{table}%

\begin{table}[htbp]
  \centering
  \begin{tabular}{lcccc}
    \toprule
    Metric & Ours(c5) & Baseline(c5) & Ours(c40) & Baseline(c40)\\
    \midrule
    BLEU-1 & 79.5 & 79.4 &94.1 &93.9 \\
    BLEU-2 &  62.7& 62.3 &86.9 &86.1\\
    BLEU-3 &47.4 & 47.0 &76.4 & 75.4\\
    BLEU-4 & 35.2& 34.9 & 64.7 & 63.7\\
    METEOR & 27.1& 26.8 & 35.5 & 35.1\\
    ROUGE & 56.7& 56.4 &71.0 & 70.5\\
    CIDEr & 110.0& 109.1 & 111.6 & 110.6\\
    \bottomrule
  \end{tabular}
  \caption{Results on COCO test, reported by the COCO
    server. Ours: Att2in+CIDER+DISC(1). Baseline: Att2in+CIDER.}\label{tab:disccap_mscoco}
\end{table}
\section{Conclusion}\label{sec:disccap_end}
We demonstrated that incorporating a discriminability objective, derived from the loss of a trained image/caption retrieval model, in training image caption generators improves the quality of resulting captions across a variety of properties and metrics. It does, as expected, lead to captions that are more discriminative, allowing both human recipients and machines to better identify an image being described, and thus arguably conveying more valuable information about the images. More surprisingly, it also yields captions that are scored higher on metrics not directly related to discriminability, such as BLEU/METEOR/ROUGE/CIDEr as well as SPICE, reflecting more descriptive captions. This suggests that richer, more diverse sources of training signal may further improve training of caption generators.




\chapter{Diversity analysis for image captioning models}\label{chap:diversity}






\section{Problem statement}


People can produce a diverse yet accurate set of captions for a given
image. Sources of diversity include syntactic variations and
paraphrasing, focus on different components of the visual scene, and
focus on different levels of abstraction, e.g., describing scene
composition vs. settings vs. more abstract, non-grounded
circumstances, observed or imagined. In this chapter, we consider the
goal of endowing automatic image caption generators with the ability to
produce diverse caption sets.


\textbf{Why is caption diversity important}, beyond the fact that ``people can
do it'' (evident in existing datasets with multiple captions per
image, such as COCO)? One caption may not be sufficient to describe
the whole image, and there is more than one way to describe an
image. Producing multiple distinct descriptions is required in tasks like image paragraph
generation and dense captioning.
In some applications (e.g., Powerpoint), human users may want to select from a diverse set of candidates.
Additionally, access to a diverse
\emph{and} accurate caption set may enable better production of a
single caption, e.g., when combined with a re-ranking method~\cite{collins2005discriminative, shen2003svm,luo2017comprehension, holtzman2018learning, park2019adversarial}.

\textbf{How is diversity achieved} in automated captioning?
Latent variable models like Conditional GAN~\cite{shetty2017speaking,dai2017towards}, Conditional VAE~\cite{wang2017diverse,jain2017creativity} and Mixture of experts~\cite{wang2016diverse,shen2019mixture}  generate a set of captions by sampling the latent variable.
However, they didn't carefully compare against different sampling methods. For example, the role of sampling temperature in controlling diversity/accuracy tradeoff was overlooked.

Some efforts have been made to explore different sampling methods to generate diverse outputs, like diverse beam search~\cite{vijayakumar2016diverse}, Top-K sampling~\cite{fan2018hierarchical, radford2019language}, Top-p/nucleus sampling~\cite{Ari2019nucleus} and etc.~\cite{batra2012diverse,gimpel2013systematic,li2016mutual,li2015diversity}. We will study these methods for captioning.


\cite{wang2019describing} has found that existing captioning models have a tradeoff between diversity and accuracy. By training a weighted combination of CIDEr reward and cross entropy loss, they can get models with a different diversity-accuracy tradeoff with different weight factors.

In this chapter, we want to take SOTA captioning models, known to achieve good accuracy, and study the diversity accuracy tradeoff with different sampling methods and hyperparameters while fixing the model parameters. Especially, we will show if CIDEr-optimized model (high accuracy low diversity) can get a better tradeoff than cross entropy trained model under different sampling settings.

\textbf{How is diversity measured?} Common metrics for captioning
tasks like BLEU, CIDEr, etc., are aimed at measuring similarity
between a pair of single captions and can't evaluate diversity. On
the other hand, some metrics have been proposed to capture diversity
but not accuracy. These include
mBLEU\rtcom{Unfortunately, can't find any reference. People seems to not cite it}, Distinct unigram(Div-1),
Distinct bigram(Div-2), Self-CIDEr~\cite{wang2019describing}, etc. We
propose a new metric, \spicen, an extension of SPICE able to measure diversity and accuracy
of a caption set w.r.t. ground truth captions at the same time.

In addition to the new metric, our primary contribution is the systematic evaluation of the role of different choices -- training objectives, hyperparameter values, sampling/decoding procedure -- play in the resulting tradeoff between accuracy and the diversity of generated caption sets. This allows us to identify some simple but effective approaches to producing diverse and accurate captions. 

\section{Related work}

Different model designs have been proposed for diverse captioning or text generation. 
\cite{jain2017creativity} presented a conditional VAE (CVAE) model for diverse visual question generation. \cite{wang2017diverse} extended \cite{jain2017creativity} and proposed AG-CVAE which replaces the original single gaussian prior used in CVAE with an additive Gaussian prior.

Generative Adversarial Networks\cite{Goodfellow2014} is also a popular option for modeling diverse captioning. \cite{shetty2017speaking, dai2017towards,li2018generating} train a conditional GAN where the generator tries to fool a discriminator that is trained to distinguish human captions from generated captions.

Mixture of experts~\cite{masoudnia2014mixture} has also been proposed for generating diverse text generation, where each expert would generate a different ``style'' of captions. \cite{wang2016diverse} proposed GroupTalk which trains 3 models sequentially in a ``boosting''~\cite{schapire2003boosting} way, resulting in each model having different specializations. \cite{shen2019mixture} studied different training strategies and model designs of mixture of experts model and showed its superior performance on diverse machine translation.

Much work also focuses on proposing different decoding/inference methods for diverse generation. 
\cite{batra2012diverse} proposed an algorithm for Diverse M-Best problem (find M most probable solutions for structured prediction tasks), by iteratively selecting a solution that has both high likelihood and large dissimilarity to previous solutions.
\cite{gimpel2013systematic} ported DivMBest to sequence generation task like machine translation, using beam search as a black-box inference algorithm.
\cite{vijayakumar2016diverse} proposed diverse beam search which modifies beam search and forces beams of different groups to have a large distance.
\cite{li2016mutual} proposed a beam search diversification
heuristic that discourages beams from sharing common
roots, implicitly resulting in diverse lists.
\cite{li2015diversity} introduced a novel decoding
objective that maximizes mutual information between inputs and predictions to penalize generic sequences.
To avoid the problem of repetition when generating long sequences with beam search, \cite{fan2018hierarchical, radford2019language,Ari2019nucleus} proposed Top-K/Top-p sampling to obtain the diversity of naive sampling while also being accurate. See \secref{sec:bg_sampling} for more details.



The most relevant work to ours is \cite{wang2019describing}. They not only proposed a new diversity metric Self-Cider, but also did a complete study on the diversity performance of multiple SOTA models. They found jointly training with CIDEr reward and cross entropy loss (MLE) can balance between diversity and accuracy. However, they mostly investigated naive sampling under temperature 1. In fact, different sampling temperatures can get us captions with different level of diversity from one single model. To gain a more comprehensive understanding of the models' behavior, we will study the impact of different hyperparameters for different decoding algorithms.


\noindent\textbf{Diversity definitions and metrics.} There are two common ``definitions'' of diversity. The first one is diversity at corpus level, looking at how different the generated captions are on the ``test'' set. The metrics include vocabulary size, \% Novel Sentences, word recall~\cite{van2018measuring} etc. The second one is image-level caption diversity. For a single image, the models are asked to generate multiple captions and are measured by how different the captions are. This category includes metrics like Oracle/Average score, mBLEU, Div-1/2, Self-CIDEr etc. In this work, we mostly focus on the second definition, but also reports metrics of the first one. Our proposed \spicen falls into the category of the second definition.
 \section{Methods for diverse captioning}
In most of our experiments we work with an attention-based LSTM model (Att2in from~\cite{rennie2017self})
with hidden layer size 512. We also conduct experiments with three other architectures: FC (non-attention LSTM model from \cite{rennie2017self}), Att2in-L (Att2in with hidden size 2048) and Trans (transformer model in \cite{vaswani2017attention, sharma2018conceptual})\footnote{We modify the original transformer to adapt to image captioning task by removing the word embedding layer and positional encoding in the transformer encoder. Instead, we directly feed the projected visual features as the encoder input.}.

\noindent Learning objectives we consider:
\begin{itemize}
\item Cross Entropy loss (\textbf{XE}): standard cross entropy (MLE) loss as described in Section \ref{sec:bg_mle}.
\item Reinforcement learning (\textbf{RL}): we first train the model with standard cross entropy loss, then finetuned on CIDEr reward with Self-Critical training~\cite{rennie2017self} (a brief introduction in Section \ref{sec:bg_rl}). Specifically we use the variant in \cite{mnih2016variational,luo2020better}.
\item \textbf{XE+RL}: we first train the model with standard cross entropy loss, then finetuned on the convex combination of cross entropy loss and CIDEr reward. This setup follows \cite{wang2019describing}.
\end{itemize}


Next, we consider different decoding methods for trained
generators. (Some of the following methods are covered in Section \ref{sec:bg_decoding}.)

\begin{itemize}
\item \textbf{SP}: Naive sampling, with temperature $\MT$,
      \[P(w_t|\I,w_{1\ldots,t-1})=\frac{\exp\left(s(w_t|\I,w_{1,\ldots,t-1})/\MT\right)}
        {\sum_w\exp\left(s(w|\I,w_{1,\ldots,t-1})/\MT\right)}
      \]
      where $s$ is the score (logit) of the model for the
      next word given image $\I$. We are
      not aware of any discussion of this approach to 
      diverse captioning in the literature.
      
\item \textbf{Top-K} sampling~\cite{fan2018hierarchical, radford2019language}: limiting sampling to top $K$ words.

\item \textbf{Top-p} (nucleus) sampling~\cite{Ari2019nucleus}: limiting
sampling to a set with high probability mass $p$.

\item \textbf{BS}: beam search, producing $m$-best list.

\item \textbf{DBS}: diverse beam search~\cite{vijayakumar2016diverse} with a diversity
parameter $\lambda$.

\end{itemize}

\noindent Each of these methods can be used with temperature $\MT$ affecting the word
posterior, as well as its ``native'' hyperparameters ($K$, $p$, $m$, $\lambda$).

\section{Measuring diversity and accuracy}
We will use the following metrics for diverse captioning:

\begin{itemize}
\item \textbf{mBLEU}: mean of BLEU-4 scores computed between each caption
in the set against the rest. Lower is more diverse.
    
\item \textbf{Self-CIDEr}~\cite{wang2019describing}: a diversity metric derived from latent 
    semantic analysis and 
    CIDEr similarity. Higher is more diverse.

\item \textbf{Oracle/Average} scores: taking the
maximum/mean of each relevant metric over 
all the candidates (of one image).

\item \textbf{Div-1, Div-2}: ratio of number of unique uni/bigrams 
    in generated captions to number of words in caption set.
    Higher is more diverse.

\item \textbf{Vocabulary Size}: number of unique words used in all
generated captions. Higher is more diverse.

\item \textbf{\% Novel Sentences}: percentage of generated captions
not seen in the training set. Higher is more diverse.
\end{itemize}

\begin{figure}[!th]
\centering
  \begin{subfigure}{0.8\linewidth}
    \includegraphics[width=\linewidth]{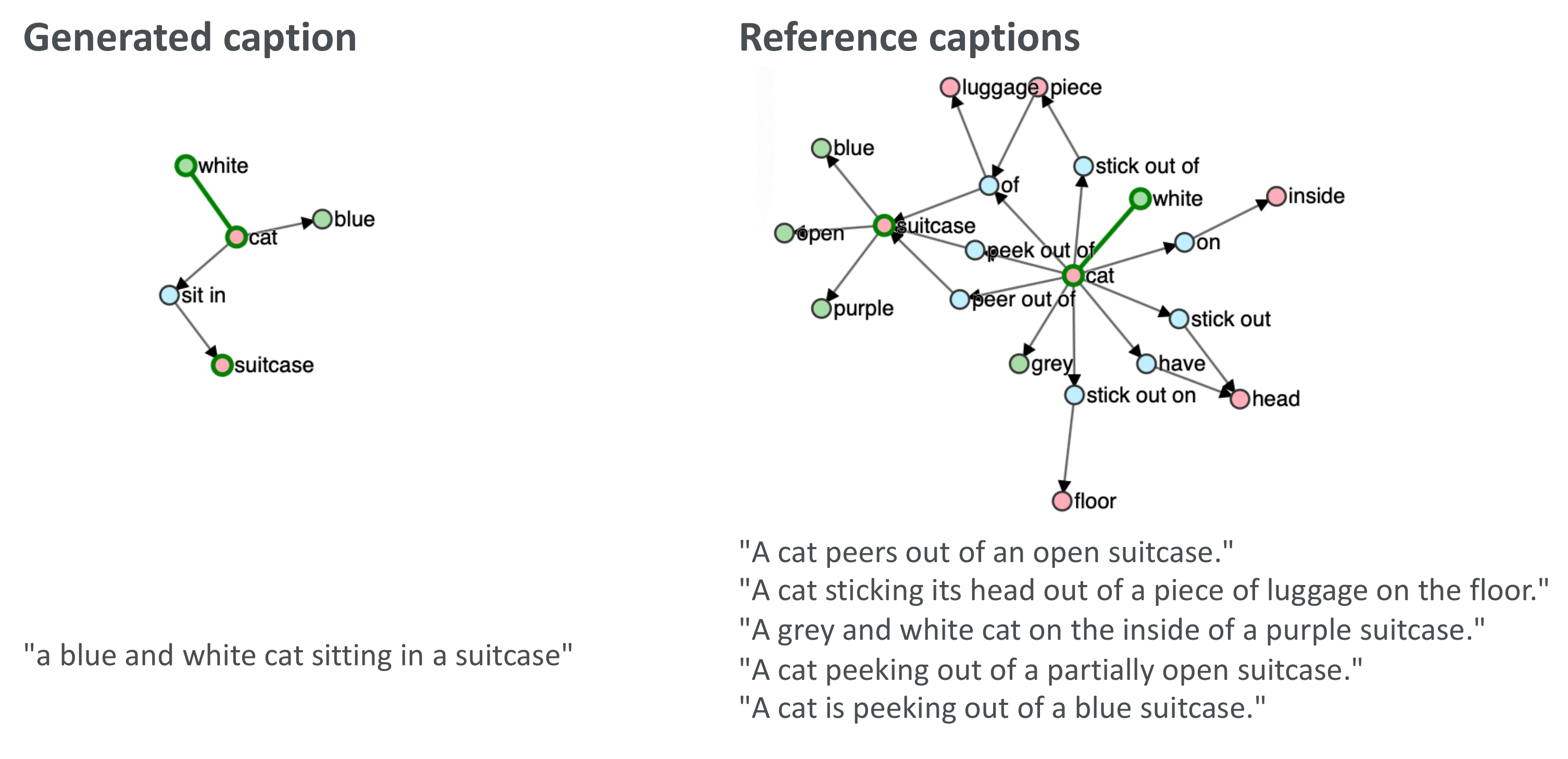}
    \caption{Graph construction of SPICE}\label{fig:div_spice}
  \end{subfigure}%
  \\
  \begin{subfigure}{0.8\linewidth}
    \includegraphics[width=\linewidth]{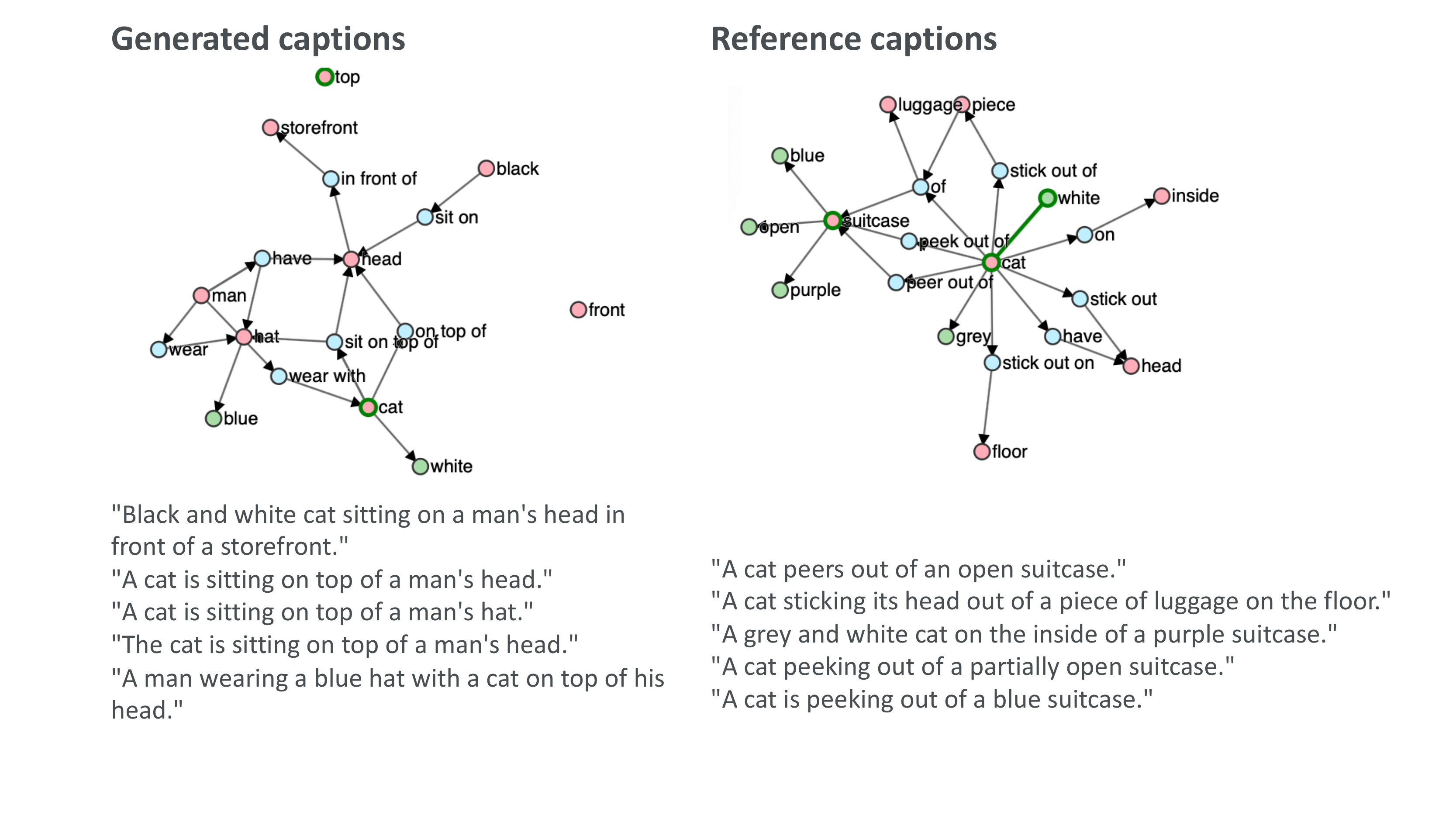}
    \caption{Graph construction of \spicen} \label{fig:div_allspice}
  \end{subfigure}%
\caption{The illustration of SPICE and \spicen. SPICE compares the scene graph of one generated caption and the scene graph of reference caption set. \spicen compares the scene graph of sampled caption set and the scene graph of reference caption set.} 
\label{fig:div_spice_allspice}
\end{figure}

We propose a new metric \spicen. It builds upon the SPICE~\cite{anderson2016spice} designed for
evaluating a single caption. To compute SPICE, a \emph{scene graph} is
constructed for the evaluated caption, with vertices corresponding to
objects, attributes, and relations; edges reflect possession of attributes or relations between objects. Another scene graph is constructed from the reference
caption; with multiple reference captions, the reference scene graph
is the union of the graphs for individual captions, combining
synonymous vertices. An example is shown in \figref{fig:div_spice}. The SPICE metric is the F-score
on matching vertices and edges between the two graphs.

\spicen builds a single graph for the
\emph{generated caption set}, in the same way that SPICE treats
reference caption sets, as shown in \figref{fig:div_allspice}. \spicen has the following properties.
\begin{itemize}
  \item Repetitions across captions
  in the set won't change the score, due to the merging of synonymous
  vertices.
  \item Adding a caption that captures part of the reference not
  captured by previous captions in the set may improve the
  score (by increasing recall).
  \item Wrong content in any caption in the set will harm the
  score (by reducing precision).
\end{itemize}
\textbf{Comparison to other metrics.}
Similar to \spicen, ``Oracle score'' also captures both accuracy and diversity. However, Oracle score only requires one caption in the set
to be good, and adding wrong captions does not affect the
metric, making it an imperfect measure of \emph{set} quality.
Metrics like mBLEU, Self-CIDEr separate diversity
considerations from those of accuracy, and can be easily cheated by random sentences.
In contrast, higher \spicen requires the samples to be both semantically diverse
\emph{and} correct; and diversity without accuracy is penalized.\footnote{AllSPICE is released at \href{https://github.com/ruotianluo/coco-caption}{https://github.com/ruotianluo/coco-caption}.}
%
%
%
%
We summarize the properties of all used metrics in Table \ref{tab:div_metrics}.

\begin{table}[ht]
    \center
    \begin{tabular}{lcccc}
                          & Diversity Type & X is more diverse & Captures accuracy \\
    \toprule
    mBLEU                 & Image-level          & Lower       & $\times$                 \\
    Self-CIDEr            & Image-level          & Higher      & $\times$                 \\
    Oracle/Average scores & Image-level          & Higher      & $\checkmark$                 \\
    AllSPICE              & Image-level          & Higher      & $\checkmark$                 \\
    Div-1,Div-2           & Image-level          & Higher      & $\times$                 \\
    Vocabulary Size       & Corpus-level         & Higher      & $\times$                 \\
    \% Novel Sentences    & Corpus-level         & Higher      & $\times$                 \\
    \bottomrule
    \end{tabular}
    \caption{The properties of different diversity metrics.}
    \label{tab:div_metrics}
\end{table}
\section{Experiments and results}

The aim of our experiments is to explore how different
methods (varying training procedures, decoding
procedures, and hyperparameter settings) navigate the
diversity/accuracy tradeoff.

\subsection{Implementation details}
\noindent\textbf{Training details.}
All of our captioning models are trained according to the following scheme. The XE-trained captioning models are trained for 30 epochs (transformer 15 epochs) with Adam~\cite{kingma2014adam}. RL-trained models are finetuned from XE-trained model, optimizing CIDEr score with the Self-Critical variant in \cite{luo2020better}. The batch size is chosen to be 250 for LSTM-based model, and 100 for transformer model. The learning rate is initialized to be 5e-4 and decayed by a factor 0.8 for every three epochs. During reinforcement learning phase, the learning rate is fixed to 4e-5.

\noindent\textbf{The visual features.}
For FC model, we used the input of the last fully connected layer of a pre-trained ResNet-101~\cite{he2016deep} as image feature.
For Att2in and transformer, the bottom-up features are used \cite{anderson2018bottom}.

\noindent\textbf{Dataset.} For most of the experiments, we train and test on COCO-Captions~\cite{chen2015microsoft}.
We also report results on a smaller captioning dataset Flickr30k~\cite{young-etal-2014-image}. We use Karpathy test split~\cite{karpathy2015deep} for both datasets.

\noindent\textbf{Generation setup.}
Unless stated otherwise, caption sets contain five generated
captions for each image. \footnote{The experiment related code is released at \href{https://github.com/ruotianluo/self-critical.pytorch/tree/master/projects/Diversity}{https://github.com/ruotianluo/self-critical.pytorch/tree/master/projects/Diversity}.} 





\subsection{Results and discussions}

\begin{figure}[t]
      \begin{minipage}[c]{.3\textwidth}
        \caption{Diversity-accuracy tradeoff. Blue: RL, decoding with different
        temperatures. Orange: XE+RL, trained with different
        CIDEr-MLE weights $\gamma$, decoding with $\MT=1$. Green: XE, decoding
        with different temperatures.}
      \label{fig:div_1}
    \end{minipage}%
    \hfill 
    \begin{minipage}[c]{.65\textwidth}
      \includegraphics[width=\linewidth]{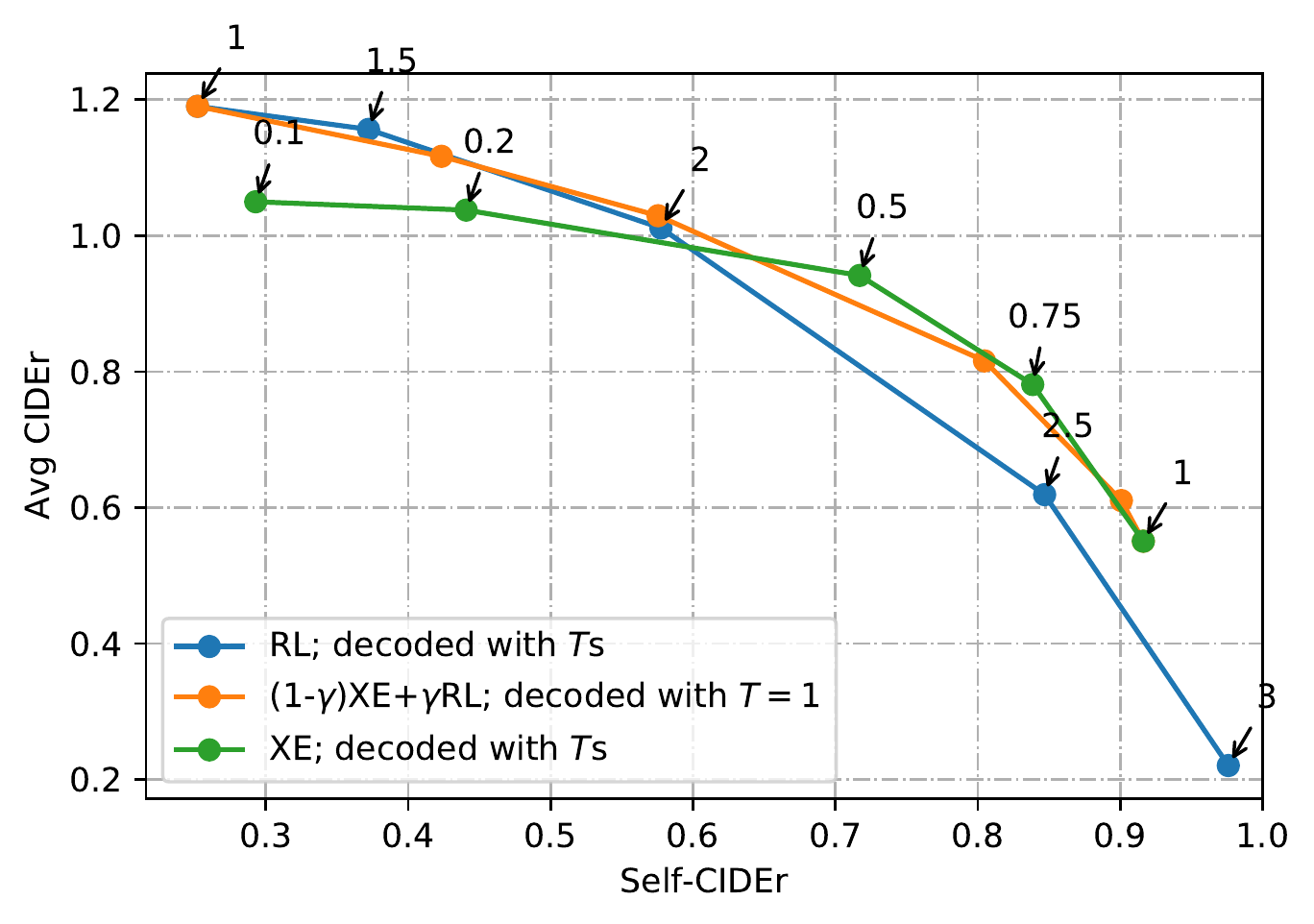}
    \end{minipage}%
  \end{figure}

\begin{figure}[th]
  \centering
\includegraphics[height=4.5cm]{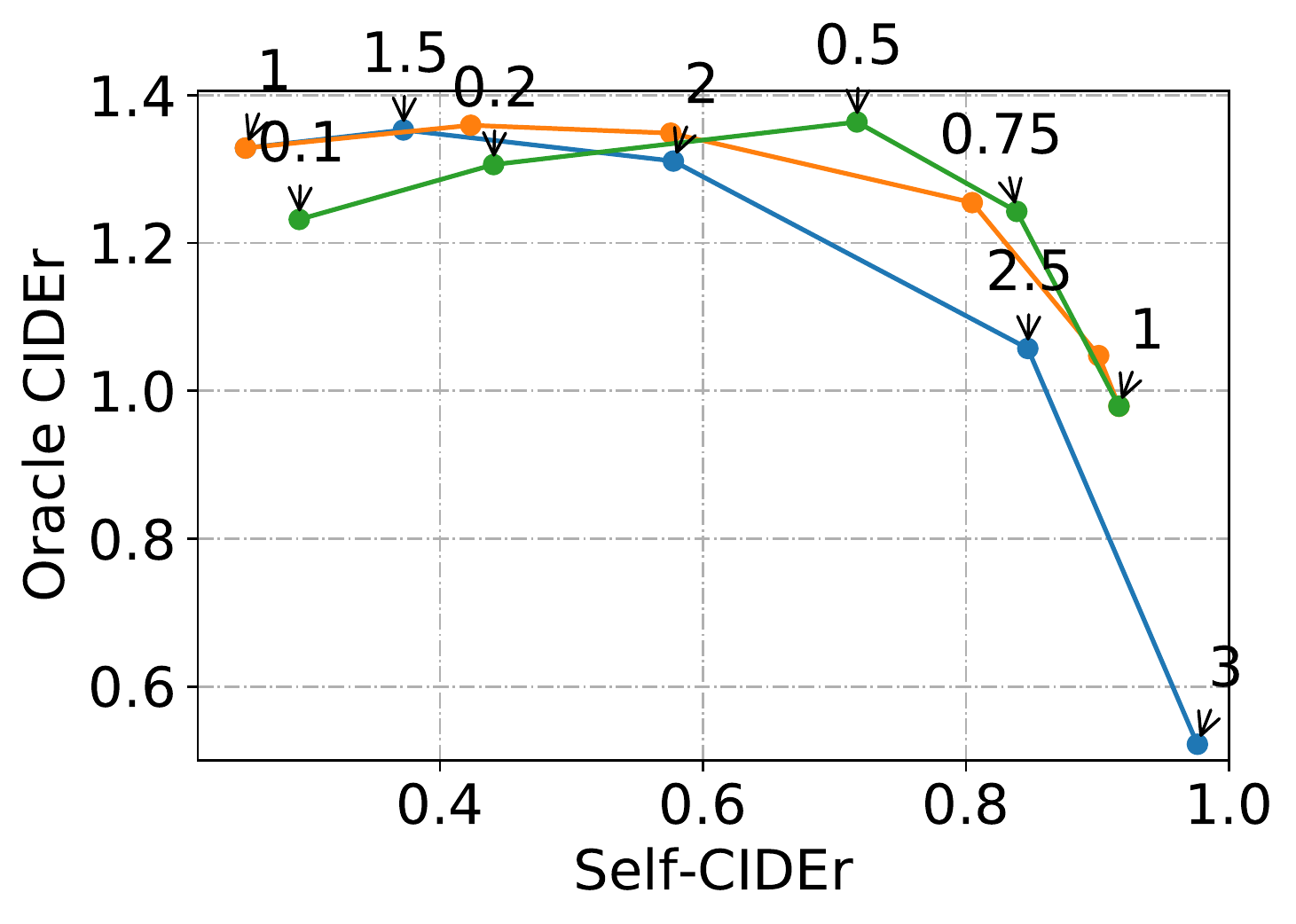}
\hspace{1em}
\includegraphics[height=4.5cm]{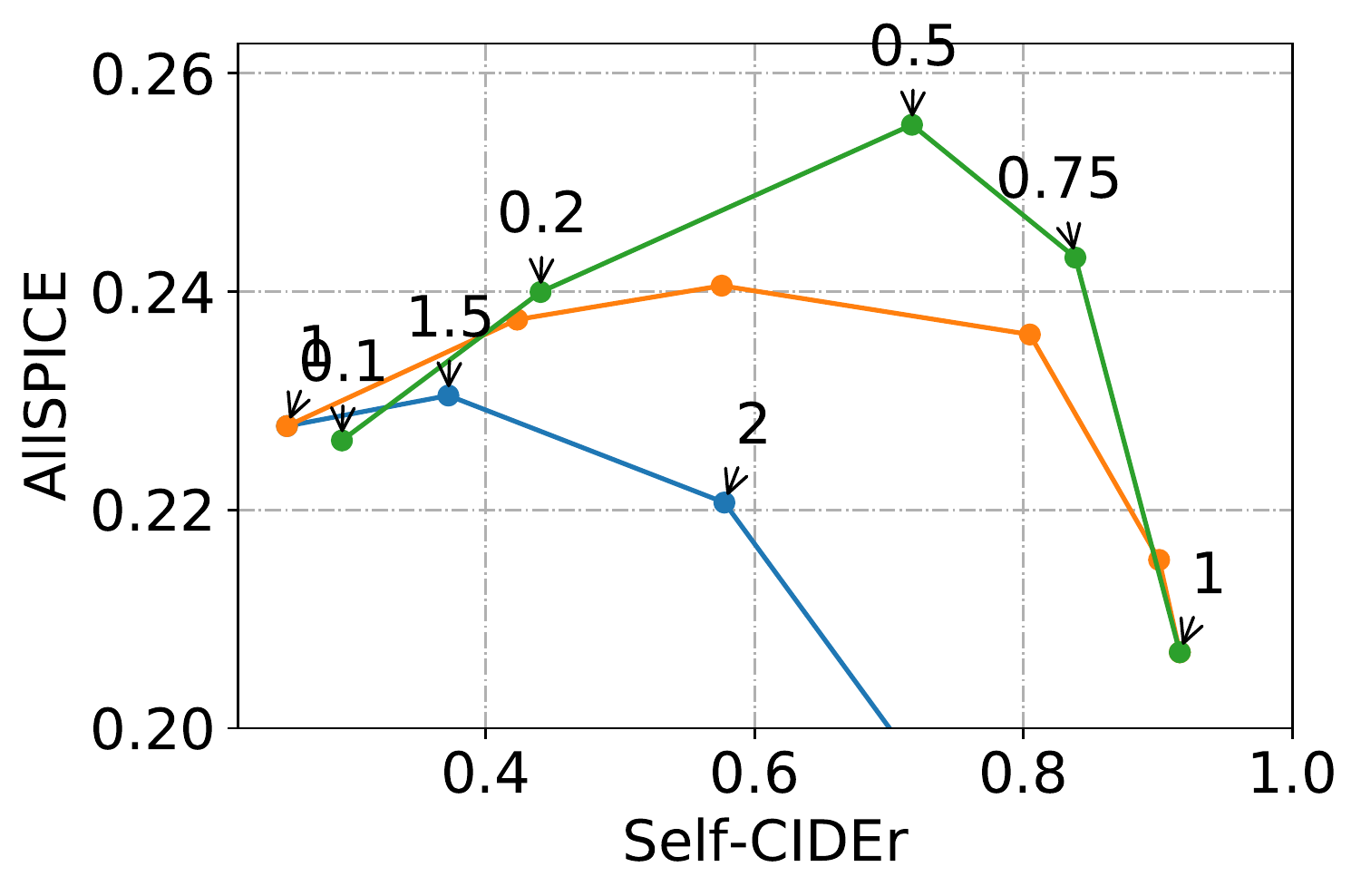}
\caption{Additional views of diversity/accuracy tradeoff: Oracle CIDEr and \spicen vs. Self-CIDEr. See legend for Figure~\ref{fig:div_1}.}
\label{fig:div_2}
\end{figure}
\noindent\textbf{Naive sampling.} \cite{wang2019describing} showed that
with temperature 1, XE-trained model will have high diversity but low
accuracy, and RL-trained model the opposite. Thus they proposed to use
a weighted combination of RL-objective and XE-loss to achieve better
tradeoff. However, Figure \ref{fig:div_1} shows that a simple alternative
for trading diversity for accuracy is to modulate the sampling
temperature. Although RL-trained model is not diverse when $\MT$=1, it
generates more diverse captions with higher
temperatures. XE-trained model can generate \emph{more accurate} captions
with lower temperatures.
In Figure \ref{fig:div_2}, we show that naive sampling with $\MT$=0.5 performs better than other settings on Oracle CIDEr and \spicen.

Comparing the two panels of Figure~\ref{fig:div_2} to each other and to Figure~\ref{fig:div_1}, we also see that measuring \spicen allows significant distinction between settings that would appear very similar under Oracle CIDEr. This suggests that \spicen is complementary to other metrics and reflects a different aspect of accuracy/diversity tradeoff.


\begin{figure}[tb!]
  \centering
\includegraphics[height=4.5cm]{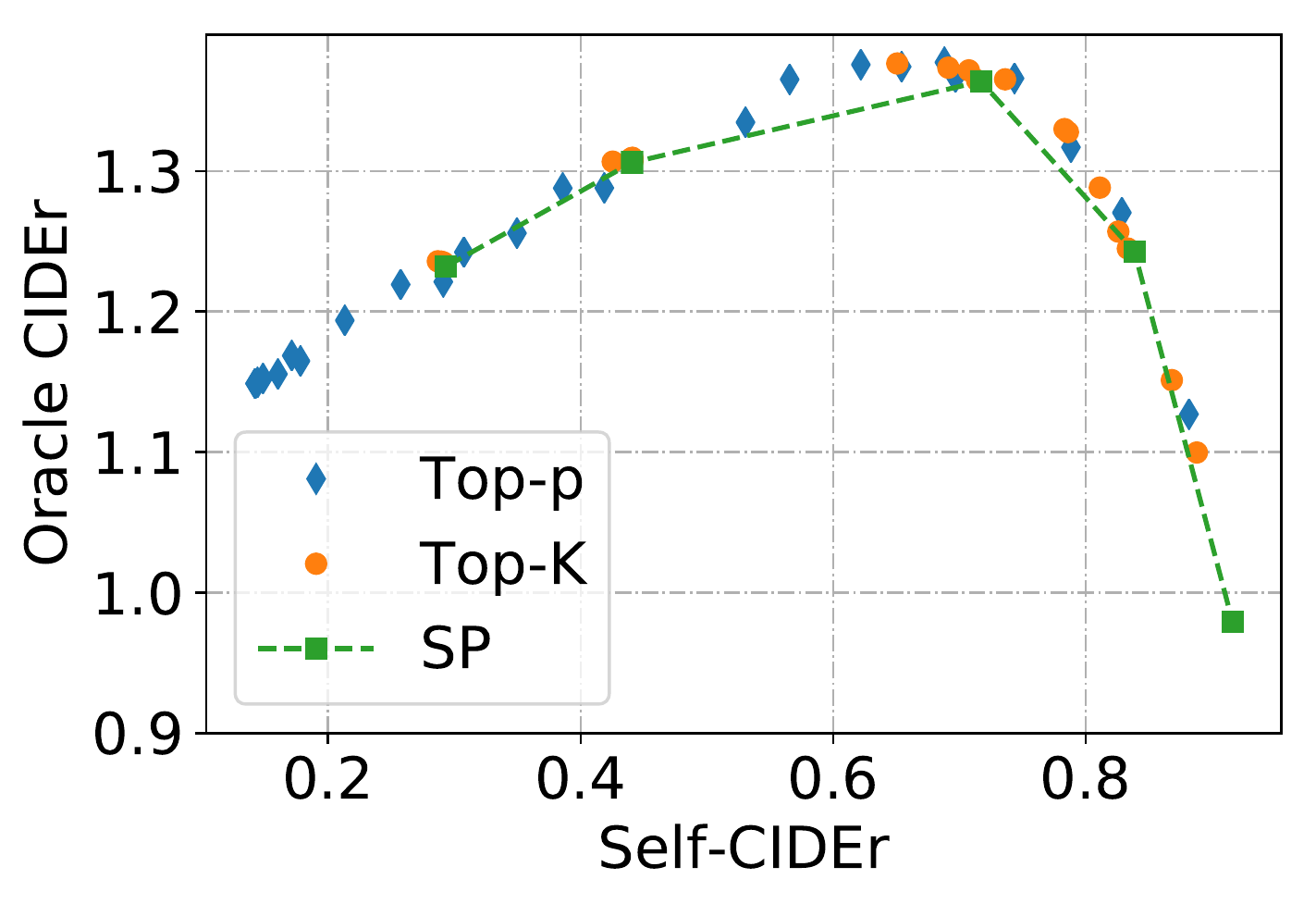}
\hspace{1em}
\includegraphics[height=4.5cm]{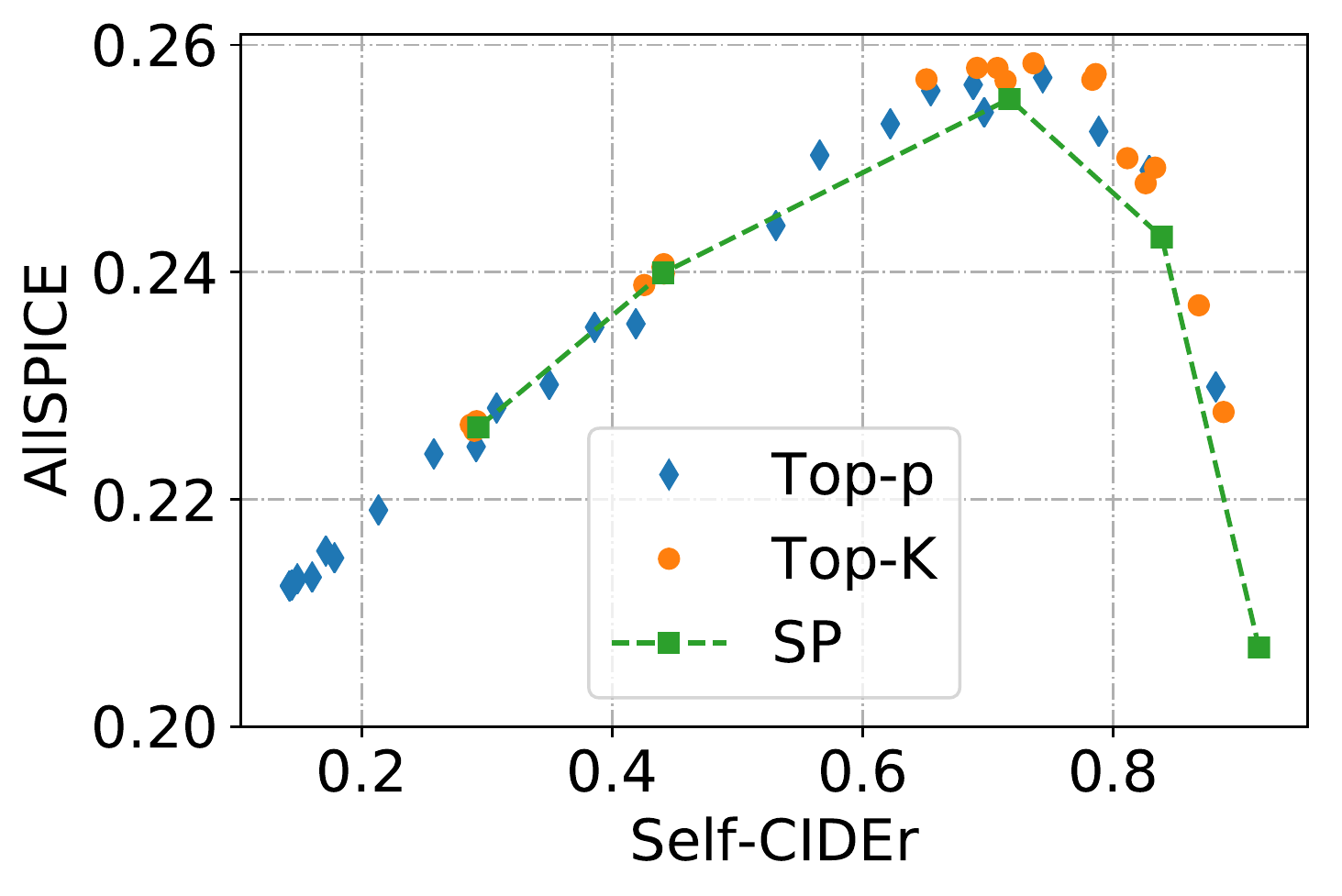}
\caption{Oracle CIDEr and \spicen of Top-K (orange) and Top-p (blue)
  sampling, compared to naive sampling with varying $\MT$
  (green), for XE-trained model. Each orange/blue dot represents a ($\MT$, K/p) pair.}
\label{fig:div_3}
\end{figure}

\noindent\textbf{Biased sampling.}
In Figure \ref{fig:div_3}, we explore Top-K and Top-p sampling for XE-trained models,
with  different thresholds and temperatures, compared against SP/temperature.
We see that they can slightly outperform SP, with no extra computation cost.

\begin{figure}[th!]
\centering
\includegraphics[height=4.5cm]{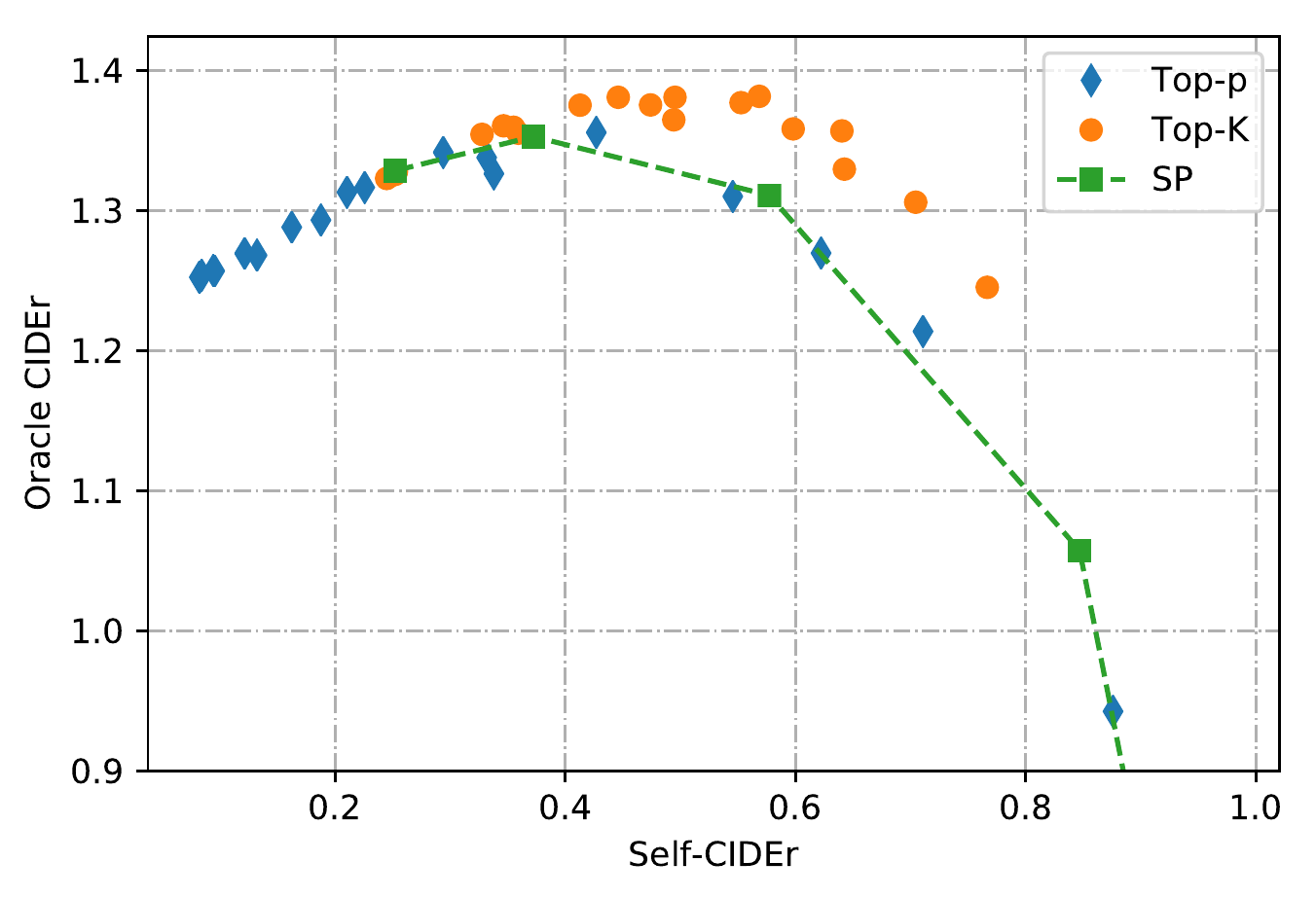}
\hspace{1em}
\includegraphics[height=4.5cm]{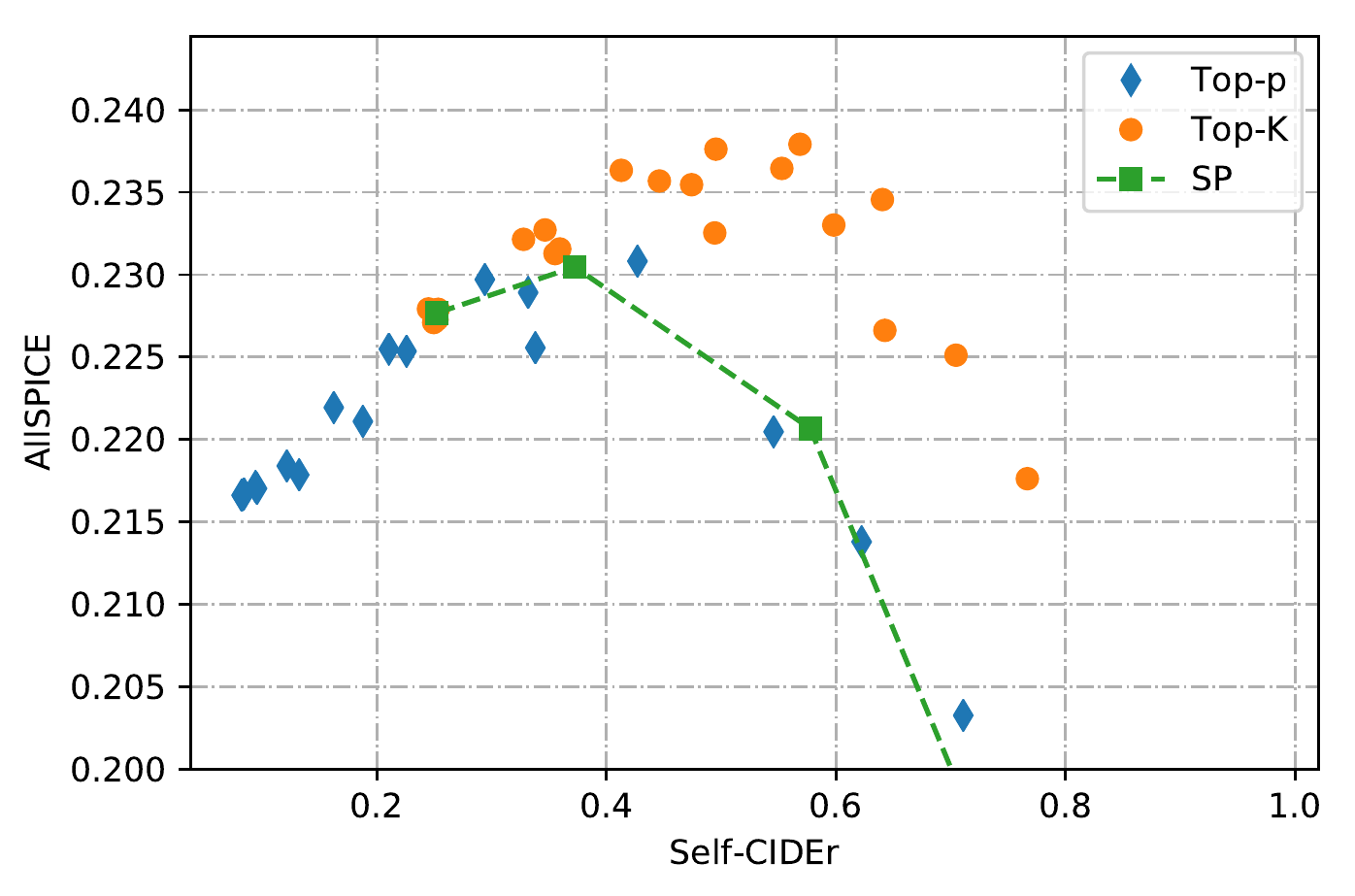}
\caption{Oracle CIDEr and \spicen of Top-K (orange) and Top-p (blue)
  sampling, compared to naive sampling with varying $\MT$
  (green), for RL-trained model.}
\label{fig:div_rl_biased}
\end{figure}

For RL-trained model, Top-K sampling can outperform SP and nucleus sampling: captions are more accurate with a similar level of diversity, shown in Figure \ref{fig:div_rl_biased}. Our intuition is CIDEr optimization makes the word posterior very peaky and the tail distribution noisier. With a large temperature, the sampling would fail because the selection will fall into the bad tail. Top-K can successfully eliminate this phenomenon and focus on correct things. Unfortunately, nucleus sampling is not performing as satisfying in this case.

\begin{figure}[th]
\centering
\includegraphics[height=4.5cm]{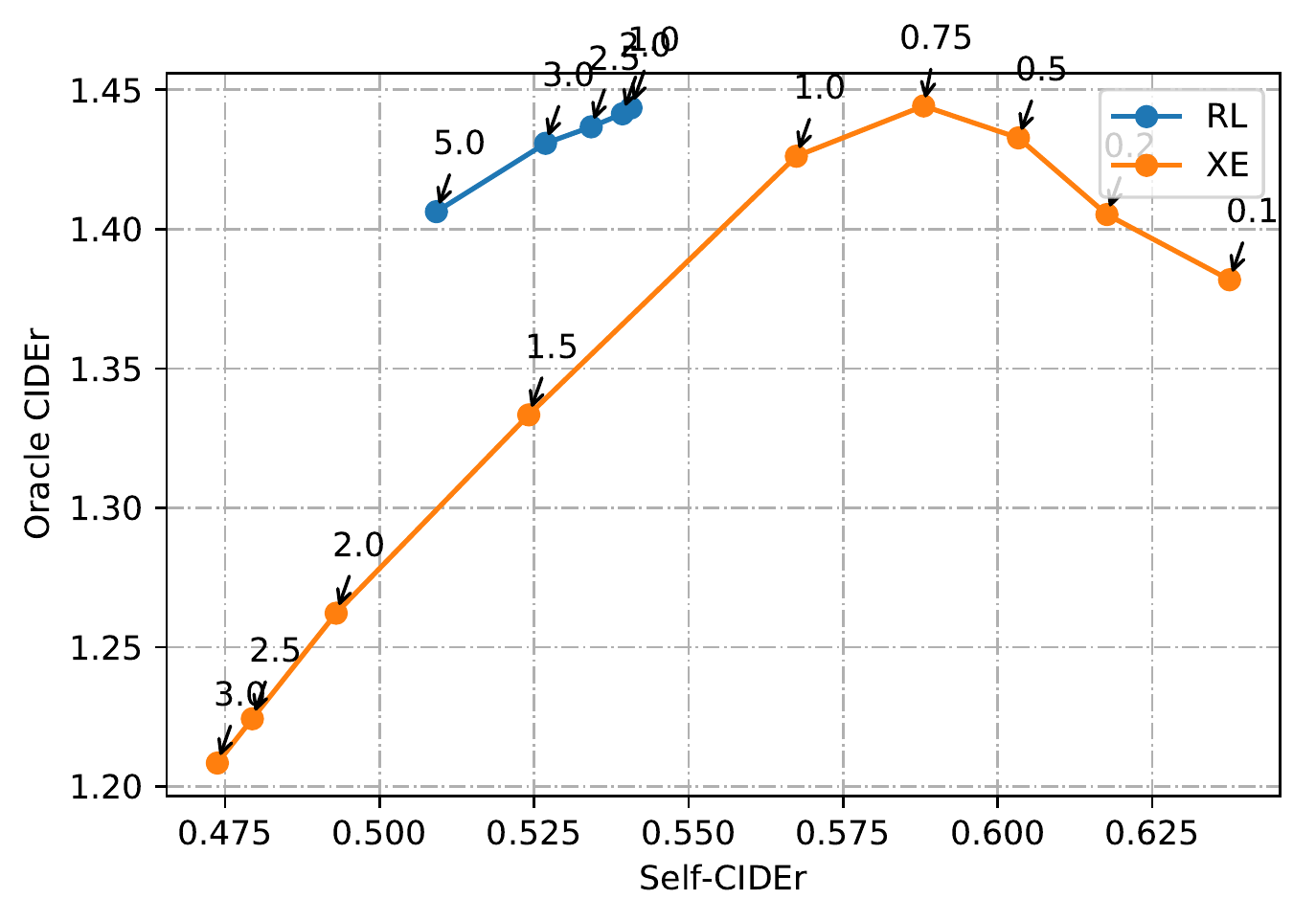}
\hspace{1em}
\includegraphics[height=4.5cm]{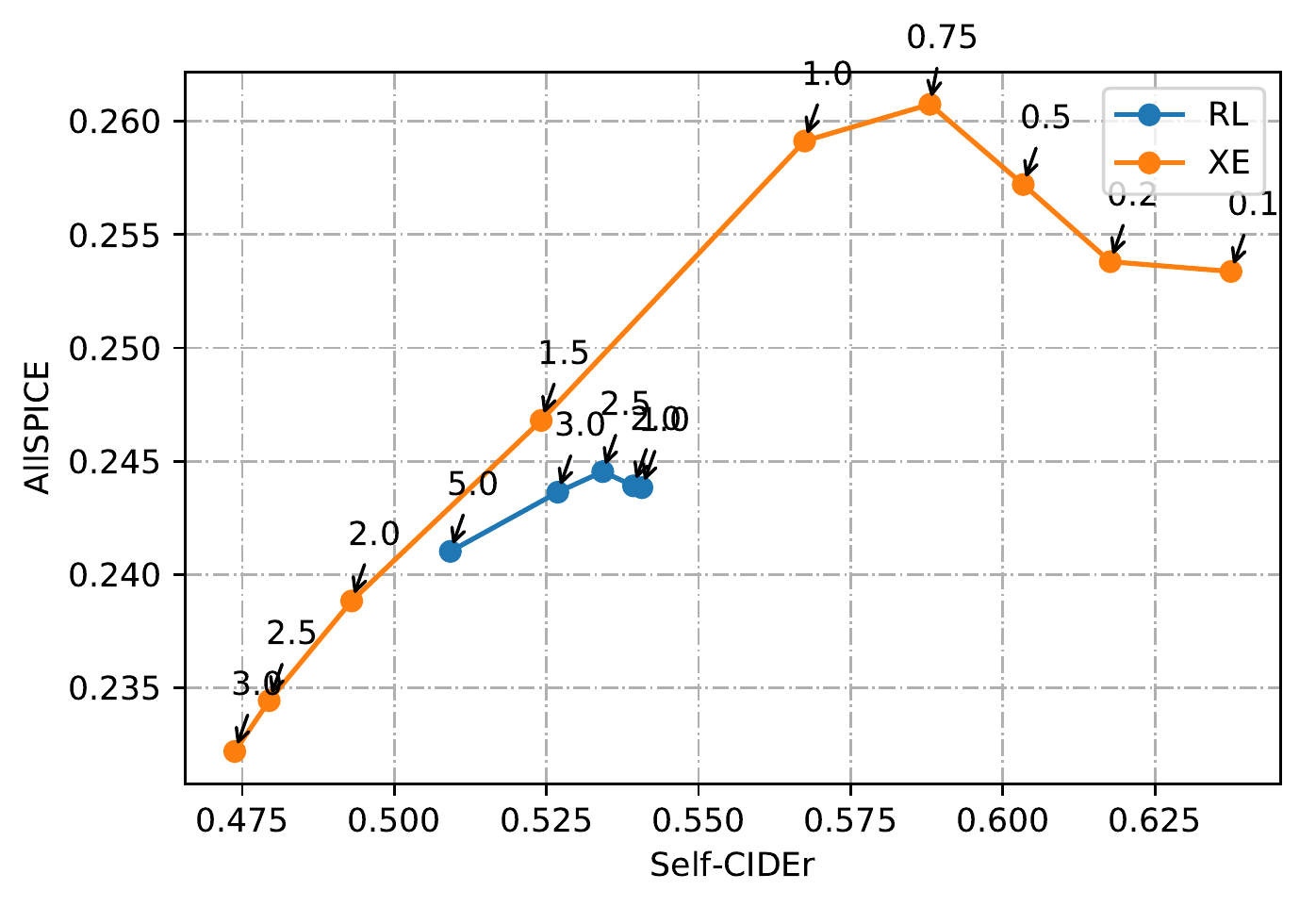}
\caption{Oracle CIDEr and \spicen of beam search with different temperatures, from RL-trained model (blue) and XE-trained model (orange). The numbers are the temperature $\MT$.}
\label{fig:div_bs}
\end{figure}

\noindent\textbf{Beam search.}
Figure \ref{fig:div_bs} shows how scores change by tuning the temperature for beam search. It's clear that no matter how to tune the BS for RL-trained model, the performance is always lower than that of XE-trained model.

Another interesting finding is unlike other methods, beam search produces more diverse set when having lower temperature. This is because lower temperature makes the new beams to have more chances to be expanded from different old beams instead of having one beam dominating (beams sharing same root).

\begin{figure}[th]
\centering
\includegraphics[height=4.5cm]{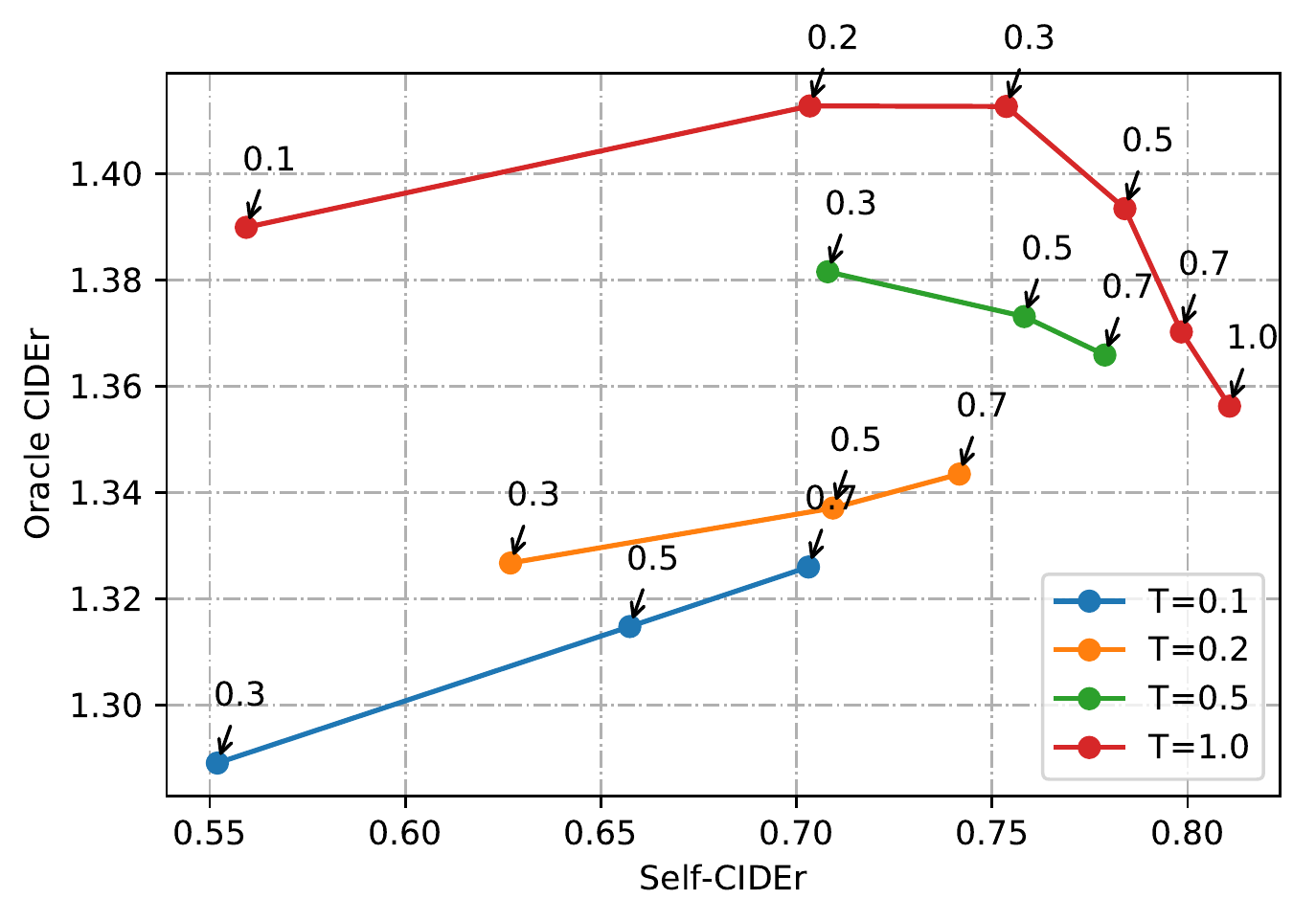}
\hspace{1em}
\includegraphics[height=4.5cm]{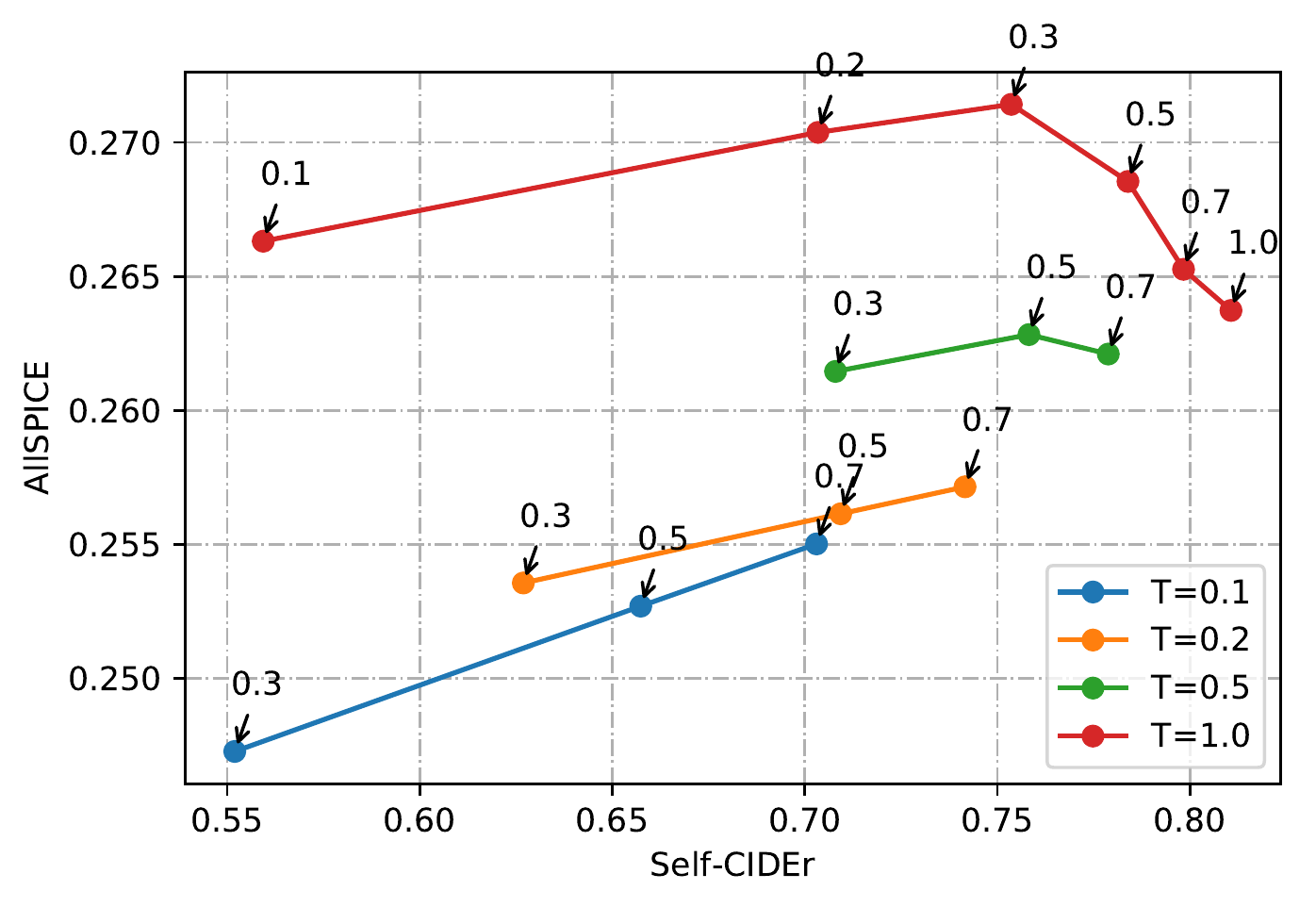}
\caption{Diverse beam search. Curve colors correspond to results with
  decoding an XE-trained model with different temperatures; each curve
tracks varying $\lambda$ in DBS.}
\label{fig:div_dbs}
\end{figure}

\begin{figure}[th]
\centering
\includegraphics[height=4.5cm]{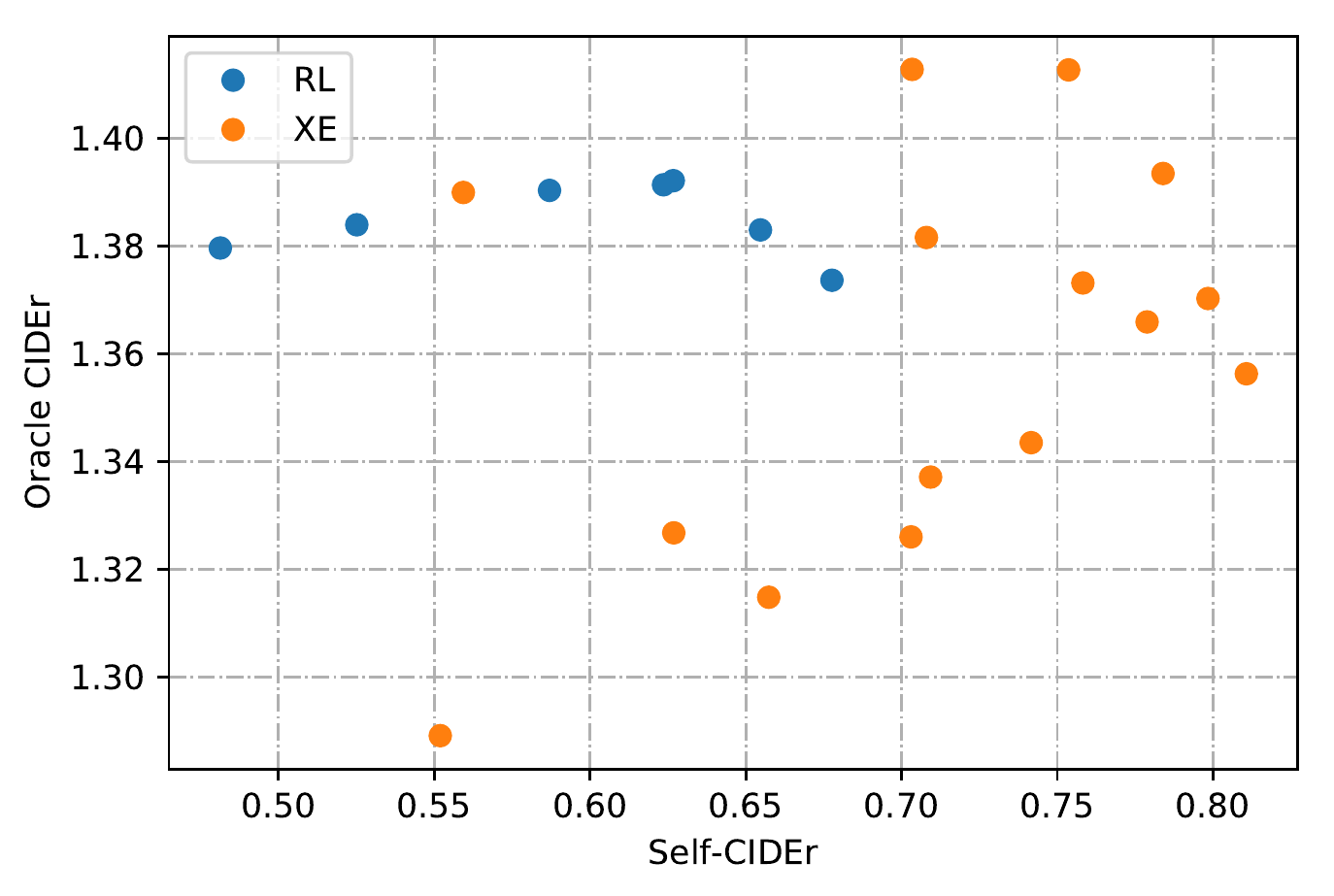}
\hspace{1em}
\includegraphics[height=4.5cm]{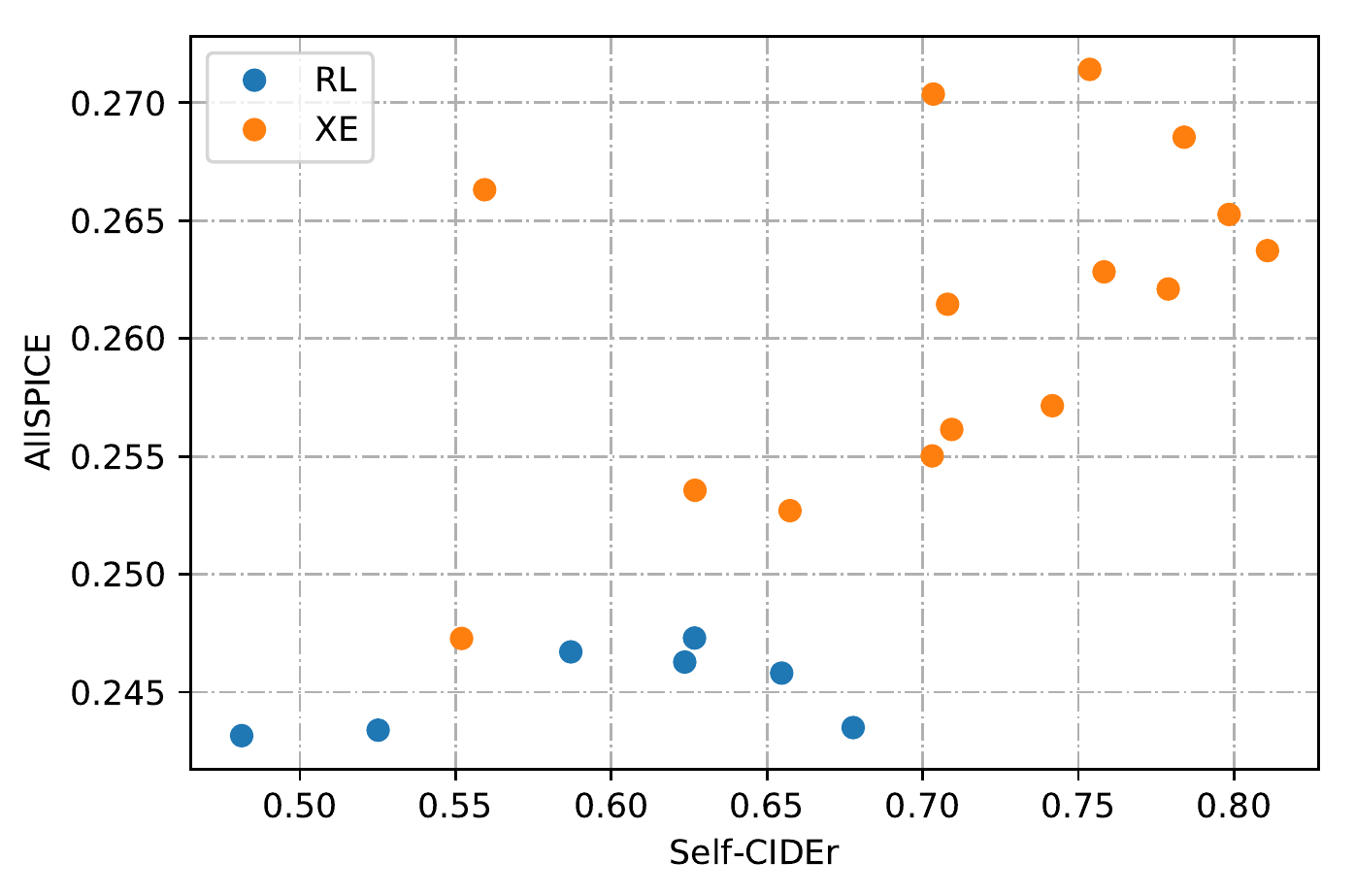}
\caption{Diverse beam search result of XE-trained model and RL-trained model. Each dot corresponds to a specific ($\lambda$, $\MT$) pair.}
\label{fig:div_dbs_rlxe}
\end{figure}

\noindent\textbf{Diverse beam search.}
Figure \ref{fig:div_dbs} shows the performance of Oracle CIDEr and \spicen with different settings of $\lambda$ and $\MT$ from an XE-trained model. Figure \ref{fig:div_dbs_rlxe} compares between XE-trained model and RL-trained model. The RL-trained model is also worse than XE-trained model when using DBS.

In summary, in \emph{all settings}, the RL-trained model performs worse than XE-trained models,
suggesting that RL-objective is detrimental to diversity.
In the remainder, we only discuss XE-trained models.

\begin{table}[tb!]
  \centering
    \begin{tabular}{lrrrr}
      \toprule
          & \multicolumn{1}{c}{Avg. CIDEr} & \multicolumn{1}{c}{Oracle CIDEr} & \multicolumn{1}{c}{\spicen} & \multicolumn{1}{c}{Self-CIDEr} \\
    \midrule
    DBS $\lambda$=0.3, T=1 & 0.919 & 1.413 & \textbf{0.271} & \textbf{0.754} \\
    BS T=0.75 & \textbf{1.073} & \textbf{1.444} & 0.261 & 0.588 \\
    Top-K K=3 T=0.75 & 0.921 & 1.365 & 0.258 & 0.736 \\
    Top-p p=0.8 T=0.75 & 0.929 & 1.366 & 0.257 & 0.744 \\
    SP T=0.5 & 0.941 & 1.364 & 0.255 & 0.717 \\
    \bottomrule
    \end{tabular}
  \caption{Performance for best hyperparameters for each method (XE-trained model).}
  \label{tab:div_methods_sp_5}%
\end{table}

\noindent\textbf{Comparing across methods (sample size 5).}
Table \ref{tab:div_methods_sp_5} shows the performance of each method under its best performing hyperparameter (under \spicen). Diverse beam search is the best algorithm with high \spicen and Self-CIDEr, indicating both semantic and syntactic diversity.

Beam search performs best on oracle CIDEr and average CIDEr, and it performs well on \spicen too. However, although all the generated captions are accurate, the syntactic diversity is missing, shown by Self-CIDEr. Qualitative results are shown in \secref{sec:div_qual}.

Sampling methods (SP, Top-K, Top-p) are much faster than beam
search and competitive in performance. This suggests them as a compelling alternative to recent methods for speeding up diverse caption generation, like~\cite{deshpande2019fast}.

\begin{figure}[th]
      \begin{minipage}[c]{.3\textwidth}
        \caption{Effect of sample size on different metrics, XE-trained model. Colors indicate decoding strategies.}
        \label{fig:div_samplen}
    \end{minipage}%
    \hfill 
    \begin{minipage}[c]{.65\textwidth}
      \centering
      \includegraphics[width=0.8\linewidth]{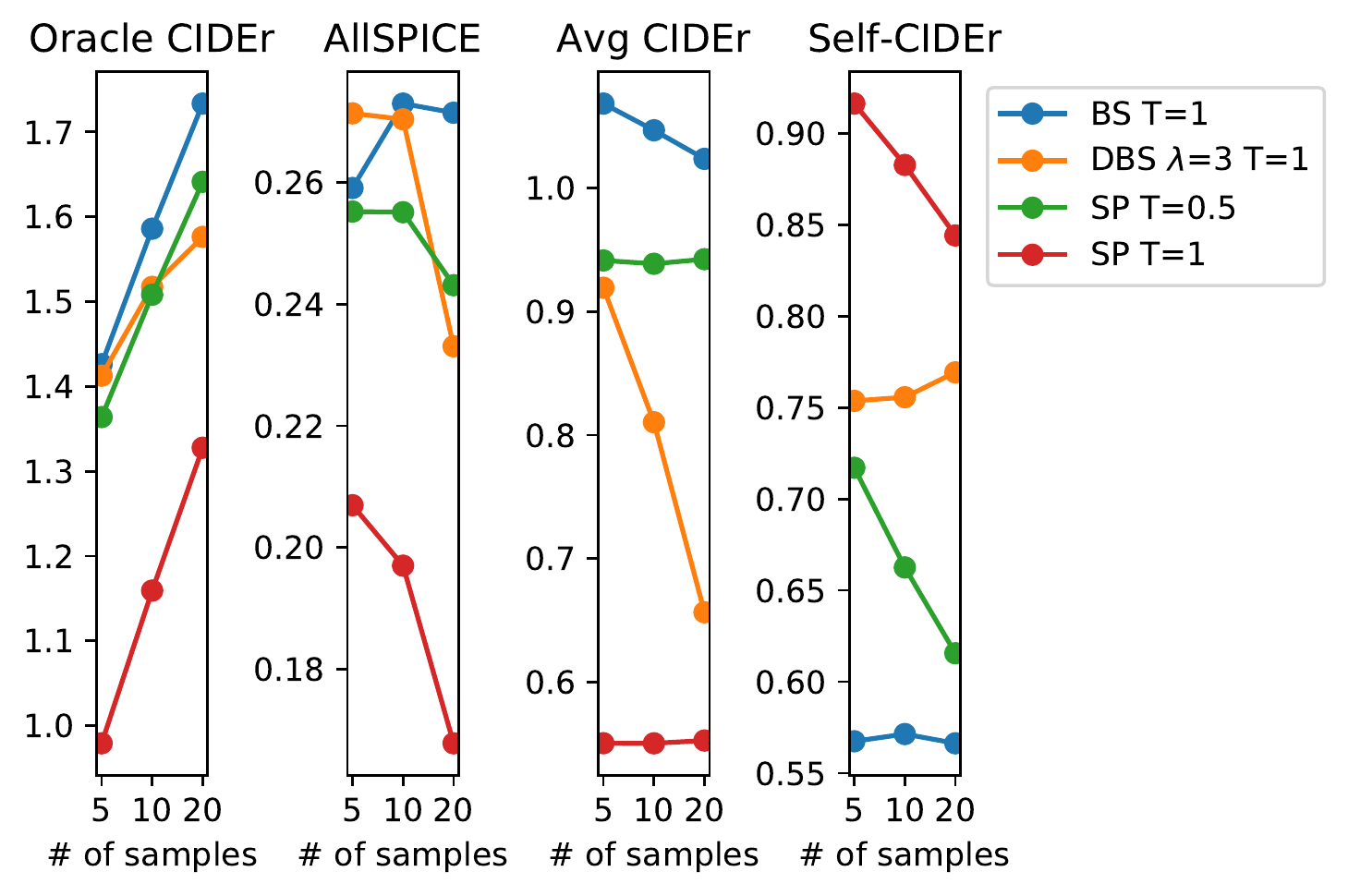}
    \end{minipage}%
  \end{figure}

\noindent\textbf{Different sample sizes} (see
Figure~\ref{fig:div_samplen}). \emph{Oracle CIDEr} tends to increase with sample size, 
as more captions mean more chances to fit the reference. 

\noindent\emph{\spicen} drops with more samples, because additional captions are more likely to hurt (say something wrong) than help (add something correct but not yet said). BS, which explores the caption space more ``cautiously'' than other methods, is initially resilient to this effect, but with enough samples its \spicen drops as well.

\noindent\emph{Average CIDEr}. Sampling methods' average scores are largely invariant to sample size.  BS and especially DBS suffer a lot with more samples, because diversity constraints and the properties of the beam search force the additional captions to be lower quality, hurting precision without improving recall.

\noindent\emph{Self-CIDEr} The relative order in Self-CIDEr is the same across different sample size; but the gap between SP/$\MT$=1 and DBS and between SP/$\MT$=0.5 and BS is closer when increasing the sample size. Because of the definition of Self-CIDEr, comparison across sample sizes is not very meaningful.

\begin{table}[!tb]
  \centering
    \begin{tabular}{lrrrrrrr}
      \toprule
          & \multicolumn{1}{c}{\multirow{1}[1]{*}{METEOR}} & \multicolumn{1}{c}{\multirow{1}[1]{*}{SPICE}} & \multicolumn{1}{c}{\multirow{1}[1]{*}{Div-1}} & \multicolumn{1}{c}{\multirow{1}[1]{*}{Div-2}} & \multicolumn{1}{c}{\multirow{1}[1]{*}{mBleu-4}} & \multicolumn{1}{c}{Vocabulary} & \multicolumn{1}{c}{\%Novel Sentence} \\
    \midrule
    Base & 0.265 & 0.186 & 0.31  & 0.44  & 0.68  & 1460  & 55.2 \\
    Adv & 0.236 & 0.166 & 0.41  & 0.55  & 0.51  & 2671  & 79.8 \\
    \midrule
    T=0.33 & 0.266 & 0.187 & 0.31  & 0.44 & 0.65 & 1219 & 58.3 \\
    T=0.5 & 0.260 & 0.183 & 0.37  & 0.55  & 0.47  & 1683  & 75.6 \\
    T=0.6 & 0.259 & 0.181 & 0.41  & 0.60  & 0.37  & 2093  & 83.7 \\
    T=0.7 & 0.250 & 0.174 & 0.45  & 0.66  & 0.28  & 2573  & 89.9 \\
    \textcolor{green!50!black}{T=0.8} & 0.240 & 0.166 & 0.49  & 0.71  & 0.21  & 3206  & 94.4 \\
    T=0.9 & 0.228 & 0.157 & 0.54  & 0.77  & 0.14  & 3969  & 97.1 \\
    T=1   & 0.214 & 0.144 & 0.59  & 0.81  & 0.08  & 4875  & 98.7 \\
    \bottomrule
    \end{tabular}%
  \caption{Base and Adv are from \cite{shetty2017speaking}. The other models are trained by us. All the methods use naive sampling decoding. }
  \label{tab:div_gan}%
\end{table}

\noindent\textbf{Comparison with adversarial loss.}
In \cite{shetty2017speaking}, the authors get more diverse but less
accurate captions with adversarial loss. Table~\ref{tab:div_gan} shows that by tuning sampling temperature, a XE-trained model can generate more diverse \emph{and} more accurate captions than adversarially trained model. Following~\cite{shetty2017speaking}, METEOR and SPICE are computed by measuring the score of the highest likelihood sample among 5 generated captions.
For a fair comparison, we use a similar non-attention LSTM model (FC model). When the temperature is 0.8, the output generations outperform the adversarial method (Adv). While \cite{shetty2017speaking} also compare their model to a base model using sampling method (Base), they get less diverse captions because they use temperature $\frac{1}{3}$ (our model with T=0.33 is performing similar to ``Base'').

\begin{table}[htbp]
  \centering
    \begin{tabular}{lrrrr}
      \toprule
          & ROUGE & METEOR & CIDEr & SPICE \\
    \midrule
    FC (XE)    & 0.543 & 0.258 & 1.006 & 0.187 \\
    Att2in (XE)  & 0.562 & 0.272 & 1.109 & 0.201 \\
    Att2in-L (XE) & 0.561 & 0.275 & 1.116 & 0.205 \\
    Trans (XE) & 0.563 & 0.278 & 1.131 & 0.208 \\
    \midrule
    FC (RL) & 0.555 & 0.263 & 1.123 & 0.197 \\
    Att2in (RL) & 0.572 & 0.274 & 1.211 & 0.209 \\
    Att2in-L (RL) & 0.584 & 0.285 & 1.267 & 0.219 \\
    Trans (RL) & 0.589 & 0.291 & 1.298 & 0.230 \\
    \bottomrule
    \end{tabular}%
  \caption{Single caption generation performance of different model architectures. Generated by beam search with beam size 5.}
  \label{tab:div_models}%
\end{table}%


\begin{figure}[th]
\centering
\includegraphics[height=4.5cm]{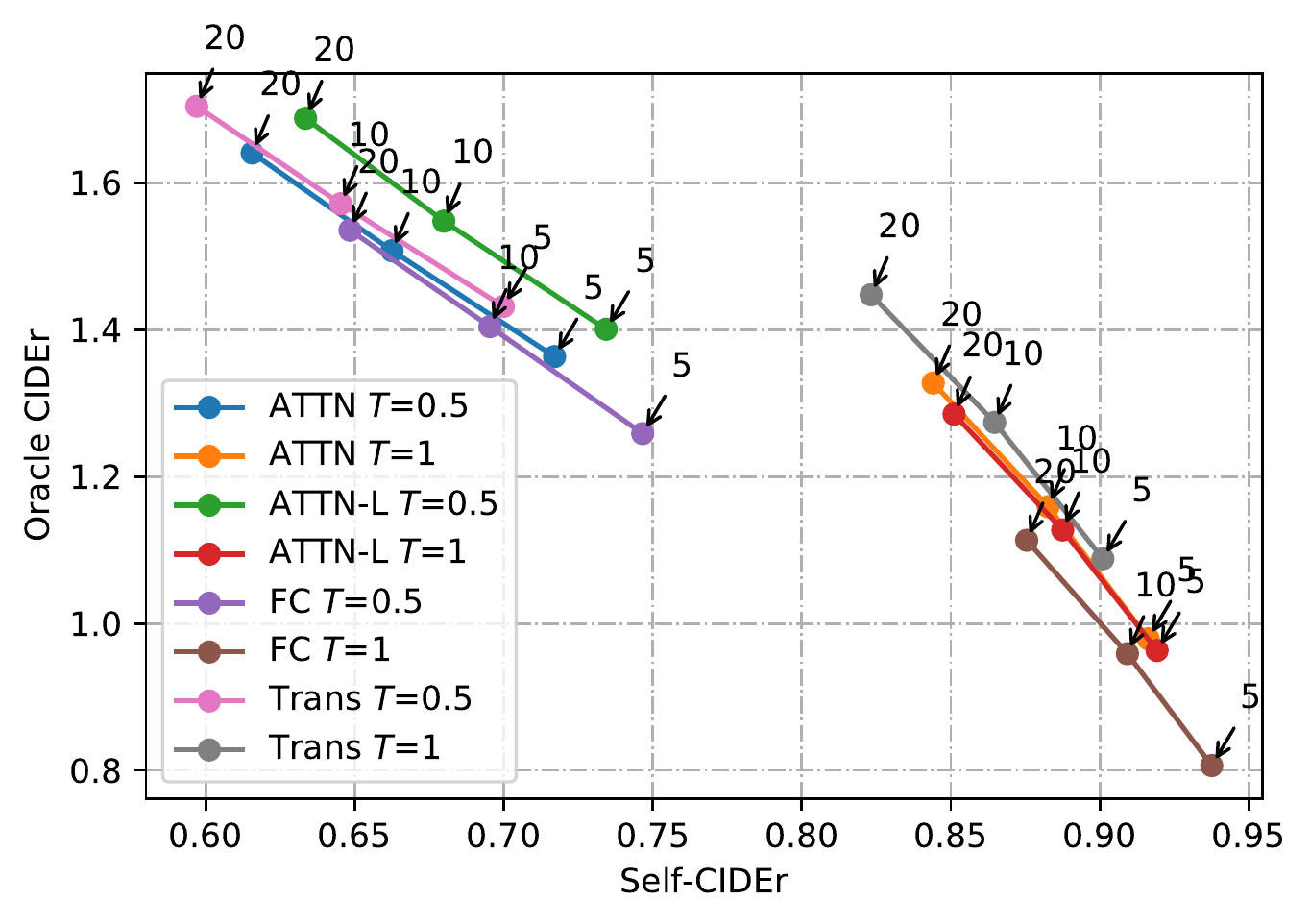}
\hspace{1em}
\includegraphics[height=4.5cm]{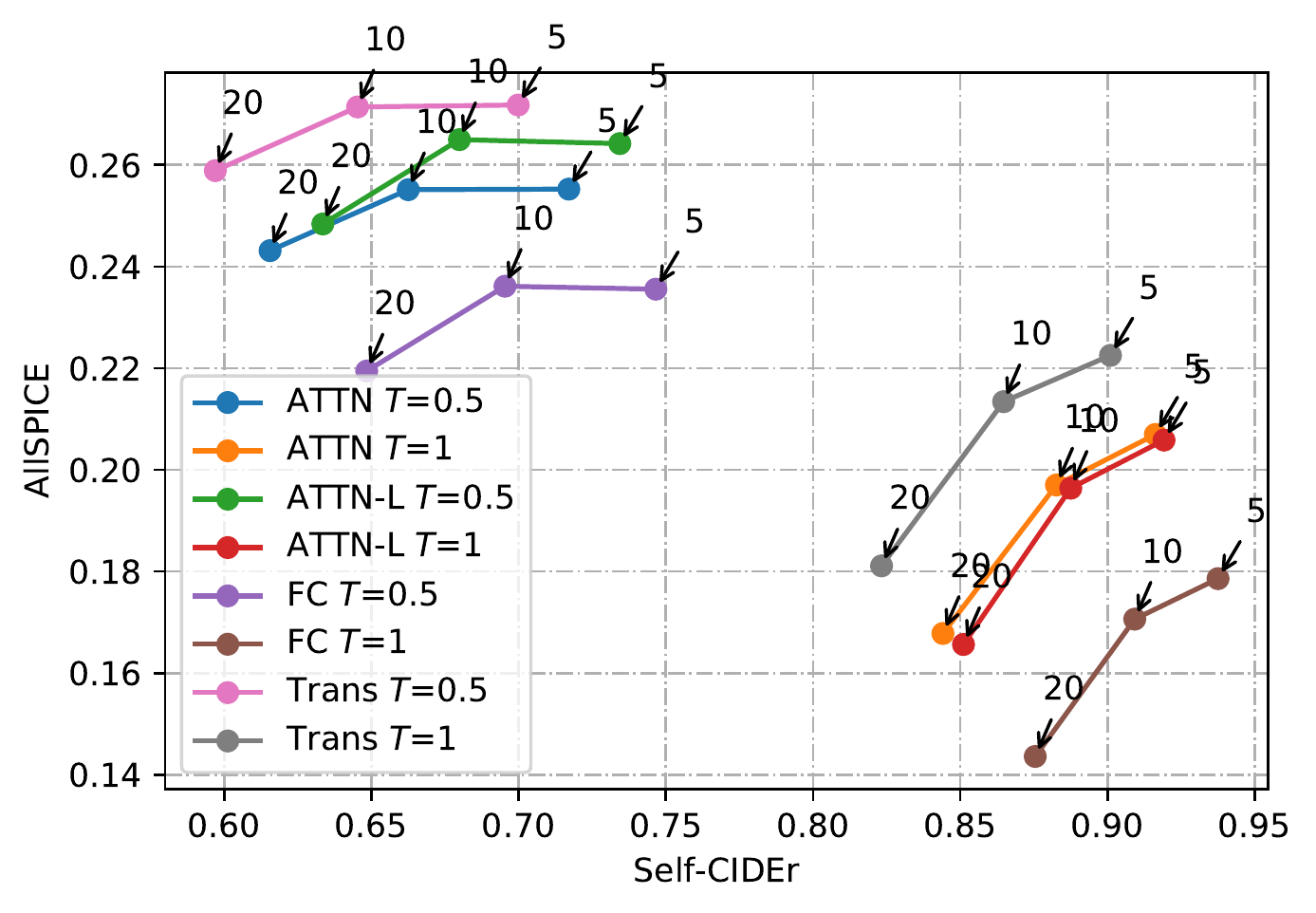}
\caption{Oracle CIDEr and \spicen of different models. The numbers with arrows indicate the sample size.}
\label{fig:div_models}
\end{figure}

\subsection{Different models}
Here we compare FC, Att2in, Att2in-L, and Trans trained with XE. 
Table \ref{tab:div_models} shows the single caption generation result of these four architectures trained with XE and RL. The captions are generated by beam search with beam size 5. The performance ranking is: FC $<$ Att2in $<$ Att2in-L $<$ Trans. Figure \ref{fig:div_models} shows the naive sampling results for different models trained with XE. \spicen performance on different models has the same order as in single caption generation performance.


\begin{figure}[th]
  \begin{minipage}[c]{.3\textwidth}
    \caption{Diversity-accuracy tradeoff of Att2in on Flickr30k. Blue: RL, decoding with different temperatures. 
    Orange: XE, decoding with different temperatures.}
    \label{fig:div_f30k1}
\end{minipage}%
\hfill 
\begin{minipage}[c]{.65\textwidth}
  \centering
  \includegraphics[width=0.8\linewidth]{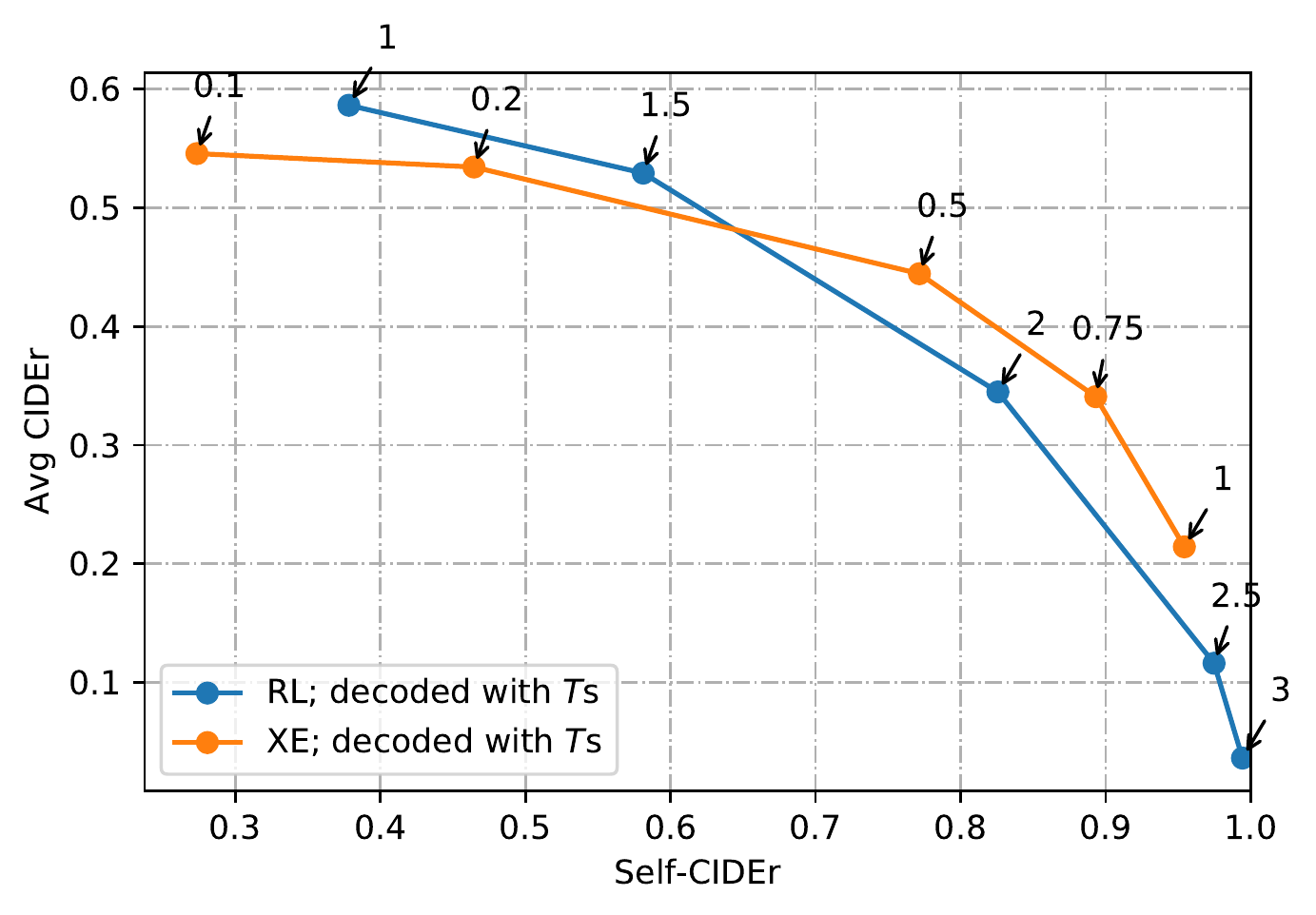}
\end{minipage}%
\end{figure}

\begin{figure}[th]
\includegraphics[height=4.5cm]{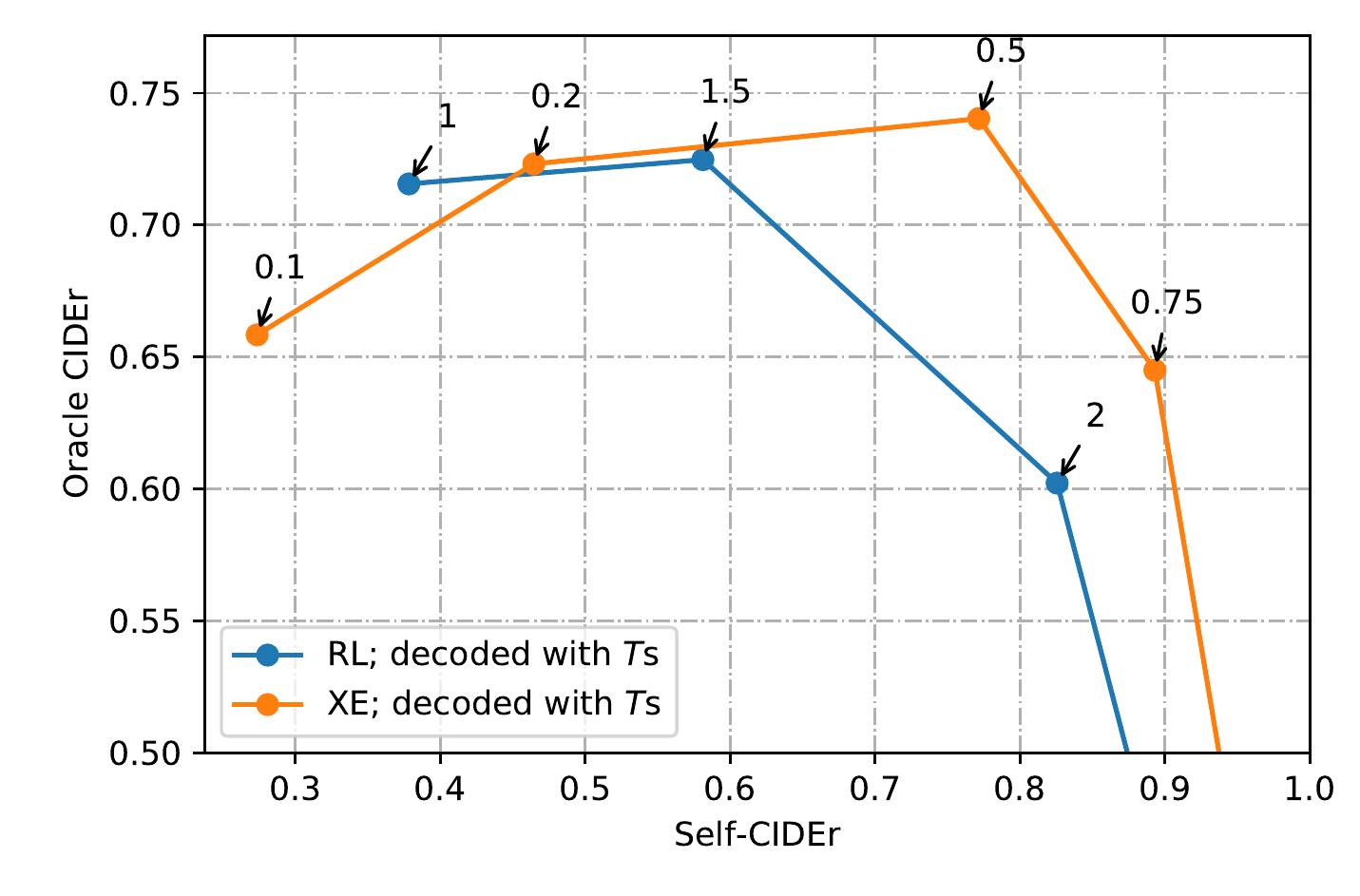}
\hspace{1em}
\includegraphics[height=4.5cm]{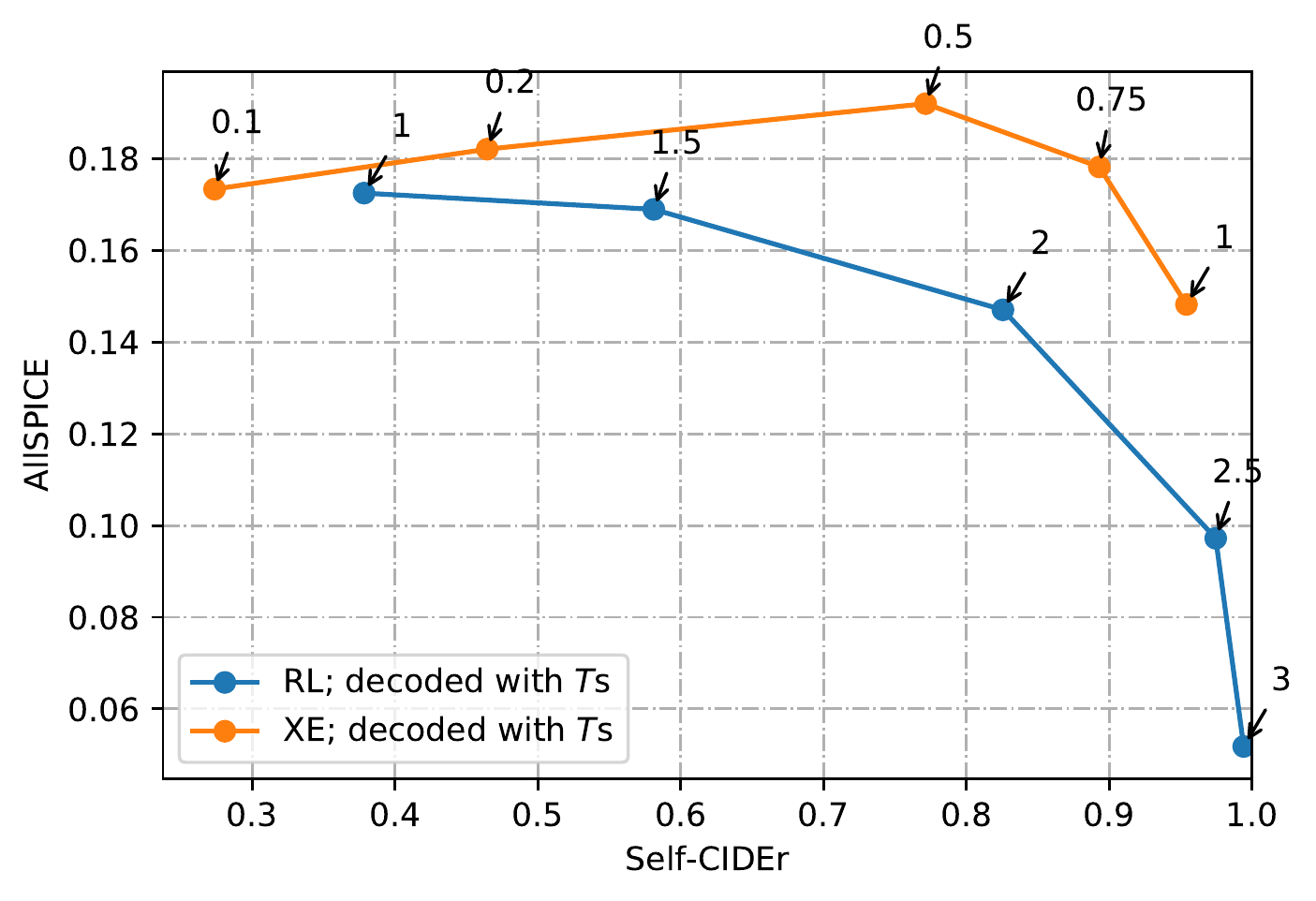}
\caption{Oracle CIDEr and \spicen vs. Self-CIDEr  on Flickr30k.}
\label{fig:div_f30k2}
\end{figure}

\begin{figure}[tb!]
\includegraphics[height=4.5cm]{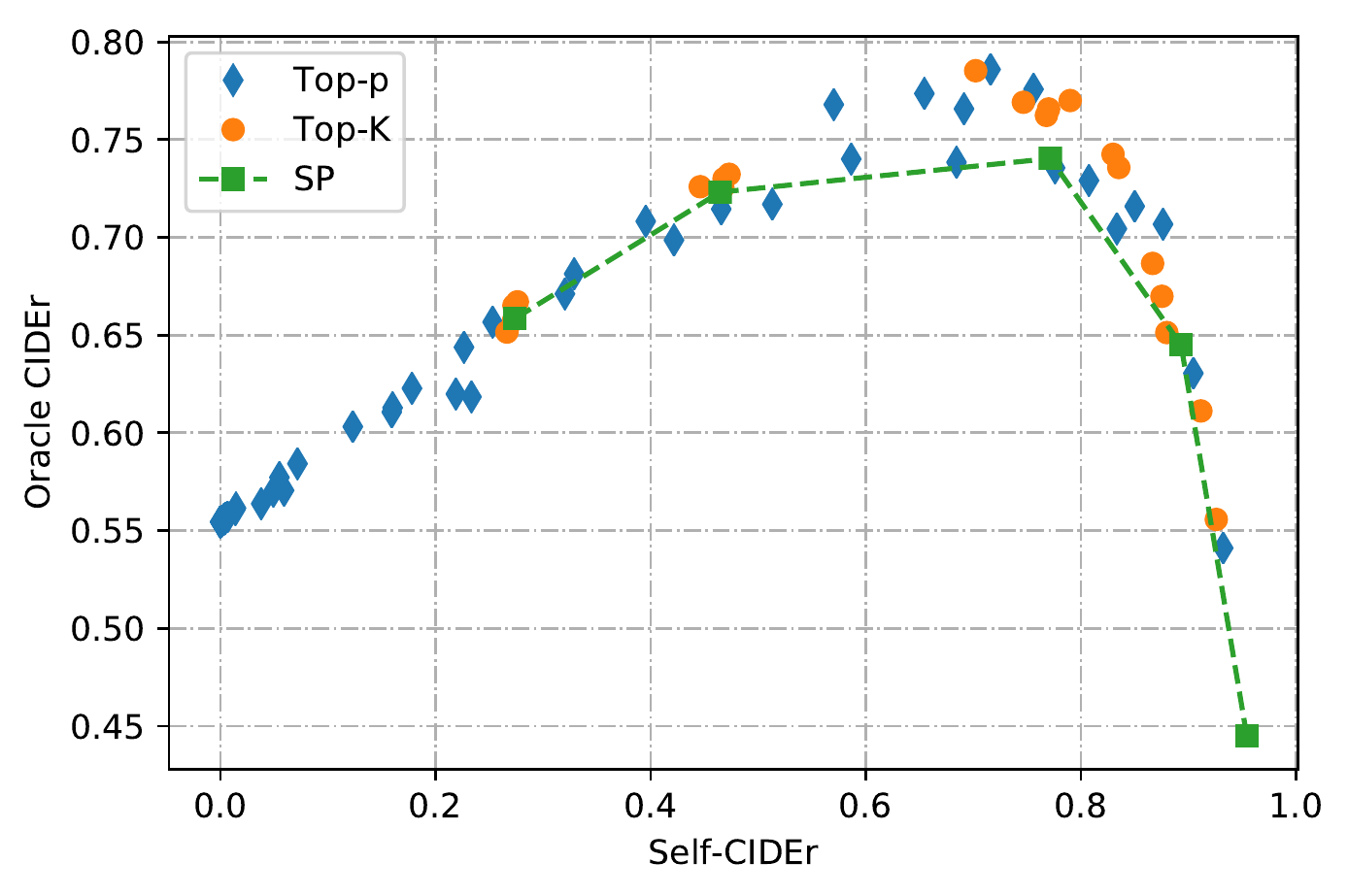}
\hspace{1em}
\includegraphics[height=4.5cm]{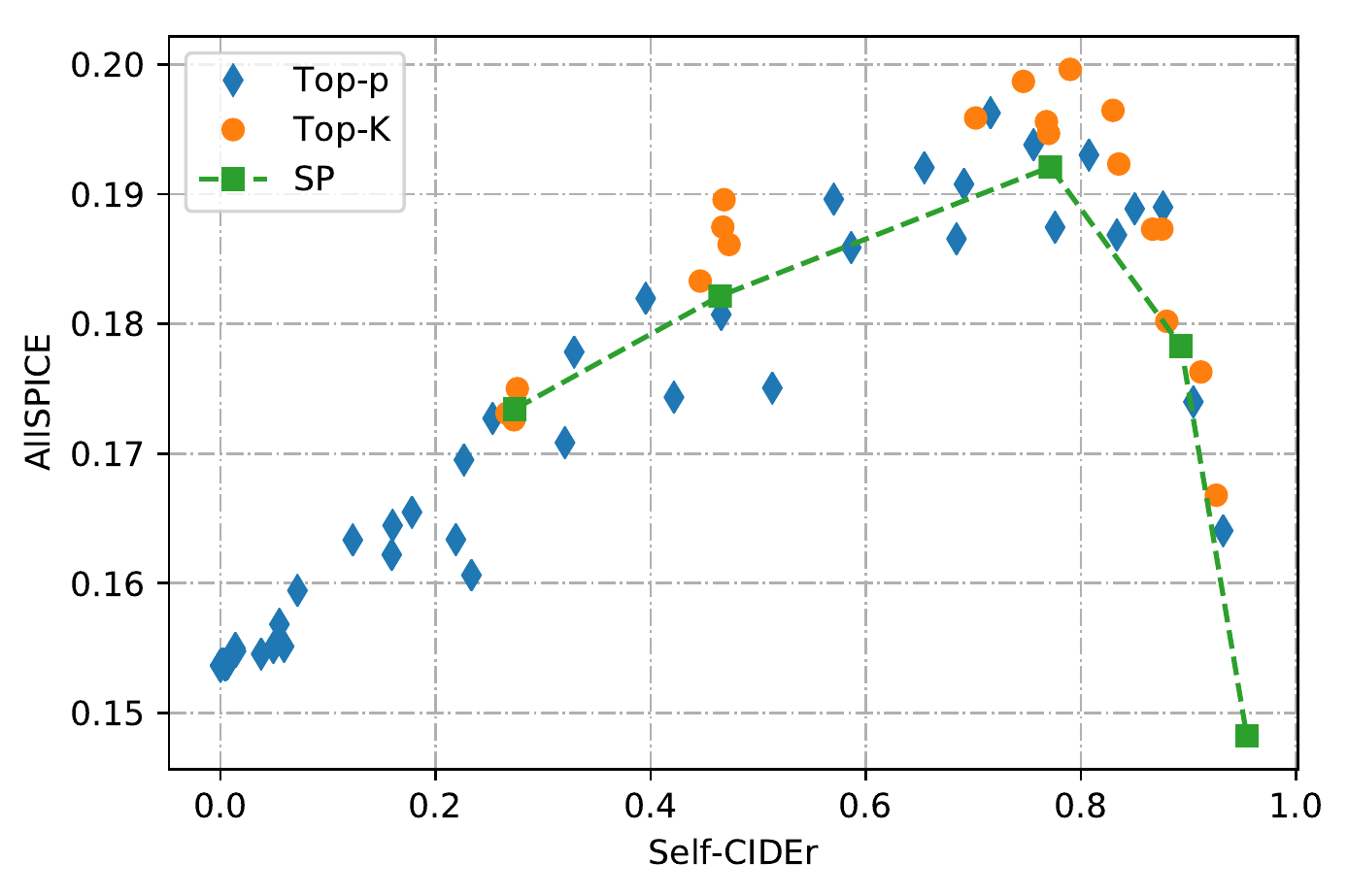}
\caption{Oracle CIDEr and \spicen of Top-K (orange) and Top-p (blue)
sampling, compared to naive sampling with varying $\MT$
(green) on Flickr30k, for XE-trained model.}
\label{fig:div_f30k3}
\end{figure}

\subsection{Results on Flickr30k}
Here we also plot similar figures of naive sampling and biased sampling for Att2in model on Flickr30k dataset~\cite{young-etal-2014-image} (see Figure \ref{fig:div_f30k1}, \ref{fig:div_f30k2}, \ref{fig:div_f30k3}). The results show that our conclusions on COCO also hold for Flickr30k.


\subsection{Qualitative results}\label{sec:div_qual}

\begin{figure}
\begin{minipage}[t]{0.45\linewidth}
\includegraphics[height=3.5cm]{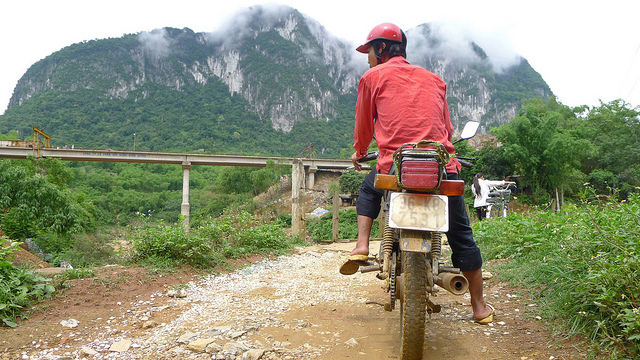}\\
{\small 
{\bf DBS $\lambda$=3 $\MT$=1}:\\
a man riding a motorcycle down a dirt road\\there is a man riding a motorcycle down the road\\man riding a motorcycle down a dirt road\\an image of a man riding a motorcycle\\the person is riding a motorcycle down the road

{\bf BS $\MT$=0.75}:\\
a man riding a motorcycle down a dirt road\\a man riding a motorcycle on a dirt road\\a man riding a motorcycle down a rural road\\a man riding a motorcycle down a road\\a man riding a motorcycle down a road next to a mountain

{\bf Top-K K=3 $\MT$=0.75}:\\
a man riding a motorcycle down a dirt road\\a person on a motor bike on a road\\a man riding a bike on a path with a mountain in the background\\a man riding a motorcycle down a dirt road with mountains in the background\\a man riding a motorcycle down a road next to a mountain

{\bf Top-p p=0.8 $\MT$=0.75}:\\
a man riding a motorcycle down a dirt road\\a person riding a motorcycle down a dirt road\\a man riding a motorcycle down a dirt road\\a man riding a motorcycle down a dirt road\\a man on a motorcycle in the middle of the road

{\bf SP $\MT$=0.5}:\\
a man riding a motorcycle down a dirt road\\a man on a motorcycle in the middle of a road\\a person riding a motorcycle down a dirt road\\a man rides a scooter down a dirt road\\a man riding a motorcycle on a dirt road
}
\end{minipage}
\hspace{1em}
\begin{minipage}[t]{0.45\linewidth}
\includegraphics[height=3.5cm]{figures/diversity/captions/f2.jpeg}\\
{\small 
{\bf DBS $\lambda$=3 $\MT$=1}:\\
a woman sitting at a table with a plate of food\\two women are sitting at a table with a cake\\the woman is cutting the cake on the table\\a woman cutting a cake with candles on it\\a couple of people that are eating some food

{\bf BS $\MT$=0.75}:\\
a woman sitting at a table with a plate of food\\a woman sitting at a table with a cake\\a woman cutting a cake with a candle on it\\a couple of women sitting at a table with a cake\\a couple of women sitting at a table with a plate of food

{\bf Top-K K=3 $\MT$=0.75}:\\
a woman sitting at a table with a cake with candles\\a woman cutting into a cake with a candle\\a woman sitting at a table with a fork and a cake\\a woman cutting a birthday cake with a knife\\a woman sitting at a table with a plate of food

{\bf Top-p p=0.8 $\MT$=0.75}:\\
two women eating a cake on a table\\a woman cutting a cake with a candle on it\\two women sitting at a table with a plate of food\\a woman cutting a cake with candles on it\\a woman is eating a cake with a fork

{\bf SP $\MT$=0.5}:\\
a woman is cutting a piece of cake on a table\\a woman is cutting a cake with a knife\\a woman is eating a piece of cake\\a woman is cutting a cake on a table\\a group of people eat a piece of cake
}
\end{minipage}
\captionof{figure}{Qualitative results.}\label{fig:div_qual}
\end{figure}

We show caption sets sampled by different decoding methods for two images in Figure \ref{fig:div_qual}. The model is Att2in model trained with XE. The hyperparameters are the same as in Table \ref{tab:div_methods_sp_5}. Additional three images are shown in Appendix \ref{app:div_more_qual}.

 \section{Conclusion}

Our results provide
both a practical guidance and a better understanding of the
diversity/accuracy tradeoff in image captioning.
We show that caption decoding methods may affect
this tradeoff more significantly than the choice of
training objectives or model. In particular, simple naive sampling, coupled with suitably low
temperature, is a competitive method with respect to speed and
diversity/accuracy tradeoff. Diverse beam search exhibits the best
tradeoff, but is also the slowest.
Among training objectives, using CIDEr-based
reward reduces diversity in a way that is not mitigated by
manipulating decoding parameters. Finally, we introduce a
metric, \spicen, for directly evaluating tradeoff between accuracy and semantic diversity.

\chapter{Controlling caption length}\label{chap:length}

\section{Problem statement}

\begin{figure}[htbp]
\centering
\includegraphics[width=0.6\textwidth]{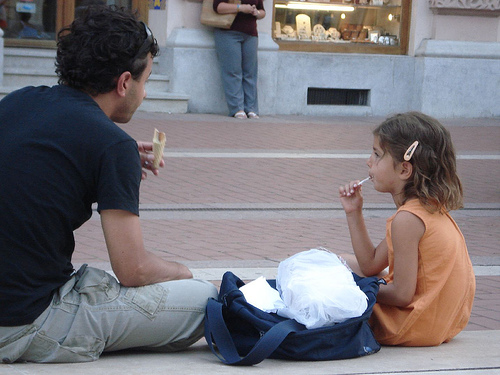}
\begin{minipage}[t]{0.8\textwidth}
A man and a girl sit on the ground and eat . \\
A man and a little girl are sitting on a sidewalk near a blue bag eating .\\ 
A man wearing a black shirt and a little girl wearing an orange dress share a treat .\\
\end{minipage}
\caption{An example of different-length captions for the same image: different lengths have different detailedness. The example is from link\protect\footnotemark.}\label{fig:length_example}
\end{figure}

Most existing captioning models learn an autoregressive model, where the captions are generated left-to-right word by word. The length of the caption is determined
only after the End Of Sentence ($\langle$eos$\rangle$) is generated.
Explicit control of such generation process
is difficult for such a model: it is hard to
know and control the length beforehand.
However, length can be an
important property of a caption.
By controlling the length, we can
influence the style and descriptiveness of the caption: short, simple
captions vs. longer, more complex and detailed descriptions for the
same image. (See \figref{fig:length_example} for an example.)

\footnotetext{\href{https://www.analyticsvidhya.com/blog/2018/04/solving-an-image-captioning-task-using-deep-learning/}{https://www.analyticsvidhya.com/blog/2018/04/solving-an-image-captioning-task-using-deep-learning/}\rtcom{not in the same page}}

To control the length of the generated caption, we build the model
borrowing existing ideas from summarization literature by injecting length
information into the model~\cite{kikuchi2016controlling,fan2017controllable}.
We also add a \emph{length prediction module} which predicts an optimal length for the input image, when no explicit length specification is given.
We show that the length models can successfully generate a caption
ranging from 7 up to 28 words long. We also show that length models
perform better than a non-controlled model (with some special decoding
methods)
when asked to generate long captions.

\section{Related work}

There exists some recent work on introducing controllability into image captioning models. \cite{deshpande2019fast} proposed a model that generates a caption by controlled POS tags. \cite{cornia2019show} control the captions by inputting a different set of image regions. Instead of selecting regions, \cite{zhong2020comprehensive} decompose a scene graph (of an image) into a set of subgraphs and control the caption generation by selecting sub-graphs where each captures a semantic component of the input image. \cite{chen2021human} proposed verb-specific semantic roles controlled captioning that is both event-compatible and sample-suitable. Other work has focused on controlling sentiment~\cite{mathews2016senticap}, emotions~\cite{mathews2018semstyle,gan2017stylenet}, and personality~\cite{shuster2019engaging}.

Control is more broadly studied in the NLP community for different text generation tasks. \cite{ficler2017controlling} train a conditional RNN for controlling style and content where the controls are fed as contexts. CTRL~\cite{keskar2019ctrl} scale \cite{ficler2017controlling} to a large model (1.6B parameters) with over 50 control codes.
\cite{hu2017toward} train variational auto-encoders for style control where the disentangled latent representations control style and content.
\cite{tu2019generating} studied diverse story continuation by controlling multiple sentence attributes including sentiment, frames, predicates. \cite{dathathri2019plug} proposed Plug and play Language model, a simple way to use pre-trained language models to generate text with specific attributes by manipulating the hidden state without changing pre-trained parameters.
Closer to us, length control has gained interest in abstract
summarization in
\cite{kikuchi2016controlling,fan2017controllable,takase2019positional}

There also exists decoding methods that can loosely control the length. For example, \cite{jia2015guiding,wu2016google} apply length normalization during beam search to encourage longer sequences for image captioning and machine translation, respectively.

Concurrent to us, \cite{deng2020length} proposed a method that can control caption length. Their work differs us in two aspects. First, their model is based on a non-autoregressive model which generates tokens simultaneously. Secondly, their model takes a length level (for example, 10-14 tokens) as input, instead of a specific length as we do. Thus our model is able to provide more fine-grained control.

\section{Models}

We consider repurposing existing methods in summarization for captioning.
In general, the length is treated as an intermediate
variable: $P(\bc|\I) =
P(\bc|\I,l)*P(l|\I)$. $\bc$, $\I$ and $l$ are caption, image and caption length,
respectively. We introduce how we build $P(\bc|\I,l)$ and $P(l|\I)$ as
follows. Note that, the following methods can be used in conjunctions
with any standard captioning model.

\subsection{LenEmb~\cite{kikuchi2016controlling}}
We take LenEmb from \cite{kikuchi2016controlling} and make a small
change according to \cite{takase2019positional}. Given desired length
$l$ and current time step $t$, we embed the remaining length $l-t$
into a vector that is the same size as the word embedding. Then the
word embedding of the previous word $w_{t-1}$, added with the length
embedding (rather than concatenated, as in \cite{kikuchi2016controlling}), is fed as the input to the rest of the LSTM model.
\begin{align}
x_t = E_w(w_{t-1}) + E_l(l-t) \\
P(w_t|\I, w_{0, \ldots, t-1}) = \text{LSTM}(x_t, h_{t-1})
\end{align}

where $E_w$ represents the word embedding and $E_l$ is the length embedding.

\noindent\textbf{Learning to predict length.} We add a length prediction
module to predict the length given the image features while no desired length is
provided. We treat it as a classification task and train it
with the reference caption length.



\subsection{Marker~\cite{fan2017controllable}}
We also implement Marker model from \cite{fan2017controllable}. The desired length is fed as a special token at the beginning of generation as the ``first word.'' (The first two tokens of the sequence will be the length token and the $\langle$BOS$\rangle$ token.) At training time, the model needs to learn to predict the length at the first step the same way as other words (no extra length predictor needed). At test time, the length token is sampled in the same way as other words if no desired length is specified.




\section{Experiments and results}

\noindent\textbf{Dataset and evaluation.}
We use COCO-Captions~\cite{chen2015microsoft} to evaluate our models. For train/val/test split we use Karpathy split. All the results are reported on the test set.

For evaluation, we use BLEU, METEOR, ROUGE, CIDEr, SPICE and bad ending rate~\cite{guo2018improving,self-critical-code}. Bad ending rate is calculated by counting the proportion of sentences that ends with a word in the ``bad ending'' set\footnote{\{a, an, the, in, for, at, of, with, before, after, on, upon, near, to, is, are, am\}}. We report bad ending rate to see if the caption of designated length is generated by sacrificing the fluency or not (an extreme case to generate a caption of the desired length is to just cut off at the desired length). To evaluate the controllability, we also report LenMSE, the mean square error between the desired length and the actual length.

When evaluating captions of different lengths, we also report mCIDEr, a modified CIDEr. The problem of the original CIDEr-D is that it promotes short and generic captions because it computes the average similarity between the generated and the references.
We modify it in two ways: 1) removing the length penalty term in the CIDEr-D; 2) combining the n-gram counts from all the reference captions to compute similarity~\cite{coco-caption-code}.

\noindent\textbf{Implementation details.}
The base captioning model is Att2in~\cite{rennie2017self}. The image features are bottom-up features~\cite{anderson2018bottom}. For the length prediction module in LenEmb, we use averaging region features as input.
We train the models with standard MLE. Unless specified
otherwise, decoding is beam search with beam size 5.






\begin{table}[htbp]
  \centering
    \begin{tabular}{lccccccc}
      \toprule
          & BLEU-4 & ROUGE & METEOR & CIDEr & SPICE & BER & LenMSE\\
    \midrule
    Att2in & 35.9 & 56.1  & 27.1  & 110.6 & 20.0  & 0.1 & N/A\\
    \midrule
    LenEmb & 34.9 & 56.2  & 27.0  & 110.0 & 20.0  & 0.0 & 0\\
    Marker & 35.2 & 56.2  & 26.9  & 109.8 & 19.9  & 0.0 & 0\\
    \bottomrule
    \end{tabular}%
  \caption{Performance on COCO Karpathy test set. BER is bad ending rate. Numbers are scaled by 100x except LenMSE.}
  \label{tab:length_pred_perf}%
\end{table}%

\subsection{Generation with predicted lengths}
For fair comparison to a non-controlled model on general image captioning task, we predict the length and generate the caption conditioned on the predicted length for length models. Results in Table \ref{tab:length_pred_perf} show that the length models are comparable to the base model.

\begin{figure}[t]
  \centering
  \begin{minipage}[b]{.6\linewidth}
  \includegraphics[width=\linewidth]{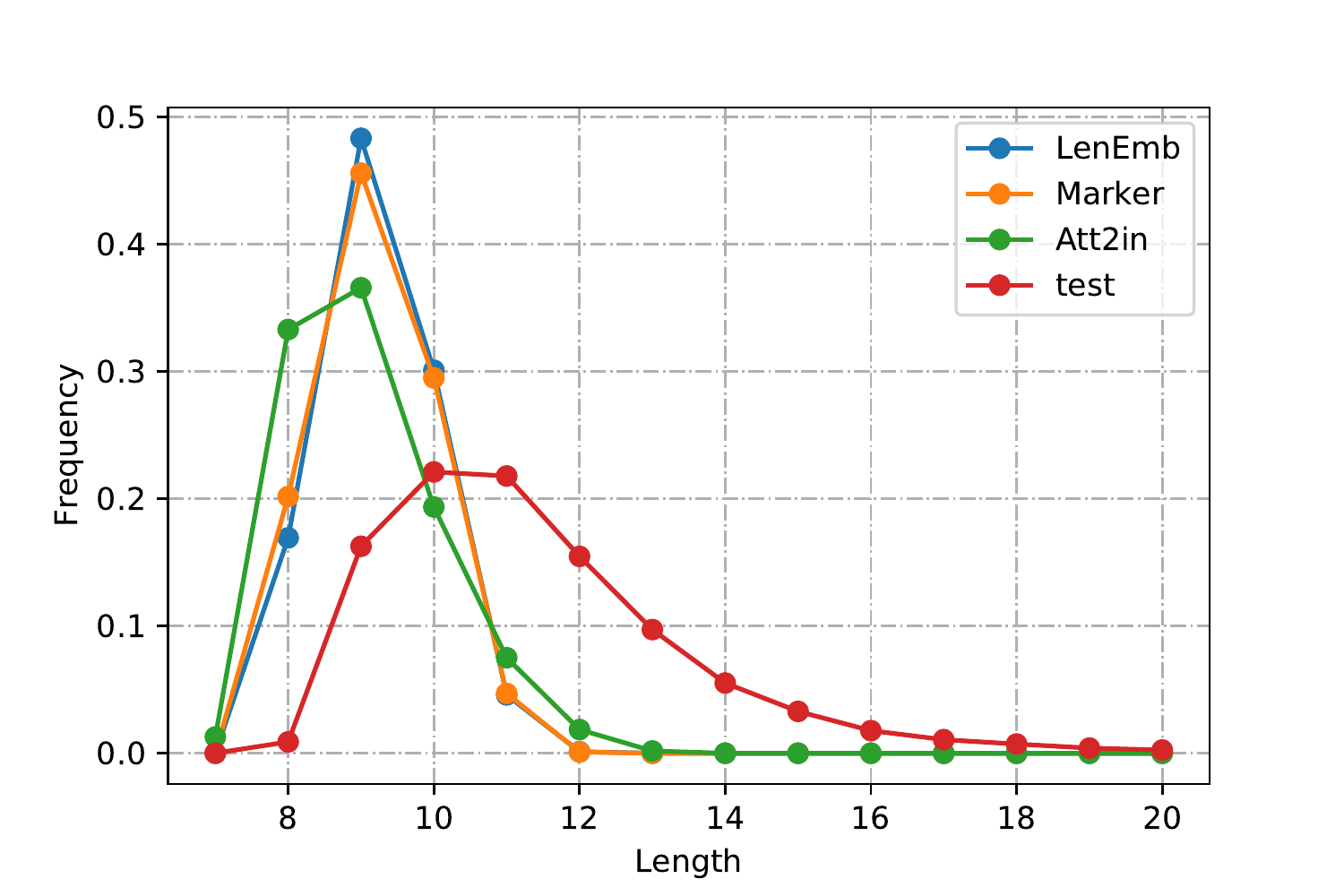}
  \end{minipage}
  \begin{minipage}[b]{.38\linewidth}
  \captionof{figure}{Length distribution of captions generated by different models. For length models, the length is obtained within the beam search process as a special token.}
  \label{fig:length_length_dist}
  \end{minipage}
  \end{figure}

\noindent\textbf{Length distribution (\figref{fig:length_length_dist}).}
While the scores are close, the length distribution is quite different. Length models tend to generate longer captions than normal auto-regressive models. However, neither is close to the real caption length distribution (``test'' in the figure).



\begin{figure}[t]
  \centering
  \begin{subfigure}{0.45\linewidth}
  \includegraphics[width=\linewidth]{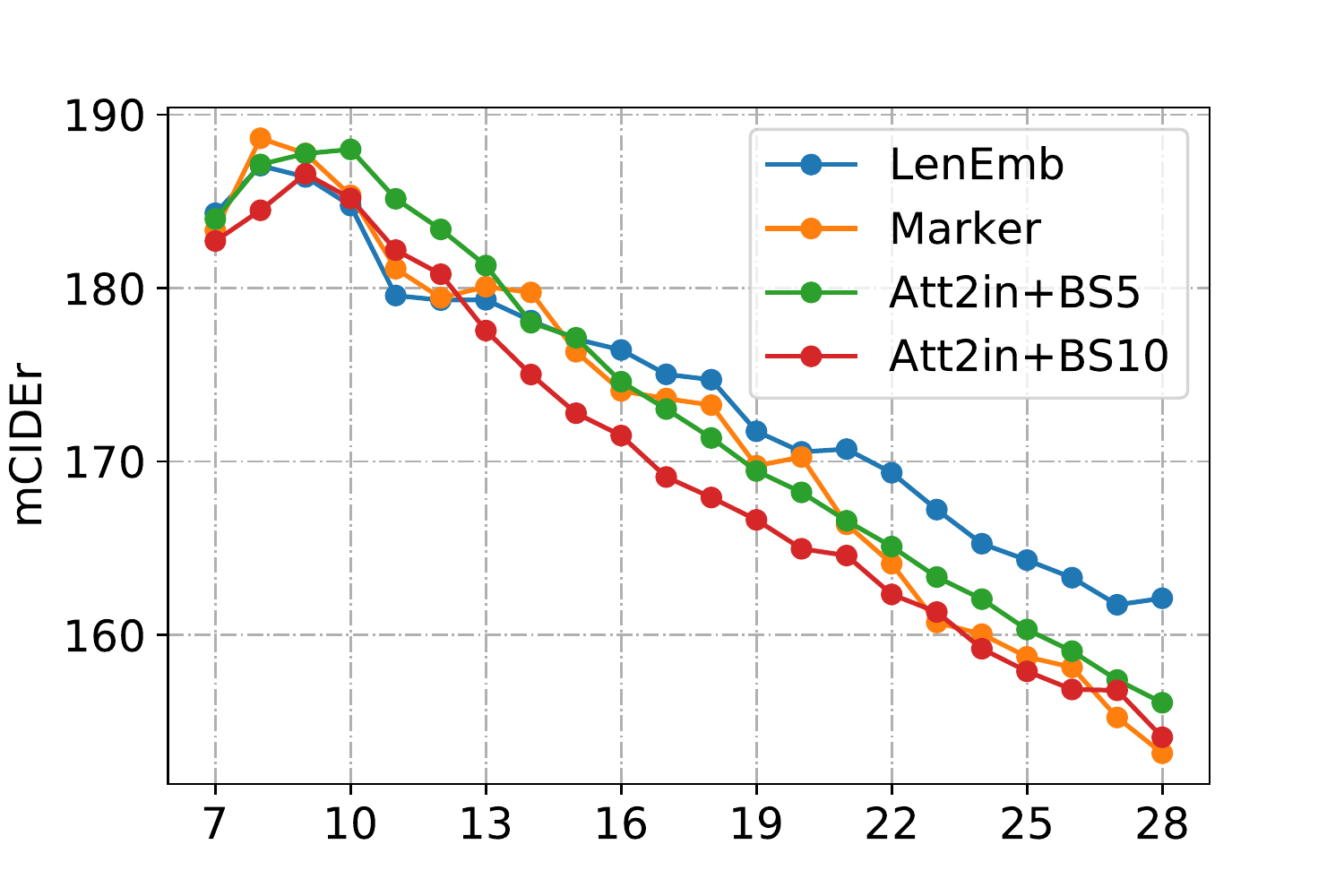}
  \caption{}\label{fig:length_cider}
  \end{subfigure}
  \begin{subfigure}{0.45\linewidth}
  \includegraphics[width=\linewidth]{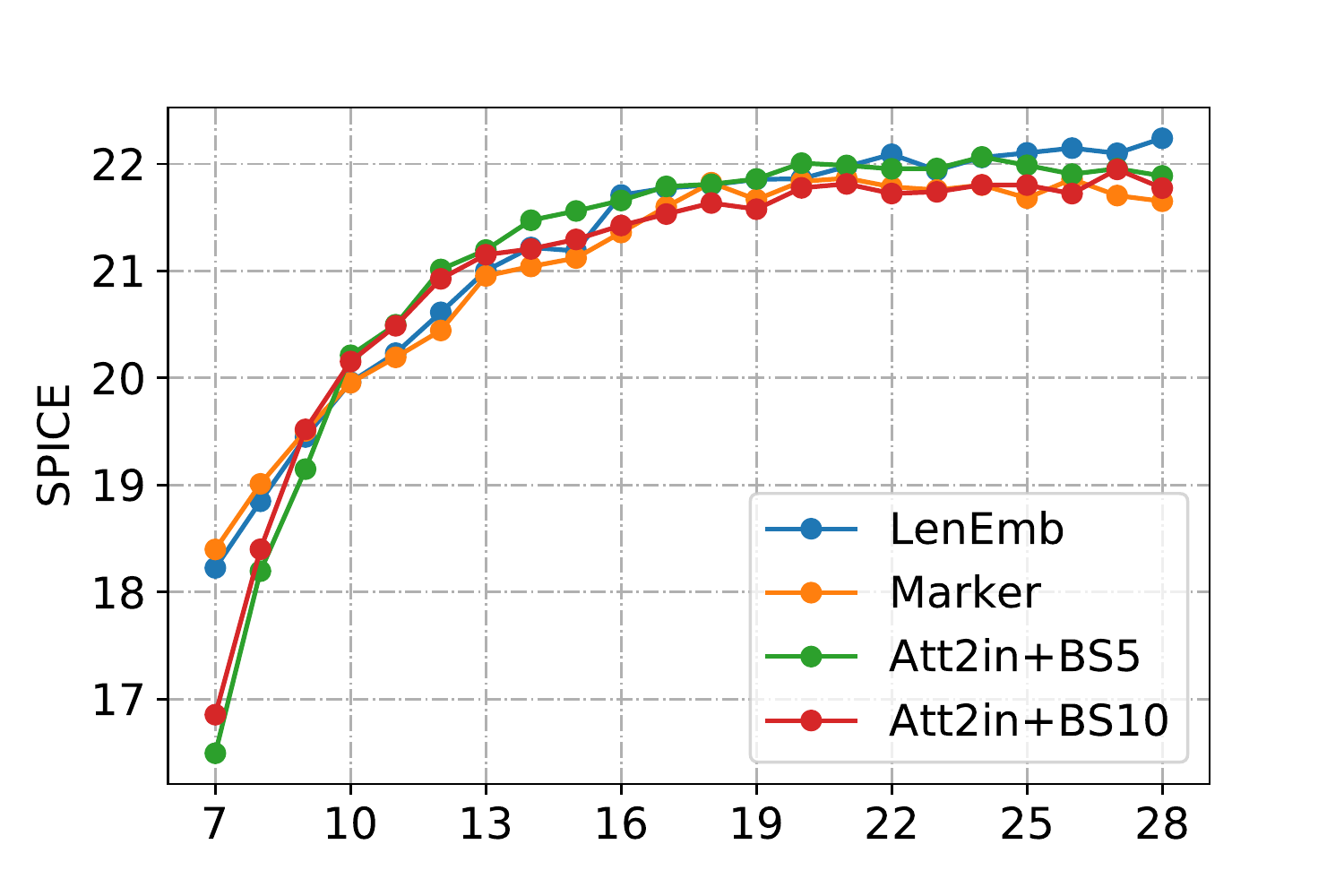}
  \caption{}\label{fig:length_spice}
  \end{subfigure}
  \begin{subfigure}{0.45\linewidth}
  \includegraphics[width=\linewidth]{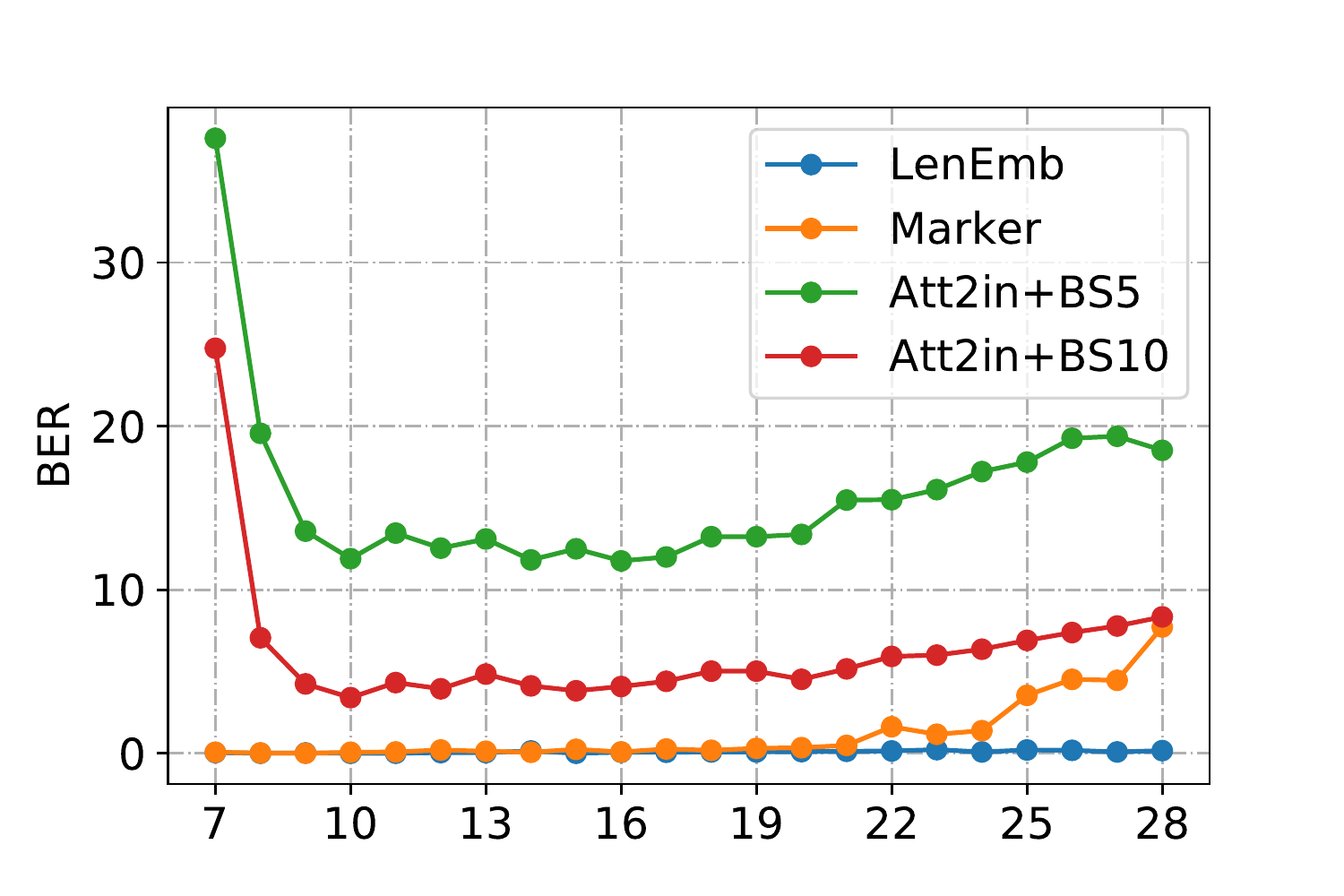}
  \caption{}\label{fig:length_ber}
  \end{subfigure}
  \begin{subfigure}{0.45\linewidth}
  \includegraphics[width=\linewidth]{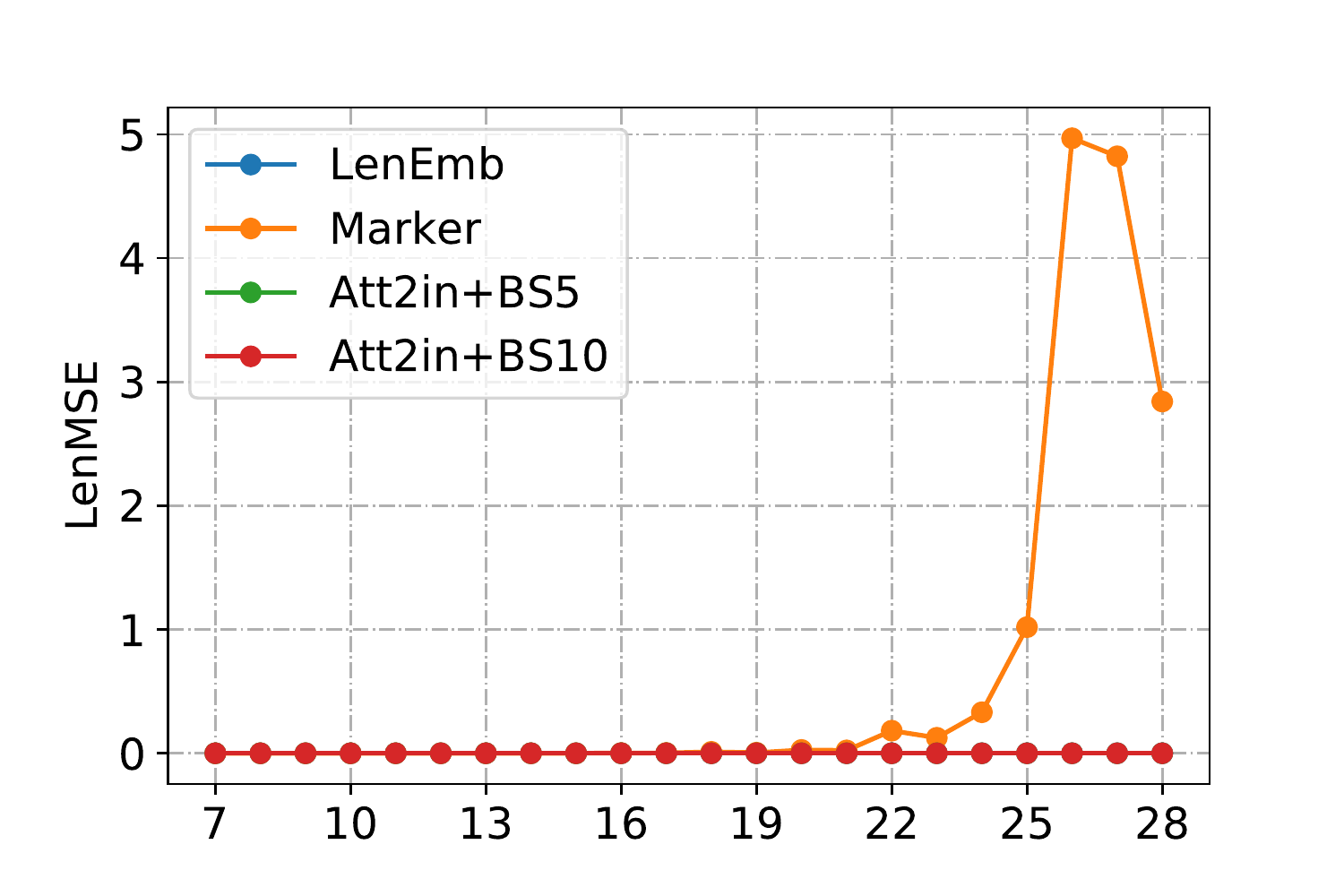}
  \caption{}\label{fig:length_lenmse}
  \end{subfigure}
  \caption{The performance of models with different desired lengths. Att2in+BSx is Att2in+fixLen with beam size x.}
  \label{fig:length_scores}
  \end{figure}
  
\subsection{Generation with controlled lengths}

For the baseline non-controlled model, we use the method fixLen in \cite{kikuchi2016controlling}. fixLen is a decoding method. During beam search, when the length of the beams exceeds the desired length, the last word is replaced with the $\langle$eos$\rangle$ token and also the score of the last word is replaced with the score of the $\langle$eos$\rangle$ tag.

\noindent\textbf{Fluency (\figref{fig:length_ber}).} The high bad ending rate for Att2in+fixLen indicates that it cannot generate fluent sentences.
When increasing beam size, the bad ending rate becomes lower, because it is more likely to have a beam that ends properly when having larger beams. For length models,
Marker performs well when the length is less than 20 but fails after, while LenEmb performs consistently well.

\noindent\textbf{Accuracy (\figref{fig:length_cider}, \ref{fig:length_spice}).} The length models perform better than base model when the length is smaller than 10. The base model performs better between 10-16 which are the most common lengths in the dataset. For larger length, the LenEmb performs the best on both mCIDEr and SPICE, indicating it's covering more information in the reference captions.


\noindent\textbf{Controllability.} When using predicted length, the length models perfectly achieve the predicted length (in Table \ref{tab:length_pred_perf}). When a desired length is fed, \figref{fig:length_lenmse} shows that LenEmb can perfectly obey the length while Marker fails for long captions probably due to poor long-term dependency.

\begin{figure}[!th]
{\footnotesize
  \centering
    \begin{tabular}{p{0.01\linewidth}p{0.95\linewidth}}
    \multicolumn{2}{c}{
\includegraphics[width=0.5\linewidth]{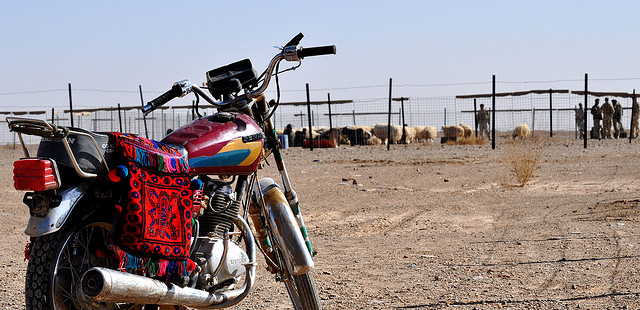}}\\ 
7 &  a motorcycle parked on a dirt road\\ 
10 &  a motorcycle is parked on the side of a road\\ 
16 &  a motorcycle parked on the side of a dirt road with a fence in the background\\ 
22 &  a motorcycle parked on the side of a dirt road in front of a fence with a group of sheep behind it\\ 
28 &  a motorcycle is parked in a dirt field with a lot of sheep on the side of the road in front of a fence on a sunny day\\ 
\hline
7 & a motorcycle parked on a dirt road\\ 
10 & a motorcycle parked on a dirt road near a fence\\ 
16 & a motorcycle parked on a dirt road in front of a group of people behind it\\ 
22 & a motorcycle parked on a dirt road in front of a group of people on a dirt road next to a fence\\ 
28 & a motorcycle parked on a dirt road in front of a group of people on a dirt road in front of a group of people in the background\\ 
    \end{tabular}
    \begin{tabular}{p{0.01\linewidth}p{0.95\linewidth}}
    \multicolumn{2}{c}{
\includegraphics[width=0.5\linewidth]{figures/length/000000348881.jpg}}\\ 
 7 &  an airplane is parked at an airport\\ 
10 &  an airplane is parked on the tarmac at an airport\\ 
16 &  an airplane is parked on a runway with a man standing on the side of it\\ 
22 &  an airplane is parked on a runway with a man standing on the side of it and a person in the background\\ 
28 &  an airplane is parked on the tarmac at an airport with a man standing on the side of the stairs and a man standing next to the plane\\
\hline
 7 & a plane is sitting on the tarmac\\ 
10 & a plane is sitting on the tarmac at an airport\\ 
16 & a plane that is sitting on the tarmac at an airport with people in the background\\ 
22 & a plane is sitting on the tarmac at an airport with people in the background and a man standing in the background\\ 
28 & a plane is sitting on the tarmac at an airport with people in the background and a man standing on the side of the road in the background\\
    \end{tabular}
}
    \caption{Generated captions of different lengths. Top: LenEmb; Bottom: Att2in+BS10}
    \label{fig:length_qualitative}
\end{figure}

\noindent\textbf{Qualitative results (\figref{fig:length_qualitative})} show that
the LenEmb model, when generating longer captions, changes the caption
structure and covers more detail, while the base model tends to have
the same prefix for different lengths and also repeats itself. More results can be browsed online~\cite{colab}.

\subsection{Failure in CIDEr optimization}
We apply SCST~\cite{rennie2017self} for length models. However, SCST doesn't work well. While the CIDEr scores can be improved, the generated captions tend to be less fluent, including bad endings (ending with ``with a'') or repetitions (like ``a a'').









\section{Conclusion}
To enable the captioning model to control its generated caption length, we propose two captioning models LenEmb and Marker based on \cite{kikuchi2016controlling,fan2017controllable}. We show that the models have the ability to follow the control and generate good captions of different lengths.

\chapter{Informative tagging}\label{chap:tagging}


\newcommand{\gal}[1]{\xxcom{red}{\{GalC: #1\}}}
\newcommand{\ake}[1]{\xxcom{blue}{[#1 -AE]}} 
\newcommand{\lior}[1]{\xxcom{orange}{[#1 -LB]}}

\section{Problem statement}
When people choose how to describe a visual scene, they effortlessly execute a computation that constitutes a critical foundation of language understanding and communication: not just being able to communicate information, but identifying \emph{which} information is important to convey in order to enable understanding. 
In describing individual objects,  people tend to be  surprisingly consistent in choosing the level of abstraction and words they use~\cite{rosch1976basic}. 

\begin{figure}[t]
    \begin{center}
    \includegraphics[width=1\linewidth, trim={0 0 0 0},clip]{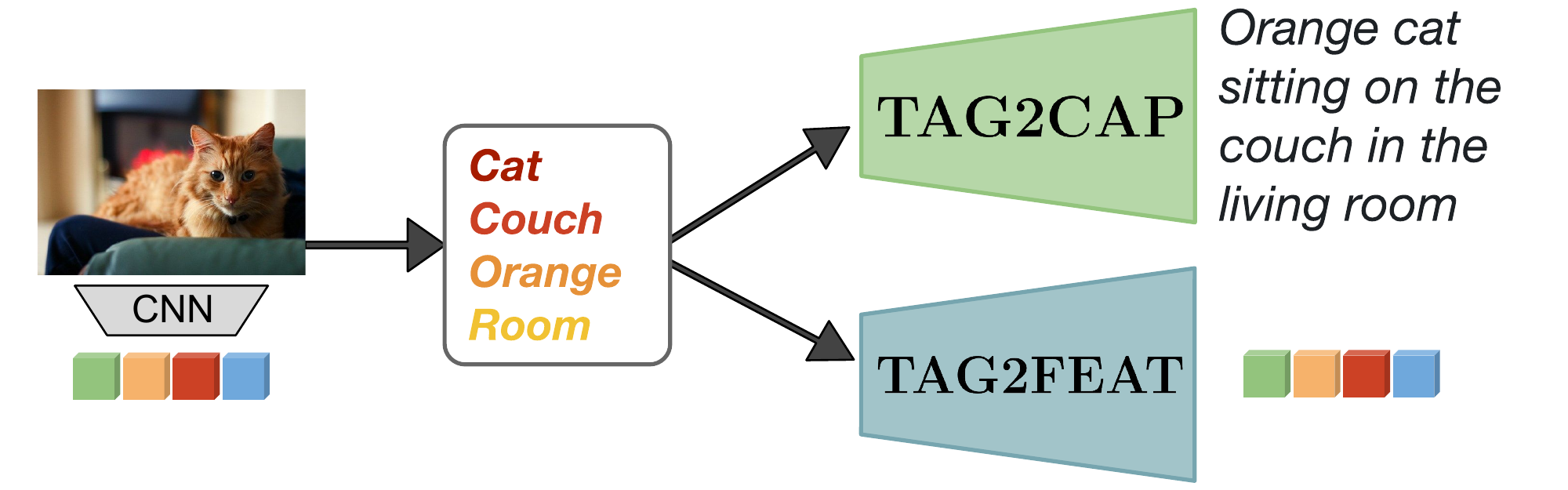}
    \end{center}
    \caption{We define information utility of a set of tags for an image as the ability (of a model) to reconstruct a ``mental picture'' of the image from the tags. In \tagcap this means reconstructing a complete caption; in \tagfeat this means reconstructing CNN-extracted image features. We use \tagcap and \tagfeat to evaluate how good a tag or a set of tags (e.g., $\{cat, couch, orange, room\}$) is at conveying information about the image.}
    \label{fig:tag2cap_tagging_full-pipeline}
\end{figure}

Unfortunately, it is
less clear what terms people may choose to describe a more complex visual scene. In this paper, we take a step towards modeling this phenomenon. We limit the scope to a set of tags, rather than fully formed statements, because it reduces the focus on syntactic fluency common in modern image captioning systems and focus on meaning, and allows us to control the evaluation better.
We consider the task of producing a set of tags that are not only correct, but also maximally informative in the sense that they best capture what the image is about. For example, in \figref{fig:tag2cap_tagging_full-pipeline}, it is intuitive to people that the most important tag for that image is \textit{cat}, but current tagging systems generally do not make this computation and can emit the tag \textit{couch} with equal probability.  

%
%

Quantifying which tags are informative and relevant is a challenging problem. While there are many approaches, we build on the idea of effective communication \cite{grice1975logic,wilson2002relevance} and evaluate tags by their value to a prospective recipient. In our context, the tags provide a summary or a ``gloss'' of what the image is about. Given the tags, the recipient can reconstruct a ``mental picture'' of the image from the gloss, using their model of the visual world and of the language.
The ability of the recipient to reconstruct an accurate mental picture from the tags is what we define as \textbf{information utility} of the tags.

\figref{fig:tag2cap_tagging_full-pipeline} illustrates this idea. A set of tags for an input image is fed to a reconstruction machine (namely, a listener)  that attempts to recover information about the image based on a set of tags (without seeing the image). We consider two objectives for such a recovery. The first, \tagcap, is to reconstruct a sentence describing the image. The sentence may or may not include the tags from which it is reconstructed. The second objective is to reconstruct high-level visual features extracted from the image by a trained convolutional network. We call this objective \tagfeat. 

To train models that predict useful tags, we leverage existing datasets of images and their captions. For evaluation, we collect a new dataset of image tags focused on the notion of tag importance and show that utility-maximization objectives produce tag sequences that are aligned with human judgment. 

Using visual tags allows us to focus on meaning and not consider other aspects of the language like syntax, style, etc. However, the word order and dependency will be broken. For example, when the tags are ``person, dog, run,'' it is unclear what the subject of ``run'' is. Nevertheless, such informative tags are helpful when combining with tag-conditioned captioning methods like \cite{raghavi2020weakly} to generate more informative captions.

 
Our main contributions are as follows (i) we propose a novel approach to study tag importance by measuring its \textit{information-utility}, and design training objectives aimed at producing tags that are not only correct, but also useful. (ii) we provide a new evaluation dataset to study human judgments of importance. (iii) Finally, we put forward two models \tagcap and \tagfeat that are optimized on selecting informative tags. We show that our method outperforms multiple baselines, for both tag-ranking and tag-generation.

\section{Related work}

\noindent\textbf{Image tagging.}
The simplest and most common image tagging method is to use one separate classifier for each category, treating each category independently~\cite{wei2014cnn,gong2013deep,mahajan2018exploring}. Some work exploits the correlations between categories by modeling the labels space \cite{yeh2017learning,lin2014multi,ghamrawi2005collective}.

Similar to our tag-generation model, there has been work that treats multi-label classification as a sequence generation task.
\cite{jin2016annotation} investigated generating image tags in sequence, learned with predefined orders.
\cite{vinyals2015order} learned the order of generating the set element by searching over possible orders during training.
\cite{yang2019deep} used policy gradient method to learn label sequence generation by directly optimizing the F1 score.
\cite{salvador2019inverse} learned intrinsic label ordering during training by applying a category-wise max-pooling across the time dimension.
\cite{yazici2020orderless} proposed to dynamically order the ground truth labels with the predicted label
sequence of the model at each iteration, with the help of Hungarian algorithm for alignment.

%
We refer the audience to \cite{pineda2019elucidating}, which surveys image-to-set frameworks, including both non-autoregressive and autoregressive set prediction.


\noindent\textbf{Gloss to sentence.}
In long-text generation, many coarse-to-fine generation strategies have been proposed
\cite{fan2018hierarchical,xu2018skeleton,yao2019plan,mao2019improving,tan2020progressive}. Usually, a list of keywords or a short prompt is first generated, serving as a summary of the original text. The prompt is then fed to some model as an input to output the full text.

While our \tagcap shares a similar input output format, we differ in two ways. First, the input to \tagcap is an unordered tag set instead of a sequence. Secondly, our goal of learning \tagcap is to use the output(sentences/captions) to select better input (tags), instead of generating better output.
Nevertheless, our tag-ranking method can select the most informative tags and certainly be combined with methods like \cite{raghavi2020weakly} to generate more informative captions.

\section{Importance-driven tag data}
To gain insight into human judgment of importance, we collect a new dataset of human-produced image tags, where importance is front and center. We emphasize that we use this dataset only for evaluation of models (including model selection) and not for training any models. 

\begin{figure*}[ht]
    \begin{minipage}[c]{.22\textwidth}
        \caption{People are asked what is the most important thing in the image, in three rounds. Each round provides a set of labels (red) as context of what is already known. We do this for 2k images and collect 20 sets  per image.}\label{table:tag2cap_tagging_rounds}
    \end{minipage}%
    \hfill 
    \begin{minipage}[c]{120mm}
        \scalebox{0.8}{ 
            \begin{tabular}{l|l|ll|ll}
                &\textbf{Round 1} & \multicolumn{2}{c}{\textbf{Round 2}} & \multicolumn{2}{c}{\textbf{Round 3}} \\\hline
                &  & \textit{\textbf{\textcolor{red}{cat}}} & \textit{\textbf{\textcolor{red}{chair}}} & \textit{\textbf{ \textcolor{red}{cat, chair}}}     & \textit{\textbf{ \textcolor{red}{cat, lazy} }}     \\ \hline
                \multirow{10}{*}{\includegraphics[width=1in]{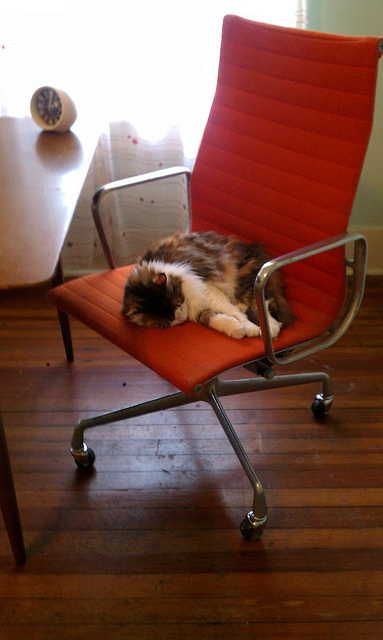}} & Cat &  chair  & Cat    & wheels  & chair    \\
                & Cat  &  chair    & cat  &   Lazy &  chair  \\
                & cat chair &  chair  &   cat   &  red & chair \\
                & cat lounging in chair  &  chair &  cat     & table & chair \\
                & Cat   & Chair  & cat  & wheel  & nap \\
                & cat on a chair & cat & Sleeping stage & chair & chair \\
                & Cat & chair &  cat              &  sleep  & chair\\
                & cat on chair      & chair      & cat & room & chair \\
                & cat & Lazy & cat  & floor  & chair  \\
                & Cat sleeping & rolling    & cat & Table & chair
            \end{tabular}
        }
    \end{minipage}
\end{figure*}

\subsection{Data collection}
We first randomly select 2K images from 5K COCO-Captions~\cite{chen2015microsoft} Karpathy test split~\cite{karpathy2015deep}.
We then present images to human annotators using Amazon Mechanical Turk. The annotators are given the following prompt: ``Please describe, in a single word or a short phrase, the most important thing about this image.''
Note that we do not restrict the annotators' vocabulary nor provide any guiding examples; we avoid these steps to minimize biasing the annotations. In this way we collect ten annotations per image. This initial round gives us an estimated distribution of human judgment on the importance of image tags.

Next, we collect a second round of annotations, in which the annotators are presented in addition to the image with a tag, and an additional instruction to avoid using that tag in their annotations. For each image, we use the two most common tags for that image in the initial round. In this way, we collect twenty annotations per image (ten for each of the two ``given'' tags).

\begin{figure}
    \centering
\includegraphics[width=0.45\linewidth]{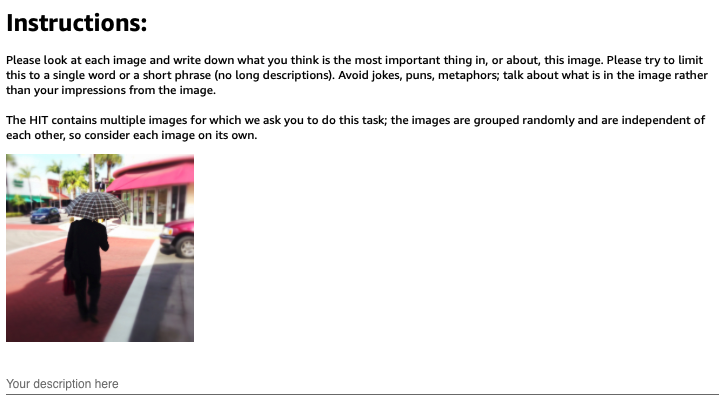}
\hspace{0.5in}
\includegraphics[width=0.45\linewidth]{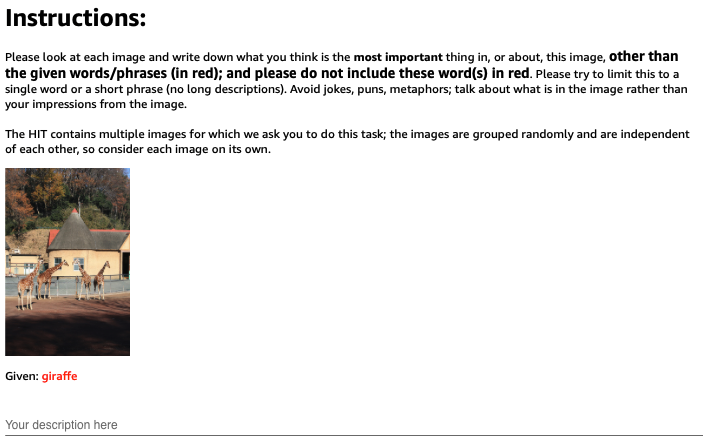}
    \caption{Amazon Mechanical Turk interface for data collection. The left is the interface for the first round. The right is the interface for the second and third round.}
    \label{fig:tagging_mechanical_turker_interface}
\end{figure}

Finally, we collect the third round, in which we present a set of two tags that the annotators must avoid. These are chosen as the two most common \emph{sets} of two tags from the first two rounds. Again, this yields twenty annotations (ten per set of two tags). We do not strictly enforce the ``single word'' rule, or (in the second and third rounds) the ban on repeating given tags. However, we find that the annotators tend to abide by these rules. Less than 1\% repeat the given tag, and 80\% of the raters describe the image with just one word. \figref{table:tag2cap_tagging_rounds} illustrates the data collection process. 
In \figref{fig:tagging_mechanical_turker_interface} shows the Amazon Mechanical Turk interface for the first and second/third round.

\subsection{Data processing}
\label{sec:tag2cap_data_processing}
We remove punctuation and stop words from the raw tags. Then, tags are lemmatized, and the un-lemmatized words as well as synonyms are mapped to a base form. For example,  \textit{bicycle} and \textit{bikes} are mapped to the base form \textit{bike}. Noun-noun pairs are processed as a single phrase (e.g., \textit{traffic light}). We then rank the tags by its frequency in the answers. 

\begin{figure}[!htp]
    \begin{center} 
    \includegraphics[width=1\linewidth, trim={0 0 0 0.3in},clip]{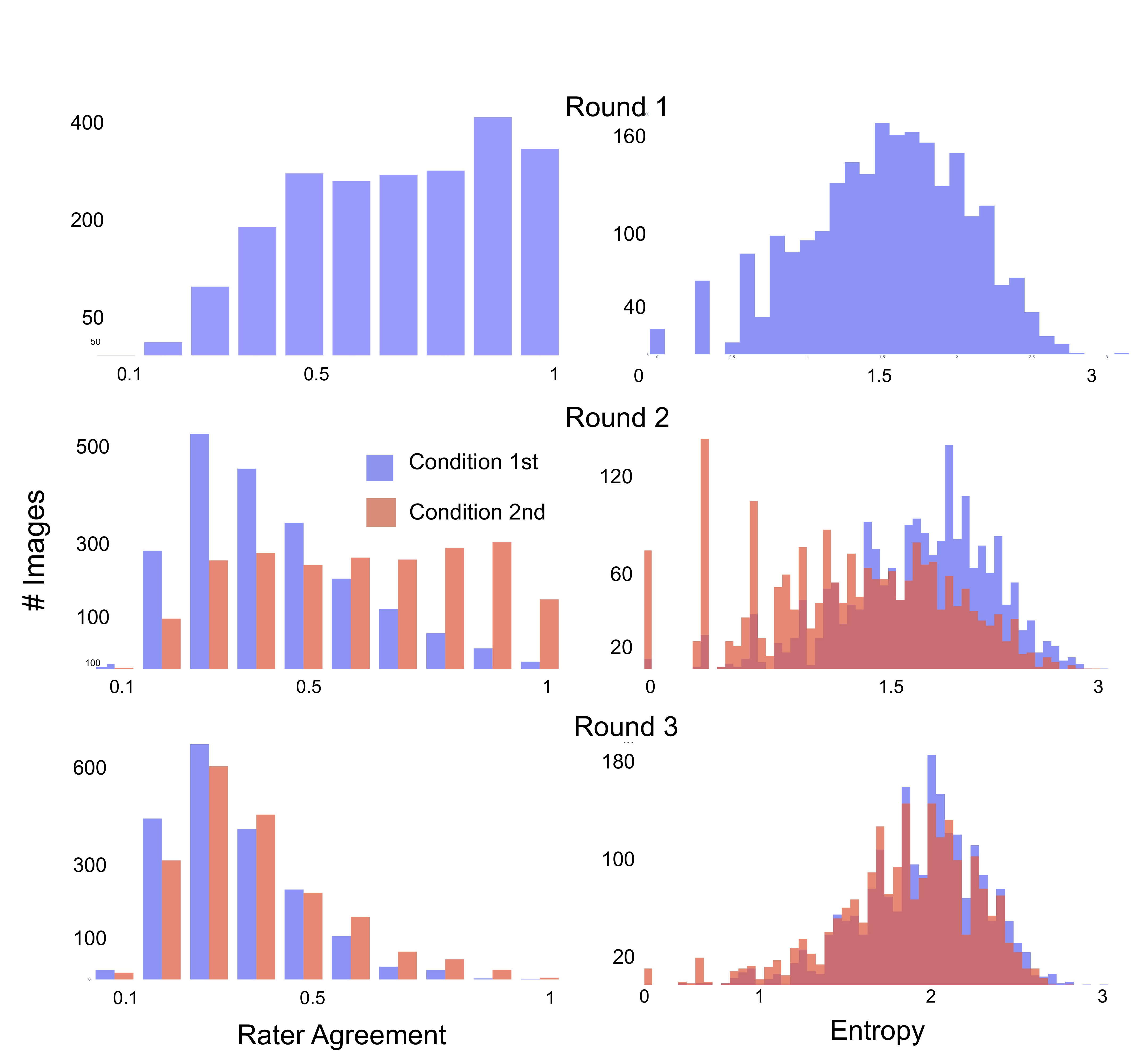}
    \end{center}\vspace{-1em}
    \caption{
    Distribution of rater agreement (left) and entropy (right) of tag distribution of tags provided by human raters in three rounds. \textbf{Top: } In round 1, the mean rater agreement is 70\%. \textbf{Middle:} Agreement increases when we condition on the second top tag (red compared to blue). It indicates that there is a consensus about what is the most important thing in the image. \textbf{Bottom: } In the third round, rater agreement is relatively low, around 37\%. }\label{fig:tag2cap_agreement}
\end{figure}

\subsection{Data properties}\label{sec:tag2cap_data_stats}
For the collected 2000 images, on average 7 tags are extracted per image. The first round yielded 3121 unique tags. When conditioning on the top answer, we get 2609 in the second and 3021 in the third. When we condition on the second-top answer, we get 2256 tags in the second round and 2993 in the third. 

\noindent\textbf{Raters agreement.} In the first round, the average raters agreement was 70\%. In the second round, the agreement is 43\% given the top tag, and 62\% given the second top tag. This is expected. An example of this effect is in \figref{table:tag2cap_tagging_rounds} where in round 2, when the given tag is  \textit{chair}, there is a broad consensus that the most important thing in the image is \textit{cat}. In the third round, the average agreement is low and is around 37\% for both conditions. 

\figref{fig:tag2cap_agreement} depicts the agreement between raters (left column) and the entropy of the distribution of answers (right column). The maximum entropy is $\log(8)$ in congruence with having seven tags on average per image.

\section{Method}

The key to our method is the task of ``mental picture reconstruction'' from image tags, as depicted in \figref{fig:tag2cap_tagging_full-pipeline}.
We do not consider reconstructing the actual image pixels from the tags, because we conjecture that not all the pixel values are important for a human interlocutor. Instead, we use the \tagcap and \tagfeat models to measure the information contained in the tags about two proxies for the image: a natural language image description, or caption (\tagcap) and a feature representation extracted from the image by a pre-trained neural network (\tagfeat). Our ultimate goal is to use these models to appraise the information utility of potential tags, rather than perform the reconstruction tasks per se.

We start by describing two distinct tasks: ranking tags, and generating tags. Then we introduce our methods for modeling these tasks; additional baselines are described in \secref{sec:tag2cap_tagging_exp}.

\subsection{Tag-ranking and generation}

\noindent\textbf{Tag-ranking.} Given an image and a set of tags  associated with that image, the task is to rank the tags according to the importance to the image. 
Formally, we are given $n$ candidate tags $T=\{t_1,\ldots, t_n\}$ for an image $\I$,
and our goal is to rank the tags that are most useful for reconstructing the image caption (or feature) in a greedy fashion. We iteratively choose a tag $t_j$ that minimize the reconstruction error, given all previously selected tags:
%
\begin{align}
    &\widehat t_1 = {\argmin}_j L_{\text{rec}}(\I,\{t_j\}) \label{eq:tag2cap_tagging_greedy_t1}\\
    &\widehat t_k = {\argmin}_j L_{\text{rec}}(\I,\{t_j,\widehat{t}_1,\ldots,\widehat{t}_{k-1}\}),
    \label{eq:tag2cap_tagging_ranking}
\end{align}
for all $k \in \{2,\ldots,n\}$, where $L_{\text{rec}}$ is a reconstruction loss associated with a ranking model. In our experiments the set $T$ is assumed to contain tags relevant for the image.



\noindent\textbf{Tag-generation.} Given an image $\I$, the goal is to produce a set of tags that describe the image correctly \emph{and} have high information utility. This is a real ``image tagging'' task, combining correctness and importance. 

In this setting, we build a model that takes an image feature vector as input and outputs an ordered sequence of tags. Specifically, we follow \cite{jin2016annotation}, and use an LSTM to generate tags sequentially. The image feature vector, mapped through a projection layer to word semantic space, is fed as the first input token of the LSTM.

To train an image tag-generation module, we generate a sequence of \emph{pseudo tags} to be used as labels for all training images. For each image, we extract all potential tags from its caption(s), and then rank these tags as in tag-ranking. 

During evaluation, the generator trained with tags ranked by the \tagcap ranker, is named ``\tagcap generator'' (and similarly, for \tagfeat generator). During training we use the standard cross entropy loss (just as an image captioning model) at every time step. At test time, we use greedy decoding to generate the tags.


\subsection{The \tagcap module}
The \tagcap module is designed to measure the quality of caption (sentence) reconstruction from a set of tags (gloss). For a caption $\bc = (w_1,\ldots,w_n)$, the utility of a tag set $T$ is measured as the conditional probability $P(w_1,\ldots,w_n|T)$, learned as described below.

\noindent\textbf{Architecture.} The module has an encoder-decoder architecture, using a \textsc{set2seq} transformer.
%
Specifically, $\mathcal{E}$ is a transformer encoder with full attention, but no position embedding~\cite{lee2019set}. The \textit{sequence decoder} $\mathcal{D}$ is a standard transformer decoder with causal attention.

\noindent\textbf{Training.} For each sentence in the training set, we extract tags and randomly sample a subset of these tags $T_s \subseteq T$. The model then reconstructs the sentence by reducing the cross entropy loss. This sampling procedure aims to reconstruct the caption from an arbitrary number of tags. 

\begin{figure}[ht]
    \begin{center}
    \includegraphics[width=1\linewidth, trim={0 0 0 0},clip]{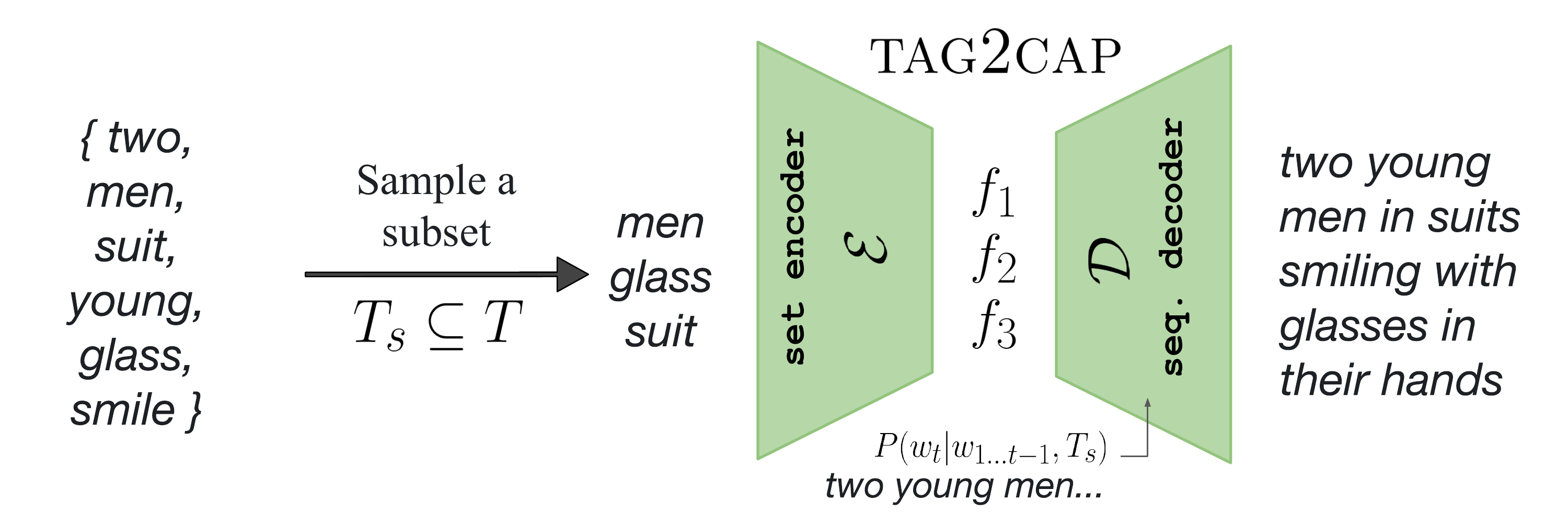}
    \end{center}
    \caption{\tagcap is a set2seq transformer that reconstructs a caption based on a set of tags. During training, we randomly sample a subset of the tags $T_s \subseteq T$. $\mathcal{E}$ encodes $T_s$ into a set of tag features $f_i$, Then the decoder $\mathcal{D}$ predicts the probability of next word based on these features. The training objective is to reconstruct the sentence by reducing the cross entropy loss.}
    \label{fig:tag2cap_tagging_tag2cap}
\end{figure}

In \figref{fig:tag2cap_tagging_tag2cap}, given a caption \textit{``Two young men in suits smiling with glasses in their hands,''} we extract \{\textit{two, man, suit, young, glass, smile}\}; we then sample a subset of arbitrary length such that \{\textit{young}\}, \{\textit{men, glass, suit}\} or \{\textit{young, suit, smile, glass}\} can be used to reconstruct the caption.  

Specifically, $\mathcal{E}$ encodes the tags $T_s$ into a set of features $f_{1\ldots k}=\mathcal{E}(T_s), k = |T_s|$, with self attention. The decoder $\mathcal{D}$ then takes the tag features $f_i$ and previous words of the caption and predicts the probability of next word, $P(w_t|w_{1\ldots t-1}, T_s) = \mathcal{D}(w_t|w_{1\ldots t-1}, \{f_i\})$.

Given an image $\I$ with caption $\bc$, we define the utility of a set of tags $\tilde T$ as the negative log probability of caption $\bc$ given $\tilde T$:
\begin{align}
    L^\tagc_{\text{rec}}(\I,\tilde T) &= -\log P(\bc|\tilde T) \\
    &=\sum_t \log P(w_t|w_{1\ldots t-1}, \tilde T)
\end{align}
The training loss is thus the expectation of the reconstruction error on the training set:
\begin{align}
    L^\tagc = \bbE_{T, \I \sim D}\bbE_{T_s \subseteq T} L^\tagc_{\text{rec}}(\I,T_s)  \quad.\label{eq:tag2cap_tagging_tagcap-train}
\end{align}

Once trained, $L_{\text{rec}}^{\tagc}$ can be used in ranking tags in \eqref{eq:tag2cap_tagging_ranking} to greedily construct a tag sequence that maximizes utility. We note that since in \eqref{eq:tag2cap_tagging_tagcap-train} we only need access to captions, and not actual images, we can train \tagcap on a language-only corpus, as a general gloss-to-sentence model. 

\subsection{\tagfeat}
Similar to \tagcap, the \tagfeat module is designed to reconstruct image features given a subset of tags. Once trained, it provides a utility score for any set of tags.

\noindent\textbf{Architecture.} We use a similar set encoder to encode the tag set. The output of the first token (a $	\langle$bos$\rangle$ token) is treated as the set feature and projected to the dimension of visual features $\widehat \bv = F(\mathcal{E}_{\tagf}(T))$, where $\mathcal{E}_\tagf$ is the set encoder and $F$ is the text-to-visual projection layer. We use ResNet-50~\cite{he2016deep} to encode all the images and treat them as the ground truth visual feature when training \tagfeat. $\bv = RN_{50}(\I), \bv\in \mathbb R ^{2048}$, where $\I$ is the image.

\noindent\textbf{Training.} We follow the same training strategy as in \tagcap. For each sample, we randomly select a subset of tags $T_s$ and feed them through the set transformer and projection.

We define reconstruction error as the cosine distance between the estimated image feature vector and the ground truth image feature vector:
\begin{align}
    L^\tagf_{\text{rec}}(\I,\tilde T) =  1- \text{cos}[F(\mathcal{E}_{\tagf}(\tilde T)), \text{RN}_{50}(\I)]
\end{align}
yielding the training loss:
\begin{align}
    L^\tagf \ = \bbE_{T, \I \sim D}\bbE_{T_s \subseteq T} L^\tagf_{\text{rec}}(\I,T_s) \quad.
\end{align}

Unlike \tagcap, we have to have an image-tag dataset to train \tagfeat, because the training involves image features. 


\section{Experiments and results}\label{sec:tag2cap_tagging_exp}
Our experiments are designed to evaluate how an automatic method for tag-ranking or generation matches human importance-driven tagging in our dataset. 

\subsection{Experimental settings}
We use a random subset of 1000 images in our dataset as a validation set for debugging, parameter tuning, and model selection. The remaining 1000 images are used as a held-out test set. We report results on the latter. 

\noindent\textbf{5-caption and 1-caption regimes.} Images in the  COCO-Captions are annotated with five captions per image, but 
other datasets such as Conceptual Captions~\cite{sharma2018conceptual} and SBU-Captions~\cite{ordonez2011im2text} only provide a single caption. Thus we consider two regimes. In the 5-caption regime, we assume that each training image and each test image is associated with five captions, and use these to extract candidate tags for ranking (in both training and evaluation). In the 1-caption regime, while still using COCO images, we simulate single caption by replicating each image five times, one per distinct caption.


%
%
%
%
%

\noindent\textbf{Evaluation metric.}
First, let SetRecall($A,B$) be the measure of coverage of elements in set $B$ by set $A$: $\text{SetRecall}(A,B)=|A\cap B|/|B|$.
Now, consider the predicted sequence of tags $\widehat{t}_1,\ldots$ and the ground truth (human reference) sequence $t_1,\ldots,t_m$. Suppose we only output $k$ predicted tags, $\widehat{t}_1,\ldots,\widehat{t}_k$. We measure its accuracy w.r.t. the reference tag sequence by 
\begin{align}
    \text{SetRecall@k} = \text{SetRecall}\bigl(&\{\widehat t_{1\ldots k}\},\nonumber\\ &\{t_{1\ldots \min[k,m]}\}\bigr)
\end{align}
%
%
%
For instance, for a ground truth sequence [{\it cat}, {\it bed}, {\it pillow}], and a predicted sequence [{\it bed}, {\it cat}, {\it sleep},  {\it wall}, {\it pillow}], we have SetRecall@k=[0, 1, .66, .66, 1].


\noindent\textbf{Ground truth tag sequence(s).}
In our dataset there is no clear single ground truth tag sequence for each image, due to multiple potentially different annotations. Our intuition is that any tag provided by at least one annotator should be considered important, but we would like to prioritize (by a higher weight in the metric) those that are more important (judging by prevalence in annotations). Thus, we propose a new metric, importance-weighted SetRecall@k that incorporates these ideas.

Let $M_1(t)$ be the count of tag $t$ appearing in the 1st round annotations for an image. For the second round, $M_2(t|t_1)$ is the count of $t$ appearing in annotations conditioned on $t_1$. Similarly, for the third round, $M_3(t|\{t_1, t_2\})$ is the count of $t$ appearing in annotations where the conditioning is on the \emph{set} $\{t_1,t_2\}$.

We define the weight of a tag sequence $\{t_1,t_2,t_3\}$ for an image as
\begin{align*}
    W(t_1,t_2,t_3) \propto M_1(t_1)\cdot M_2(t_2|t_1)\cdot 
    (M_3(t|\{t_1,t_2\}))
\end{align*}
normalized to sum 1 across all possible tag sequences for that image.

Finally, the weighted SetRecall@k for an image with predicted tag sequence $\widehat{\bf t}$ is:
\begin{align}
    &\text{SetRecall@k}(\widehat {\bf t}) = \nonumber\\
    &\sum_{t_1,t_2,t_3} \text{SetRecall@k}(\widehat {\bf t}, \{t_1,t_2,t_3\}) \cdot W(t_1,t_2,t_3)
\end{align}
%
%
%
%
%
%
%
%
%
%
%
%
\subsection{Implementation details}
We initialize our \tagcap with BART~\cite{lewis2019bart}, a pretrained transformer seq2seq model. We remove the positional embedding in the encoder to make it a set encoder. For better utilizing the pretrained model and for easier cross-dataset generalization, we use the pretrained tokenizer of BART to tokenize both the tags and captions/sentences.

For \tagfeat, we train it from scratch with a BERT-like architecture, removing the positional embedding in the encoder. We found using pretrained BERT hurts the performance.

For image tagging model, the word embedding and hidden size of LSTM is set to 512.

\noindent\textbf{Tag extraction during training.} Similar to human result processing, we first process the raw captions. Then we create an allowed tag vocabulary of 1000 most frequent non-stop words. For each image, tag set $T(\I)$ consists of all the tokens in the image captions that are also in the tag vocabulary.  


\subsection{Baselines}
\noindent\textbf{Tag-ranking baselines.}
We consider a number of baselines that rank tags given the ground truth caption(s) $\bc(\I)$ for the image at hand.
\begin{itemize}[noitemsep,topsep=0pt,leftmargin=*]
    \item \textbf{Term Frequency (TF):} Tags are ranked by the descending order of frequency in $\bc(\I)$. 
    \item \textbf{TFIDF:} TF multiplied by log of inverse normalized frequency of a tag across all captions in the dataset. 
    \item \textbf{TagOrder:} Rank the tags according to their order appeared in $\bc(\I)$. 
\end{itemize}
Note that TF is only meaningful in the 5-caption regime. In 1-caption regime, TFIDF degenerates to ranking based only on inverse caption frequency. TagOrder is only well-defined in 1-caption regime. 

We also consider the following baselines that do not use ground truth captions for ranking.
\begin{itemize}[noitemsep,topsep=0pt,leftmargin=*]
    \item \textbf{Frequency (Freq):} Rank the tags according to their frequency in the training set.
    \item \textbf{\tagtag:} Rank tags by how well they can predict the other tags, using a multi-label classifier based on tag set encoder. The classifier is trained on COCO training set. 
\end{itemize}

~\newline
\noindent\textbf{Tag-generation baselines.}
For any ranking method X, we can train a generator using pseudo-tags from X. We refer to the resulting generator as ``X'' in reporting the results. 
In addition we include a generation-specific baseline: PLA~\cite{yazici2020orderless}.
PLA is a training method for recurrent models to do multi-label classification. The ``ground truth'' order to generate is dynamically determined during training.

\begin{table}[!tbp]
    \centering
    \begin{tabular}{lccccc}
    \toprule
         SetRecall@k, k= & 1 & 2 & 3 & 4 & 5 \\ 
    \midrule
COCO & 30.7 & 33.8 & 30.3 & 33.9 & 37.6 \\
Conceptual   & 43.8 & 51.4 & 44.6 & 48.8 & 52.3 \\
WikiText & 43.0 & 52.5 & 46.3 & 51.3 & 54.1 \\
CNN  & \textbf{47.1} & \textbf{53.9} & \textbf{47.4} & \textbf{52.2} & \textbf{55.2} \\
    \bottomrule
    \end{tabular}
  \caption{\tagcap performance in tag-ranking (5-caption) is affected by the dataset used for pre-training. }\label{tab:tag2cap_tagging_dataset_ablation_ranking}%
\end{table}

\begin{figure}[!htp]
  \begin{minipage}[c]{0.49\linewidth}
  \includegraphics[width=\linewidth, trim={1in 1in 0in 1in},clip]{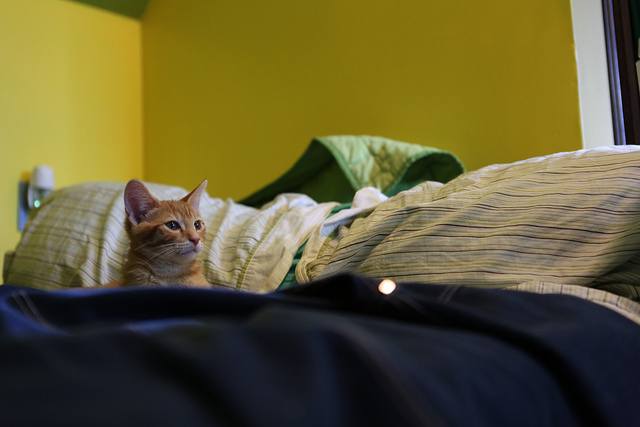}
  \end{minipage}
  \begin{minipage}[c]{0.49\linewidth}
  \textbf{\tagcap(COCO)} ranked tags: \\ \textbf{pillow}, lie, lay, bed, cat\\
  \textbf{Reconstructed caption from ``pillow''}: \\\quad a cat laying on top of a bed next to a window\\
  \textbf{\tagcap(CNN)} ranked tags: \\ \quad
  cat, bed, white, lie, orange\\
  \textbf{Turker:} \\ \quad cat, bed, pillow
  \end{minipage}
  \caption{An example of how \tagcap trained on COCO is biased. Due to the unusually high correlation between pillow and cat, ``pillow'' becomes the most informative tag and downgrades ``cat'' by a lot. The caption predicted by \tagcap(COCO) also reflects this bias. We also present the tags ranked by \tagcap(CNN) and the turker annotation for comparison.}
  \label{fig:tag2cap_tagging_coco_biased_tag2cap}
\end{figure}%

\subsection{Training \tagcap / \tagfeat}
We consider pre-training \tagcap on four corpora: two image captioning datasets:(1) COCO captions (Karpathy train split) (2) Conceptual Captions~\cite{sharma2018conceptual}; and two non-image-related text datasets, (3) CNN News~\cite{hermann2015teaching} and (4) WikiText-103~\cite{merity2016pointer}. Table~\ref{tab:tag2cap_tagging_dataset_ablation_ranking} shows the resulting performance of ranking (5-caption regime) by \tagcap models trained on different datasets. All models trained on out-of-domain datasets perform superior to that trained on COCO. Among those, CNN News leads to the best data. In the rest of the paper, we use the \tagcap model trained on CNN News. Specifically, we use a subset of CNN News, including 10K documents and 157K sentences, following \cite{tan2020progressive}. 

It is perhaps surprising that training \tagcap on a dataset ostensibly unrelated to image descriptions is better than training on COCO captions. We believe one reason is overfitting to COCO-specific biases when training on COCO, as illustrated in \figref{fig:tag2cap_tagging_coco_biased_tag2cap}. For \tagfeat, we train on COCO as it requires image features during training.%

\subsection{Results}

\begin{table}[!htp]
    \centering
    \begin{tabular}{lccccc}
    \toprule
         SetRecall@k, k= & 1 & 2 & 3 & 4 & 5 \\ 
    \midrule
Freq & 3.1  & 12.1 & 16.7 & 25.0   & 33.0   \\
\tagtag  & 19.1 & 21.4 & 20.6 & 23.4 & 26.0   \\
TFIDF  & 39.7 & 53.9 & 48.4 & 53.9 & 57.7 \\
TF      & 41.5 & 54.5 & 48.5 & 53.8 & 57.9 \\ 
\midrule
\tagcap & 47.1 & 53.9 & 47.4 & 52.2 & 55.2 \\
\tagfeat & 31.1 & 30.8 & 28.9 & 32.7 & 36.1 \\
\midrule
\midrule
\tagcap+\tagfeat & 48.3 & 53.6 & 46.8 & 51.8 & 55.4 \\
\tagcap+TF       & 45.0 & 55.8 & 49.6 & 54.3 & 57.7 \\
\tagcap+TFIDF   & 44.0 & 56.4 & 49.1 & 54.3 & 57.7 \\
\midrule
\midrule
\tagc+\tagf+TF       & \textbf{50.3} & \textbf{57.3} & \textbf{50.0}   & \textbf{54.6} & \textbf{58.2}  \\
\tagc+\tagf+TFIDF &47.0	&56.6	&49.4	&54.2	&57.8 \\
    \bottomrule
    \end{tabular}
  \caption{Tag-ranking (5-caption). We measure SetRecall@k of tag sequences produced by the different ranking models. The last two blocks show the performance of model ensembles.
  }
  \label{tab:tag2cap_tagging_ranking_5cap}%
\end{table}

\begin{table}[ht]
    \centering
    \begin{tabular}{lccccc}
    \toprule
         SetRecall@k, k= & 1 & 2 & 3 & 4 & 5 \\ 
    \midrule
Freq    & 3.1  & 12.1 & 16.7 & 25.0   & 33.0   \\
\tagtag     & 9.9  & 12.7 & 14.9 & 18.6 & 22.0   \\
TFIDF  & 12.5 & 24.7 & 30.6 & 39.8 & 45.1 \\
TagOrder         & 20.2 & 30.2 & 31.2 & 39.2 & 44.7 \\
\midrule
    \tagcap & 27.7 & 34.5 & 34.2 & \textbf{41.9} & \textbf{47.9} \\
\tagfeat    & 32.9 & 32.4 & 30.3 & 34.1 & 37.2 \\
\midrule
\midrule
\tagcap+\tagfeat & \textbf{39.3} & \textbf{40.8} & \textbf{36.5} & \textbf{41.9} & 45.6 \\
    \bottomrule
    \end{tabular}
  \caption{Tag-ranking (1-caption). For single models (top 2 blocks), \tagfeat outperforms all for k=1 and \tagcap is the best for $k>1$. Combining both leads to the best results. }
  \label{tab:tag2cap_tagging_ranking_1cap}%
\end{table}
First, we evaluate the ranking methods in 5-caption regime (Table~\ref{tab:tag2cap_tagging_ranking_5cap}) and 1-caption regime (Table~\ref{tab:tag2cap_tagging_ranking_1cap}). In the 5-caption regime, \tagcap is competitive with the TF-based baselines which effectively have access to importance via the proxy of term frequency (if something is mentioned in multiple captions, it is likely to be important). \tagfeat is dominated by both \tagcap and the TF-based baselines. Other baselines are less successful. Combining \tagcap with those baselines, however, produces further improvement, indicating that \tagcap provides a benefit even when having access to frequency as a proxy for importance. The best overall results are with a combination that includes \tagfeat, although on its own it does not produce particularly impressive rankings.

In the more general 1-caption ranking experiments, there is no longer an easy way to ``cheat'' by looking up term frequency.
TagOrder is competitive with \tagtag and TFIDF, reflecting the bias towards putting more important words (subjects) earlier in the captions.
\tagcap outperforms all baselines on all metrics, except that \tagfeat is better on SetRecall@1. Again best results are obtained by combining \tagcap and \tagfeat, improving a lot upon individual models.

\begin{table}[!ht]
    \centering
    \begin{tabular}{lccccc}
    \toprule
         SetRecall@k, k= & 1 & 2 & 3 & 4 & 5 \\ 
    \midrule
        Freq  & 2.5  & 10.2 & 13.1 & 20.2 & 27.2 \\
        TF    & 43.3 & 48.1 & 41.0 & 44.2 & 46.9 \\
        TFIDF & 44.5 & 47.7 & 41.3 & 44.9 & 47.5 \\
        PLA   & 15.7 & 22.5 & 25.0 & 36.8 & 43.8 \\
        \midrule
        \tagcap & 46.0 & 46.8 & 40.0 & 43.4 & 46.1 \\
        \tagfeat & 33.9 & 32.3 & 29.4 & 33.6 & 36.7 \\
        \midrule
        \midrule
        \tagcap+TF    & 46.0 & 49.1 & 41.3 & 45.2 & 47.9 \\
        \tagcap+TFIDF & 47.0 & 49.0 & 41.5 & 45.3 & 47.8 \\
        \tagcap+\tagfeat      & 46.6                 & 48.0                 & 40.7                 & 44.6                 & 47.1                 \\
        \midrule
        \midrule
        \tagc+\tagf+TF    & 46.4 & 48.8 & 41.8 & 45.4 & 47.6 \\
        \tagc+\tagf+TFIDF & \textbf{48.1} & \textbf{50.5} & \textbf{42.6} & \textbf{46.1} & \textbf{48.7} \\
\tagc+\tagf+TFIDF+PLA & 47.0&	50.0&	42.2&	45.7&	48.3 \\
    \bottomrule
    \end{tabular}
  \caption{Tag-generation (5-caption). Among single models, \tagcap outperforms other baselines at $k=1$ but performance decreases for larger k. For ensembles we find that a combination of \tagcap (\tagc), \tagfeat (\tagf) and TFIDF gives the best results.}
  \label{tab:tag2cap_tagging_generation_5cap}%
\end{table}

\begin{table}[!ht]
    \centering
    \begin{tabular}{lccccc}
    \toprule
         SetRecall@k, k= & 1 & 2 & 3 & 4 & 5 \\ 
    \midrule
Freq        & 16.4 & 32.4 & 34.4 & 40.8 & 43.9 \\
TagOrder    & 27.0 & 36.7 & 35.7 & 41.8 & 44.9 \\
TFIDF      & 26.3 & 36.3 & 36.3 & 41.4 & 43.5 \\
PLA         & 27.6 & 41.7 & 39.3 & \textbf{46.2} & 47.2 \\
\midrule
\tagcap     & 41.3 & 43.7 & 38.6 & 42.4 & 44.2 \\
\tagfeat    & 42.0 & 43.5 & 38.2 & 41.6 & 43.7 \\
\midrule
\midrule
\tagcap+PLA & \textbf{45.6} & 45.5 & 39.0 & 44.2 & \textbf{48.8} \\
\tagfeat+PLA         & 44.4 & 47.2 & 40.1 & 44.1 & 46.2 \\
\tagcap+\tagfeat     & 44.0 & 45.6 & 39.8 & 42.5 & 44.7 \\
\midrule
\midrule
\tagc+\tagf+PLA & 45.4 & \textbf{49.0} & \textbf{40.8} & 45.0 & 46.4 \\
    \bottomrule
    \end{tabular}
  \caption{Tag-generation (1-caption). Among single models, \tagcap and \tagfeat are better compared to the other baselines on SetRecall@1,2. An ensemble of \tagcap, \tagfeat, and PLA gives the best results.}
  \label{tab:tag2cap_tagging_generation_1cap}%
\end{table}

We next consider tag-generation. Table~\ref{tab:tag2cap_tagging_generation_5cap} shows that in the 5-caption regime, the effect of different ranking methods on the resulting tag sequence generation generally follows the standing in Table~\ref{tab:tag2cap_tagging_ranking_5cap}. The overall best performance for all $k$ is obtained by the combination including \tagcap, \tagfeat, and the TFIDF baseline, again showing a benefit in modeling information utility beyond observed frequency in captions, and despite the disappointing individual performance of the \tagfeat-trained generator.

Finally, we consider the arguably most realistic and general settings of generation in the 1-caption regime. Results are shown in Table \ref{tab:tag2cap_tagging_generation_1cap}. The PLA baseline dominates other individual objectives for $k\ge 3$, and
\tagcap and \tagfeat perform significantly better for $k=1$ and slightly better for $k=2$.
The winning combinations for most values of $k$ are to combine \tagcap and/or \tagfeat with PLA. 

To summarize our experimental results,  \tagcap and \tagfeat are successful in predicting the tags that are human raters find important for an image. \tagcap achieves high SetRecall@1 under both 5-caption and 1-caption regimes. Integrating \tagfeat is helpful for predicting the top tag ($k=1$) under both regimes, and its benefit is larger when the number of captions is limited. The overall best result, is achieved when combining \tagcap and \tagfeat with methods that perform well with a large number of tags ($k\ge3$) like TF/TFIDF/PLA.


\section{Conclusion}

We tackle the problem of producing an informative gloss for an image, in the form of a sequence of tags, with the goal to optimize information utility of those tags. To this end, we collect a dataset of human-produced image tags driven by perceived importance. We propose an approach for capturing information utility of tags based on the capacity of a model to recover information about the image from these tags.  We show that including this approach in the training objective for image tag generation makes the generated image annotations more faithful to human judgments of tag importance, in a variety of settings.

\chapter{Conclusion}\label{chap:finalconclusions}

In this thesis, we first considered to build a better referring expression generation system. In Chapter \ref{chap:ref_exp}, we aimed to optimize the end goal of referring expression directly: better at referring. To achieve this, we trained a referring expression comprehension model and proposed two ways to use the comprehension model to guide generating better referring expressions, either during training or at test time. We then extended this idea to the image captioning domain: increasing the discriminability of the generated captions. In Chapter \ref{chap:disc_cap}, we trained an image caption retrieval system. We used the retrieval loss to measure discriminability and trained the image captioning model with the negative retrieval loss as part of the reward.Chapter \ref{chap:diversity} studied the diversity accuracy tradeoff of different models trained with different objectives and decoded with different algorithms. We also proposed a novel metric, \spicen, for evaluating diverse image captioning. In Chapter \ref{chap:length}, we proposed two models that can take a designated length as input and output a caption of that length, to achieve different levels of detailedness and style. Finally, in Chapter \ref{chap:tagging}, we studied informative tagging. Instead of outputting everything in the image, our goal was to generate the tags that can retain the most information of the image. We were able to borrow pre-trained seq2seq models and pure language data to figure out what tags are semantically informative.

The idea of incorporating goals into text generation is not limited to what is presented in this thesis. In image captioning, there are many other goals to consider, for example, describing the images to better assist the blind. Beyond image captioning, goal driven generation can be used for other tasks such as visual question generation, vision language navigation instruction generation, dialogue systems, abstract summarization, etc. To obtain better AI systems that can serve humans, more people in the AI community should consider integrating goals into the system training and deployment.

\bibliography{ref_new}

\begin{thebibliography}{292}
\providecommand{\natexlab}[1]{#1}
\providecommand{\url}[1]{\texttt{#1}}
\expandafter\ifx\csname urlstyle\endcsname\relax
  \providecommand{\doi}[1]{doi: #1}\else
  \providecommand{\doi}{doi: \begingroup \urlstyle{rm}\Url}\fi

\bibitem[Dai and Lin(2017)]{dai2017contrastive}
Bo~Dai and Dahua Lin.
\newblock Contrastive learning for image captioning.
\newblock In Isabelle Guyon, Ulrike von Luxburg, Samy Bengio, Hanna~M. Wallach,
  Rob Fergus, S.~V.~N. Vishwanathan, and Roman Garnett, editors, \emph{Advances
  in Neural Information Processing Systems 30: Annual Conference on Neural
  Information Processing Systems 2017, December 4-9, 2017, Long Beach, CA,
  {USA}}, pages 898--907, 2017.
\newblock URL
  \url{https://proceedings.neurips.cc/paper/2017/hash/46922a0880a8f11f8f69cbb52b1396be-Abstract.html}.

\bibitem[Dai et~al.(2017)Dai, Fidler, Urtasun, and Lin]{dai2017towards}
Bo~Dai, Sanja Fidler, Raquel Urtasun, and Dahua Lin.
\newblock Towards diverse and natural image descriptions via a conditional
  {GAN}.
\newblock In \emph{{IEEE} International Conference on Computer Vision, {ICCV}
  2017, Venice, Italy, October 22-29, 2017}, pages 2989--2998. {IEEE} Computer
  Society, 2017.
\newblock \doi{10.1109/ICCV.2017.323}.
\newblock URL \url{https://doi.org/10.1109/ICCV.2017.323}.

\bibitem[Shetty et~al.(2017)Shetty, Rohrbach, Hendricks, Fritz, and
  Schiele]{shetty2017speaking}
Rakshith Shetty, Marcus Rohrbach, Lisa~Anne Hendricks, Mario Fritz, and Bernt
  Schiele.
\newblock Speaking the same language: Matching machine to human captions by
  adversarial training.
\newblock In \emph{{IEEE} International Conference on Computer Vision, {ICCV}
  2017, Venice, Italy, October 22-29, 2017}, pages 4155--4164. {IEEE} Computer
  Society, 2017.
\newblock \doi{10.1109/ICCV.2017.445}.
\newblock URL \url{http://doi.ieeecomputersociety.org/10.1109/ICCV.2017.445}.

\bibitem[Rennie et~al.(2017)Rennie, Marcheret, Mroueh, Ross, and
  Goel]{rennie2017self}
Steven~J. Rennie, Etienne Marcheret, Youssef Mroueh, Jerret Ross, and Vaibhava
  Goel.
\newblock Self-critical sequence training for image captioning.
\newblock In \emph{2017 {IEEE} Conference on Computer Vision and Pattern
  Recognition, {CVPR} 2017, Honolulu, HI, USA, July 21-26, 2017}, pages
  1179--1195. {IEEE} Computer Society, 2017.
\newblock \doi{10.1109/CVPR.2017.131}.
\newblock URL \url{https://doi.org/10.1109/CVPR.2017.131}.

\bibitem[Grice(1975)]{grice1975logic}
Herbert~P Grice.
\newblock Logic and conversation.
\newblock In \emph{Speech acts}, pages 41--58. Brill, 1975.

\bibitem[Farhadi et~al.(2010)Farhadi, Hejrati, Sadeghi, Young, Rashtchian,
  Hockenmaier, and Forsyth]{farhadi2010every}
Ali Farhadi, Mohsen Hejrati, Mohammad~Amin Sadeghi, Peter Young, Cyrus
  Rashtchian, Julia Hockenmaier, and David Forsyth.
\newblock Every picture tells a story: Generating sentences from images.
\newblock In \emph{European conference on computer vision}, pages 15--29.
  Springer, 2010.

\bibitem[Li et~al.(2011)Li, Kulkarni, Berg, Berg, and Choi]{li2011composing}
Siming Li, Girish Kulkarni, Tamara~L Berg, Alexander~C Berg, and Yejin Choi.
\newblock Composing simple image descriptions using web-scale n-grams.
\newblock In \emph{Proceedings of the Fifteenth Conference on Computational
  Natural Language Learning}, pages 220--228, Portland, Oregon, USA, 2011.
  Association for Computational Linguistics.
\newblock URL \url{https://www.aclweb.org/anthology/W11-0326}.

\bibitem[Kulkarni et~al.(2011)Kulkarni, Premraj, Dhar, Li, Choi, Berg, and
  Berg]{kulkarni2013babytalk}
Girish Kulkarni, Visruth Premraj, Sagnik Dhar, Siming Li, Yejin Choi,
  Alexander~C. Berg, and Tamara~L. Berg.
\newblock Baby talk: Understanding and generating simple image descriptions.
\newblock In \emph{The 24th {IEEE} Conference on Computer Vision and Pattern
  Recognition, {CVPR} 2011, Colorado Springs, CO, USA, 20-25 June 2011}, pages
  1601--1608. {IEEE} Computer Society, 2011.
\newblock \doi{10.1109/CVPR.2011.5995466}.
\newblock URL \url{https://doi.org/10.1109/CVPR.2011.5995466}.

\bibitem[Gong et~al.(2014{\natexlab{a}})Gong, Wang, Hodosh, Hockenmaier, and
  Lazebnik]{gong2014improving}
Yunchao Gong, Liwei Wang, Micah Hodosh, Julia Hockenmaier, and Svetlana
  Lazebnik.
\newblock Improving image-sentence embeddings using large weakly annotated
  photo collections.
\newblock In \emph{European conference on computer vision}, pages 529--545.
  Springer, 2014{\natexlab{a}}.

\bibitem[Hodosh et~al.(2013)Hodosh, Young, and Hockenmaier]{hodosh2013framing}
Micah Hodosh, Peter Young, and Julia Hockenmaier.
\newblock Framing image description as a ranking task: Data, models and
  evaluation metrics.
\newblock \emph{Journal of Artificial Intelligence Research}, 47:\penalty0
  853--899, 2013.

\bibitem[Ordonez et~al.(2011)Ordonez, Kulkarni, and Berg]{ordonez2011im2text}
Vicente Ordonez, Girish Kulkarni, and Tamara~L. Berg.
\newblock Im2text: Describing images using 1 million captioned photographs.
\newblock In John Shawe{-}Taylor, Richard~S. Zemel, Peter~L. Bartlett, Fernando
  C.~N. Pereira, and Kilian~Q. Weinberger, editors, \emph{Advances in Neural
  Information Processing Systems 24: 25th Annual Conference on Neural
  Information Processing Systems 2011. Proceedings of a meeting held 12-14
  December 2011, Granada, Spain}, pages 1143--1151, 2011.
\newblock URL
  \url{https://proceedings.neurips.cc/paper/2011/hash/5dd9db5e033da9c6fb5ba83c7a7ebea9-Abstract.html}.

\bibitem[Sun et~al.(2015)Sun, Gan, and Nevatia]{sun2015automatic}
Chen Sun, Chuang Gan, and Ram Nevatia.
\newblock Automatic concept discovery from parallel text and visual corpora.
\newblock In \emph{2015 {IEEE} International Conference on Computer Vision,
  {ICCV} 2015, Santiago, Chile, December 7-13, 2015}, pages 2596--2604. {IEEE}
  Computer Society, 2015.
\newblock \doi{10.1109/ICCV.2015.298}.
\newblock URL \url{https://doi.org/10.1109/ICCV.2015.298}.

\bibitem[Lu et~al.(2018)Lu, Yang, Batra, and Parikh]{lu2018neural}
Jiasen Lu, Jianwei Yang, Dhruv Batra, and Devi Parikh.
\newblock Neural baby talk.
\newblock In \emph{2018 {IEEE} Conference on Computer Vision and Pattern
  Recognition, {CVPR} 2018, Salt Lake City, UT, USA, June 18-22, 2018}, pages
  7219--7228. {IEEE} Computer Society, 2018.
\newblock \doi{10.1109/CVPR.2018.00754}.
\newblock URL
  \url{http://openaccess.thecvf.com/content\_cvpr\_2018/html/Lu\_Neural\_Baby\_Talk\_CVPR\_2018\_paper.html}.

\bibitem[Bahdanau et~al.(2015{\natexlab{a}})Bahdanau, Cho, and
  Bengio]{bahdanau2014neural}
Dzmitry Bahdanau, Kyunghyun Cho, and Yoshua Bengio.
\newblock Neural machine translation by jointly learning to align and
  translate.
\newblock In Yoshua Bengio and Yann LeCun, editors, \emph{3rd International
  Conference on Learning Representations, {ICLR} 2015, San Diego, CA, USA, May
  7-9, 2015, Conference Track Proceedings}, 2015{\natexlab{a}}.
\newblock URL \url{http://arxiv.org/abs/1409.0473}.

\bibitem[Cho et~al.(2014)Cho, van Merri{\"e}nboer, Gulcehre, Bahdanau,
  Bougares, Schwenk, and Bengio]{cho2014learning}
Kyunghyun Cho, Bart van Merri{\"e}nboer, Caglar Gulcehre, Dzmitry Bahdanau,
  Fethi Bougares, Holger Schwenk, and Yoshua Bengio.
\newblock Learning phrase representations using {RNN} encoder{--}decoder for
  statistical machine translation.
\newblock In \emph{Proceedings of the 2014 Conference on Empirical Methods in
  Natural Language Processing ({EMNLP})}, pages 1724--1734, Doha, Qatar, 2014.
  Association for Computational Linguistics.
\newblock \doi{10.3115/v1/D14-1179}.
\newblock URL \url{https://www.aclweb.org/anthology/D14-1179}.

\bibitem[Karpathy and Li(2015{\natexlab{a}})]{karpathy2015deep}
Andrej Karpathy and Fei{-}Fei Li.
\newblock Deep visual-semantic alignments for generating image descriptions.
\newblock In \emph{{IEEE} Conference on Computer Vision and Pattern
  Recognition, {CVPR} 2015, Boston, MA, USA, June 7-12, 2015}, pages
  3128--3137. {IEEE} Computer Society, 2015{\natexlab{a}}.
\newblock \doi{10.1109/CVPR.2015.7298932}.
\newblock URL \url{https://doi.org/10.1109/CVPR.2015.7298932}.

\bibitem[Donahue et~al.(2015)Donahue, Hendricks, Guadarrama, Rohrbach,
  Venugopalan, Darrell, and Saenko]{donahue2015long}
Jeff Donahue, Lisa~Anne Hendricks, Sergio Guadarrama, Marcus Rohrbach,
  Subhashini Venugopalan, Trevor Darrell, and Kate Saenko.
\newblock Long-term recurrent convolutional networks for visual recognition and
  description.
\newblock In \emph{{IEEE} Conference on Computer Vision and Pattern
  Recognition, {CVPR} 2015, Boston, MA, USA, June 7-12, 2015}, pages
  2625--2634. {IEEE} Computer Society, 2015.
\newblock \doi{10.1109/CVPR.2015.7298878}.
\newblock URL \url{https://doi.org/10.1109/CVPR.2015.7298878}.

\bibitem[Vinyals et~al.(2015)Vinyals, Toshev, Bengio, and Erhan]{Vinyals2014}
Oriol Vinyals, Alexander Toshev, Samy Bengio, and Dumitru Erhan.
\newblock Show and tell: {A} neural image caption generator.
\newblock In \emph{{IEEE} Conference on Computer Vision and Pattern
  Recognition, {CVPR} 2015, Boston, MA, USA, June 7-12, 2015}, pages
  3156--3164. {IEEE} Computer Society, 2015.
\newblock \doi{10.1109/CVPR.2015.7298935}.
\newblock URL \url{https://doi.org/10.1109/CVPR.2015.7298935}.

\bibitem[Mao et~al.(2015{\natexlab{a}})Mao, Xu, Yang, Wang, and
  Yuille]{Mao2015}
Junhua Mao, Wei Xu, Yi~Yang, Jiang Wang, and Alan~L. Yuille.
\newblock Deep captioning with multimodal recurrent neural networks (m-rnn).
\newblock In Yoshua Bengio and Yann LeCun, editors, \emph{3rd International
  Conference on Learning Representations, {ICLR} 2015, San Diego, CA, USA, May
  7-9, 2015, Conference Track Proceedings}, 2015{\natexlab{a}}.
\newblock URL \url{http://arxiv.org/abs/1412.6632}.

\bibitem[Xu et~al.(2015)Xu, Ba, Kiros, Cho, Courville, Salakhutdinov, Zemel,
  and Bengio]{Xu2015}
Kelvin Xu, Jimmy Ba, Ryan Kiros, Kyunghyun Cho, Aaron~C. Courville, Ruslan
  Salakhutdinov, Richard~S. Zemel, and Yoshua Bengio.
\newblock Show, attend and tell: Neural image caption generation with visual
  attention.
\newblock In Francis~R. Bach and David~M. Blei, editors, \emph{Proceedings of
  the 32nd International Conference on Machine Learning, {ICML} 2015, Lille,
  France, 6-11 July 2015}, volume~37 of \emph{{JMLR} Workshop and Conference
  Proceedings}, pages 2048--2057. JMLR.org, 2015.
\newblock URL \url{http://proceedings.mlr.press/v37/xuc15.html}.

\bibitem[Karpathy and Li(2015{\natexlab{b}})]{Karpathy2015}
Andrej Karpathy and Fei{-}Fei Li.
\newblock Deep visual-semantic alignments for generating image descriptions.
\newblock In \emph{{IEEE} Conference on Computer Vision and Pattern
  Recognition, {CVPR} 2015, Boston, MA, USA, June 7-12, 2015}, pages
  3128--3137. {IEEE} Computer Society, 2015{\natexlab{b}}.
\newblock \doi{10.1109/CVPR.2015.7298932}.
\newblock URL \url{https://doi.org/10.1109/CVPR.2015.7298932}.

\bibitem[Lu et~al.(2017{\natexlab{a}})Lu, Xiong, Parikh, and
  Socher]{lu2016knowing}
Jiasen Lu, Caiming Xiong, Devi Parikh, and Richard Socher.
\newblock Knowing when to look: Adaptive attention via a visual sentinel for
  image captioning.
\newblock In \emph{2017 {IEEE} Conference on Computer Vision and Pattern
  Recognition, {CVPR} 2017, Honolulu, HI, USA, July 21-26, 2017}, pages
  3242--3250. {IEEE} Computer Society, 2017{\natexlab{a}}.
\newblock \doi{10.1109/CVPR.2017.345}.
\newblock URL \url{https://doi.org/10.1109/CVPR.2017.345}.

\bibitem[Yang et~al.(2016)Yang, Yuan, Wu, Cohen, and
  Salakhutdinov]{yang2016review}
Zhilin Yang, Ye~Yuan, Yuexin Wu, William~W. Cohen, and Ruslan Salakhutdinov.
\newblock Review networks for caption generation.
\newblock In Daniel~D. Lee, Masashi Sugiyama, Ulrike von Luxburg, Isabelle
  Guyon, and Roman Garnett, editors, \emph{Advances in Neural Information
  Processing Systems 29: Annual Conference on Neural Information Processing
  Systems 2016, December 5-10, 2016, Barcelona, Spain}, pages 2361--2369, 2016.
\newblock URL
  \url{https://proceedings.neurips.cc/paper/2016/hash/9996535e07258a7bbfd8b132435c5962-Abstract.html}.

\bibitem[Anderson et~al.(2018)Anderson, He, Buehler, Teney, Johnson, Gould, and
  Zhang]{anderson2018bottom}
Peter Anderson, Xiaodong He, Chris Buehler, Damien Teney, Mark Johnson, Stephen
  Gould, and Lei Zhang.
\newblock Bottom-up and top-down attention for image captioning and visual
  question answering.
\newblock In \emph{2018 {IEEE} Conference on Computer Vision and Pattern
  Recognition, {CVPR} 2018, Salt Lake City, UT, USA, June 18-22, 2018}, pages
  6077--6086. {IEEE} Computer Society, 2018.
\newblock \doi{10.1109/CVPR.2018.00636}.
\newblock URL
  \url{http://openaccess.thecvf.com/content\_cvpr\_2018/html/Anderson\_Bottom-Up\_and\_Top-Down\_CVPR\_2018\_paper.html}.

\bibitem[Wang et~al.(2017{\natexlab{a}})Wang, Lin, Shen, Cohen, and
  Cottrell]{wang2017skeleton}
Yufei Wang, Zhe Lin, Xiaohui Shen, Scott Cohen, and Garrison~W. Cottrell.
\newblock Skeleton key: Image captioning by skeleton-attribute decomposition.
\newblock In \emph{2017 {IEEE} Conference on Computer Vision and Pattern
  Recognition, {CVPR} 2017, Honolulu, HI, USA, July 21-26, 2017}, pages
  7378--7387. {IEEE} Computer Society, 2017{\natexlab{a}}.
\newblock \doi{10.1109/CVPR.2017.780}.
\newblock URL \url{https://doi.org/10.1109/CVPR.2017.780}.

\bibitem[Lu et~al.(2017{\natexlab{b}})Lu, Xiong, Parikh, and
  Socher]{lu2017knowing}
Jiasen Lu, Caiming Xiong, Devi Parikh, and Richard Socher.
\newblock Knowing when to look: Adaptive attention via a visual sentinel for
  image captioning.
\newblock In \emph{2017 {IEEE} Conference on Computer Vision and Pattern
  Recognition, {CVPR} 2017, Honolulu, HI, USA, July 21-26, 2017}, pages
  3242--3250. {IEEE} Computer Society, 2017{\natexlab{b}}.
\newblock \doi{10.1109/CVPR.2017.345}.
\newblock URL \url{https://doi.org/10.1109/CVPR.2017.345}.

\bibitem[Gu et~al.(2018{\natexlab{a}})Gu, Cai, Wang, and Chen]{gu2018stack}
Jiuxiang Gu, Jianfei Cai, Gang Wang, and Tsuhan Chen.
\newblock Stack-captioning: Coarse-to-fine learning for image captioning.
\newblock In Sheila~A. McIlraith and Kilian~Q. Weinberger, editors,
  \emph{Proceedings of the Thirty-Second {AAAI} Conference on Artificial
  Intelligence, (AAAI-18), the 30th innovative Applications of Artificial
  Intelligence (IAAI-18), and the 8th {AAAI} Symposium on Educational Advances
  in Artificial Intelligence (EAAI-18), New Orleans, Louisiana, USA, February
  2-7, 2018}, pages 6837--6844. {AAAI} Press, 2018{\natexlab{a}}.
\newblock URL
  \url{https://www.aaai.org/ocs/index.php/AAAI/AAAI18/paper/view/16465}.

\bibitem[Cornia et~al.(2020)Cornia, Stefanini, Baraldi, and
  Cucchiara]{cornia2020meshed}
Marcella Cornia, Matteo Stefanini, Lorenzo Baraldi, and Rita Cucchiara.
\newblock Meshed-memory transformer for image captioning.
\newblock In \emph{2020 {IEEE/CVF} Conference on Computer Vision and Pattern
  Recognition, {CVPR} 2020, Seattle, WA, USA, June 13-19, 2020}, pages
  10575--10584. {IEEE}, 2020.
\newblock \doi{10.1109/CVPR42600.2020.01059}.
\newblock URL \url{https://doi.org/10.1109/CVPR42600.2020.01059}.

\bibitem[Yang et~al.(2019{\natexlab{a}})Yang, Tang, Zhang, and
  Cai]{yang2019auto}
Xu~Yang, Kaihua Tang, Hanwang Zhang, and Jianfei Cai.
\newblock Auto-encoding scene graphs for image captioning.
\newblock In \emph{{IEEE} Conference on Computer Vision and Pattern
  Recognition, {CVPR} 2019, Long Beach, CA, USA, June 16-20, 2019}, pages
  10685--10694. Computer Vision Foundation / {IEEE}, 2019{\natexlab{a}}.
\newblock \doi{10.1109/CVPR.2019.01094}.
\newblock URL
  \url{http://openaccess.thecvf.com/content\_CVPR\_2019/html/Yang\_Auto-Encoding\_Scene\_Graphs\_for\_Image\_Captioning\_CVPR\_2019\_paper.html}.

\bibitem[Huang et~al.(2019)Huang, Wang, Chen, and Wei]{huang2019attention}
Lun Huang, Wenmin Wang, Jie Chen, and Xiaoyong Wei.
\newblock Attention on attention for image captioning.
\newblock In \emph{2019 {IEEE/CVF} International Conference on Computer Vision,
  {ICCV} 2019, Seoul, Korea (South), October 27 - November 2, 2019}, pages
  4633--4642. {IEEE}, 2019.
\newblock \doi{10.1109/ICCV.2019.00473}.
\newblock URL \url{https://doi.org/10.1109/ICCV.2019.00473}.

\bibitem[Liu et~al.(2017{\natexlab{a}})Liu, Mao, Sha, and Yuille]{Liu2016}
Chenxi Liu, Junhua Mao, Fei Sha, and Alan~L. Yuille.
\newblock Attention correctness in neural image captioning.
\newblock In Satinder~P. Singh and Shaul Markovitch, editors, \emph{Proceedings
  of the Thirty-First {AAAI} Conference on Artificial Intelligence, February
  4-9, 2017, San Francisco, California, {USA}}, pages 4176--4182. {AAAI} Press,
  2017{\natexlab{a}}.
\newblock URL \url{http://aaai.org/ocs/index.php/AAAI/AAAI17/paper/view/14246}.

\bibitem[Pan et~al.(2020)Pan, Yao, Li, and Mei]{pan2020x}
Yingwei Pan, Ting Yao, Yehao Li, and Tao Mei.
\newblock X-linear attention networks for image captioning.
\newblock In \emph{2020 {IEEE/CVF} Conference on Computer Vision and Pattern
  Recognition, {CVPR} 2020, Seattle, WA, USA, June 13-19, 2020}, pages
  10968--10977. {IEEE}, 2020.
\newblock \doi{10.1109/CVPR42600.2020.01098}.
\newblock URL \url{https://doi.org/10.1109/CVPR42600.2020.01098}.

\bibitem[He et~al.(2016)He, Zhang, Ren, and Sun]{he2016deep}
Kaiming He, Xiangyu Zhang, Shaoqing Ren, and Jian Sun.
\newblock Deep residual learning for image recognition.
\newblock In \emph{2016 {IEEE} Conference on Computer Vision and Pattern
  Recognition, {CVPR} 2016, Las Vegas, NV, USA, June 27-30, 2016}, pages
  770--778. {IEEE} Computer Society, 2016.
\newblock \doi{10.1109/CVPR.2016.90}.
\newblock URL \url{https://doi.org/10.1109/CVPR.2016.90}.

\bibitem[Szegedy et~al.(2015)Szegedy, Liu, Jia, Sermanet, Reed, Anguelov,
  Erhan, Vanhoucke, and Rabinovich]{szegedy2015going}
Christian Szegedy, Wei Liu, Yangqing Jia, Pierre Sermanet, Scott~E. Reed,
  Dragomir Anguelov, Dumitru Erhan, Vincent Vanhoucke, and Andrew Rabinovich.
\newblock Going deeper with convolutions.
\newblock In \emph{{IEEE} Conference on Computer Vision and Pattern
  Recognition, {CVPR} 2015, Boston, MA, USA, June 7-12, 2015}, pages 1--9.
  {IEEE} Computer Society, 2015.
\newblock \doi{10.1109/CVPR.2015.7298594}.
\newblock URL \url{https://doi.org/10.1109/CVPR.2015.7298594}.

\bibitem[Krishna et~al.(2017{\natexlab{a}})Krishna, Zhu, Groth, Johnson, Hata,
  Kravitz, Chen, Kalantidis, Li, Shamma, et~al.]{krishna2017visual}
Ranjay Krishna, Yuke Zhu, Oliver Groth, Justin Johnson, Kenji Hata, Joshua
  Kravitz, Stephanie Chen, Yannis Kalantidis, Li-Jia Li, David~A Shamma, et~al.
\newblock Visual genome: Connecting language and vision using crowdsourced
  dense image annotations.
\newblock \emph{International Journal of Computer Vision}, 123\penalty0
  (1):\penalty0 32--73, 2017{\natexlab{a}}.

\bibitem[Zhang et~al.(2021)Zhang, Li, Hu, Yang, Zhang, Wang, Choi, and
  Gao]{zhang2021vinvl}
Pengchuan Zhang, Xiujun Li, Xiaowei Hu, Jianwei Yang, Lei Zhang, Lijuan Wang,
  Yejin Choi, and Jianfeng Gao.
\newblock Vinvl: Revisiting visual representations in vision-language models.
\newblock In \emph{Proceedings of the IEEE/CVF Conference on Computer Vision
  and Pattern Recognition}, pages 5579--5588, 2021.

\bibitem[Ranzato et~al.(2016)Ranzato, Chopra, Auli, and Zaremba]{Ranzato2016}
Marc'Aurelio Ranzato, Sumit Chopra, Michael Auli, and Wojciech Zaremba.
\newblock Sequence level training with recurrent neural networks.
\newblock In Yoshua Bengio and Yann LeCun, editors, \emph{4th International
  Conference on Learning Representations, {ICLR} 2016, San Juan, Puerto Rico,
  May 2-4, 2016, Conference Track Proceedings}, 2016.
\newblock URL \url{http://arxiv.org/abs/1511.06732}.

\bibitem[Zhang et~al.(2017)Zhang, Sung, Liu, Xiang, Gong, Yang, and
  Hospedales]{zhang2017actor}
Li~Zhang, Flood Sung, Feng Liu, Tao Xiang, Shaogang Gong, Yongxin Yang, and
  Timothy~M Hospedales.
\newblock Actor-critic sequence training for image captioning.
\newblock \emph{arXiv preprint arXiv:1706.09601}, 2017.

\bibitem[Liu et~al.(2017{\natexlab{b}})Liu, Zhu, Ye, Guadarrama, and
  Murphy]{liu2017improved}
Siqi Liu, Zhenhai Zhu, Ning Ye, Sergio Guadarrama, and Kevin Murphy.
\newblock Improved image captioning via policy gradient optimization of spider.
\newblock In \emph{{IEEE} International Conference on Computer Vision, {ICCV}
  2017, Venice, Italy, October 22-29, 2017}, pages 873--881. {IEEE} Computer
  Society, 2017{\natexlab{b}}.
\newblock \doi{10.1109/ICCV.2017.100}.
\newblock URL \url{https://doi.org/10.1109/ICCV.2017.100}.

\bibitem[Peters et~al.(2018)Peters, Neumann, Iyyer, Gardner, Clark, Lee, and
  Zettlemoyer]{peters2018deep}
Matthew Peters, Mark Neumann, Mohit Iyyer, Matt Gardner, Christopher Clark,
  Kenton Lee, and Luke Zettlemoyer.
\newblock Deep contextualized word representations.
\newblock In \emph{Proceedings of the 2018 Conference of the North {A}merican
  Chapter of the Association for Computational Linguistics: Human Language
  Technologies, Volume 1 (Long Papers)}, pages 2227--2237, New Orleans,
  Louisiana, 2018. Association for Computational Linguistics.
\newblock \doi{10.18653/v1/N18-1202}.
\newblock URL \url{https://www.aclweb.org/anthology/N18-1202}.

\bibitem[Devlin et~al.(2019)Devlin, Chang, Lee, and Toutanova]{devlin2018bert}
Jacob Devlin, Ming-Wei Chang, Kenton Lee, and Kristina Toutanova.
\newblock {BERT}: Pre-training of deep bidirectional transformers for language
  understanding.
\newblock In \emph{Proceedings of the 2019 Conference of the North {A}merican
  Chapter of the Association for Computational Linguistics: Human Language
  Technologies, Volume 1 (Long and Short Papers)}, pages 4171--4186,
  Minneapolis, Minnesota, 2019. Association for Computational Linguistics.
\newblock \doi{10.18653/v1/N19-1423}.
\newblock URL \url{https://www.aclweb.org/anthology/N19-1423}.

\bibitem[Radford et~al.(2018)Radford, Narasimhan, Salimans, and
  Sutskever]{radford2018improving}
Alec Radford, Karthik Narasimhan, Tim Salimans, and Ilya Sutskever.
\newblock Improving language understanding by generative pre-training.
\newblock \emph{URL https://s3-us-west-2. amazonaws.
  com/openai-assets/research-covers/languageunsupervised/language understanding
  paper. pdf}, 2018.

\bibitem[Zhou et~al.(2020)Zhou, Palangi, Zhang, Hu, Corso, and
  Gao]{zhou2020unified}
Luowei Zhou, Hamid Palangi, Lei Zhang, Houdong Hu, Jason Corso, and Jianfeng
  Gao.
\newblock Unified vision-language pre-training for image captioning and vqa.
\newblock In \emph{Proceedings of the AAAI Conference on Artificial
  Intelligence}, volume~34, pages 13041--13049, 2020.

\bibitem[Li et~al.(2020)Li, Yin, Li, Zhang, Hu, Zhang, Wang, Hu, Dong, Wei,
  et~al.]{li2020oscar}
Xiujun Li, Xi~Yin, Chunyuan Li, Pengchuan Zhang, Xiaowei Hu, Lei Zhang, Lijuan
  Wang, Houdong Hu, Li~Dong, Furu Wei, et~al.
\newblock Oscar: Object-semantics aligned pre-training for vision-language
  tasks.
\newblock In \emph{European Conference on Computer Vision}, pages 121--137.
  Springer, 2020.

\bibitem[Stefanini et~al.(2021)Stefanini, Cornia, Baraldi, Cascianelli,
  Fiameni, and Cucchiara]{stefanini2021show}
Matteo Stefanini, Marcella Cornia, Lorenzo Baraldi, Silvia Cascianelli,
  Giuseppe Fiameni, and Rita Cucchiara.
\newblock From show to tell: A survey on image captioning.
\newblock \emph{arXiv preprint arXiv:2107.06912}, 2021.

\bibitem[Hossain et~al.(2019)Hossain, Sohel, Shiratuddin, and
  Laga]{hossain2019comprehensive}
MD~Zakir Hossain, Ferdous Sohel, Mohd~Fairuz Shiratuddin, and Hamid Laga.
\newblock A comprehensive survey of deep learning for image captioning.
\newblock \emph{ACM Computing Surveys (CSUR)}, 51\penalty0 (6):\penalty0 1--36,
  2019.

\bibitem[Liu et~al.(2019{\natexlab{a}})Liu, Xu, and Wang]{liu2019survey}
Xiaoxiao Liu, Qingyang Xu, and Ning Wang.
\newblock A survey on deep neural network-based image captioning.
\newblock \emph{The Visual Computer}, 35\penalty0 (3):\penalty0 445--470,
  2019{\natexlab{a}}.

\bibitem[Bai and An(2018)]{bai2018survey}
Shuang Bai and Shan An.
\newblock A survey on automatic image caption generation.
\newblock \emph{Neurocomputing}, 311:\penalty0 291--304, 2018.

\bibitem[Sadovnik et~al.(2012)Sadovnik, Chiu, Snavely, Edelman, and
  Chen]{sadovnik2012image}
Amir Sadovnik, Yi{-}I Chiu, Noah Snavely, Shimon Edelman, and Tsuhan Chen.
\newblock Image description with a goal: Building efficient discriminating
  expressions for images.
\newblock In \emph{2012 {IEEE} Conference on Computer Vision and Pattern
  Recognition, Providence, RI, USA, June 16-21, 2012}, pages 2791--2798. {IEEE}
  Computer Society, 2012.
\newblock \doi{10.1109/CVPR.2012.6248003}.
\newblock URL \url{https://doi.org/10.1109/CVPR.2012.6248003}.

\bibitem[Park et~al.(2019{\natexlab{a}})Park, Darrell, and
  Rohrbach]{park2019robust}
Dong~Huk Park, Trevor Darrell, and Anna Rohrbach.
\newblock Robust change captioning.
\newblock In \emph{2019 {IEEE/CVF} International Conference on Computer Vision,
  {ICCV} 2019, Seoul, Korea (South), October 27 - November 2, 2019}, pages
  4623--4632. {IEEE}, 2019{\natexlab{a}}.
\newblock \doi{10.1109/ICCV.2019.00472}.
\newblock URL \url{https://doi.org/10.1109/ICCV.2019.00472}.

\bibitem[Jhamtani and Berg-Kirkpatrick(2018)]{jhamtani2018learning}
Harsh Jhamtani and Taylor Berg-Kirkpatrick.
\newblock Learning to describe differences between pairs of similar images.
\newblock In \emph{Proceedings of the 2018 Conference on Empirical Methods in
  Natural Language Processing}, pages 4024--4034, Brussels, Belgium, 2018.
  Association for Computational Linguistics.
\newblock \doi{10.18653/v1/D18-1436}.
\newblock URL \url{https://www.aclweb.org/anthology/D18-1436}.

\bibitem[Gilton et~al.(2020)Gilton, Luo, Willett, and
  Shakhnarovich]{gilton2020detection}
Davis Gilton, Ruotian Luo, Rebecca Willett, and Greg Shakhnarovich.
\newblock Detection and description of change in visual streams.
\newblock \emph{arXiv preprint arXiv:2003.12633}, 2020.

\bibitem[Mao et~al.(2015{\natexlab{b}})Mao, Wei, Yang, Wang, Huang, and
  Yuille]{mao2015learning}
Junhua Mao, Xu~Wei, Yi~Yang, Jiang Wang, Zhiheng Huang, and Alan~L. Yuille.
\newblock Learning like a child: Fast novel visual concept learning from
  sentence descriptions of images.
\newblock In \emph{2015 {IEEE} International Conference on Computer Vision,
  {ICCV} 2015, Santiago, Chile, December 7-13, 2015}, pages 2533--2541. {IEEE}
  Computer Society, 2015{\natexlab{b}}.
\newblock \doi{10.1109/ICCV.2015.291}.
\newblock URL \url{https://doi.org/10.1109/ICCV.2015.291}.

\bibitem[Hendricks et~al.(2016{\natexlab{a}})Hendricks, Venugopalan, Rohrbach,
  Mooney, Saenko, and Darrell]{hendricks2016deep}
Lisa~Anne Hendricks, Subhashini Venugopalan, Marcus Rohrbach, Raymond~J.
  Mooney, Kate Saenko, and Trevor Darrell.
\newblock Deep compositional captioning: Describing novel object categories
  without paired training data.
\newblock In \emph{2016 {IEEE} Conference on Computer Vision and Pattern
  Recognition, {CVPR} 2016, Las Vegas, NV, USA, June 27-30, 2016}, pages 1--10.
  {IEEE} Computer Society, 2016{\natexlab{a}}.
\newblock \doi{10.1109/CVPR.2016.8}.
\newblock URL \url{https://doi.org/10.1109/CVPR.2016.8}.

\bibitem[Agrawal et~al.(2019)Agrawal, Anderson, Desai, Wang, Chen, Jain,
  Johnson, Batra, Parikh, and Lee]{agrawal2019nocaps}
Harsh Agrawal, Peter Anderson, Karan Desai, Yufei Wang, Xinlei Chen, Rishabh
  Jain, Mark Johnson, Dhruv Batra, Devi Parikh, and Stefan Lee.
\newblock nocaps: novel object captioning at scale.
\newblock In \emph{2019 {IEEE/CVF} International Conference on Computer Vision,
  {ICCV} 2019, Seoul, Korea (South), October 27 - November 2, 2019}, pages
  8947--8956. {IEEE}, 2019.
\newblock \doi{10.1109/ICCV.2019.00904}.
\newblock URL \url{https://doi.org/10.1109/ICCV.2019.00904}.

\bibitem[Krasin et~al.(2017)Krasin, Duerig, Alldrin, Ferrari, Abu-El-Haija,
  Kuznetsova, Rom, Uijlings, Popov, Veit, et~al.]{krasin2017openimages}
Ivan Krasin, Tom Duerig, Neil Alldrin, Vittorio Ferrari, Sami Abu-El-Haija,
  Alina Kuznetsova, Hassan Rom, Jasper Uijlings, Stefan Popov, Andreas Veit,
  et~al.
\newblock Openimages: A public dataset for large-scale multi-label and
  multi-class image classification.
\newblock \emph{Dataset available from https://github. com/openimages},
  2\penalty0 (3):\penalty0 18, 2017.

\bibitem[Gurari et~al.(2020)Gurari, Zhao, Zhang, and
  Bhattacharya]{gurari2020captioning}
Danna Gurari, Yinan Zhao, Meng Zhang, and Nilavra Bhattacharya.
\newblock Captioning images taken by people who are blind.
\newblock In \emph{European Conference on Computer Vision}, pages 417--434.
  Springer, 2020.

\bibitem[Sidorov et~al.(2020)Sidorov, Hu, Rohrbach, and
  Singh]{sidorov2020textcaps}
Oleksii Sidorov, Ronghang Hu, Marcus Rohrbach, and Amanpreet Singh.
\newblock Textcaps: a dataset for image captioning with reading comprehension.
\newblock In \emph{European Conference on Computer Vision}, pages 742--758.
  Springer, 2020.

\bibitem[Gan et~al.(2017)Gan, Gan, He, Gao, and Deng]{gan2017stylenet}
Chuang Gan, Zhe Gan, Xiaodong He, Jianfeng Gao, and Li~Deng.
\newblock Stylenet: Generating attractive visual captions with styles.
\newblock In \emph{2017 {IEEE} Conference on Computer Vision and Pattern
  Recognition, {CVPR} 2017, Honolulu, HI, USA, July 21-26, 2017}, pages
  955--964. {IEEE} Computer Society, 2017.
\newblock \doi{10.1109/CVPR.2017.108}.
\newblock URL \url{https://doi.org/10.1109/CVPR.2017.108}.

\bibitem[Shuster et~al.(2019)Shuster, Humeau, Hu, Bordes, and
  Weston]{shuster2019engaging}
Kurt Shuster, Samuel Humeau, Hexiang Hu, Antoine Bordes, and Jason Weston.
\newblock Engaging image captioning via personality.
\newblock In \emph{{IEEE} Conference on Computer Vision and Pattern
  Recognition, {CVPR} 2019, Long Beach, CA, USA, June 16-20, 2019}, pages
  12516--12526. Computer Vision Foundation / {IEEE}, 2019.
\newblock \doi{10.1109/CVPR.2019.01280}.
\newblock URL
  \url{http://openaccess.thecvf.com/content\_CVPR\_2019/html/Shuster\_Engaging\_Image\_Captioning\_via\_Personality\_CVPR\_2019\_paper.html}.

\bibitem[You et~al.(2018)You, Jin, and Luo]{you2018image}
Quanzeng You, Hailin Jin, and Jiebo Luo.
\newblock Image captioning at will: A versatile scheme for effectively
  injecting sentiments into image descriptions.
\newblock \emph{arXiv preprint arXiv:1801.10121}, 2018.

\bibitem[Mathews et~al.(2018)Mathews, Xie, and He]{mathews2018semstyle}
Alexander~Patrick Mathews, Lexing Xie, and Xuming He.
\newblock Semstyle: Learning to generate stylised image captions using
  unaligned text.
\newblock In \emph{2018 {IEEE} Conference on Computer Vision and Pattern
  Recognition, {CVPR} 2018, Salt Lake City, UT, USA, June 18-22, 2018}, pages
  8591--8600. {IEEE} Computer Society, 2018.
\newblock \doi{10.1109/CVPR.2018.00896}.
\newblock URL
  \url{http://openaccess.thecvf.com/content\_cvpr\_2018/html/Mathews\_SemStyle\_Learning\_to\_CVPR\_2018\_paper.html}.

\bibitem[Chen and Dolan(2011)]{chen2011collecting}
David Chen and William Dolan.
\newblock Collecting highly parallel data for paraphrase evaluation.
\newblock In \emph{Proceedings of the 49th Annual Meeting of the Association
  for Computational Linguistics: Human Language Technologies}, pages 190--200,
  Portland, Oregon, USA, 2011. Association for Computational Linguistics.
\newblock URL \url{https://www.aclweb.org/anthology/P11-1020}.

\bibitem[Xu et~al.(2016)Xu, Mei, Yao, and Rui]{xu2016msr}
Jun Xu, Tao Mei, Ting Yao, and Yong Rui.
\newblock {MSR-VTT:} {A} large video description dataset for bridging video and
  language.
\newblock In \emph{2016 {IEEE} Conference on Computer Vision and Pattern
  Recognition, {CVPR} 2016, Las Vegas, NV, USA, June 27-30, 2016}, pages
  5288--5296. {IEEE} Computer Society, 2016.
\newblock \doi{10.1109/CVPR.2016.571}.
\newblock URL \url{https://doi.org/10.1109/CVPR.2016.571}.

\bibitem[Wang et~al.(2019)Wang, Wu, Chen, Li, Wang, and Wang]{wang2019vatex}
Xin Wang, Jiawei Wu, Junkun Chen, Lei Li, Yuan{-}Fang Wang, and William~Yang
  Wang.
\newblock Vatex: {A} large-scale, high-quality multilingual dataset for
  video-and-language research.
\newblock In \emph{2019 {IEEE/CVF} International Conference on Computer Vision,
  {ICCV} 2019, Seoul, Korea (South), October 27 - November 2, 2019}, pages
  4580--4590. {IEEE}, 2019.
\newblock \doi{10.1109/ICCV.2019.00468}.
\newblock URL \url{https://doi.org/10.1109/ICCV.2019.00468}.

\bibitem[Johnson et~al.(2016)Johnson, Karpathy, and
  Fei{-}Fei]{johnson2016densecap}
Justin Johnson, Andrej Karpathy, and Li~Fei{-}Fei.
\newblock Densecap: Fully convolutional localization networks for dense
  captioning.
\newblock In \emph{2016 {IEEE} Conference on Computer Vision and Pattern
  Recognition, {CVPR} 2016, Las Vegas, NV, USA, June 27-30, 2016}, pages
  4565--4574. {IEEE} Computer Society, 2016.
\newblock \doi{10.1109/CVPR.2016.494}.
\newblock URL \url{https://doi.org/10.1109/CVPR.2016.494}.

\bibitem[Krishna et~al.(2017{\natexlab{b}})Krishna, Hata, Ren, Fei{-}Fei, and
  Niebles]{krishna2017dense}
Ranjay Krishna, Kenji Hata, Frederic Ren, Li~Fei{-}Fei, and Juan~Carlos
  Niebles.
\newblock Dense-captioning events in videos.
\newblock In \emph{{IEEE} International Conference on Computer Vision, {ICCV}
  2017, Venice, Italy, October 22-29, 2017}, pages 706--715. {IEEE} Computer
  Society, 2017{\natexlab{b}}.
\newblock \doi{10.1109/ICCV.2017.83}.
\newblock URL \url{https://doi.org/10.1109/ICCV.2017.83}.

\bibitem[Zhou et~al.(2018{\natexlab{a}})Zhou, Xu, and Corso]{zhou2018towards}
Luowei Zhou, Chenliang Xu, and Jason~J. Corso.
\newblock Towards automatic learning of procedures from web instructional
  videos.
\newblock In Sheila~A. McIlraith and Kilian~Q. Weinberger, editors,
  \emph{Proceedings of the Thirty-Second {AAAI} Conference on Artificial
  Intelligence, (AAAI-18), the 30th innovative Applications of Artificial
  Intelligence (IAAI-18), and the 8th {AAAI} Symposium on Educational Advances
  in Artificial Intelligence (EAAI-18), New Orleans, Louisiana, USA, February
  2-7, 2018}, pages 7590--7598. {AAAI} Press, 2018{\natexlab{a}}.
\newblock URL
  \url{https://www.aaai.org/ocs/index.php/AAAI/AAAI18/paper/view/17344}.

\bibitem[Lei et~al.(2020)Lei, Yu, Berg, and Bansal]{lei2020tvr}
Jie Lei, Licheng Yu, Tamara~L Berg, and Mohit Bansal.
\newblock Tvr: A large-scale dataset for video-subtitle moment retrieval.
\newblock In \emph{Computer Vision--ECCV 2020: 16th European Conference,
  Glasgow, UK, August 23--28, 2020, Proceedings, Part XXI 16}, pages 447--463.
  Springer, 2020.

\bibitem[Huang et~al.(2016)Huang, Ferraro, Mostafazadeh, Misra, Agrawal,
  Devlin, Girshick, He, Kohli, Batra, Zitnick, Parikh, Vanderwende, Galley, and
  Mitchell]{huang2016visual}
Ting-Hao~Kenneth Huang, Francis Ferraro, Nasrin Mostafazadeh, Ishan Misra,
  Aishwarya Agrawal, Jacob Devlin, Ross Girshick, Xiaodong He, Pushmeet Kohli,
  Dhruv Batra, C.~Lawrence Zitnick, Devi Parikh, Lucy Vanderwende, Michel
  Galley, and Margaret Mitchell.
\newblock Visual storytelling.
\newblock In \emph{Proceedings of the 2016 Conference of the North {A}merican
  Chapter of the Association for Computational Linguistics: Human Language
  Technologies}, pages 1233--1239, San Diego, California, 2016. Association for
  Computational Linguistics.
\newblock \doi{10.18653/v1/N16-1147}.
\newblock URL \url{https://www.aclweb.org/anthology/N16-1147}.

\bibitem[Tariq and Foroosh(2016)]{tariq2016context}
Amara Tariq and Hassan Foroosh.
\newblock A context-driven extractive framework for generating realistic image
  descriptions.
\newblock \emph{IEEE Transactions on Image Processing}, 26\penalty0
  (2):\penalty0 619--632, 2016.

\bibitem[Feng and Lapata(2010)]{feng2010topic}
Yansong Feng and Mirella Lapata.
\newblock Topic models for image annotation and text illustration.
\newblock In \emph{Human Language Technologies: The 2010 Annual Conference of
  the North {A}merican Chapter of the Association for Computational
  Linguistics}, pages 831--839, Los Angeles, California, 2010. Association for
  Computational Linguistics.
\newblock URL \url{https://www.aclweb.org/anthology/N10-1125}.

\bibitem[Biten et~al.(2019)Biten, G{\'{o}}mez, Rusi{\~{n}}ol, and
  Karatzas]{biten2019good}
Ali~Furkan Biten, Llu{\'{\i}}s G{\'{o}}mez, Mar{\c{c}}al Rusi{\~{n}}ol, and
  Dimosthenis Karatzas.
\newblock Good news, everyone! context driven entity-aware captioning for news
  images.
\newblock In \emph{{IEEE} Conference on Computer Vision and Pattern
  Recognition, {CVPR} 2019, Long Beach, CA, USA, June 16-20, 2019}, pages
  12466--12475. Computer Vision Foundation / {IEEE}, 2019.
\newblock \doi{10.1109/CVPR.2019.01275}.
\newblock URL
  \url{http://openaccess.thecvf.com/content\_CVPR\_2019/html/Biten\_Good\_News\_Everyone\_Context\_Driven\_Entity-Aware\_Captioning\_for\_News\_Images\_CVPR\_2019\_paper.html}.

\bibitem[Liu et~al.(2020)Liu, Wang, Wang, and Ordonez]{liu2020visualnews}
Fuxiao Liu, Yinghan Wang, Tianlu Wang, and Vicente Ordonez.
\newblock Visualnews: Benchmark and challenges in entity-aware image
  captioning.
\newblock \emph{arXiv preprint arXiv:2010.03743}, 2020.

\bibitem[Tan et~al.(2019)Tan, Yu, and Bansal]{tan2019learning}
Hao Tan, Licheng Yu, and Mohit Bansal.
\newblock Learning to navigate unseen environments: Back translation with
  environmental dropout.
\newblock In \emph{Proceedings of the 2019 Conference of the North {A}merican
  Chapter of the Association for Computational Linguistics: Human Language
  Technologies, Volume 1 (Long and Short Papers)}, pages 2610--2621,
  Minneapolis, Minnesota, 2019. Association for Computational Linguistics.
\newblock \doi{10.18653/v1/N19-1268}.
\newblock URL \url{https://www.aclweb.org/anthology/N19-1268}.

\bibitem[Fried et~al.(2018)Fried, Hu, Cirik, Rohrbach, Andreas, Morency,
  Berg{-}Kirkpatrick, Saenko, Klein, and Darrell]{fried2018speaker}
Daniel Fried, Ronghang Hu, Volkan Cirik, Anna Rohrbach, Jacob Andreas,
  Louis{-}Philippe Morency, Taylor Berg{-}Kirkpatrick, Kate Saenko, Dan Klein,
  and Trevor Darrell.
\newblock Speaker-follower models for vision-and-language navigation.
\newblock In Samy Bengio, Hanna~M. Wallach, Hugo Larochelle, Kristen Grauman,
  Nicol{\`{o}} Cesa{-}Bianchi, and Roman Garnett, editors, \emph{Advances in
  Neural Information Processing Systems 31: Annual Conference on Neural
  Information Processing Systems 2018, NeurIPS 2018, December 3-8, 2018,
  Montr{\'{e}}al, Canada}, pages 3318--3329, 2018.
\newblock URL
  \url{https://proceedings.neurips.cc/paper/2018/hash/6a81681a7af700c6385d36577ebec359-Abstract.html}.

\bibitem[Rashtchian et~al.(2010)Rashtchian, Young, Hodosh, and
  Hockenmaier]{rashtchian2010collecting}
Cyrus Rashtchian, Peter Young, Micah Hodosh, and Julia Hockenmaier.
\newblock Collecting image annotations using {A}mazon{'}s {M}echanical {T}urk.
\newblock In \emph{Proceedings of the {NAACL} {HLT} 2010 Workshop on Creating
  Speech and Language Data with {A}mazon{'}s Mechanical Turk}, pages 139--147,
  Los Angeles, 2010. Association for Computational Linguistics.
\newblock URL \url{https://www.aclweb.org/anthology/W10-0721}.

\bibitem[Hoiem et~al.(2009)Hoiem, Divvala, and Hays]{hoiem2009pascal}
Derek Hoiem, Santosh~K Divvala, and James~H Hays.
\newblock Pascal voc 2008 challenge.
\newblock \emph{World Literature Today}, 2009.

\bibitem[Young et~al.(2014)Young, Lai, Hodosh, and
  Hockenmaier]{young-etal-2014-image}
Peter Young, Alice Lai, Micah Hodosh, and Julia Hockenmaier.
\newblock From image descriptions to visual denotations: New similarity metrics
  for semantic inference over event descriptions.
\newblock \emph{Transactions of the Association for Computational Linguistics},
  2:\penalty0 67--78, 2014.
\newblock \doi{10.1162/tacl_a_00166}.
\newblock URL \url{https://www.aclweb.org/anthology/Q14-1006}.

\bibitem[Chen et~al.(2015)Chen, Fang, Lin, Vedantam, Gupta, Doll{\'a}r, and
  Zitnick]{chen2015microsoft}
Xinlei Chen, Hao Fang, Tsung-Yi Lin, Ramakrishna Vedantam, Saurabh Gupta, Piotr
  Doll{\'a}r, and C~Lawrence Zitnick.
\newblock Microsoft coco captions: Data collection and evaluation server.
\newblock \emph{arXiv preprint arXiv:1504.00325}, 2015.

\bibitem[Mao et~al.(2016{\natexlab{a}})Mao, Xu, Jing, and
  Yuille]{mao2016training}
Junhua Mao, Jiajing Xu, Yushi Jing, and Alan~L. Yuille.
\newblock Training and evaluating multimodal word embeddings with large-scale
  web annotated images.
\newblock In Daniel~D. Lee, Masashi Sugiyama, Ulrike von Luxburg, Isabelle
  Guyon, and Roman Garnett, editors, \emph{Advances in Neural Information
  Processing Systems 29: Annual Conference on Neural Information Processing
  Systems 2016, December 5-10, 2016, Barcelona, Spain}, pages 442--450,
  2016{\natexlab{a}}.
\newblock URL
  \url{https://proceedings.neurips.cc/paper/2016/hash/c24cd76e1ce41366a4bbe8a49b02a028-Abstract.html}.

\bibitem[Sharma et~al.(2018)Sharma, Ding, Goodman, and
  Soricut]{sharma2018conceptual}
Piyush Sharma, Nan Ding, Sebastian Goodman, and Radu Soricut.
\newblock Conceptual captions: A cleaned, hypernymed, image alt-text dataset
  for automatic image captioning.
\newblock In \emph{Proceedings of the 56th Annual Meeting of the Association
  for Computational Linguistics (Volume 1: Long Papers)}, pages 2556--2565,
  Melbourne, Australia, 2018. Association for Computational Linguistics.
\newblock \doi{10.18653/v1/P18-1238}.
\newblock URL \url{https://www.aclweb.org/anthology/P18-1238}.

\bibitem[Lu et~al.(2019)Lu, Batra, Parikh, and Lee]{lu2019vilbert}
Jiasen Lu, Dhruv Batra, Devi Parikh, and Stefan Lee.
\newblock Vilbert: Pretraining task-agnostic visiolinguistic representations
  for vision-and-language tasks.
\newblock In Hanna~M. Wallach, Hugo Larochelle, Alina Beygelzimer, Florence
  d'Alch{\'{e}}{-}Buc, Emily~B. Fox, and Roman Garnett, editors, \emph{Advances
  in Neural Information Processing Systems 32: Annual Conference on Neural
  Information Processing Systems 2019, NeurIPS 2019, December 8-14, 2019,
  Vancouver, BC, Canada}, pages 13--23, 2019.
\newblock URL
  \url{https://proceedings.neurips.cc/paper/2019/hash/c74d97b01eae257e44aa9d5bade97baf-Abstract.html}.

\bibitem[Lu et~al.(2020)Lu, Goswami, Rohrbach, Parikh, and Lee]{lu202012}
Jiasen Lu, Vedanuj Goswami, Marcus Rohrbach, Devi Parikh, and Stefan Lee.
\newblock 12-in-1: Multi-task vision and language representation learning.
\newblock In \emph{2020 {IEEE/CVF} Conference on Computer Vision and Pattern
  Recognition, {CVPR} 2020, Seattle, WA, USA, June 13-19, 2020}, pages
  10434--10443. {IEEE}, 2020.
\newblock \doi{10.1109/CVPR42600.2020.01045}.
\newblock URL \url{https://doi.org/10.1109/CVPR42600.2020.01045}.

\bibitem[Chen et~al.(2019{\natexlab{a}})Chen, Li, Yu, El~Kholy, Ahmed, Gan,
  Cheng, and Liu]{chen2019uniter}
Yen-Chun Chen, Linjie Li, Licheng Yu, Ahmed El~Kholy, Faisal Ahmed, Zhe Gan,
  Yu~Cheng, and Jingjing Liu.
\newblock Uniter: Learning universal image-text representations.
\newblock 2019{\natexlab{a}}.

\bibitem[Gan et~al.(2020)Gan, Chen, Li, Zhu, Cheng, and Liu]{gan2020large}
Zhe Gan, Yen{-}Chun Chen, Linjie Li, Chen Zhu, Yu~Cheng, and Jingjing Liu.
\newblock Large-scale adversarial training for vision-and-language
  representation learning.
\newblock In Hugo Larochelle, Marc'Aurelio Ranzato, Raia Hadsell,
  Maria{-}Florina Balcan, and Hsuan{-}Tien Lin, editors, \emph{Advances in
  Neural Information Processing Systems 33: Annual Conference on Neural
  Information Processing Systems 2020, NeurIPS 2020, December 6-12, 2020,
  virtual}, 2020.
\newblock URL
  \url{https://proceedings.neurips.cc/paper/2020/hash/49562478de4c54fafd4ec46fdb297de5-Abstract.html}.

\bibitem[Changpinyo et~al.(2021)Changpinyo, Sharma, Ding, and
  Soricut]{changpinyo2021conceptual}
Soravit Changpinyo, Piyush Sharma, Nan Ding, and Radu Soricut.
\newblock Conceptual 12m: Pushing web-scale image-text pre-training to
  recognize long-tail visual concepts.
\newblock In \emph{Proceedings of the IEEE/CVF Conference on Computer Vision
  and Pattern Recognition}, pages 3558--3568, 2021.

\bibitem[Radford et~al.(2021)Radford, Kim, Hallacy, Ramesh, Goh, Agarwal,
  Sastry, Askell, Mishkin, Clark, et~al.]{radford2021learning}
Alec Radford, Jong~Wook Kim, Chris Hallacy, Aditya Ramesh, Gabriel Goh,
  Sandhini Agarwal, Girish Sastry, Amanda Askell, Pamela Mishkin, Jack Clark,
  et~al.
\newblock Learning transferable visual models from natural language
  supervision.
\newblock \emph{arXiv preprint arXiv:2103.00020}, 2021.

\bibitem[Jia et~al.(2021)Jia, Yang, Xia, Chen, Parekh, Pham, Le, Sung, Li, and
  Duerig]{jia2021scaling}
Chao Jia, Yinfei Yang, Ye~Xia, Yi-Ting Chen, Zarana Parekh, Hieu Pham, Quoc~V
  Le, Yunhsuan Sung, Zhen Li, and Tom Duerig.
\newblock Scaling up visual and vision-language representation learning with
  noisy text supervision.
\newblock \emph{arXiv preprint arXiv:2102.05918}, 2021.

\bibitem[Gr{\"{u}}binger et~al.(2006)Gr{\"{u}}binger, Clough, M{\"{u}}ller, and
  Deselaers]{Grubinger2006}
M~Gr{\"{u}}binger, P~Clough, H~M{\"{u}}ller, and T~Deselaers.
\newblock {The IAPR TC-12 Benchmark: A New Evaluation Resource for Visual
  Information Systems}.
\newblock \emph{LREC Workshop OntoImage Language Resources for Content-Based
  Image Retrieval}, pages 13--23, 2006.

\bibitem[Krause et~al.(2017)Krause, Johnson, Krishna, and
  Fei{-}Fei]{krause2017hierarchical}
Jonathan Krause, Justin Johnson, Ranjay Krishna, and Li~Fei{-}Fei.
\newblock A hierarchical approach for generating descriptive image paragraphs.
\newblock In \emph{2017 {IEEE} Conference on Computer Vision and Pattern
  Recognition, {CVPR} 2017, Honolulu, HI, USA, July 21-26, 2017}, pages
  3337--3345. {IEEE} Computer Society, 2017.
\newblock \doi{10.1109/CVPR.2017.356}.
\newblock URL \url{https://doi.org/10.1109/CVPR.2017.356}.

\bibitem[Pont-Tuset et~al.(2020)Pont-Tuset, Uijlings, Changpinyo, Soricut, and
  Ferrari]{pont2020connecting}
Jordi Pont-Tuset, Jasper Uijlings, Soravit Changpinyo, Radu Soricut, and
  Vittorio Ferrari.
\newblock Connecting vision and language with localized narratives.
\newblock In \emph{European Conference on Computer Vision}, pages 647--664.
  Springer, 2020.

\bibitem[Zitnick et~al.(2014)Zitnick, Vedantam, and
  Parikh]{zitnick2014adopting}
C~Lawrence Zitnick, Ramakrishna Vedantam, and Devi Parikh.
\newblock Adopting abstract images for semantic scene understanding.
\newblock \emph{IEEE transactions on pattern analysis and machine
  intelligence}, 38\penalty0 (4):\penalty0 627--638, 2014.

\bibitem[Johnson et~al.(2017)Johnson, Hariharan, van~der Maaten, Fei{-}Fei,
  Zitnick, and Girshick]{johnson2017clevr}
Justin Johnson, Bharath Hariharan, Laurens van~der Maaten, Li~Fei{-}Fei,
  C.~Lawrence Zitnick, and Ross~B. Girshick.
\newblock {CLEVR:} {A} diagnostic dataset for compositional language and
  elementary visual reasoning.
\newblock In \emph{2017 {IEEE} Conference on Computer Vision and Pattern
  Recognition, {CVPR} 2017, Honolulu, HI, USA, July 21-26, 2017}, pages
  1988--1997. {IEEE} Computer Society, 2017.
\newblock \doi{10.1109/CVPR.2017.215}.
\newblock URL \url{https://doi.org/10.1109/CVPR.2017.215}.

\bibitem[Kottur et~al.(2019)Kottur, Moura, Parikh, Batra, and
  Rohrbach]{kottur2019clevr}
Satwik Kottur, Jos{\'e} M.~F. Moura, Devi Parikh, Dhruv Batra, and Marcus
  Rohrbach.
\newblock {CLEVR}-dialog: A diagnostic dataset for multi-round reasoning in
  visual dialog.
\newblock In \emph{Proceedings of the 2019 Conference of the North {A}merican
  Chapter of the Association for Computational Linguistics: Human Language
  Technologies, Volume 1 (Long and Short Papers)}, pages 582--595, Minneapolis,
  Minnesota, 2019. Association for Computational Linguistics.
\newblock \doi{10.18653/v1/N19-1058}.
\newblock URL \url{https://www.aclweb.org/anthology/N19-1058}.

\bibitem[Hessel et~al.(2021)Hessel, Holtzman, Forbes, Bras, and
  Choi]{hessel2021clipscore}
Jack Hessel, Ari Holtzman, Maxwell Forbes, Ronan~Le Bras, and Yejin Choi.
\newblock Clipscore: A reference-free evaluation metric for image captioning.
\newblock \emph{arXiv preprint arXiv:2104.08718}, 2021.

\bibitem[Papineni et~al.(2002)Papineni, Roukos, Ward, and
  Zhu]{papineni2002bleu}
Kishore Papineni, Salim Roukos, Todd Ward, and Wei-Jing Zhu.
\newblock {B}leu: a method for automatic evaluation of machine translation.
\newblock In \emph{Proceedings of the 40th Annual Meeting of the Association
  for Computational Linguistics}, pages 311--318, Philadelphia, Pennsylvania,
  USA, 2002. Association for Computational Linguistics.
\newblock \doi{10.3115/1073083.1073135}.
\newblock URL \url{https://www.aclweb.org/anthology/P02-1040}.

\bibitem[Denkowski and Lavie(2014)]{denkowski2014meteor}
Michael Denkowski and Alon Lavie.
\newblock Meteor universal: Language specific translation evaluation for any
  target language.
\newblock In \emph{Proceedings of the Ninth Workshop on Statistical Machine
  Translation}, pages 376--380, Baltimore, Maryland, USA, 2014. Association for
  Computational Linguistics.
\newblock \doi{10.3115/v1/W14-3348}.
\newblock URL \url{https://www.aclweb.org/anthology/W14-3348}.

\bibitem[Lin(2004)]{lin2004rouge}
Chin-Yew Lin.
\newblock {ROUGE}: A package for automatic evaluation of summaries.
\newblock In \emph{Text Summarization Branches Out}, pages 74--81, Barcelona,
  Spain, 2004. Association for Computational Linguistics.
\newblock URL \url{https://www.aclweb.org/anthology/W04-1013}.

\bibitem[Vedantam et~al.(2015)Vedantam, Zitnick, and Parikh]{vedantam2015cider}
Ramakrishna Vedantam, C.~Lawrence Zitnick, and Devi Parikh.
\newblock Cider: Consensus-based image description evaluation.
\newblock In \emph{{IEEE} Conference on Computer Vision and Pattern
  Recognition, {CVPR} 2015, Boston, MA, USA, June 7-12, 2015}, pages
  4566--4575. {IEEE} Computer Society, 2015.
\newblock \doi{10.1109/CVPR.2015.7299087}.
\newblock URL \url{https://doi.org/10.1109/CVPR.2015.7299087}.

\bibitem[Anderson et~al.(2016)Anderson, Fernando, Johnson, and
  Gould]{anderson2016spice}
Peter Anderson, Basura Fernando, Mark Johnson, and Stephen Gould.
\newblock Spice: Semantic propositional image caption evaluation.
\newblock In \emph{European Conference on Computer Vision}, pages 382--398.
  Springer, 2016.

\bibitem[Yi et~al.(2020)Yi, Deng, and Hu]{yi2020improving}
Yanzhi Yi, Hangyu Deng, and Jinglu Hu.
\newblock Improving image captioning evaluation by considering inter references
  variance.
\newblock In \emph{Proceedings of the 58th Annual Meeting of the Association
  for Computational Linguistics}, pages 985--994, Online, 2020. Association for
  Computational Linguistics.
\newblock \doi{10.18653/v1/2020.acl-main.93}.
\newblock URL \url{https://www.aclweb.org/anthology/2020.acl-main.93}.

\bibitem[Feinglass and Yang(2021)]{feinglass2021smurf}
Joshua Feinglass and Yezhou Yang.
\newblock Smurf: Semantic and linguistic understanding fusion for caption
  evaluation via typicality analysis.
\newblock \emph{arXiv preprint arXiv:2106.01444}, 2021.

\bibitem[Kusner et~al.(2015)Kusner, Sun, Kolkin, and
  Weinberger]{kusner2015word}
Matt~J. Kusner, Yu~Sun, Nicholas~I. Kolkin, and Kilian~Q. Weinberger.
\newblock From word embeddings to document distances.
\newblock In Francis~R. Bach and David~M. Blei, editors, \emph{Proceedings of
  the 32nd International Conference on Machine Learning, {ICML} 2015, Lille,
  France, 6-11 July 2015}, volume~37 of \emph{{JMLR} Workshop and Conference
  Proceedings}, pages 957--966. JMLR.org, 2015.
\newblock URL \url{http://proceedings.mlr.press/v37/kusnerb15.html}.

\bibitem[Kilickaya et~al.(2017)Kilickaya, Erdem, Ikizler-Cinbis, and
  Erdem]{kilickaya2016re}
Mert Kilickaya, Aykut Erdem, Nazli Ikizler-Cinbis, and Erkut Erdem.
\newblock Re-evaluating automatic metrics for image captioning.
\newblock In \emph{Proceedings of the 15th Conference of the {E}uropean Chapter
  of the Association for Computational Linguistics: Volume 1, Long Papers},
  pages 199--209, Valencia, Spain, 2017. Association for Computational
  Linguistics.
\newblock URL \url{https://www.aclweb.org/anthology/E17-1019}.

\bibitem[Snover et~al.(2005)Snover, Dorr, Schwartz, Makhoul, Micciulla, and
  Weischedel]{snover2005study}
Mathew Snover, Bonnie Dorr, Richard Schwartz, John Makhoul, Linnea Micciulla,
  and Ralph Weischedel.
\newblock A study of translation error rate with targeted human annotation.
\newblock In \emph{Proceedings of the 7th Conference of the Association for
  Machine Translation in the Americas (AMTA 06)}, pages 223--231, 2005.

\bibitem[Elliott and Keller(2014)]{elliott2014comparing}
Desmond Elliott and Frank Keller.
\newblock Comparing automatic evaluation measures for image description.
\newblock In \emph{Proceedings of the 52nd Annual Meeting of the Association
  for Computational Linguistics (Volume 2: Short Papers)}, pages 452--457,
  Baltimore, Maryland, 2014. Association for Computational Linguistics.
\newblock \doi{10.3115/v1/P14-2074}.
\newblock URL \url{https://www.aclweb.org/anthology/P14-2074}.

\bibitem[Sharif et~al.(2020)Sharif, White, Bennamoun, Liu, and
  Shah]{sharif2020wembsim}
Naeha Sharif, Lyndon White, Mohammed Bennamoun, Wei Liu, and Syed Afaq~Ali
  Shah.
\newblock Wembsim: A simple yet effective metric for image captioning.
\newblock In \emph{2020 Digital Image Computing: Techniques and Applications
  (DICTA)}, pages 1--8. IEEE, 2020.

\bibitem[Jiang et~al.(2019{\natexlab{a}})Jiang, Hu, Huang, Zhang, Diesner, and
  Gao]{jiang2019reo}
Ming Jiang, Junjie Hu, Qiuyuan Huang, Lei Zhang, Jana Diesner, and Jianfeng
  Gao.
\newblock {REO}-relevance, extraness, omission: A fine-grained evaluation for
  image captioning.
\newblock In \emph{Proceedings of the 2019 Conference on Empirical Methods in
  Natural Language Processing and the 9th International Joint Conference on
  Natural Language Processing (EMNLP-IJCNLP)}, pages 1475--1480, Hong Kong,
  China, 2019{\natexlab{a}}. Association for Computational Linguistics.
\newblock \doi{10.18653/v1/D19-1156}.
\newblock URL \url{https://www.aclweb.org/anthology/D19-1156}.

\bibitem[Cui et~al.(2018)Cui, Yang, Veit, Huang, and Belongie]{cui2018learning}
Yin Cui, Guandao Yang, Andreas Veit, Xun Huang, and Serge~J. Belongie.
\newblock Learning to evaluate image captioning.
\newblock In \emph{2018 {IEEE} Conference on Computer Vision and Pattern
  Recognition, {CVPR} 2018, Salt Lake City, UT, USA, June 18-22, 2018}, pages
  5804--5812. {IEEE} Computer Society, 2018.
\newblock \doi{10.1109/CVPR.2018.00608}.
\newblock URL
  \url{http://openaccess.thecvf.com/content\_cvpr\_2018/html/Cui\_Learning\_to\_Evaluate\_CVPR\_2018\_paper.html}.

\bibitem[Jiang et~al.(2019{\natexlab{b}})Jiang, Huang, Zhang, Wang, Zhang, Gan,
  Diesner, and Gao]{jiang2019tiger}
Ming Jiang, Qiuyuan Huang, Lei Zhang, Xin Wang, Pengchuan Zhang, Zhe Gan, Jana
  Diesner, and Jianfeng Gao.
\newblock {TIGE}r: Text-to-image grounding for image caption evaluation.
\newblock In \emph{Proceedings of the 2019 Conference on Empirical Methods in
  Natural Language Processing and the 9th International Joint Conference on
  Natural Language Processing (EMNLP-IJCNLP)}, pages 2141--2152, Hong Kong,
  China, 2019{\natexlab{b}}. Association for Computational Linguistics.
\newblock \doi{10.18653/v1/D19-1220}.
\newblock URL \url{https://www.aclweb.org/anthology/D19-1220}.

\bibitem[Lee et~al.(2020)Lee, Yoon, Dernoncourt, Kim, Bui, and
  Jung]{lee2020vilbertscore}
Hwanhee Lee, Seunghyun Yoon, Franck Dernoncourt, Doo~Soon Kim, Trung Bui, and
  Kyomin Jung.
\newblock {V}i{LBERTS}core: Evaluating image caption using vision-and-language
  {BERT}.
\newblock In \emph{Proceedings of the First Workshop on Evaluation and
  Comparison of NLP Systems}, pages 34--39, Online, 2020. Association for
  Computational Linguistics.
\newblock \doi{10.18653/v1/2020.eval4nlp-1.4}.
\newblock URL \url{https://www.aclweb.org/anthology/2020.eval4nlp-1.4}.

\bibitem[Wang et~al.(2021)Wang, Yao, Wang, Wu, and Chen]{wang2021faier}
Sijin Wang, Ziwei Yao, Ruiping Wang, Zhongqin Wu, and Xilin Chen.
\newblock Faier: Fidelity and adequacy ensured image caption evaluation.
\newblock In \emph{Proceedings of the IEEE/CVF Conference on Computer Vision
  and Pattern Recognition}, pages 14050--14059, 2021.

\bibitem[Madhyastha et~al.(2019)Madhyastha, Wang, and
  Specia]{madhyastha2019vifidel}
Pranava Madhyastha, Josiah Wang, and Lucia Specia.
\newblock {VIFIDEL}: Evaluating the visual fidelity of image descriptions.
\newblock In \emph{Proceedings of the 57th Annual Meeting of the Association
  for Computational Linguistics}, pages 6539--6550, Florence, Italy, 2019.
  Association for Computational Linguistics.
\newblock \doi{10.18653/v1/P19-1654}.
\newblock URL \url{https://www.aclweb.org/anthology/P19-1654}.

\bibitem[Lee et~al.(2021)Lee, Yoon, Dernoncourt, Bui, and Jung]{lee2021umic}
Hwanhee Lee, Seunghyun Yoon, Franck Dernoncourt, Trung Bui, and Kyomin Jung.
\newblock Umic: An unreferenced metric for image captioning via contrastive
  learning.
\newblock \emph{arXiv preprint arXiv:2106.14019}, 2021.

\bibitem[Luo et~al.(2018)Luo, Price, Cohen, and
  Shakhnarovich]{luo2018discriminability}
Ruotian Luo, Brian~L. Price, Scott Cohen, and Gregory Shakhnarovich.
\newblock Discriminability objective for training descriptive captions.
\newblock In \emph{2018 {IEEE} Conference on Computer Vision and Pattern
  Recognition, {CVPR} 2018, Salt Lake City, UT, USA, June 18-22, 2018}, pages
  6964--6974. {IEEE} Computer Society, 2018.
\newblock \doi{10.1109/CVPR.2018.00728}.
\newblock URL
  \url{http://openaccess.thecvf.com/content\_cvpr\_2018/html/Luo\_Discriminability\_Objective\_for\_CVPR\_2018\_paper.html}.

\bibitem[Rohrbach et~al.(2018)Rohrbach, Hendricks, Burns, Darrell, and
  Saenko]{rohrbach2018object}
Anna Rohrbach, Lisa~Anne Hendricks, Kaylee Burns, Trevor Darrell, and Kate
  Saenko.
\newblock Object hallucination in image captioning.
\newblock In \emph{Proceedings of the 2018 Conference on Empirical Methods in
  Natural Language Processing}, pages 4035--4045, Brussels, Belgium, 2018.
  Association for Computational Linguistics.
\newblock \doi{10.18653/v1/D18-1437}.
\newblock URL \url{https://www.aclweb.org/anthology/D18-1437}.

\bibitem[Levinboim et~al.(2021)Levinboim, Thapliyal, Sharma, and
  Soricut]{levinboim2019quality}
Tomer Levinboim, Ashish~V. Thapliyal, Piyush Sharma, and Radu Soricut.
\newblock Quality estimation for image captions based on large-scale human
  evaluations.
\newblock In \emph{Proceedings of the 2021 Conference of the North American
  Chapter of the Association for Computational Linguistics: Human Language
  Technologies}, pages 3157--3166, Online, 2021. Association for Computational
  Linguistics.
\newblock URL \url{https://www.aclweb.org/anthology/2021.naacl-main.253}.

\bibitem[Vedantam et~al.(2017)Vedantam, Bengio, Murphy, Parikh, and
  Chechik]{vedantam2017context}
Ramakrishna Vedantam, Samy Bengio, Kevin Murphy, Devi Parikh, and Gal Chechik.
\newblock Context-aware captions from context-agnostic supervision.
\newblock In \emph{2017 {IEEE} Conference on Computer Vision and Pattern
  Recognition, {CVPR} 2017, Honolulu, HI, USA, July 21-26, 2017}, pages
  1070--1079. {IEEE} Computer Society, 2017.
\newblock \doi{10.1109/CVPR.2017.120}.
\newblock URL \url{https://doi.org/10.1109/CVPR.2017.120}.

\bibitem[Winograd(1972)]{winograd1972understanding}
Terry Winograd.
\newblock Understanding natural language.
\newblock \emph{Cognitive psychology}, 3\penalty0 (1):\penalty0 1--191, 1972.

\bibitem[Mitchell et~al.(2013)Mitchell, van Deemter, and
  Reiter]{mitchell2013generating}
Margaret Mitchell, Kees van Deemter, and Ehud Reiter.
\newblock Generating expressions that refer to visible objects.
\newblock In \emph{Proceedings of the 2013 Conference of the North {A}merican
  Chapter of the Association for Computational Linguistics: Human Language
  Technologies}, pages 1174--1184, Atlanta, Georgia, 2013. Association for
  Computational Linguistics.
\newblock URL \url{https://www.aclweb.org/anthology/N13-1137}.

\bibitem[Golland et~al.(2010)Golland, Liang, and Klein]{golland2010game}
Dave Golland, Percy Liang, and Dan Klein.
\newblock A game-theoretic approach to generating spatial descriptions.
\newblock In \emph{Proceedings of the 2010 Conference on Empirical Methods in
  Natural Language Processing}, pages 410--419, Cambridge, MA, 2010.
  Association for Computational Linguistics.
\newblock URL \url{https://www.aclweb.org/anthology/D10-1040}.

\bibitem[Krahmer and van Deemter(2012)]{krahmer2012computational}
Emiel Krahmer and Kees van Deemter.
\newblock Computational generation of referring expressions: A survey.
\newblock \emph{Computational Linguistics}, 38\penalty0 (1):\penalty0 173--218,
  2012.
\newblock \doi{10.1162/COLI_a_00088}.
\newblock URL \url{https://www.aclweb.org/anthology/J12-1006}.

\bibitem[FitzGerald et~al.(2013)FitzGerald, Artzi, and
  Zettlemoyer]{fitzgerald2013learning}
Nicholas FitzGerald, Yoav Artzi, and Luke Zettlemoyer.
\newblock Learning distributions over logical forms for referring expression
  generation.
\newblock In \emph{Proceedings of the 2013 Conference on Empirical Methods in
  Natural Language Processing}, pages 1914--1925, Seattle, Washington, USA,
  2013. Association for Computational Linguistics.
\newblock URL \url{https://www.aclweb.org/anthology/D13-1197}.

\bibitem[Kazemzadeh et~al.(2014)Kazemzadeh, Ordonez, Matten, and
  Berg]{kazemzadeh2014referitgame}
Sahar Kazemzadeh, Vicente Ordonez, Mark Matten, and Tamara Berg.
\newblock {R}efer{I}t{G}ame: Referring to objects in photographs of natural
  scenes.
\newblock In \emph{Proceedings of the 2014 Conference on Empirical Methods in
  Natural Language Processing ({EMNLP})}, pages 787--798, Doha, Qatar, 2014.
  Association for Computational Linguistics.
\newblock \doi{10.3115/v1/D14-1086}.
\newblock URL \url{https://www.aclweb.org/anthology/D14-1086}.

\bibitem[Yu et~al.(2016)Yu, Poirson, Yang, Berg, and Berg]{yu2016modeling}
Licheng Yu, Patrick Poirson, Shan Yang, Alexander~C Berg, and Tamara~L Berg.
\newblock Modeling context in referring expressions.
\newblock In \emph{European Conference on Computer Vision}, pages 69--85.
  Springer, 2016.

\bibitem[Lin et~al.(2014{\natexlab{a}})Lin, Maire, Belongie, Hays, Perona,
  Ramanan, Doll{\'a}r, and Zitnick]{lin2014microsoft}
Tsung-Yi Lin, Michael Maire, Serge Belongie, James Hays, Pietro Perona, Deva
  Ramanan, Piotr Doll{\'a}r, and C~Lawrence Zitnick.
\newblock Microsoft coco: Common objects in context.
\newblock In \emph{European Conference on Computer Vision}, pages 740--755.
  Springer, 2014{\natexlab{a}}.

\bibitem[Mao et~al.(2016{\natexlab{b}})Mao, Huang, Toshev, Camburu, Yuille, and
  Murphy]{Mao2016}
Junhua Mao, Jonathan Huang, Alexander Toshev, Oana Camburu, Alan~L. Yuille, and
  Kevin Murphy.
\newblock Generation and comprehension of unambiguous object descriptions.
\newblock In \emph{2016 {IEEE} Conference on Computer Vision and Pattern
  Recognition, {CVPR} 2016, Las Vegas, NV, USA, June 27-30, 2016}, pages
  11--20. {IEEE} Computer Society, 2016{\natexlab{b}}.
\newblock \doi{10.1109/CVPR.2016.9}.
\newblock URL \url{https://doi.org/10.1109/CVPR.2016.9}.

\bibitem[Liu et~al.(2019{\natexlab{b}})Liu, Liu, Bai, and Yuille]{liu2019clevr}
Runtao Liu, Chenxi Liu, Yutong Bai, and Alan~L. Yuille.
\newblock Clevr-ref+: Diagnosing visual reasoning with referring expressions.
\newblock In \emph{{IEEE} Conference on Computer Vision and Pattern
  Recognition, {CVPR} 2019, Long Beach, CA, USA, June 16-20, 2019}, pages
  4185--4194. Computer Vision Foundation / {IEEE}, 2019{\natexlab{b}}.
\newblock \doi{10.1109/CVPR.2019.00431}.
\newblock URL
  \url{http://openaccess.thecvf.com/content\_CVPR\_2019/html/Liu\_CLEVR-Ref\_Diagnosing\_Visual\_Reasoning\_With\_Referring\_Expressions\_CVPR\_2019\_paper.html}.

\bibitem[Tanaka et~al.(2019)Tanaka, Itamochi, Narioka, Sato, Ushiku, and
  Harada]{tanaka2019generating}
Mikihiro Tanaka, Takayuki Itamochi, Kenichi Narioka, Ikuro Sato, Yoshitaka
  Ushiku, and Tatsuya Harada.
\newblock Generating easy-to-understand referring expressions for target
  identifications.
\newblock In \emph{2019 {IEEE/CVF} International Conference on Computer Vision,
  {ICCV} 2019, Seoul, Korea (South), October 27 - November 2, 2019}, pages
  5793--5802. {IEEE}, 2019.
\newblock \doi{10.1109/ICCV.2019.00589}.
\newblock URL \url{https://doi.org/10.1109/ICCV.2019.00589}.

\bibitem[Krishna et~al.(2019)Krishna, Bernstein, and
  Fei{-}Fei]{krishna2019information}
Ranjay Krishna, Michael Bernstein, and Li~Fei{-}Fei.
\newblock Information maximizing visual question generation.
\newblock In \emph{{IEEE} Conference on Computer Vision and Pattern
  Recognition, {CVPR} 2019, Long Beach, CA, USA, June 16-20, 2019}, pages
  2008--2018. Computer Vision Foundation / {IEEE}, 2019.
\newblock \doi{10.1109/CVPR.2019.00211}.
\newblock URL
  \url{http://openaccess.thecvf.com/content\_CVPR\_2019/html/Krishna\_Information\_Maximizing\_Visual\_Question\_Generation\_CVPR\_2019\_paper.html}.

\bibitem[Hendricks et~al.(2016{\natexlab{b}})Hendricks, Akata, Rohrbach,
  Donahue, Schiele, and Darrell]{hendricks2016generating}
Lisa~Anne Hendricks, Zeynep Akata, Marcus Rohrbach, Jeff Donahue, Bernt
  Schiele, and Trevor Darrell.
\newblock Generating visual explanations.
\newblock In \emph{European Conference on Computer Vision}, pages 3--19.
  Springer, 2016{\natexlab{b}}.

\bibitem[Chen and Zitnick(2015)]{chen2015mind}
Xinlei Chen and C.~Lawrence Zitnick.
\newblock Mind's eye: {A} recurrent visual representation for image caption
  generation.
\newblock In \emph{{IEEE} Conference on Computer Vision and Pattern
  Recognition, {CVPR} 2015, Boston, MA, USA, June 7-12, 2015}, pages
  2422--2431. {IEEE} Computer Society, 2015.
\newblock \doi{10.1109/CVPR.2015.7298856}.
\newblock URL \url{https://doi.org/10.1109/CVPR.2015.7298856}.

\bibitem[Andreas and Klein(2016{\natexlab{a}})]{andreas2016reasoning}
Jacob Andreas and Dan Klein.
\newblock Reasoning about pragmatics with neural listeners and speakers.
\newblock In \emph{Proceedings of the 2016 Conference on Empirical Methods in
  Natural Language Processing}, pages 1173--1182, Austin, Texas,
  2016{\natexlab{a}}. Association for Computational Linguistics.
\newblock \doi{10.18653/v1/D16-1125}.
\newblock URL \url{https://www.aclweb.org/anthology/D16-1125}.

\bibitem[Cohn-Gordon et~al.(2018)Cohn-Gordon, Goodman, and
  Potts]{cohn2018pragmatically}
Reuben Cohn-Gordon, Noah Goodman, and Christopher Potts.
\newblock Pragmatically informative image captioning with character-level
  inference.
\newblock In \emph{Proceedings of the 2018 Conference of the North {A}merican
  Chapter of the Association for Computational Linguistics: Human Language
  Technologies, Volume 2 (Short Papers)}, pages 439--443, New Orleans,
  Louisiana, 2018. Association for Computational Linguistics.
\newblock \doi{10.18653/v1/N18-2070}.
\newblock URL \url{https://www.aclweb.org/anthology/N18-2070}.

\bibitem[Liu et~al.(2018{\natexlab{a}})Liu, Li, Shao, Chen, and
  Wang]{liu2018show}
Xihui Liu, Hongsheng Li, Jing Shao, Dapeng Chen, and Xiaogang Wang.
\newblock Show, tell and discriminate: Image captioning by self-retrieval with
  partially labeled data.
\newblock In \emph{Proceedings of the European Conference on Computer Vision
  (ECCV)}, pages 338--354, 2018{\natexlab{a}}.

\bibitem[Das et~al.(2017{\natexlab{a}})Das, Kottur, Moura, Lee, and
  Batra]{das2017learning}
Abhishek Das, Satwik Kottur, Jos{\'{e}} M.~F. Moura, Stefan Lee, and Dhruv
  Batra.
\newblock Learning cooperative visual dialog agents with deep reinforcement
  learning.
\newblock In \emph{{IEEE} International Conference on Computer Vision, {ICCV}
  2017, Venice, Italy, October 22-29, 2017}, pages 2970--2979. {IEEE} Computer
  Society, 2017{\natexlab{a}}.
\newblock \doi{10.1109/ICCV.2017.321}.
\newblock URL \url{https://doi.org/10.1109/ICCV.2017.321}.

\bibitem[Holtzman et~al.(2018)Holtzman, Buys, Forbes, Bosselut, Golub, and
  Choi]{holtzman2018learning}
Ari Holtzman, Jan Buys, Maxwell Forbes, Antoine Bosselut, David Golub, and
  Yejin Choi.
\newblock Learning to write with cooperative discriminators.
\newblock In \emph{Proceedings of the 56th Annual Meeting of the Association
  for Computational Linguistics (Volume 1: Long Papers)}, pages 1638--1649,
  Melbourne, Australia, 2018. Association for Computational Linguistics.
\newblock \doi{10.18653/v1/P18-1152}.
\newblock URL \url{https://www.aclweb.org/anthology/P18-1152}.

\bibitem[Bracha and Chechik(2019)]{bracha2019informative}
Lior Bracha and Gal Chechik.
\newblock Informative object annotations: Tell me something {I} don't know.
\newblock In \emph{{IEEE} Conference on Computer Vision and Pattern
  Recognition, {CVPR} 2019, Long Beach, CA, USA, June 16-20, 2019}, pages
  12507--12515. Computer Vision Foundation / {IEEE}, 2019.
\newblock \doi{10.1109/CVPR.2019.01279}.
\newblock URL
  \url{http://openaccess.thecvf.com/content\_CVPR\_2019/html/Bracha\_Informative\_Object\_Annotations\_Tell\_Me\_Something\_I\_Dont\_Know\_CVPR\_2019\_paper.html}.

\bibitem[Sun et~al.(2020)Sun, Fan, Sun, Meng, Wu, and Li]{sun2020summarize}
Xiaofei Sun, Chun Fan, Zijun Sun, Yuxian Meng, Fei Wu, and Jiwei Li.
\newblock Summarize, outline, and elaborate: Long-text generation via
  hierarchical supervision from extractive summaries.
\newblock \emph{arXiv preprint arXiv:2010.07074}, 2020.

\bibitem[Fisch et~al.(2020)Fisch, Lee, Chang, Clark, and
  Barzilay]{fisch2020capwap}
Adam Fisch, Kenton Lee, Ming-Wei Chang, Jonathan~H Clark, and Regina Barzilay.
\newblock Capwap: Captioning with a purpose.
\newblock \emph{arXiv preprint arXiv:2011.04264}, 2020.

\bibitem[Hochreiter and Schmidhuber(1997)]{hochreiter1997long}
Sepp Hochreiter and J{\"u}rgen Schmidhuber.
\newblock Long short-term memory.
\newblock \emph{Neural computation}, 9\penalty0 (8):\penalty0 1735--1780, 1997.

\bibitem[Vaswani et~al.(2017)Vaswani, Shazeer, Parmar, Uszkoreit, Jones, Gomez,
  Kaiser, and Polosukhin]{vaswani2017attention}
Ashish Vaswani, Noam Shazeer, Niki Parmar, Jakob Uszkoreit, Llion Jones,
  Aidan~N. Gomez, Lukasz Kaiser, and Illia Polosukhin.
\newblock Attention is all you need.
\newblock In Isabelle Guyon, Ulrike von Luxburg, Samy Bengio, Hanna~M. Wallach,
  Rob Fergus, S.~V.~N. Vishwanathan, and Roman Garnett, editors, \emph{Advances
  in Neural Information Processing Systems 30: Annual Conference on Neural
  Information Processing Systems 2017, December 4-9, 2017, Long Beach, CA,
  {USA}}, pages 5998--6008, 2017.
\newblock URL
  \url{https://proceedings.neurips.cc/paper/2017/hash/3f5ee243547dee91fbd053c1c4a845aa-Abstract.html}.

\bibitem[Li et~al.(2019)Li, Zhu, Liu, and Yang]{li2019entangled}
Guang Li, Linchao Zhu, Ping Liu, and Yi~Yang.
\newblock Entangled transformer for image captioning.
\newblock In \emph{2019 {IEEE/CVF} International Conference on Computer Vision,
  {ICCV} 2019, Seoul, Korea (South), October 27 - November 2, 2019}, pages
  8927--8936. {IEEE}, 2019.
\newblock \doi{10.1109/ICCV.2019.00902}.
\newblock URL \url{https://doi.org/10.1109/ICCV.2019.00902}.

\bibitem[Herdade et~al.(2019)Herdade, Kappeler, Boakye, and
  Soares]{herdade2019image}
Simao Herdade, Armin Kappeler, Kofi Boakye, and Joao Soares.
\newblock Image captioning: Transforming objects into words.
\newblock In Hanna~M. Wallach, Hugo Larochelle, Alina Beygelzimer, Florence
  d'Alch{\'{e}}{-}Buc, Emily~B. Fox, and Roman Garnett, editors, \emph{Advances
  in Neural Information Processing Systems 32: Annual Conference on Neural
  Information Processing Systems 2019, NeurIPS 2019, December 8-14, 2019,
  Vancouver, BC, Canada}, pages 11135--11145, 2019.
\newblock URL
  \url{https://proceedings.neurips.cc/paper/2019/hash/680390c55bbd9ce416d1d69a9ab4760d-Abstract.html}.

\bibitem[Aneja et~al.(2018)Aneja, Deshpande, and
  Schwing]{aneja2018convolutional}
Jyoti Aneja, Aditya Deshpande, and Alexander~G. Schwing.
\newblock Convolutional image captioning.
\newblock In \emph{2018 {IEEE} Conference on Computer Vision and Pattern
  Recognition, {CVPR} 2018, Salt Lake City, UT, USA, June 18-22, 2018}, pages
  5561--5570. {IEEE} Computer Society, 2018.
\newblock \doi{10.1109/CVPR.2018.00583}.
\newblock URL
  \url{http://openaccess.thecvf.com/content\_cvpr\_2018/html/Aneja\_Convolutional\_Image\_Captioning\_CVPR\_2018\_paper.html}.

\bibitem[Gu et~al.(2017)Gu, Wang, Cai, and Chen]{gu2017empirical}
Jiuxiang Gu, Gang Wang, Jianfei Cai, and Tsuhan Chen.
\newblock An empirical study of language {CNN} for image captioning.
\newblock In \emph{{IEEE} International Conference on Computer Vision, {ICCV}
  2017, Venice, Italy, October 22-29, 2017}, pages 1231--1240. {IEEE} Computer
  Society, 2017.
\newblock \doi{10.1109/ICCV.2017.138}.
\newblock URL \url{https://doi.org/10.1109/ICCV.2017.138}.

\bibitem[Bengio et~al.(2015)Bengio, Vinyals, Jaitly, and Shazeer]{Bengio2015}
Samy Bengio, Oriol Vinyals, Navdeep Jaitly, and Noam Shazeer.
\newblock Scheduled sampling for sequence prediction with recurrent neural
  networks.
\newblock In Corinna Cortes, Neil~D. Lawrence, Daniel~D. Lee, Masashi Sugiyama,
  and Roman Garnett, editors, \emph{Advances in Neural Information Processing
  Systems 28: Annual Conference on Neural Information Processing Systems 2015,
  December 7-12, 2015, Montreal, Quebec, Canada}, pages 1171--1179, 2015.
\newblock URL
  \url{https://proceedings.neurips.cc/paper/2015/hash/e995f98d56967d946471af29d7bf99f1-Abstract.html}.

\bibitem[Szegedy et~al.(2016)Szegedy, Vanhoucke, Ioffe, Shlens, and
  Wojna]{szegedy2016rethinking}
Christian Szegedy, Vincent Vanhoucke, Sergey Ioffe, Jonathon Shlens, and
  Zbigniew Wojna.
\newblock Rethinking the inception architecture for computer vision.
\newblock In \emph{2016 {IEEE} Conference on Computer Vision and Pattern
  Recognition, {CVPR} 2016, Las Vegas, NV, USA, June 27-30, 2016}, pages
  2818--2826. {IEEE} Computer Society, 2016.
\newblock \doi{10.1109/CVPR.2016.308}.
\newblock URL \url{https://doi.org/10.1109/CVPR.2016.308}.

\bibitem[M{\"{u}}ller et~al.(2019)M{\"{u}}ller, Kornblith, and
  Hinton]{muller2019does}
Rafael M{\"{u}}ller, Simon Kornblith, and Geoffrey~E. Hinton.
\newblock When does label smoothing help?
\newblock In Hanna~M. Wallach, Hugo Larochelle, Alina Beygelzimer, Florence
  d'Alch{\'{e}}{-}Buc, Emily~B. Fox, and Roman Garnett, editors, \emph{Advances
  in Neural Information Processing Systems 32: Annual Conference on Neural
  Information Processing Systems 2019, NeurIPS 2019, December 8-14, 2019,
  Vancouver, BC, Canada}, pages 4696--4705, 2019.
\newblock URL
  \url{https://proceedings.neurips.cc/paper/2019/hash/f1748d6b0fd9d439f71450117eba2725-Abstract.html}.

\bibitem[Yao et~al.(2017{\natexlab{a}})Yao, Pan, Li, Qiu, and
  Mei]{yao2017boosting}
Ting Yao, Yingwei Pan, Yehao Li, Zhaofan Qiu, and Tao Mei.
\newblock Boosting image captioning with attributes.
\newblock In \emph{{IEEE} International Conference on Computer Vision, {ICCV}
  2017, Venice, Italy, October 22-29, 2017}, pages 4904--4912. {IEEE} Computer
  Society, 2017{\natexlab{a}}.
\newblock \doi{10.1109/ICCV.2017.524}.
\newblock URL \url{https://doi.org/10.1109/ICCV.2017.524}.

\bibitem[Dognin et~al.(2019)Dognin, Melnyk, Mroueh, Ross, and
  Sercu]{dognin2019adversarial}
Pierre~L. Dognin, Igor Melnyk, Youssef Mroueh, Jerret Ross, and Tom Sercu.
\newblock Adversarial semantic alignment for improved image captions.
\newblock In \emph{{IEEE} Conference on Computer Vision and Pattern
  Recognition, {CVPR} 2019, Long Beach, CA, USA, June 16-20, 2019}, pages
  10463--10471. Computer Vision Foundation / {IEEE}, 2019.
\newblock \doi{10.1109/CVPR.2019.01071}.
\newblock URL
  \url{http://openaccess.thecvf.com/content\_CVPR\_2019/html/Dognin\_Adversarial\_Semantic\_Alignment\_for\_Improved\_Image\_Captions\_CVPR\_2019\_paper.html}.

\bibitem[Jiang et~al.(2018)Jiang, Ma, Jiang, Liu, and
  Zhang]{jiang2018recurrent}
Wenhao Jiang, Lin Ma, Yu-Gang Jiang, Wei Liu, and Tong Zhang.
\newblock Recurrent fusion network for image captioning.
\newblock In \emph{Proceedings of the European Conference on Computer Vision
  (ECCV)}, pages 499--515, 2018.

\bibitem[Ling and Fidler(2017{\natexlab{a}})]{ling2017teaching}
Huan Ling and Sanja Fidler.
\newblock Teaching machines to describe images via natural language feedback.
\newblock In \emph{Proceedings of the 31st International Conference on Neural
  Information Processing Systems}, pages 5075--5085. Curran Associates Inc.,
  2017{\natexlab{a}}.

\bibitem[Liu et~al.(2018{\natexlab{b}})Liu, Zha, Zhang, Zhang, and
  Wu]{liu2018context}
Daqing Liu, Zheng{-}Jun Zha, Hanwang Zhang, Yongdong Zhang, and Feng Wu.
\newblock Context-aware visual policy network for sequence-level image
  captioning.
\newblock In \emph{2018 {ACM} Multimedia Conference on Multimedia Conference,
  {MM} 2018, Seoul, Republic of Korea, October 22-26, 2018}, pages 1416--1424,
  2018{\natexlab{b}}.
\newblock \doi{10.1145/3240508.3240632}.
\newblock URL \url{https://doi.org/10.1145/3240508.3240632}.

\bibitem[Chen et~al.(2019{\natexlab{b}})Chen, Mu, Xiao, Ye, Wu, and
  Ju]{chen2019improving}
Chen Chen, Shuai Mu, Wanpeng Xiao, Zexiong Ye, Liesi Wu, and Qi~Ju.
\newblock Improving image captioning with conditional generative adversarial
  nets.
\newblock In \emph{The Thirty-Third {AAAI} Conference on Artificial
  Intelligence, {AAAI} 2019, The Thirty-First Innovative Applications of
  Artificial Intelligence Conference, {IAAI} 2019, The Ninth {AAAI} Symposium
  on Educational Advances in Artificial Intelligence, {EAAI} 2019, Honolulu,
  Hawaii, USA, January 27 - February 1, 2019}, pages 8142--8150. {AAAI} Press,
  2019{\natexlab{b}}.
\newblock \doi{10.1609/aaai.v33i01.33018142}.
\newblock URL \url{https://doi.org/10.1609/aaai.v33i01.33018142}.

\bibitem[Zhao et~al.(2018)Zhao, Wang, Ye, Yang, Zhao, Luo, and
  Qiao]{zhao2018multi}
Wei Zhao, Benyou Wang, Jianbo Ye, Min Yang, Zhou Zhao, Ruotian Luo, and
  Yu~Qiao.
\newblock A multi-task learning approach for image captioning.
\newblock In J{\'{e}}r{\^{o}}me Lang, editor, \emph{Proceedings of the
  Twenty-Seventh International Joint Conference on Artificial Intelligence,
  {IJCAI} 2018, July 13-19, 2018, Stockholm, Sweden}, pages 1205--1211.
  ijcai.org, 2018.
\newblock \doi{10.24963/ijcai.2018/168}.
\newblock URL \url{https://doi.org/10.24963/ijcai.2018/168}.

\bibitem[Li et~al.(2018{\natexlab{a}})Li, Yao, Pan, Chao, and
  Mei]{li2018jointly}
Yehao Li, Ting Yao, Yingwei Pan, Hongyang Chao, and Tao Mei.
\newblock Jointly localizing and describing events for dense video captioning.
\newblock In \emph{2018 {IEEE} Conference on Computer Vision and Pattern
  Recognition, {CVPR} 2018, Salt Lake City, UT, USA, June 18-22, 2018}, pages
  7492--7500. {IEEE} Computer Society, 2018{\natexlab{a}}.
\newblock \doi{10.1109/CVPR.2018.00782}.
\newblock URL
  \url{http://openaccess.thecvf.com/content\_cvpr\_2018/html/Li\_Jointly\_Localizing\_and\_CVPR\_2018\_paper.html}.

\bibitem[Chen et~al.(2018)Chen, Wang, Zhang, and Huang]{chen2018less}
Yangyu Chen, Shuhui Wang, Weigang Zhang, and Qingming Huang.
\newblock Less is more: Picking informative frames for video captioning.
\newblock In \emph{Proceedings of the European Conference on Computer Vision
  (ECCV)}, pages 358--373, 2018.

\bibitem[Li and Gong(2019)]{li2019end}
Lijun Li and Boqing Gong.
\newblock End-to-end video captioning with multitask reinforcement learning.
\newblock In \emph{2019 IEEE Winter Conference on Applications of Computer
  Vision (WACV)}, pages 339--348. IEEE, 2019.

\bibitem[Hu et~al.(2018)Hu, Peng, Huang, Qiu, Wei, and Zhou]{hu2017reinforced}
Minghao Hu, Yuxing Peng, Zhen Huang, Xipeng Qiu, Furu Wei, and Ming Zhou.
\newblock Reinforced mnemonic reader for machine reading comprehension.
\newblock In J{\'{e}}r{\^{o}}me Lang, editor, \emph{Proceedings of the
  Twenty-Seventh International Joint Conference on Artificial Intelligence,
  {IJCAI} 2018, July 13-19, 2018, Stockholm, Sweden}, pages 4099--4106.
  ijcai.org, 2018.
\newblock \doi{10.24963/ijcai.2018/570}.
\newblock URL \url{https://doi.org/10.24963/ijcai.2018/570}.

\bibitem[Celikyilmaz et~al.(2018)Celikyilmaz, Bosselut, He, and
  Choi]{celikyilmaz2018deep}
Asli Celikyilmaz, Antoine Bosselut, Xiaodong He, and Yejin Choi.
\newblock Deep communicating agents for abstractive summarization.
\newblock In \emph{Proceedings of the 2018 Conference of the North {A}merican
  Chapter of the Association for Computational Linguistics: Human Language
  Technologies, Volume 1 (Long Papers)}, pages 1662--1675, New Orleans,
  Louisiana, 2018. Association for Computational Linguistics.
\newblock \doi{10.18653/v1/N18-1150}.
\newblock URL \url{https://www.aclweb.org/anthology/N18-1150}.

\bibitem[Paulus et~al.(2018)Paulus, Xiong, and Socher]{paulus2017deep}
Romain Paulus, Caiming Xiong, and Richard Socher.
\newblock A deep reinforced model for abstractive summarization.
\newblock In \emph{6th International Conference on Learning Representations,
  {ICLR} 2018, Vancouver, BC, Canada, April 30 - May 3, 2018, Conference Track
  Proceedings}. OpenReview.net, 2018.
\newblock URL \url{https://openreview.net/forum?id=HkAClQgA-}.

\bibitem[Wang et~al.(2018)Wang, Yao, Tao, Zhong, Liu, and
  Du]{wang2018reinforced}
Li~Wang, Junlin Yao, Yunzhe Tao, Li~Zhong, Wei Liu, and Qiang Du.
\newblock A reinforced topic-aware convolutional sequence-to-sequence model for
  abstractive text summarization.
\newblock In J{\'{e}}r{\^{o}}me Lang, editor, \emph{Proceedings of the
  Twenty-Seventh International Joint Conference on Artificial Intelligence,
  {IJCAI} 2018, July 13-19, 2018, Stockholm, Sweden}, pages 4453--4460.
  ijcai.org, 2018.
\newblock \doi{10.24963/ijcai.2018/619}.
\newblock URL \url{https://doi.org/10.24963/ijcai.2018/619}.

\bibitem[Pasunuru and Bansal(2018)]{pasunuru2018multi}
Ramakanth Pasunuru and Mohit Bansal.
\newblock Multi-reward reinforced summarization with saliency and entailment.
\newblock In \emph{Proceedings of the 2018 Conference of the North {A}merican
  Chapter of the Association for Computational Linguistics: Human Language
  Technologies, Volume 2 (Short Papers)}, pages 646--653, New Orleans,
  Louisiana, 2018. Association for Computational Linguistics.
\newblock \doi{10.18653/v1/N18-2102}.
\newblock URL \url{https://www.aclweb.org/anthology/N18-2102}.

\bibitem[Melas-Kyriazi et~al.(2018)Melas-Kyriazi, Rush, and
  Han]{melas2018training}
Luke Melas-Kyriazi, Alexander Rush, and George Han.
\newblock Training for diversity in image paragraph captioning.
\newblock In \emph{Proceedings of the 2018 Conference on Empirical Methods in
  Natural Language Processing}, pages 757--761, Brussels, Belgium, 2018.
  Association for Computational Linguistics.
\newblock \doi{10.18653/v1/D18-1084}.
\newblock URL \url{https://www.aclweb.org/anthology/D18-1084}.

\bibitem[Zhou et~al.(2018{\natexlab{b}})Zhou, Xiong, and
  Socher]{zhou2018improving}
Yingbo Zhou, Caiming Xiong, and Richard Socher.
\newblock Improving end-to-end speech recognition with policy learning.
\newblock In \emph{2018 {IEEE} International Conference on Acoustics, Speech
  and Signal Processing, {ICASSP} 2018, Calgary, AB, Canada, April 15-20,
  2018}, pages 5819--5823. {IEEE}, 2018{\natexlab{b}}.
\newblock \doi{10.1109/ICASSP.2018.8462361}.
\newblock URL \url{https://doi.org/10.1109/ICASSP.2018.8462361}.

\bibitem[Gao et~al.(2019)Gao, Wang, Wang, Ma, and Gao]{gao2019self}
Junlong Gao, Shiqi Wang, Shanshe Wang, Siwei Ma, and Wen Gao.
\newblock Self-critical n-step training for image captioning.
\newblock In \emph{{IEEE} Conference on Computer Vision and Pattern
  Recognition, {CVPR} 2019, Long Beach, CA, USA, June 16-20, 2019}, pages
  6300--6308. Computer Vision Foundation / {IEEE}, 2019.
\newblock \doi{10.1109/CVPR.2019.00646}.
\newblock URL
  \url{http://openaccess.thecvf.com/content\_CVPR\_2019/html/Gao\_Self-Critical\_N-Step\_Training\_for\_Image\_Captioning\_CVPR\_2019\_paper.html}.

\bibitem[Luo(2020)]{luo2020better}
Ruotian Luo.
\newblock A better variant of self-critical sequence training.
\newblock \emph{arXiv preprint arXiv:2003.09971}, 2020.

\bibitem[Ding and Soricut(2017)]{ding2017cold}
Nan Ding and Radu Soricut.
\newblock Cold-start reinforcement learning with softmax policy gradient.
\newblock In Isabelle Guyon, Ulrike von Luxburg, Samy Bengio, Hanna~M. Wallach,
  Rob Fergus, S.~V.~N. Vishwanathan, and Roman Garnett, editors, \emph{Advances
  in Neural Information Processing Systems 30: Annual Conference on Neural
  Information Processing Systems 2017, December 4-9, 2017, Long Beach, CA,
  {USA}}, pages 2817--2826, 2017.
\newblock URL
  \url{https://proceedings.neurips.cc/paper/2017/hash/faafda66202d234463057972460c04f5-Abstract.html}.

\bibitem[Seo et~al.(2020)Seo, Sharma, Levinboim, Han, and
  Soricut]{seo2020reinforcing}
Paul~Hongsuck Seo, Piyush Sharma, Tomer Levinboim, Bohyung Han, and Radu
  Soricut.
\newblock Reinforcing an image caption generator using off-line human feedback.
\newblock In \emph{Proceedings of the AAAI Conference on Artificial
  Intelligence}, volume~34, pages 2693--2700, 2020.

\bibitem[Williams(1992)]{williams1992simple}
Ronald~J Williams.
\newblock Simple statistical gradient-following algorithms for connectionist
  reinforcement learning.
\newblock \emph{Machine learning}, 8\penalty0 (3-4):\penalty0 229--256, 1992.

\bibitem[Yao et~al.(2017{\natexlab{b}})Yao, Pan, Li, Qiu, and
  Mei]{yao2016boosting}
Ting Yao, Yingwei Pan, Yehao Li, Zhaofan Qiu, and Tao Mei.
\newblock Boosting image captioning with attributes.
\newblock In \emph{{IEEE} International Conference on Computer Vision, {ICCV}
  2017, Venice, Italy, October 22-29, 2017}, pages 4904--4912. {IEEE} Computer
  Society, 2017{\natexlab{b}}.
\newblock \doi{10.1109/ICCV.2017.524}.
\newblock URL \url{https://doi.org/10.1109/ICCV.2017.524}.

\bibitem[Edunov et~al.(2018)Edunov, Ott, Auli, Grangier, and
  Ranzato]{edunov2017classical}
Sergey Edunov, Myle Ott, Michael Auli, David Grangier, and Marc{'}Aurelio
  Ranzato.
\newblock Classical structured prediction losses for sequence to sequence
  learning.
\newblock In \emph{Proceedings of the 2018 Conference of the North {A}merican
  Chapter of the Association for Computational Linguistics: Human Language
  Technologies, Volume 1 (Long Papers)}, pages 355--364, New Orleans,
  Louisiana, 2018. Association for Computational Linguistics.
\newblock \doi{10.18653/v1/N18-1033}.
\newblock URL \url{https://www.aclweb.org/anthology/N18-1033}.

\bibitem[Welleck et~al.(2020)Welleck, Kulikov, Roller, Dinan, Cho, and
  Weston]{welleck2019neural}
Sean Welleck, Ilia Kulikov, Stephen Roller, Emily Dinan, Kyunghyun Cho, and
  Jason Weston.
\newblock Neural text generation with unlikelihood training.
\newblock In \emph{8th International Conference on Learning Representations,
  {ICLR} 2020, Addis Ababa, Ethiopia, April 26-30, 2020}. OpenReview.net, 2020.
\newblock URL \url{https://openreview.net/forum?id=SJeYe0NtvH}.

\bibitem[Gu et~al.(2018{\natexlab{b}})Gu, Bradbury, Xiong, Li, and
  Socher]{gu2017non}
Jiatao Gu, James Bradbury, Caiming Xiong, Victor O.~K. Li, and Richard Socher.
\newblock Non-autoregressive neural machine translation.
\newblock In \emph{6th International Conference on Learning Representations,
  {ICLR} 2018, Vancouver, BC, Canada, April 30 - May 3, 2018, Conference Track
  Proceedings}. OpenReview.net, 2018{\natexlab{b}}.
\newblock URL \url{https://openreview.net/forum?id=B1l8BtlCb}.

\bibitem[Wang and Cho(2019)]{wang2019bert}
Alex Wang and Kyunghyun Cho.
\newblock {BERT} has a mouth, and it must speak: {BERT} as a {M}arkov random
  field language model.
\newblock In \emph{Proceedings of the Workshop on Methods for Optimizing and
  Evaluating Neural Language Generation}, pages 30--36, Minneapolis, Minnesota,
  2019. Association for Computational Linguistics.
\newblock \doi{10.18653/v1/W19-2304}.
\newblock URL \url{https://www.aclweb.org/anthology/W19-2304}.

\bibitem[Ghazvininejad et~al.(2019)Ghazvininejad, Levy, Liu, and
  Zettlemoyer]{ghazvininejad2019mask}
Marjan Ghazvininejad, Omer Levy, Yinhan Liu, and Luke Zettlemoyer.
\newblock Mask-predict: Parallel decoding of conditional masked language
  models.
\newblock In \emph{Proceedings of the 2019 Conference on Empirical Methods in
  Natural Language Processing and the 9th International Joint Conference on
  Natural Language Processing (EMNLP-IJCNLP)}, pages 6112--6121, Hong Kong,
  China, 2019. Association for Computational Linguistics.
\newblock \doi{10.18653/v1/D19-1633}.
\newblock URL \url{https://www.aclweb.org/anthology/D19-1633}.

\bibitem[Ma et~al.(2019)Ma, Zhou, Li, Neubig, and Hovy]{ma2019flowseq}
Xuezhe Ma, Chunting Zhou, Xian Li, Graham Neubig, and Eduard Hovy.
\newblock {F}low{S}eq: Non-autoregressive conditional sequence generation with
  generative flow.
\newblock In \emph{Proceedings of the 2019 Conference on Empirical Methods in
  Natural Language Processing and the 9th International Joint Conference on
  Natural Language Processing (EMNLP-IJCNLP)}, pages 4282--4292, Hong Kong,
  China, 2019. Association for Computational Linguistics.
\newblock \doi{10.18653/v1/D19-1437}.
\newblock URL \url{https://www.aclweb.org/anthology/D19-1437}.

\bibitem[Tu et~al.(2020)Tu, Pang, Wiseman, and Gimpel]{tu2020engine}
Lifu Tu, Richard~Yuanzhe Pang, Sam Wiseman, and Kevin Gimpel.
\newblock {ENGINE}: Energy-based inference networks for non-autoregressive
  machine translation.
\newblock In \emph{Proceedings of the 58th Annual Meeting of the Association
  for Computational Linguistics}, pages 2819--2826, Online, 2020. Association
  for Computational Linguistics.
\newblock \doi{10.18653/v1/2020.acl-main.251}.
\newblock URL \url{https://www.aclweb.org/anthology/2020.acl-main.251}.

\bibitem[Lee et~al.(2018)Lee, Mansimov, and Cho]{lee2018deterministic}
Jason Lee, Elman Mansimov, and Kyunghyun Cho.
\newblock Deterministic non-autoregressive neural sequence modeling by
  iterative refinement.
\newblock In \emph{Proceedings of the 2018 Conference on Empirical Methods in
  Natural Language Processing}, pages 1173--1182, Brussels, Belgium, 2018.
  Association for Computational Linguistics.
\newblock \doi{10.18653/v1/D18-1149}.
\newblock URL \url{https://www.aclweb.org/anthology/D18-1149}.

\bibitem[Deng et~al.(2020)Deng, Ding, Tan, and Wu]{deng2020length}
Chaorui Deng, Ning Ding, Mingkui Tan, and Qi~Wu.
\newblock Length-controllable image captioning.
\newblock \emph{arXiv preprint arXiv:2007.09580}, 2020.

\bibitem[Medress et~al.(1977)Medress, Cooper, Forgie, Green, Klatt, O'Malley,
  Neuburg, Newell, Reddy, Ritea, et~al.]{medress1977speech}
Mark~F. Medress, Franklin~S Cooper, Jim~W. Forgie, CC~Green, Dennis~H. Klatt,
  Michael~H. O'Malley, Edward~P Neuburg, Allen Newell, DR~Reddy, B~Ritea,
  et~al.
\newblock Speech understanding systems: Report of a steering committee.
\newblock \emph{Artificial Intelligence}, 9\penalty0 (3):\penalty0 307--316,
  1977.

\bibitem[Chen et~al.(2021{\natexlab{a}})Chen, Tworek, Jun, Yuan, Ponde, Kaplan,
  Edwards, Burda, Joseph, Brockman, et~al.]{chen2021evaluating}
Mark Chen, Jerry Tworek, Heewoo Jun, Qiming Yuan, Henrique Ponde, Jared Kaplan,
  Harri Edwards, Yura Burda, Nicholas Joseph, Greg Brockman, et~al.
\newblock Evaluating large language models trained on code.
\newblock \emph{arXiv preprint arXiv:2107.03374}, 2021{\natexlab{a}}.

\bibitem[Radford et~al.(2019)Radford, Wu, Child, Luan, Amodei, and
  Sutskever]{radford2019language}
Alec Radford, Jeffrey Wu, Rewon Child, David Luan, Dario Amodei, and Ilya
  Sutskever.
\newblock Language models are unsupervised multitask learners.
\newblock \emph{OpenAI Blog}, 1:\penalty0 8, 2019.

\bibitem[Brown et~al.(2020)Brown, Mann, Ryder, Subbiah, Kaplan, Dhariwal,
  Neelakantan, Shyam, Sastry, Askell, Agarwal, Herbert{-}Voss, Krueger,
  Henighan, Child, Ramesh, Ziegler, Wu, Winter, Hesse, Chen, Sigler, Litwin,
  Gray, Chess, Clark, Berner, McCandlish, Radford, Sutskever, and
  Amodei]{brown2020language}
Tom~B. Brown, Benjamin Mann, Nick Ryder, Melanie Subbiah, Jared Kaplan,
  Prafulla Dhariwal, Arvind Neelakantan, Pranav Shyam, Girish Sastry, Amanda
  Askell, Sandhini Agarwal, Ariel Herbert{-}Voss, Gretchen Krueger, Tom
  Henighan, Rewon Child, Aditya Ramesh, Daniel~M. Ziegler, Jeffrey Wu, Clemens
  Winter, Christopher Hesse, Mark Chen, Eric Sigler, Mateusz Litwin, Scott
  Gray, Benjamin Chess, Jack Clark, Christopher Berner, Sam McCandlish, Alec
  Radford, Ilya Sutskever, and Dario Amodei.
\newblock Language models are few-shot learners.
\newblock In Hugo Larochelle, Marc'Aurelio Ranzato, Raia Hadsell,
  Maria{-}Florina Balcan, and Hsuan{-}Tien Lin, editors, \emph{Advances in
  Neural Information Processing Systems 33: Annual Conference on Neural
  Information Processing Systems 2020, NeurIPS 2020, December 6-12, 2020,
  virtual}, 2020.
\newblock URL
  \url{https://proceedings.neurips.cc/paper/2020/hash/1457c0d6bfcb4967418bfb8ac142f64a-Abstract.html}.

\bibitem[Holtzman et~al.(2020)Holtzman, Buys, Du, Forbes, and
  Choi]{Ari2019nucleus}
Ari Holtzman, Jan Buys, Li~Du, Maxwell Forbes, and Yejin Choi.
\newblock The curious case of neural text degeneration.
\newblock In \emph{8th International Conference on Learning Representations,
  {ICLR} 2020, Addis Ababa, Ethiopia, April 26-30, 2020}. OpenReview.net, 2020.
\newblock URL \url{https://openreview.net/forum?id=rygGQyrFvH}.

\bibitem[Fan et~al.(2018{\natexlab{a}})Fan, Lewis, and
  Dauphin]{fan2018hierarchical}
Angela Fan, Mike Lewis, and Yann Dauphin.
\newblock Hierarchical neural story generation.
\newblock In \emph{Proceedings of the 56th Annual Meeting of the Association
  for Computational Linguistics (Volume 1: Long Papers)}, pages 889--898,
  Melbourne, Australia, 2018{\natexlab{a}}. Association for Computational
  Linguistics.
\newblock \doi{10.18653/v1/P18-1082}.
\newblock URL \url{https://www.aclweb.org/anthology/P18-1082}.

\bibitem[Hu et~al.(2016)Hu, Xu, Rohrbach, Feng, Saenko, and Darrell]{Hu2015}
Ronghang Hu, Huazhe Xu, Marcus Rohrbach, Jiashi Feng, Kate Saenko, and Trevor
  Darrell.
\newblock Natural language object retrieval.
\newblock In \emph{2016 {IEEE} Conference on Computer Vision and Pattern
  Recognition, {CVPR} 2016, Las Vegas, NV, USA, June 27-30, 2016}, pages
  4555--4564. {IEEE} Computer Society, 2016.
\newblock \doi{10.1109/CVPR.2016.493}.
\newblock URL \url{https://doi.org/10.1109/CVPR.2016.493}.

\bibitem[Goodfellow et~al.(2014)Goodfellow, Pouget{-}Abadie, Mirza, Xu,
  Warde{-}Farley, Ozair, Courville, and Bengio]{Goodfellow2014}
Ian~J. Goodfellow, Jean Pouget{-}Abadie, Mehdi Mirza, Bing Xu, David
  Warde{-}Farley, Sherjil Ozair, Aaron~C. Courville, and Yoshua Bengio.
\newblock Generative adversarial nets.
\newblock In Zoubin Ghahramani, Max Welling, Corinna Cortes, Neil~D. Lawrence,
  and Kilian~Q. Weinberger, editors, \emph{Advances in Neural Information
  Processing Systems 27: Annual Conference on Neural Information Processing
  Systems 2014, December 8-13 2014, Montreal, Quebec, Canada}, pages
  2672--2680, 2014.
\newblock URL
  \url{https://proceedings.neurips.cc/paper/2014/hash/5ca3e9b122f61f8f06494c97b1afccf3-Abstract.html}.

\bibitem[Radford et~al.(2016)Radford, Metz, and Chintala]{Radford2015}
Alec Radford, Luke Metz, and Soumith Chintala.
\newblock Unsupervised representation learning with deep convolutional
  generative adversarial networks.
\newblock In Yoshua Bengio and Yann LeCun, editors, \emph{4th International
  Conference on Learning Representations, {ICLR} 2016, San Juan, Puerto Rico,
  May 2-4, 2016, Conference Track Proceedings}, 2016.
\newblock URL \url{http://arxiv.org/abs/1511.06434}.

\bibitem[Konda and Tsitsiklis(2000)]{konda2000actor}
Vijay~R Konda and John~N Tsitsiklis.
\newblock Actor-critic algorithms.
\newblock In \emph{Advances in neural information processing systems}, pages
  1008--1014, 2000.

\bibitem[Andreas and Klein(2016{\natexlab{b}})]{Andreas2016}
Jacob Andreas and Dan Klein.
\newblock Reasoning about pragmatics with neural listeners and speakers.
\newblock In \emph{Proceedings of the 2016 Conference on Empirical Methods in
  Natural Language Processing}, pages 1173--1182, Austin, Texas,
  2016{\natexlab{b}}. Association for Computational Linguistics.
\newblock \doi{10.18653/v1/D16-1125}.
\newblock URL \url{https://www.aclweb.org/anthology/D16-1125}.

\bibitem[Vered et~al.(2019)Vered, Oren, Atzmon, and Chechik]{vered2019joint}
Gilad Vered, Gal Oren, Yuval Atzmon, and Gal Chechik.
\newblock Joint optimization for cooperative image captioning.
\newblock In \emph{2019 {IEEE/CVF} International Conference on Computer Vision,
  {ICCV} 2019, Seoul, Korea (South), October 27 - November 2, 2019}, pages
  8897--8906. {IEEE}, 2019.
\newblock \doi{10.1109/ICCV.2019.00899}.
\newblock URL \url{https://doi.org/10.1109/ICCV.2019.00899}.

\bibitem[Nagaraja et~al.(2016)Nagaraja, Morariu, and Davis]{Nagaraja2016}
Varun~K Nagaraja, Vlad~I Morariu, and Larry~S Davis.
\newblock {Modeling Context Between Objects for Referring Expression
  Understanding}.
\newblock \emph{Eccv}, 2016.

\bibitem[Yu et~al.(2017{\natexlab{a}})Yu, Zhang, Wang, and Yu]{Yu2016a}
Lantao Yu, Weinan Zhang, Jun Wang, and Yong Yu.
\newblock Seqgan: Sequence generative adversarial nets with policy gradient.
\newblock In Satinder~P. Singh and Shaul Markovitch, editors, \emph{Proceedings
  of the Thirty-First {AAAI} Conference on Artificial Intelligence, February
  4-9, 2017, San Francisco, California, {USA}}, pages 2852--2858. {AAAI} Press,
  2017{\natexlab{a}}.
\newblock URL \url{http://aaai.org/ocs/index.php/AAAI/AAAI17/paper/view/14344}.

\bibitem[Rohrbach et~al.(2015)Rohrbach, Rohrbach, Hu, Darrell, and
  Schiele]{Rohrbach2015}
Anna Rohrbach, Marcus Rohrbach, Ronghang Hu, Trevor Darrell, and Bernt Schiele.
\newblock {Grounding of Textual Phrases in Images by Reconstruction}.
\newblock \emph{1511.03745V1}, 1:\penalty0 1--10, 2015.
\newblock URL \url{http://arxiv.org/abs/1511.03745}.

\bibitem[Fukui et~al.(2016)Fukui, Park, Yang, Rohrbach, Darrell, and
  Rohrbach]{Fukui2016}
Akira Fukui, Dong~Huk Park, Daylen Yang, Anna Rohrbach, Trevor Darrell, and
  Marcus Rohrbach.
\newblock Multimodal compact bilinear pooling for visual question answering and
  visual grounding.
\newblock In \emph{Proceedings of the 2016 Conference on Empirical Methods in
  Natural Language Processing}, pages 457--468, Austin, Texas, 2016.
  Association for Computational Linguistics.
\newblock \doi{10.18653/v1/D16-1044}.
\newblock URL \url{https://www.aclweb.org/anthology/D16-1044}.

\bibitem[Yu et~al.(2017{\natexlab{b}})Yu, Tan, Bansal, and Berg]{yu2017joint}
Licheng Yu, Hao Tan, Mohit Bansal, and Tamara~L. Berg.
\newblock A joint speaker-listener-reinforcer model for referring expressions.
\newblock In \emph{2017 {IEEE} Conference on Computer Vision and Pattern
  Recognition, {CVPR} 2017, Honolulu, HI, USA, July 21-26, 2017}, pages
  3521--3529. {IEEE} Computer Society, 2017{\natexlab{b}}.
\newblock \doi{10.1109/CVPR.2017.375}.
\newblock URL \url{https://doi.org/10.1109/CVPR.2017.375}.

\bibitem[Simonyan and Zisserman(2015)]{simonyan2014very}
Karen Simonyan and Andrew Zisserman.
\newblock Very deep convolutional networks for large-scale image recognition.
\newblock In Yoshua Bengio and Yann LeCun, editors, \emph{3rd International
  Conference on Learning Representations, {ICLR} 2015, San Diego, CA, USA, May
  7-9, 2015, Conference Track Proceedings}, 2015.
\newblock URL \url{http://arxiv.org/abs/1409.1556}.

\bibitem[Graves et~al.(2013)Graves, Mohamed, and Hinton]{Graves2013}
Alex Graves, Abdel-rahman Mohamed, and Geoffrey Hinton.
\newblock Speech recognition with deep recurrent neural networks.
\newblock In \emph{2013 IEEE international conference on acoustics, speech and
  signal processing}, pages 6645--6649. IEEE, 2013.

\bibitem[Dhingra et~al.(2017)Dhingra, Liu, Yang, Cohen, and
  Salakhutdinov]{Dhingra2016}
Bhuwan Dhingra, Hanxiao Liu, Zhilin Yang, William Cohen, and Ruslan
  Salakhutdinov.
\newblock Gated-attention readers for text comprehension.
\newblock In \emph{Proceedings of the 55th Annual Meeting of the Association
  for Computational Linguistics (Volume 1: Long Papers)}, pages 1832--1846,
  Vancouver, Canada, 2017. Association for Computational Linguistics.
\newblock \doi{10.18653/v1/P17-1168}.
\newblock URL \url{https://www.aclweb.org/anthology/P17-1168}.

\bibitem[Bahdanau et~al.(2017)Bahdanau, Brakel, Xu, Goyal, Lowe, Pineau,
  Courville, and Bengio]{Bahdanau2016}
Dzmitry Bahdanau, Philemon Brakel, Kelvin Xu, Anirudh Goyal, Ryan Lowe, Joelle
  Pineau, Aaron~C. Courville, and Yoshua Bengio.
\newblock An actor-critic algorithm for sequence prediction.
\newblock In \emph{5th International Conference on Learning Representations,
  {ICLR} 2017, Toulon, France, April 24-26, 2017, Conference Track
  Proceedings}. OpenReview.net, 2017.
\newblock URL \url{https://openreview.net/forum?id=SJDaqqveg}.

\bibitem[Tu and Gimpel(2018)]{tu2018learning}
Lifu Tu and Kevin Gimpel.
\newblock Learning approximate inference networks for structured prediction.
\newblock In \emph{6th International Conference on Learning Representations,
  {ICLR} 2018, Vancouver, BC, Canada, April 30 - May 3, 2018, Conference Track
  Proceedings}. OpenReview.net, 2018.
\newblock URL \url{https://openreview.net/forum?id=H1WgVz-AZ}.

\bibitem[Bahdanau et~al.(2015{\natexlab{b}})Bahdanau, Cho, and
  Bengio]{DzmitryBahdana2014}
Dzmitry Bahdanau, Kyunghyun Cho, and Yoshua Bengio.
\newblock Neural machine translation by jointly learning to align and
  translate.
\newblock In Yoshua Bengio and Yann LeCun, editors, \emph{3rd International
  Conference on Learning Representations, {ICLR} 2015, San Diego, CA, USA, May
  7-9, 2015, Conference Track Proceedings}, 2015{\natexlab{b}}.
\newblock URL \url{http://arxiv.org/abs/1409.0473}.

\bibitem[Goodfellow et~al.(2015)Goodfellow, Shlens, and
  Szegedy]{goodfellow2014explaining}
Ian~J. Goodfellow, Jonathon Shlens, and Christian Szegedy.
\newblock Explaining and harnessing adversarial examples.
\newblock In Yoshua Bengio and Yann LeCun, editors, \emph{3rd International
  Conference on Learning Representations, {ICLR} 2015, San Diego, CA, USA, May
  7-9, 2015, Conference Track Proceedings}, 2015.
\newblock URL \url{http://arxiv.org/abs/1412.6572}.

\bibitem[Lewis et~al.(2017)Lewis, Yarats, Dauphin, Parikh, and
  Batra]{lewis2017deal}
Mike Lewis, Denis Yarats, Yann~N Dauphin, Devi Parikh, and Dhruv Batra.
\newblock Deal or no deal? end-to-end learning for negotiation dialogues.
\newblock \emph{arXiv preprint arXiv:1706.05125}, 2017.

\bibitem[Escalante et~al.(2010)Escalante, Hern{\'{a}}ndez, Gonzalez,
  L{\'{o}}pez-L{\'{o}}pez, Montes, Morales, {Enrique Sucar}, Villase{\~{n}}or,
  and Grubinger]{Escalante2010}
Hugo~Jair Escalante, Carlos~A. Hern{\'{a}}ndez, Jesus~A. Gonzalez,
  A.~L{\'{o}}pez-L{\'{o}}pez, Manuel Montes, Eduardo~F. Morales, L.~{Enrique
  Sucar}, Luis Villase{\~{n}}or, and Michael Grubinger.
\newblock {The segmented and annotated IAPR TC-12 benchmark}.
\newblock \emph{Computer Vision and Image Understanding}, 114\penalty0
  (4):\penalty0 419--428, 2010.
\newblock ISSN 10773142.
\newblock \doi{10.1016/j.cviu.2009.03.008}.
\newblock URL \url{http://dx.doi.org/10.1016/j.cviu.2009.03.008}.

\bibitem[Girshick(2015)]{girshick2015fast}
Ross~B. Girshick.
\newblock Fast {R-CNN}.
\newblock In \emph{2015 {IEEE} International Conference on Computer Vision,
  {ICCV} 2015, Santiago, Chile, December 7-13, 2015}, pages 1440--1448. {IEEE}
  Computer Society, 2015.
\newblock \doi{10.1109/ICCV.2015.169}.
\newblock URL \url{https://doi.org/10.1109/ICCV.2015.169}.

\bibitem[Zitnick and
  Dollar(2014)]{edge-boxes-locating-object-proposals-from-edges}
Larry Zitnick and Piotr Dollar.
\newblock Edge boxes: Locating object proposals from edges.
\newblock In \emph{ECCV}. European Conference on Computer Vision, 2014.
\newblock URL
  \url{https://www.microsoft.com/en-us/research/publication/edge-boxes-locating-object-proposals-from-edges/}.

\bibitem[Kingma and Ba(2015)]{kingma2014adam}
Diederik~P. Kingma and Jimmy Ba.
\newblock Adam: {A} method for stochastic optimization.
\newblock In Yoshua Bengio and Yann LeCun, editors, \emph{3rd International
  Conference on Learning Representations, {ICLR} 2015, San Diego, CA, USA, May
  7-9, 2015, Conference Track Proceedings}, 2015.
\newblock URL \url{http://arxiv.org/abs/1412.6980}.

\bibitem[Ma et~al.(2015)Ma, Lu, Shang, and Li]{Ma2016}
Lin Ma, Zhengdong Lu, Lifeng Shang, and Hang Li.
\newblock Multimodal convolutional neural networks for matching image and
  sentence.
\newblock In \emph{2015 {IEEE} International Conference on Computer Vision,
  {ICCV} 2015, Santiago, Chile, December 7-13, 2015}, pages 2623--2631. {IEEE}
  Computer Society, 2015.
\newblock \doi{10.1109/ICCV.2015.301}.
\newblock URL \url{https://doi.org/10.1109/ICCV.2015.301}.

\bibitem[Frome et~al.(2013)Frome, Corrado, Shlens, Bengio, Dean, Ranzato, and
  Mikolov]{frome2013devise}
Andrea Frome, Gregory~S. Corrado, Jonathon Shlens, Samy Bengio, Jeffrey Dean,
  Marc'Aurelio Ranzato, and Tom{\'{a}}s Mikolov.
\newblock Devise: {A} deep visual-semantic embedding model.
\newblock In Christopher J.~C. Burges, L{\'{e}}on Bottou, Zoubin Ghahramani,
  and Kilian~Q. Weinberger, editors, \emph{Advances in Neural Information
  Processing Systems 26: 27th Annual Conference on Neural Information
  Processing Systems 2013. Proceedings of a meeting held December 5-8, 2013,
  Lake Tahoe, Nevada, United States}, pages 2121--2129, 2013.
\newblock URL
  \url{https://proceedings.neurips.cc/paper/2013/hash/7cce53cf90577442771720a370c3c723-Abstract.html}.

\bibitem[Wang et~al.(2016{\natexlab{a}})Wang, Li, and Lazebnik]{Wang2016a}
Liwei Wang, Yin Li, and Svetlana Lazebnik.
\newblock Learning deep structure-preserving image-text embeddings.
\newblock In \emph{2016 {IEEE} Conference on Computer Vision and Pattern
  Recognition, {CVPR} 2016, Las Vegas, NV, USA, June 27-30, 2016}, pages
  5005--5013. {IEEE} Computer Society, 2016{\natexlab{a}}.
\newblock \doi{10.1109/CVPR.2016.541}.
\newblock URL \url{https://doi.org/10.1109/CVPR.2016.541}.

\bibitem[Weston et~al.(2011)Weston, Bengio, and Usunier]{WSABIE}
Jason Weston, Samy Bengio, and Nicolas Usunier.
\newblock {WSABIE:} scaling up to large vocabulary image annotation.
\newblock In Toby Walsh, editor, \emph{{IJCAI} 2011, Proceedings of the 22nd
  International Joint Conference on Artificial Intelligence, Barcelona,
  Catalonia, Spain, July 16-22, 2011}, pages 2764--2770. {IJCAI/AAAI}, 2011.
\newblock \doi{10.5591/978-1-57735-516-8/IJCAI11-460}.
\newblock URL \url{https://doi.org/10.5591/978-1-57735-516-8/IJCAI11-460}.

\bibitem[Wang et~al.(2017{\natexlab{b}})Wang, Li, and
  Lazebnik]{wang2017learning}
Liwei Wang, Yin Li, and Svetlana Lazebnik.
\newblock Learning two-branch neural networks for image-text matching tasks.
\newblock \emph{arXiv preprint arXiv:1704.03470}, 2017{\natexlab{b}}.

\bibitem[Nam et~al.(2017)Nam, Ha, and Kim]{nam2016dual}
Hyeonseob Nam, Jung{-}Woo Ha, and Jeonghee Kim.
\newblock Dual attention networks for multimodal reasoning and matching.
\newblock In \emph{2017 {IEEE} Conference on Computer Vision and Pattern
  Recognition, {CVPR} 2017, Honolulu, HI, USA, July 21-26, 2017}, pages
  2156--2164. {IEEE} Computer Society, 2017.
\newblock \doi{10.1109/CVPR.2017.232}.
\newblock URL \url{https://doi.org/10.1109/CVPR.2017.232}.

\bibitem[Vendrov et~al.(2016)Vendrov, Kiros, Fidler, and Urtasun]{Vendrov2015}
Ivan Vendrov, Ryan Kiros, Sanja Fidler, and Raquel Urtasun.
\newblock Order-embeddings of images and language.
\newblock In Yoshua Bengio and Yann LeCun, editors, \emph{4th International
  Conference on Learning Representations, {ICLR} 2016, San Juan, Puerto Rico,
  May 2-4, 2016, Conference Track Proceedings}, 2016.
\newblock URL \url{http://arxiv.org/abs/1511.06361}.

\bibitem[Yu et~al.(2017{\natexlab{c}})Yu, Tan, Bansal, and Berg]{yu2016joint}
Licheng Yu, Hao Tan, Mohit Bansal, and Tamara~L. Berg.
\newblock A joint speaker-listener-reinforcer model for referring expressions.
\newblock In \emph{2017 {IEEE} Conference on Computer Vision and Pattern
  Recognition, {CVPR} 2017, Honolulu, HI, USA, July 21-26, 2017}, pages
  3521--3529. {IEEE} Computer Society, 2017{\natexlab{c}}.
\newblock \doi{10.1109/CVPR.2017.375}.
\newblock URL \url{https://doi.org/10.1109/CVPR.2017.375}.

\bibitem[Luo and Shakhnarovich(2017)]{luo2017comprehension}
Ruotian Luo and Gregory Shakhnarovich.
\newblock Comprehension-guided referring expressions.
\newblock In \emph{2017 {IEEE} Conference on Computer Vision and Pattern
  Recognition, {CVPR} 2017, Honolulu, HI, USA, July 21-26, 2017}, pages
  3125--3134. {IEEE} Computer Society, 2017.
\newblock \doi{10.1109/CVPR.2017.333}.
\newblock URL \url{https://doi.org/10.1109/CVPR.2017.333}.

\bibitem[Geman et~al.(2015)Geman, Geman, Hallonquist, and
  Younes]{geman2015visual}
Donald Geman, Stuart Geman, Neil Hallonquist, and Laurent Younes.
\newblock Visual turing test for computer vision systems.
\newblock \emph{Proceedings of the National Academy of Sciences}, 112\penalty0
  (12):\penalty0 3618--3623, 2015.

\bibitem[Das et~al.(2017{\natexlab{b}})Das, Kottur, Gupta, Singh, Yadav, Moura,
  Parikh, and Batra]{das2016visual}
Abhishek Das, Satwik Kottur, Khushi Gupta, Avi Singh, Deshraj Yadav, Jos{\'{e}}
  M.~F. Moura, Devi Parikh, and Dhruv Batra.
\newblock Visual dialog.
\newblock In \emph{2017 {IEEE} Conference on Computer Vision and Pattern
  Recognition, {CVPR} 2017, Honolulu, HI, USA, July 21-26, 2017}, pages
  1080--1089. {IEEE} Computer Society, 2017{\natexlab{b}}.
\newblock \doi{10.1109/CVPR.2017.121}.
\newblock URL \url{https://doi.org/10.1109/CVPR.2017.121}.

\bibitem[de~Vries et~al.(2017)de~Vries, Strub, Chandar, Pietquin, Larochelle,
  and Courville]{de2016guesswhat}
Harm de~Vries, Florian Strub, Sarath Chandar, Olivier Pietquin, Hugo
  Larochelle, and Aaron~C. Courville.
\newblock Guesswhat?! visual object discovery through multi-modal dialogue.
\newblock In \emph{2017 {IEEE} Conference on Computer Vision and Pattern
  Recognition, {CVPR} 2017, Honolulu, HI, USA, July 21-26, 2017}, pages
  4466--4475. {IEEE} Computer Society, 2017.
\newblock \doi{10.1109/CVPR.2017.475}.
\newblock URL \url{https://doi.org/10.1109/CVPR.2017.475}.

\bibitem[Li et~al.(2017)Li, Huang, Tang, and Loy]{li2017learning}
Yining Li, Chen Huang, Xiaoou Tang, and Chen~Change Loy.
\newblock Learning to disambiguate by asking discriminative questions.
\newblock In \emph{{IEEE} International Conference on Computer Vision, {ICCV}
  2017, Venice, Italy, October 22-29, 2017}, pages 3439--3448. {IEEE} Computer
  Society, 2017.
\newblock \doi{10.1109/ICCV.2017.370}.
\newblock URL \url{https://doi.org/10.1109/ICCV.2017.370}.

\bibitem[Lu et~al.(2017{\natexlab{c}})Lu, Kannan, Yang, Parikh, and
  Batra]{lu2017best}
Jiasen Lu, Anitha Kannan, Jianwei Yang, Devi Parikh, and Dhruv Batra.
\newblock Best of both worlds: Transferring knowledge from discriminative
  learning to a generative visual dialog model.
\newblock In Isabelle Guyon, Ulrike von Luxburg, Samy Bengio, Hanna~M. Wallach,
  Rob Fergus, S.~V.~N. Vishwanathan, and Roman Garnett, editors, \emph{Advances
  in Neural Information Processing Systems 30: Annual Conference on Neural
  Information Processing Systems 2017, December 4-9, 2017, Long Beach, CA,
  {USA}}, pages 314--324, 2017{\natexlab{c}}.
\newblock URL
  \url{https://proceedings.neurips.cc/paper/2017/hash/077e29b11be80ab57e1a2ecabb7da330-Abstract.html}.

\bibitem[Wang et~al.(2020)Wang, Xu, Wang, and Chan]{wang2020compare}
Jiuniu Wang, Wenjia Xu, Qingzhong Wang, and Antoni~B Chan.
\newblock Compare and reweight: Distinctive image captioning using similar
  images sets.
\newblock In \emph{European Conference on Computer Vision}, pages 370--386.
  Springer, 2020.

\bibitem[Faghri et~al.(2017)Faghri, Fleet, Kiros, and Fidler]{faghri2017vse++}
Fartash Faghri, David~J Fleet, Jamie~Ryan Kiros, and Sanja Fidler.
\newblock Vse++: Improved visual-semantic embeddings.
\newblock \emph{arXiv preprint arXiv:1707.05612}, 2017.

\bibitem[Schroff et~al.(2015)Schroff, Kalenichenko, and
  Philbin]{schroff2015facenet}
Florian Schroff, Dmitry Kalenichenko, and James Philbin.
\newblock Facenet: {A} unified embedding for face recognition and clustering.
\newblock In \emph{{IEEE} Conference on Computer Vision and Pattern
  Recognition, {CVPR} 2015, Boston, MA, USA, June 7-12, 2015}, pages 815--823.
  {IEEE} Computer Society, 2015.
\newblock \doi{10.1109/CVPR.2015.7298682}.
\newblock URL \url{https://doi.org/10.1109/CVPR.2015.7298682}.

\bibitem[Ling and Fidler(2017{\natexlab{b}})]{LingNIPS2017}
Huan Ling and Sanja Fidler.
\newblock Teaching machines to describe images via natural language feedback.
\newblock In \emph{NIPS}, 2017{\natexlab{b}}.

\bibitem[Jang et~al.(2017)Jang, Gu, and Poole]{jang2016categorical}
Eric Jang, Shixiang Gu, and Ben Poole.
\newblock Categorical reparameterization with gumbel-softmax.
\newblock In \emph{5th International Conference on Learning Representations,
  {ICLR} 2017, Toulon, France, April 24-26, 2017, Conference Track
  Proceedings}. OpenReview.net, 2017.
\newblock URL \url{https://openreview.net/forum?id=rkE3y85ee}.

\bibitem[Maddison et~al.(2017)Maddison, Mnih, and Teh]{maddison2016concrete}
Chris~J. Maddison, Andriy Mnih, and Yee~Whye Teh.
\newblock The concrete distribution: {A} continuous relaxation of discrete
  random variables.
\newblock In \emph{5th International Conference on Learning Representations,
  {ICLR} 2017, Toulon, France, April 24-26, 2017, Conference Track
  Proceedings}. OpenReview.net, 2017.
\newblock URL \url{https://openreview.net/forum?id=S1jE5L5gl}.

\bibitem[Ren et~al.(2015)Ren, He, Girshick, and Sun]{renNIPS15fasterrcnn}
Shaoqing Ren, Kaiming He, Ross~B. Girshick, and Jian Sun.
\newblock Faster {R-CNN:} towards real-time object detection with region
  proposal networks.
\newblock In Corinna Cortes, Neil~D. Lawrence, Daniel~D. Lee, Masashi Sugiyama,
  and Roman Garnett, editors, \emph{Advances in Neural Information Processing
  Systems 28: Annual Conference on Neural Information Processing Systems 2015,
  December 7-12, 2015, Montreal, Quebec, Canada}, pages 91--99, 2015.
\newblock URL
  \url{https://proceedings.neurips.cc/paper/2015/hash/14bfa6bb14875e45bba028a21ed38046-Abstract.html}.

\bibitem[Collins(2000)]{collins2005discriminative}
Michael Collins.
\newblock Discriminative reranking for natural language parsing.
\newblock In Pat Langley, editor, \emph{Proceedings of the Seventeenth
  International Conference on Machine Learning {(ICML} 2000), Stanford
  University, Stanford, CA, USA, June 29 - July 2, 2000}, pages 175--182.
  Morgan Kaufmann, 2000.

\bibitem[Shen and Joshi(2003)]{shen2003svm}
Libin Shen and Aravind~K. Joshi.
\newblock An {SVM}-based voting algorithm with application to parse reranking.
\newblock In \emph{Proceedings of the Seventh Conference on Natural Language
  Learning at {HLT}-{NAACL} 2003}, pages 9--16, 2003.
\newblock URL \url{https://www.aclweb.org/anthology/W03-0402}.

\bibitem[Park et~al.(2019{\natexlab{b}})Park, Rohrbach, Darrell, and
  Rohrbach]{park2019adversarial}
Jae~Sung Park, Marcus Rohrbach, Trevor Darrell, and Anna Rohrbach.
\newblock Adversarial inference for multi-sentence video description.
\newblock In \emph{{IEEE} Conference on Computer Vision and Pattern
  Recognition, {CVPR} 2019, Long Beach, CA, USA, June 16-20, 2019}, pages
  6598--6608. Computer Vision Foundation / {IEEE}, 2019{\natexlab{b}}.
\newblock \doi{10.1109/CVPR.2019.00676}.
\newblock URL
  \url{http://openaccess.thecvf.com/content\_CVPR\_2019/html/Park\_Adversarial\_Inference\_for\_Multi-Sentence\_Video\_Description\_CVPR\_2019\_paper.html}.

\bibitem[Wang et~al.(2017{\natexlab{c}})Wang, Schwing, and
  Lazebnik]{wang2017diverse}
Liwei Wang, Alexander~G. Schwing, and Svetlana Lazebnik.
\newblock Diverse and accurate image description using a variational
  auto-encoder with an additive gaussian encoding space.
\newblock In Isabelle Guyon, Ulrike von Luxburg, Samy Bengio, Hanna~M. Wallach,
  Rob Fergus, S.~V.~N. Vishwanathan, and Roman Garnett, editors, \emph{Advances
  in Neural Information Processing Systems 30: Annual Conference on Neural
  Information Processing Systems 2017, December 4-9, 2017, Long Beach, CA,
  {USA}}, pages 5756--5766, 2017{\natexlab{c}}.
\newblock URL
  \url{https://proceedings.neurips.cc/paper/2017/hash/4b21cf96d4cf612f239a6c322b10c8fe-Abstract.html}.

\bibitem[Jain et~al.(2017)Jain, Zhang, and Schwing]{jain2017creativity}
Unnat Jain, Ziyu Zhang, and Alexander~G. Schwing.
\newblock Creativity: Generating diverse questions using variational
  autoencoders.
\newblock In \emph{2017 {IEEE} Conference on Computer Vision and Pattern
  Recognition, {CVPR} 2017, Honolulu, HI, USA, July 21-26, 2017}, pages
  5415--5424. {IEEE} Computer Society, 2017.
\newblock \doi{10.1109/CVPR.2017.575}.
\newblock URL \url{https://doi.org/10.1109/CVPR.2017.575}.

\bibitem[Wang et~al.(2016{\natexlab{b}})Wang, Wu, Lu, Xiao, Li, Zhang, and
  Zhuang]{wang2016diverse}
Zhuhao Wang, Fei Wu, Weiming Lu, Jun Xiao, Xi~Li, Zitong Zhang, and Yueting
  Zhuang.
\newblock Diverse image captioning via grouptalk.
\newblock In Subbarao Kambhampati, editor, \emph{Proceedings of the
  Twenty-Fifth International Joint Conference on Artificial Intelligence,
  {IJCAI} 2016, New York, NY, USA, 9-15 July 2016}, pages 2957--2964.
  {IJCAI/AAAI} Press, 2016{\natexlab{b}}.
\newblock URL \url{http://www.ijcai.org/Abstract/16/420}.

\bibitem[Shen et~al.(2019)Shen, Ott, Auli, and Ranzato]{shen2019mixture}
Tianxiao Shen, Myle Ott, Michael Auli, and Marc'Aurelio Ranzato.
\newblock Mixture models for diverse machine translation: Tricks of the trade.
\newblock In Kamalika Chaudhuri and Ruslan Salakhutdinov, editors,
  \emph{Proceedings of the 36th International Conference on Machine Learning,
  {ICML} 2019, 9-15 June 2019, Long Beach, California, {USA}}, volume~97 of
  \emph{Proceedings of Machine Learning Research}, pages 5719--5728. {PMLR},
  2019.
\newblock URL \url{http://proceedings.mlr.press/v97/shen19c.html}.

\bibitem[Vijayakumar et~al.(2016)Vijayakumar, Cogswell, Selvaraju, Sun, Lee,
  Crandall, and Batra]{vijayakumar2016diverse}
Ashwin~K Vijayakumar, Michael Cogswell, Ramprasath~R Selvaraju, Qing Sun,
  Stefan Lee, David Crandall, and Dhruv Batra.
\newblock Diverse beam search: Decoding diverse solutions from neural sequence
  models.
\newblock \emph{arXiv preprint arXiv:1610.02424}, 2016.

\bibitem[Batra et~al.(2012)Batra, Yadollahpour, Guzman-Rivera, and
  Shakhnarovich]{batra2012diverse}
Dhruv Batra, Payman Yadollahpour, Abner Guzman-Rivera, and Gregory
  Shakhnarovich.
\newblock Diverse m-best solutions in markov random fields.
\newblock In \emph{European Conference on Computer Vision}, pages 1--16.
  Springer, 2012.

\bibitem[Gimpel et~al.(2013)Gimpel, Batra, Dyer, and
  Shakhnarovich]{gimpel2013systematic}
Kevin Gimpel, Dhruv Batra, Chris Dyer, and Gregory Shakhnarovich.
\newblock A systematic exploration of diversity in machine translation.
\newblock In \emph{Proceedings of the 2013 Conference on Empirical Methods in
  Natural Language Processing}, pages 1100--1111, Seattle, Washington, USA,
  2013. Association for Computational Linguistics.
\newblock URL \url{https://www.aclweb.org/anthology/D13-1111}.

\bibitem[Li and Jurafsky(2016)]{li2016mutual}
Jiwei Li and Dan Jurafsky.
\newblock Mutual information and diverse decoding improve neural machine
  translation.
\newblock \emph{arXiv preprint arXiv:1601.00372}, 2016.

\bibitem[Li et~al.(2016)Li, Galley, Brockett, Gao, and Dolan]{li2015diversity}
Jiwei Li, Michel Galley, Chris Brockett, Jianfeng Gao, and Bill Dolan.
\newblock A diversity-promoting objective function for neural conversation
  models.
\newblock In \emph{Proceedings of the 2016 Conference of the North {A}merican
  Chapter of the Association for Computational Linguistics: Human Language
  Technologies}, pages 110--119, San Diego, California, 2016. Association for
  Computational Linguistics.
\newblock \doi{10.18653/v1/N16-1014}.
\newblock URL \url{https://www.aclweb.org/anthology/N16-1014}.

\bibitem[Wang and Chan(2019)]{wang2019describing}
Qingzhong Wang and Antoni~B. Chan.
\newblock Describing like humans: On diversity in image captioning.
\newblock In \emph{{IEEE} Conference on Computer Vision and Pattern
  Recognition, {CVPR} 2019, Long Beach, CA, USA, June 16-20, 2019}, pages
  4195--4203. Computer Vision Foundation / {IEEE}, 2019.
\newblock \doi{10.1109/CVPR.2019.00432}.
\newblock URL
  \url{http://openaccess.thecvf.com/content\_CVPR\_2019/html/Wang\_Describing\_Like\_Humans\_On\_Diversity\_in\_Image\_Captioning\_CVPR\_2019\_paper.html}.

\bibitem[Li et~al.(2018{\natexlab{b}})Li, Huang, He, Zhang, and
  Sun]{li2018generating}
Dianqi Li, Qiuyuan Huang, Xiaodong He, Lei Zhang, and Ming-Ting Sun.
\newblock Generating diverse and accurate visual captions by comparative
  adversarial learning.
\newblock \emph{arXiv preprint arXiv:1804.00861}, 2018{\natexlab{b}}.

\bibitem[Masoudnia and Ebrahimpour(2014)]{masoudnia2014mixture}
Saeed Masoudnia and Reza Ebrahimpour.
\newblock Mixture of experts: a literature survey.
\newblock \emph{Artificial Intelligence Review}, 42\penalty0 (2):\penalty0
  275--293, 2014.

\bibitem[Schapire(2003)]{schapire2003boosting}
Robert~E Schapire.
\newblock The boosting approach to machine learning: An overview.
\newblock \emph{Nonlinear estimation and classification}, pages 149--171, 2003.

\bibitem[van Miltenburg et~al.(2018)van Miltenburg, Elliott, and
  Vossen]{van2018measuring}
Emiel van Miltenburg, Desmond Elliott, and Piek Vossen.
\newblock Measuring the diversity of automatic image descriptions.
\newblock In \emph{Proceedings of the 27th International Conference on
  Computational Linguistics}, pages 1730--1741, Santa Fe, New Mexico, USA,
  2018. Association for Computational Linguistics.
\newblock URL \url{https://www.aclweb.org/anthology/C18-1147}.

\bibitem[Mnih and Rezende(2016)]{mnih2016variational}
Andriy Mnih and Danilo~Jimenez Rezende.
\newblock Variational inference for monte carlo objectives.
\newblock In Maria{-}Florina Balcan and Kilian~Q. Weinberger, editors,
  \emph{Proceedings of the 33nd International Conference on Machine Learning,
  {ICML} 2016, New York City, NY, USA, June 19-24, 2016}, volume~48 of
  \emph{{JMLR} Workshop and Conference Proceedings}, pages 2188--2196.
  JMLR.org, 2016.
\newblock URL \url{http://proceedings.mlr.press/v48/mnihb16.html}.

\bibitem[Deshpande et~al.(2019)Deshpande, Aneja, Wang, Schwing, and
  Forsyth]{deshpande2019fast}
Aditya Deshpande, Jyoti Aneja, Liwei Wang, Alexander~G. Schwing, and David~A.
  Forsyth.
\newblock Fast, diverse and accurate image captioning guided by part-of-speech.
\newblock In \emph{{IEEE} Conference on Computer Vision and Pattern
  Recognition, {CVPR} 2019, Long Beach, CA, USA, June 16-20, 2019}, pages
  10695--10704. Computer Vision Foundation / {IEEE}, 2019.
\newblock \doi{10.1109/CVPR.2019.01095}.
\newblock URL
  \url{http://openaccess.thecvf.com/content\_CVPR\_2019/html/Deshpande\_Fast\_Diverse\_and\_Accurate\_Image\_Captioning\_Guided\_by\_Part-Of-Speech\_CVPR\_2019\_paper.html}.

\bibitem[Kikuchi et~al.(2016)Kikuchi, Neubig, Sasano, Takamura, and
  Okumura]{kikuchi2016controlling}
Yuta Kikuchi, Graham Neubig, Ryohei Sasano, Hiroya Takamura, and Manabu
  Okumura.
\newblock Controlling output length in neural encoder-decoders.
\newblock In \emph{Proceedings of the 2016 Conference on Empirical Methods in
  Natural Language Processing}, pages 1328--1338, Austin, Texas, 2016.
  Association for Computational Linguistics.
\newblock \doi{10.18653/v1/D16-1140}.
\newblock URL \url{https://www.aclweb.org/anthology/D16-1140}.

\bibitem[Fan et~al.(2018{\natexlab{b}})Fan, Grangier, and
  Auli]{fan2017controllable}
Angela Fan, David Grangier, and Michael Auli.
\newblock Controllable abstractive summarization.
\newblock In \emph{Proceedings of the 2nd Workshop on Neural Machine
  Translation and Generation}, pages 45--54, Melbourne, Australia,
  2018{\natexlab{b}}. Association for Computational Linguistics.
\newblock \doi{10.18653/v1/W18-2706}.
\newblock URL \url{https://www.aclweb.org/anthology/W18-2706}.

\bibitem[Cornia et~al.(2019)Cornia, Baraldi, and Cucchiara]{cornia2019show}
Marcella Cornia, Lorenzo Baraldi, and Rita Cucchiara.
\newblock Show, control and tell: {A} framework for generating controllable and
  grounded captions.
\newblock In \emph{{IEEE} Conference on Computer Vision and Pattern
  Recognition, {CVPR} 2019, Long Beach, CA, USA, June 16-20, 2019}, pages
  8307--8316. Computer Vision Foundation / {IEEE}, 2019.
\newblock \doi{10.1109/CVPR.2019.00850}.
\newblock URL
  \url{http://openaccess.thecvf.com/content\_CVPR\_2019/html/Cornia\_Show\_Control\_and\_Tell\_A\_Framework\_for\_Generating\_Controllable\_and\_CVPR\_2019\_paper.html}.

\bibitem[Zhong et~al.(2020)Zhong, Wang, Chen, Yu, and
  Li]{zhong2020comprehensive}
Yiwu Zhong, Liwei Wang, Jianshu Chen, Dong Yu, and Yin Li.
\newblock Comprehensive image captioning via scene graph decomposition.
\newblock In \emph{European Conference on Computer Vision}, pages 211--229.
  Springer, 2020.

\bibitem[Chen et~al.(2021{\natexlab{b}})Chen, Jiang, Xiao, and
  Liu]{chen2021human}
Long Chen, Zhihong Jiang, Jun Xiao, and Wei Liu.
\newblock Human-like controllable image captioning with verb-specific semantic
  roles.
\newblock In \emph{Proceedings of the IEEE/CVF Conference on Computer Vision
  and Pattern Recognition}, pages 16846--16856, 2021{\natexlab{b}}.

\bibitem[Mathews et~al.(2016)Mathews, Xie, and He]{mathews2016senticap}
Alexander~Patrick Mathews, Lexing Xie, and Xuming He.
\newblock Senticap: Generating image descriptions with sentiments.
\newblock In Dale Schuurmans and Michael~P. Wellman, editors, \emph{Proceedings
  of the Thirtieth {AAAI} Conference on Artificial Intelligence, February
  12-17, 2016, Phoenix, Arizona, {USA}}, pages 3574--3580. {AAAI} Press, 2016.
\newblock URL
  \url{http://www.aaai.org/ocs/index.php/AAAI/AAAI16/paper/view/12501}.

\bibitem[Ficler and Goldberg(2017)]{ficler2017controlling}
Jessica Ficler and Yoav Goldberg.
\newblock Controlling linguistic style aspects in neural language generation.
\newblock In \emph{Proceedings of the Workshop on Stylistic Variation}, pages
  94--104, Copenhagen, Denmark, 2017. Association for Computational
  Linguistics.
\newblock \doi{10.18653/v1/W17-4912}.
\newblock URL \url{https://www.aclweb.org/anthology/W17-4912}.

\bibitem[Keskar et~al.(2019)Keskar, McCann, Varshney, Xiong, and
  Socher]{keskar2019ctrl}
Nitish~Shirish Keskar, Bryan McCann, Lav~R Varshney, Caiming Xiong, and Richard
  Socher.
\newblock Ctrl: A conditional transformer language model for controllable
  generation.
\newblock \emph{arXiv preprint arXiv:1909.05858}, 2019.

\bibitem[Hu et~al.(2017)Hu, Yang, Liang, Salakhutdinov, and Xing]{hu2017toward}
Zhiting Hu, Zichao Yang, Xiaodan Liang, Ruslan Salakhutdinov, and Eric~P. Xing.
\newblock Toward controlled generation of text.
\newblock In Doina Precup and Yee~Whye Teh, editors, \emph{Proceedings of the
  34th International Conference on Machine Learning, {ICML} 2017, Sydney, NSW,
  Australia, 6-11 August 2017}, volume~70 of \emph{Proceedings of Machine
  Learning Research}, pages 1587--1596. {PMLR}, 2017.
\newblock URL \url{http://proceedings.mlr.press/v70/hu17e.html}.

\bibitem[Tu et~al.(2019)Tu, Ding, Yu, and Gimpel]{tu2019generating}
Lifu Tu, Xiaoan Ding, Dong Yu, and Kevin Gimpel.
\newblock Generating diverse story continuations with controllable semantics.
\newblock In \emph{Proceedings of the 3rd Workshop on Neural Generation and
  Translation}, pages 44--58, Hong Kong, 2019. Association for Computational
  Linguistics.
\newblock \doi{10.18653/v1/D19-5605}.
\newblock URL \url{https://www.aclweb.org/anthology/D19-5605}.

\bibitem[Dathathri et~al.(2020)Dathathri, Madotto, Lan, Hung, Frank, Molino,
  Yosinski, and Liu]{dathathri2019plug}
Sumanth Dathathri, Andrea Madotto, Janice Lan, Jane Hung, Eric Frank, Piero
  Molino, Jason Yosinski, and Rosanne Liu.
\newblock Plug and play language models: {A} simple approach to controlled text
  generation.
\newblock In \emph{8th International Conference on Learning Representations,
  {ICLR} 2020, Addis Ababa, Ethiopia, April 26-30, 2020}. OpenReview.net, 2020.
\newblock URL \url{https://openreview.net/forum?id=H1edEyBKDS}.

\bibitem[Takase and Okazaki(2019)]{takase2019positional}
Sho Takase and Naoaki Okazaki.
\newblock Positional encoding to control output sequence length.
\newblock In \emph{Proceedings of the 2019 Conference of the North {A}merican
  Chapter of the Association for Computational Linguistics: Human Language
  Technologies, Volume 1 (Long and Short Papers)}, pages 3999--4004,
  Minneapolis, Minnesota, 2019. Association for Computational Linguistics.
\newblock \doi{10.18653/v1/N19-1401}.
\newblock URL \url{https://www.aclweb.org/anthology/N19-1401}.

\bibitem[Jia et~al.(2015)Jia, Gavves, Fernando, and Tuytelaars]{jia2015guiding}
Xu~Jia, Efstratios Gavves, Basura Fernando, and Tinne Tuytelaars.
\newblock Guiding the long-short term memory model for image caption
  generation.
\newblock In \emph{2015 {IEEE} International Conference on Computer Vision,
  {ICCV} 2015, Santiago, Chile, December 7-13, 2015}, pages 2407--2415. {IEEE}
  Computer Society, 2015.
\newblock \doi{10.1109/ICCV.2015.277}.
\newblock URL \url{https://doi.org/10.1109/ICCV.2015.277}.

\bibitem[Wu et~al.(2016)Wu, Schuster, Chen, Le, Norouzi, Macherey, Krikun, Cao,
  Gao, Macherey, et~al.]{wu2016google}
Yonghui Wu, Mike Schuster, Zhifeng Chen, Quoc~V Le, Mohammad Norouzi, Wolfgang
  Macherey, Maxim Krikun, Yuan Cao, Qin Gao, Klaus Macherey, et~al.
\newblock Google's neural machine translation system: Bridging the gap between
  human and machine translation.
\newblock \emph{arXiv preprint arXiv:1609.08144}, 2016.

\bibitem[Guo et~al.(2018)Guo, Chang, Yu, and Bai]{guo2018improving}
Tszhang Guo, Shiyu Chang, Mo~Yu, and Kun Bai.
\newblock Improving reinforcement learning based image captioning with natural
  language prior.
\newblock In \emph{Proceedings of the 2018 Conference on Empirical Methods in
  Natural Language Processing}, pages 751--756, Brussels, Belgium, 2018.
  Association for Computational Linguistics.
\newblock \doi{10.18653/v1/D18-1083}.
\newblock URL \url{https://www.aclweb.org/anthology/D18-1083}.

\bibitem[sel()]{self-critical-code}
Bad ending rate evaluation.
\newblock
  \url{https://github.com/ruotianluo/self-critical.pytorch/blob/master/eval_utils.py\#L31}.

\bibitem[coc()]{coco-caption-code}
Modified cider.
\newblock \url{https://github.com/ruotianluo/coco-caption/commit/f415e0}.

\bibitem[Luo and Shakhnarovich()]{colab}
Ruotian Luo and Greg Shakhnarovich.
\newblock Colab notebook: Controlling length in image captioning.
\newblock
  \url{https://colab.research.google.com/drive/1TM_KZBixY-L47gHRfXiavgU_T-WfzI3I?usp=sharing}.

\bibitem[Rosch et~al.(1976)Rosch, Mervis, Gray, Johnson, and
  Boyes-Braem]{rosch1976basic}
Eleanor Rosch, Carolyn~B Mervis, Wayne~D Gray, David~M Johnson, and Penny
  Boyes-Braem.
\newblock Basic objects in natural categories.
\newblock \emph{Cognitive psychology}, 8\penalty0 (3):\penalty0 382--439, 1976.

\bibitem[Wilson and Sperber(2002)]{wilson2002relevance}
Deirdre Wilson and Dan Sperber.
\newblock Relevance theory, 2002.

\bibitem[Raghavi~Chandu et~al.(2020)Raghavi~Chandu, Sharma, Changpinyo,
  Thapliyal, and Soricut]{raghavi2020weakly}
Khyathi Raghavi~Chandu, Piyush Sharma, Soravit Changpinyo, Ashish Thapliyal,
  and Radu Soricut.
\newblock Weakly supervised content selection for improved image captioning.
\newblock \emph{arXiv e-prints}, pages arXiv--2009, 2020.

\bibitem[Wei et~al.(2014)Wei, Xia, Huang, Ni, Dong, Zhao, and Yan]{wei2014cnn}
Yunchao Wei, Wei Xia, Junshi Huang, Bingbing Ni, Jian Dong, Yao Zhao, and
  Shuicheng Yan.
\newblock Cnn: Single-label to multi-label.
\newblock \emph{arXiv preprint arXiv:1406.5726}, 2014.

\bibitem[Gong et~al.(2014{\natexlab{b}})Gong, Jia, Leung, Toshev, and
  Ioffe]{gong2013deep}
Yunchao Gong, Yangqing Jia, Thomas Leung, Alexander Toshev, and Sergey Ioffe.
\newblock Deep convolutional ranking for multilabel image annotation.
\newblock In Yoshua Bengio and Yann LeCun, editors, \emph{2nd International
  Conference on Learning Representations, {ICLR} 2014, Banff, AB, Canada, April
  14-16, 2014, Conference Track Proceedings}, 2014{\natexlab{b}}.
\newblock URL \url{http://arxiv.org/abs/1312.4894}.

\bibitem[Mahajan et~al.(2018)Mahajan, Girshick, Ramanathan, He, Paluri, Li,
  Bharambe, and Van Der~Maaten]{mahajan2018exploring}
Dhruv Mahajan, Ross Girshick, Vignesh Ramanathan, Kaiming He, Manohar Paluri,
  Yixuan Li, Ashwin Bharambe, and Laurens Van Der~Maaten.
\newblock Exploring the limits of weakly supervised pretraining.
\newblock In \emph{Proceedings of the European Conference on Computer Vision
  (ECCV)}, pages 181--196, 2018.

\bibitem[Yeh et~al.(2017)Yeh, Wu, Ko, and Wang]{yeh2017learning}
Chih{-}Kuan Yeh, Wei{-}Chieh Wu, Wei{-}Jen Ko, and Yu{-}Chiang~Frank Wang.
\newblock Learning deep latent space for multi-label classification.
\newblock In Satinder~P. Singh and Shaul Markovitch, editors, \emph{Proceedings
  of the Thirty-First {AAAI} Conference on Artificial Intelligence, February
  4-9, 2017, San Francisco, California, {USA}}, pages 2838--2844. {AAAI} Press,
  2017.
\newblock URL \url{http://aaai.org/ocs/index.php/AAAI/AAAI17/paper/view/14166}.

\bibitem[Lin et~al.(2014{\natexlab{b}})Lin, Ding, Hu, and Wang]{lin2014multi}
Zijia Lin, Guiguang Ding, Mingqing Hu, and Jianmin Wang.
\newblock Multi-label classification via feature-aware implicit label space
  encoding.
\newblock In \emph{Proceedings of the 31th International Conference on Machine
  Learning, {ICML} 2014, Beijing, China, 21-26 June 2014}, volume~32 of
  \emph{{JMLR} Workshop and Conference Proceedings}, pages 325--333. JMLR.org,
  2014{\natexlab{b}}.
\newblock URL \url{http://proceedings.mlr.press/v32/linc14.html}.

\bibitem[Ghamrawi and McCallum(2005)]{ghamrawi2005collective}
Nadia Ghamrawi and Andrew McCallum.
\newblock Collective multi-label classification.
\newblock In \emph{Proceedings of the 14th ACM international conference on
  Information and knowledge management}, pages 195--200, 2005.

\bibitem[Jin and Nakayama(2016)]{jin2016annotation}
Jiren Jin and Hideki Nakayama.
\newblock Annotation order matters: Recurrent image annotator for arbitrary
  length image tagging.
\newblock In \emph{2016 23rd International Conference on Pattern Recognition
  (ICPR)}, pages 2452--2457. IEEE, 2016.

\bibitem[Vinyals et~al.(2016)Vinyals, Bengio, and Kudlur]{vinyals2015order}
Oriol Vinyals, Samy Bengio, and Manjunath Kudlur.
\newblock Order matters: Sequence to sequence for sets.
\newblock In Yoshua Bengio and Yann LeCun, editors, \emph{4th International
  Conference on Learning Representations, {ICLR} 2016, San Juan, Puerto Rico,
  May 2-4, 2016, Conference Track Proceedings}, 2016.
\newblock URL \url{http://arxiv.org/abs/1511.06391}.

\bibitem[Yang et~al.(2019{\natexlab{b}})Yang, Luo, Ma, Lin, and
  Sun]{yang2019deep}
Pengcheng Yang, Fuli Luo, Shuming Ma, Junyang Lin, and Xu~Sun.
\newblock A deep reinforced sequence-to-set model for multi-label
  classification.
\newblock In \emph{Proceedings of the 57th Annual Meeting of the Association
  for Computational Linguistics}, pages 5252--5258, Florence, Italy,
  2019{\natexlab{b}}. Association for Computational Linguistics.
\newblock \doi{10.18653/v1/P19-1518}.
\newblock URL \url{https://www.aclweb.org/anthology/P19-1518}.

\bibitem[Salvador et~al.(2019)Salvador, Drozdzal, Gir{\'{o}}{-}i{-}Nieto, and
  Romero]{salvador2019inverse}
Amaia Salvador, Michal Drozdzal, Xavier Gir{\'{o}}{-}i{-}Nieto, and Adriana
  Romero.
\newblock Inverse cooking: Recipe generation from food images.
\newblock In \emph{{IEEE} Conference on Computer Vision and Pattern
  Recognition, {CVPR} 2019, Long Beach, CA, USA, June 16-20, 2019}, pages
  10453--10462. Computer Vision Foundation / {IEEE}, 2019.
\newblock \doi{10.1109/CVPR.2019.01070}.
\newblock URL
  \url{http://openaccess.thecvf.com/content\_CVPR\_2019/html/Salvador\_Inverse\_Cooking\_Recipe\_Generation\_From\_Food\_Images\_CVPR\_2019\_paper.html}.

\bibitem[Yazici et~al.(2020)Yazici, Gonzalez{-}Garcia, Ramisa, Twardowski, and
  van~de Weijer]{yazici2020orderless}
Vacit~Oguz Yazici, Abel Gonzalez{-}Garcia, Arnau Ramisa, Bartlomiej Twardowski,
  and Joost van~de Weijer.
\newblock Orderless recurrent models for multi-label classification.
\newblock In \emph{2020 {IEEE/CVF} Conference on Computer Vision and Pattern
  Recognition, {CVPR} 2020, Seattle, WA, USA, June 13-19, 2020}, pages
  13437--13446. {IEEE}, 2020.
\newblock \doi{10.1109/CVPR42600.2020.01345}.
\newblock URL \url{https://doi.org/10.1109/CVPR42600.2020.01345}.

\bibitem[Pineda et~al.(2019)Pineda, Salvador, Drozdzal, and
  Romero]{pineda2019elucidating}
Luis Pineda, Amaia Salvador, Michal Drozdzal, and Adriana Romero.
\newblock Elucidating image-to-set prediction: An analysis of models, losses
  and datasets.
\newblock \emph{arXiv preprint arXiv:1904.05709}, 2019.

\bibitem[Xu et~al.(2018)Xu, Ren, Zhang, Zeng, Cai, and Sun]{xu2018skeleton}
Jingjing Xu, Xuancheng Ren, Yi~Zhang, Qi~Zeng, Xiaoyan Cai, and Xu~Sun.
\newblock A skeleton-based model for promoting coherence among sentences in
  narrative story generation.
\newblock In \emph{Proceedings of the 2018 Conference on Empirical Methods in
  Natural Language Processing}, pages 4306--4315, Brussels, Belgium, 2018.
  Association for Computational Linguistics.
\newblock \doi{10.18653/v1/D18-1462}.
\newblock URL \url{https://www.aclweb.org/anthology/D18-1462}.

\bibitem[Yao et~al.(2019)Yao, Peng, Weischedel, Knight, Zhao, and
  Yan]{yao2019plan}
Lili Yao, Nanyun Peng, Ralph Weischedel, Kevin Knight, Dongyan Zhao, and Rui
  Yan.
\newblock Plan-and-write: Towards better automatic storytelling.
\newblock In \emph{Proceedings of the AAAI Conference on Artificial
  Intelligence}, volume~33, pages 7378--7385, 2019.

\bibitem[Mao et~al.(2019)Mao, Majumder, McAuley, and
  Cottrell]{mao2019improving}
Huanru~Henry Mao, Bodhisattwa~Prasad Majumder, Julian McAuley, and Garrison
  Cottrell.
\newblock Improving neural story generation by targeted common sense grounding.
\newblock In \emph{Proceedings of the 2019 Conference on Empirical Methods in
  Natural Language Processing and the 9th International Joint Conference on
  Natural Language Processing (EMNLP-IJCNLP)}, pages 5988--5993, Hong Kong,
  China, 2019. Association for Computational Linguistics.
\newblock \doi{10.18653/v1/D19-1615}.
\newblock URL \url{https://www.aclweb.org/anthology/D19-1615}.

\bibitem[Tan et~al.(2020)Tan, Yang, AI-Shedivat, Xing, and
  Hu]{tan2020progressive}
Bowen Tan, Zichao Yang, Maruan AI-Shedivat, Eric~P Xing, and Zhiting Hu.
\newblock Progressive generation of long text.
\newblock \emph{arXiv preprint arXiv:2006.15720}, 2020.

\bibitem[Lee et~al.(2019)Lee, Lee, Kim, Kosiorek, Choi, and Teh]{lee2019set}
Juho Lee, Yoonho Lee, Jungtaek Kim, Adam~R. Kosiorek, Seungjin Choi, and
  Yee~Whye Teh.
\newblock Set transformer: {A} framework for attention-based
  permutation-invariant neural networks.
\newblock In Kamalika Chaudhuri and Ruslan Salakhutdinov, editors,
  \emph{Proceedings of the 36th International Conference on Machine Learning,
  {ICML} 2019, 9-15 June 2019, Long Beach, California, {USA}}, volume~97 of
  \emph{Proceedings of Machine Learning Research}, pages 3744--3753. {PMLR},
  2019.
\newblock URL \url{http://proceedings.mlr.press/v97/lee19d.html}.

\bibitem[Lewis et~al.(2020)Lewis, Liu, Goyal, Ghazvininejad, Mohamed, Levy,
  Stoyanov, and Zettlemoyer]{lewis2019bart}
Mike Lewis, Yinhan Liu, Naman Goyal, Marjan Ghazvininejad, Abdelrahman Mohamed,
  Omer Levy, Veselin Stoyanov, and Luke Zettlemoyer.
\newblock {BART}: Denoising sequence-to-sequence pre-training for natural
  language generation, translation, and comprehension.
\newblock In \emph{Proceedings of the 58th Annual Meeting of the Association
  for Computational Linguistics}, pages 7871--7880, Online, 2020. Association
  for Computational Linguistics.
\newblock \doi{10.18653/v1/2020.acl-main.703}.
\newblock URL \url{https://www.aclweb.org/anthology/2020.acl-main.703}.

\bibitem[Hermann et~al.(2015)Hermann, Kocisk{\'{y}}, Grefenstette, Espeholt,
  Kay, Suleyman, and Blunsom]{hermann2015teaching}
Karl~Moritz Hermann, Tom{\'{a}}s Kocisk{\'{y}}, Edward Grefenstette, Lasse
  Espeholt, Will Kay, Mustafa Suleyman, and Phil Blunsom.
\newblock Teaching machines to read and comprehend.
\newblock In Corinna Cortes, Neil~D. Lawrence, Daniel~D. Lee, Masashi Sugiyama,
  and Roman Garnett, editors, \emph{Advances in Neural Information Processing
  Systems 28: Annual Conference on Neural Information Processing Systems 2015,
  December 7-12, 2015, Montreal, Quebec, Canada}, pages 1693--1701, 2015.
\newblock URL
  \url{https://proceedings.neurips.cc/paper/2015/hash/afdec7005cc9f14302cd0474fd0f3c96-Abstract.html}.

\bibitem[Merity et~al.(2017)Merity, Xiong, Bradbury, and
  Socher]{merity2016pointer}
Stephen Merity, Caiming Xiong, James Bradbury, and Richard Socher.
\newblock Pointer sentinel mixture models.
\newblock In \emph{5th International Conference on Learning Representations,
  {ICLR} 2017, Toulon, France, April 24-26, 2017, Conference Track
  Proceedings}. OpenReview.net, 2017.
\newblock URL \url{https://openreview.net/forum?id=Byj72udxe}.

\end{thebibliography}

\appendix

\chapter{Appendix: Discriminability in image captioning}\label{app:disccap}

\section{Comparison to \cite{dai2017contrastive,dai2017towards,shetty2017speaking}} \label{app:disccap_comparison}
First, while the losses in \cite{dai2017contrastive,dai2017towards,shetty2017speaking} and our
discriminability objective all deal with matched vs. mismatched
image/caption pairs, the actual formulations differ.

If $(\I,\bc)$ is the matched image/caption pair, $\I',\bc'$ are
other images/captions, and $\widehat{\bc}$ the generated caption, the
losses are related to the following notions of ``contrast'' :
\begin{align*}
  &\text{\cite{dai2017contrastive}:}~(\I,\bc)\,\text{vs}\,(\I,\bc'),\\
  &\text{\cite{dai2017towards,shetty2017speaking}}:~(\I,\bc)\,\text{vs}\,(\I,\widehat{\bc}),\,(\I,\bc)\,\text{vs}\,(\I,\bc')\\
  &\text{Ours:}~(\I,\widehat{\bc})\,\text{vs}\,(\I,\widehat{\bc}'),\,
    (\I,\widehat{\bc})\,\text{vs}\,(\I',\widehat{\bc})
\end{align*}

Second, compared to \cite{dai2017towards,shetty2017speaking}, we have a different focus and different evaluation results.
~\cite{dai2017towards,shetty2017speaking} focuses on generating a diverse set of natural captions; in
contrast, we go after discriminability (accuracy) with a single
caption.  We demonstrate improvement in \textbf{both} standard metrics and
our target (discriminability), while they have to sacrifice the former
to improve the performance under their target (diversity).

On SPICE subcategory score, count and size are improved after applying the GAN training in~\cite{shetty2017speaking}, while our discriminability objective helps color, attribute and count. That implies that our discriminability objective is in favor of different aspects of captions from the discriminator in GAN.

Third, the aims in \cite{dai2017contrastive} are more aligned with ours,
but \cite{dai2017contrastive} do not
explicitly incorporate the discriminative task into learning, while we
do. It is also not clear that \cite{dai2017contrastive} could allow for CIDEr optimization,
which we do allow for, and which appears to lead to significant
improvement over MLE.

Finally, none of \cite{dai2017contrastive,dai2017towards,shetty2017speaking} address discriminability of resulting
captions, or report discriminability results comparable to ours.

\section{More qualitative results}\label{app:disccap_qual}
\begin{figure}[htbp]
\begin{minipage}[t]{.45\linewidth}
\includegraphics[height=4cm]{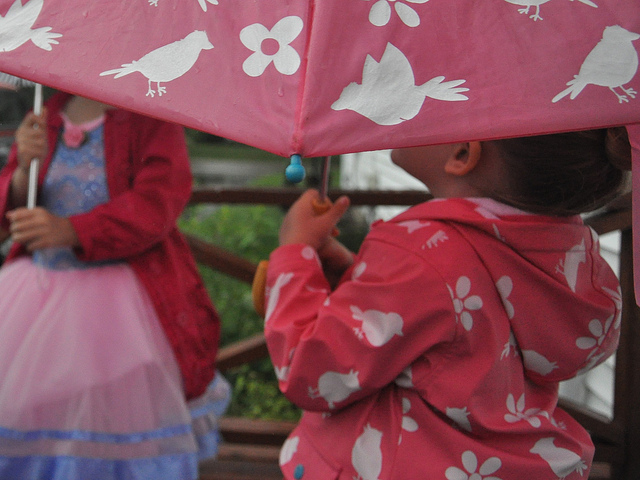}\\
{\footnotesize
{\bf Human}: a young child holding an umbrella with birds and flowers
\\
{\bf Att2in+CIDER~\cite{rennie2017self}}: a group of people standing in the rain with an umbrella\\
{\bf Ours}:a little girl holding an umbrella in the rain
}
\end{minipage}%
\hspace{1em}
\begin{minipage}[t]{.45\linewidth}
\includegraphics[height=4cm]{figures/disccap/COCO_train2014_000000231878.jpg}\\
{\footnotesize
{\bf Human}: costumed wait staff standing in front of a restaurant awaiting customers\\
{\bf Att2in+CIDER~\cite{rennie2017self}}: a group of people standing in the rain with an umbrella\\
{\bf Ours}: a group of people standing in front of a building
}
\end{minipage}

\begin{minipage}[t]{.45\linewidth}
\includegraphics[height=4cm]{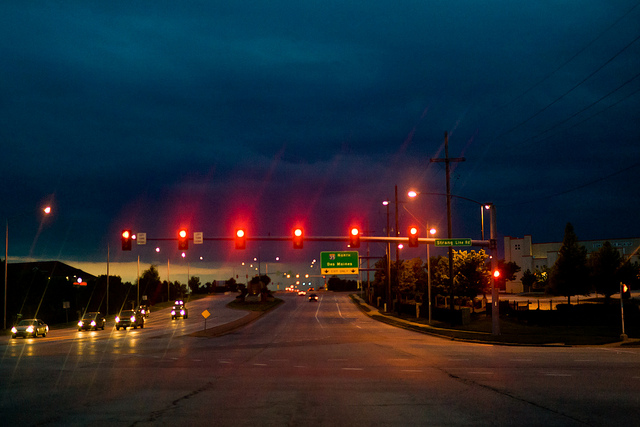}\\
{\footnotesize
{\bf Human}: a street that goes on to a high way with the light on red
\\
{\bf Att2in+CIDER~\cite{rennie2017self}}: a traffic light on the side of a city street\\
{\bf Ours}:a street at night with traffic lights at night
}
\end{minipage}%
\hspace{1em}
\begin{minipage}[t]{.45\linewidth}
\includegraphics[height=4cm]{figures/disccap/COCO_train2014_000000554439.jpg}\\
{\footnotesize
{\bf Human}: view of tourist tower behind a traffic signal\\
{\bf Att2in+CIDER~\cite{rennie2017self}}: a traffic light on the side of a city street\\
{\bf Ours}: a traffic light sitting on the side of a city
}
\end{minipage}

\begin{minipage}[t]{.45\linewidth}
\includegraphics[height=4cm]{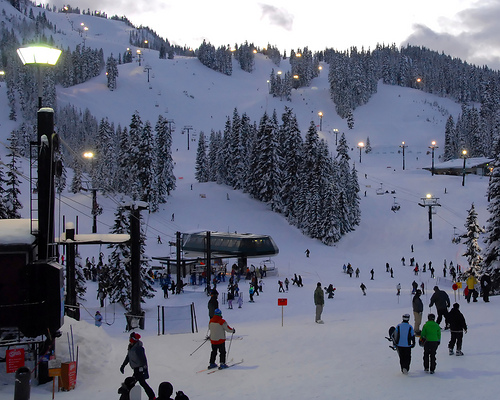}\\
{\footnotesize
{\bf Human}: people skiing in the snow on the mountainside
\\
{\bf Att2in+CIDER~\cite{rennie2017self}}: a group of people standing on skis in the snow\\
{\bf Ours}:a group of people skiing down a snow covered slope
}
\end{minipage}%
\hspace{1em}
\begin{minipage}[t]{.45\linewidth}
\includegraphics[height=4cm]{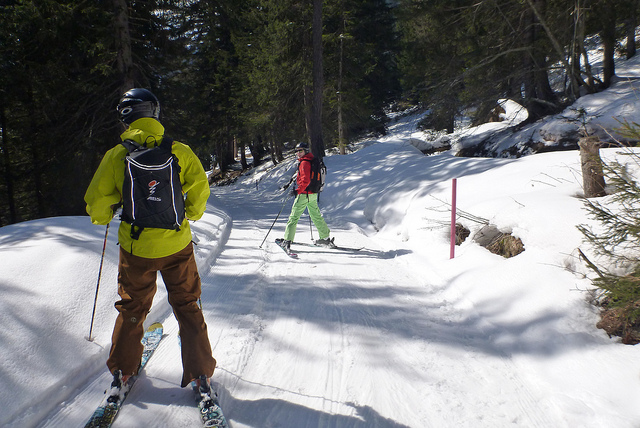}\\
{\footnotesize
{\bf Human}: two skiers travel along a snowy path towards trees\\
{\bf Att2in+CIDER~\cite{rennie2017self}}: a group of people standing on skis in the snow\\
{\bf Ours}: two people standing on skis in the snow
}
\end{minipage}
\caption{The human caption, caption generated by Att2in+CIDER, and caption generated by 
Att2in+CIDER+DISC(1) (denoted as Ours in the figure) for each image.}
\label{fig:disccap_qual+}
\end{figure}

In Figure \ref{fig:disccap_qual+}, we show pairs of images that have the same caption
generated by Att2in+CIDER~\cite{rennie2017self}, where our method (Att2in+CIDER+DISC(1)) can generate
more specific captions.

\section{More results}
\begin{figure}
\begin{minipage}[t]{\linewidth}
\begin{minipage}[t]{.45\linewidth}
\includegraphics[height=4cm]{figures/disccap/COCO_val2014_000000358750.jpg}
\end{minipage}%
\hspace{1em}
\begin{minipage}[t]{.45\linewidth}
\includegraphics[height=4cm]{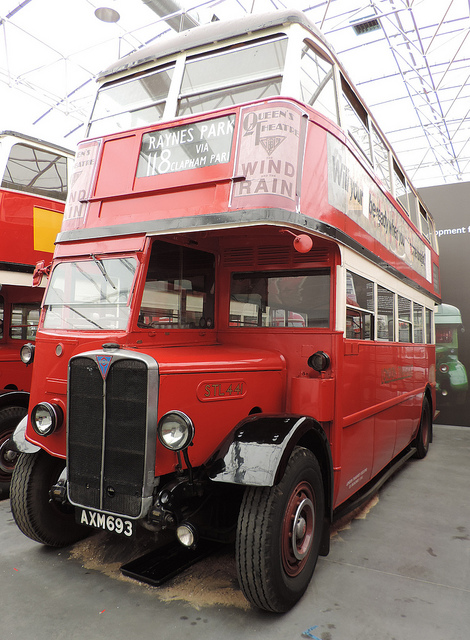}
\end{minipage}
\end{minipage}
{\footnotesize
{\bf Att2in+MLE}: a double decker bus parked in front of a building
\\
{\bf Att2in+CIDER~\cite{rennie2017self}}: a double decker bus parked in front of a building\\
{\bf Att2in+CIDER+DISC(1)}: a \textcolor{green!50!black}{blue} double decker bus parked in front of a building\\
{\bf Att2in+CIDER+DISC(10)}: a \textcolor{green!50!black}{blue} double decker bus parked in front of a \textcolor{green!50!black}{brick} building\\
}
\begin{minipage}[t]{\linewidth}
\begin{minipage}[t]{.45\linewidth}
\includegraphics[height=4cm]{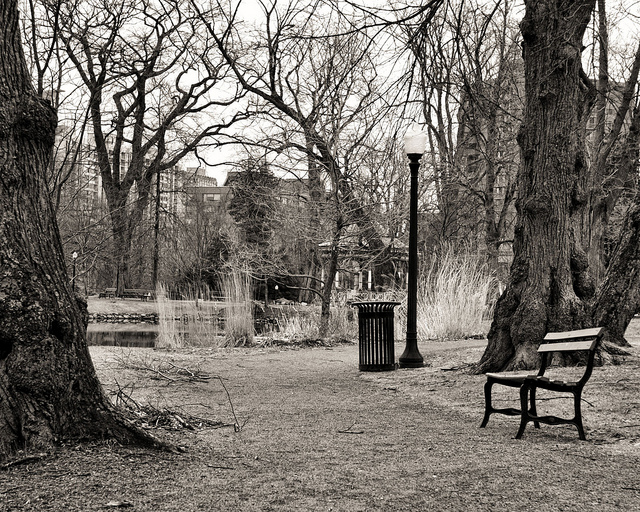}
\end{minipage}%
\hspace{1em}
\begin{minipage}[t]{.45\linewidth}
\includegraphics[height=4cm]{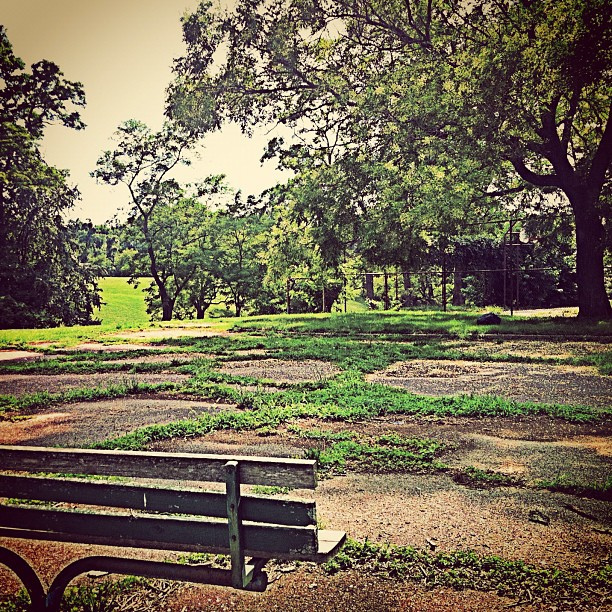}
\end{minipage}
\end{minipage}
{\footnotesize
{\bf Att2in+MLE}: a park bench sitting next to a tree
\\
{\bf Att2in+CIDER~\cite{rennie2017self}}: a bench sitting in the middle of a tree\\
{\bf Att2in+CIDER+DISC(1)}: a \textcolor{green!50!black}{black and white} photo of a bench in the park\\
{\bf Att2in+CIDER+DISC(10)}: a \textcolor{green!50!black}{black and white} photo of a park with a tree\\
}
\begin{minipage}[t]{\linewidth}
\begin{minipage}[t]{.45\linewidth}
\includegraphics[height=4cm]{figures/disccap/COCO_train2014_000000047084.jpg}
\end{minipage}%
\hspace{1em}
\begin{minipage}[t]{.45\linewidth}
\includegraphics[height=4cm]{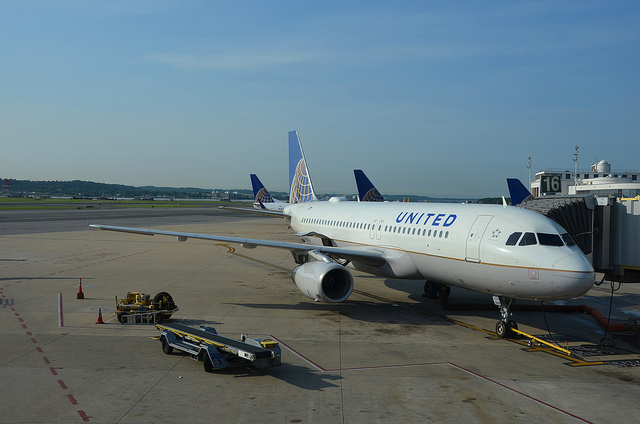}
\end{minipage}
\end{minipage}
{\footnotesize
{\bf Att2in+MLE}: a large jetliner sitting on top of an airport tarmac
\\
{\bf Att2in+CIDER~\cite{rennie2017self}}: a large airplane sitting on the runway at an airport\\
{\bf Att2in+CIDER+DISC(1)}: a \textcolor{green!50!black}{blue} airplane sitting on the tarmac at an airport\\
{\bf Att2in+CIDER+DISC(10)}: a \textcolor{green!50!black}{blue} \textcolor{red}{and blue} airplane sitting on a tarmac with a plane\\
}
\caption{Examples of cases where our method improves the color accuracy of generated captions.}
\label{fig:disccap_color}
\end{figure}

\begin{figure}
\begin{minipage}[t]{\linewidth}
\begin{minipage}[t]{.45\linewidth}
\includegraphics[height=4cm]{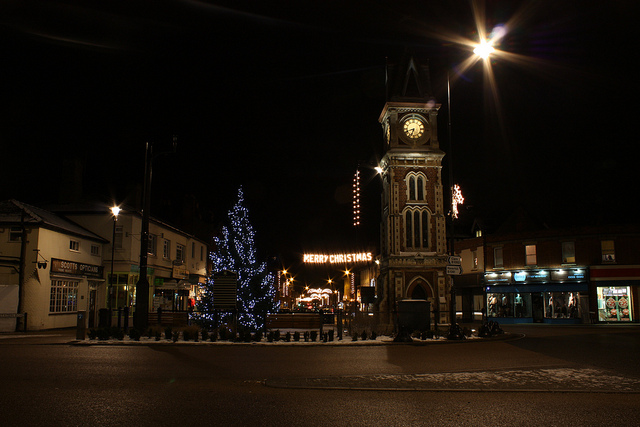}
\end{minipage}%
\hspace{2.5em}
\begin{minipage}[t]{.45\linewidth}
\includegraphics[height=4cm]{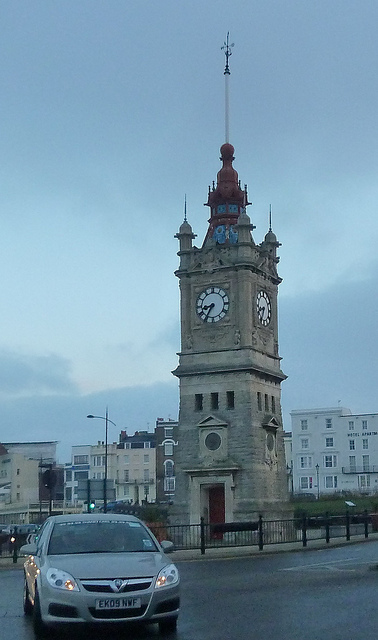}
\end{minipage}
\end{minipage}
{\footnotesize
{\bf Att2in+MLE}: a large clock tower towering over a city street
\\
{\bf Att2in+CIDER~\cite{rennie2017self}}: a clock tower in the middle of a city street\\
{\bf Att2in+CIDER+DISC(1)}: a city street with a clock tower \textcolor{green!50!black}{at night}\\
{\bf Att2in+CIDER+DISC(10)}: a lit building with a clock tower \textcolor{green!50!black}{at night} \textcolor{red}{at night}\\
}
\begin{minipage}[t]{\linewidth}
\begin{minipage}[t]{.45\linewidth}
\includegraphics[height=4cm]{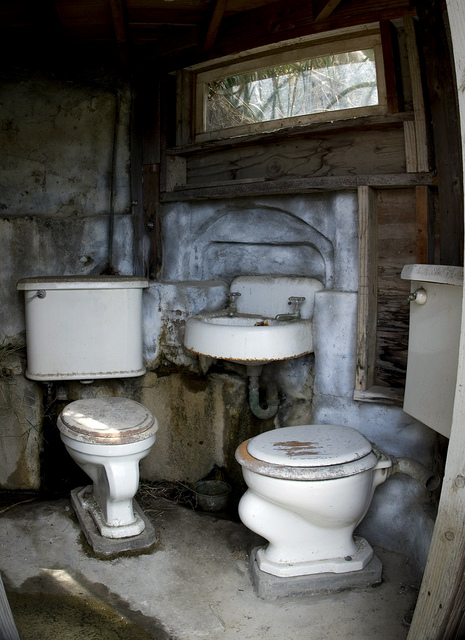}
\end{minipage}%
\hspace{1em}
\begin{minipage}[t]{.45\linewidth}
\includegraphics[height=4cm]{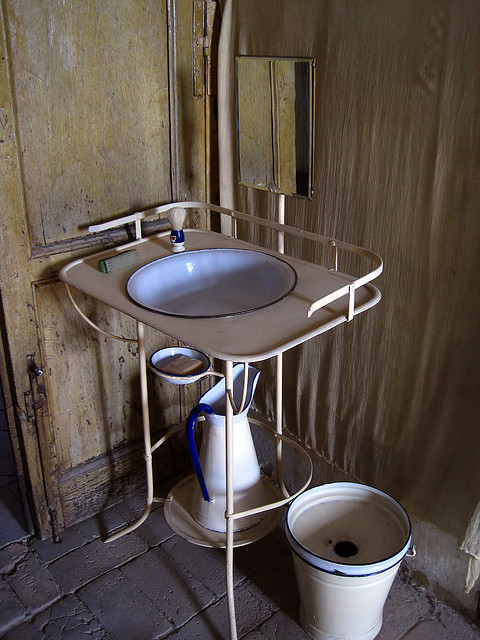}
\end{minipage}
\end{minipage}
{\footnotesize
{\bf Att2in+MLE}: a bathroom with a toilet and a toilet
\\
{\bf Att2in+CIDER~\cite{rennie2017self}}: a bathroom with a toilet and sink in the\\
{\bf Att2in+CIDER+DISC(1)}: a \textcolor{green!50!black}{dirty} bathroom with a toilet and a window\\
{\bf Att2in+CIDER+DISC(10)}: a \textcolor{green!50!black}{dirty} bathroom with a toilet and \textcolor{red}{a dirty}\\
}
\caption{Examples of cases where our method improves the attribute accuracy of generated captions.}
\label{fig:disccap_attribute}
\end{figure}

\begin{figure}
\begin{minipage}[t]{\linewidth}
\begin{minipage}[t]{.45\linewidth}
\includegraphics[height=4cm]{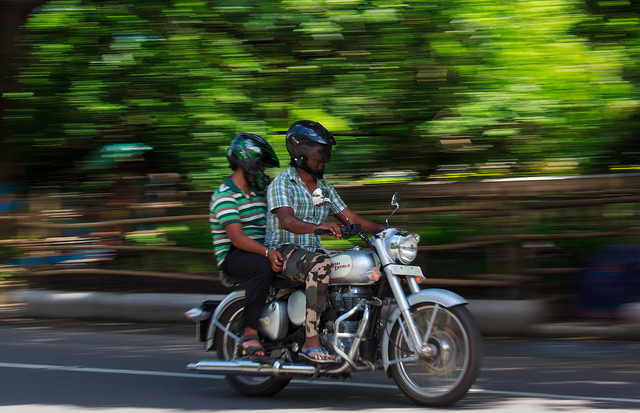}
\end{minipage}%
\hspace{3em}
\begin{minipage}[t]{.45\linewidth}
\includegraphics[height=4cm]{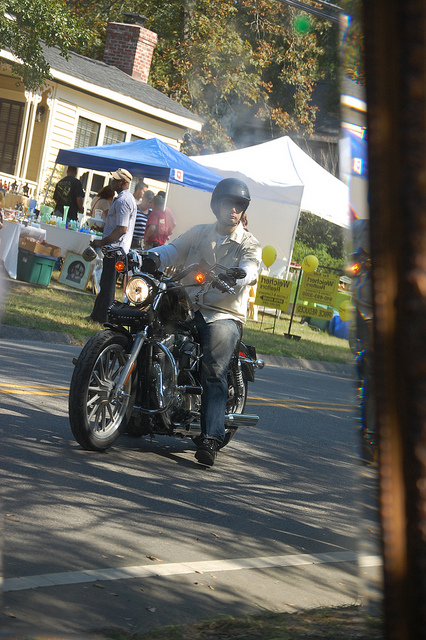}
\end{minipage}
\end{minipage}
{\footnotesize
{\bf Att2in+MLE}: a couple of people riding on the back of a motorcycle
\\
{\bf Att2in+CIDER~\cite{rennie2017self}}: a man riding a motorcycle on the street\\
{\bf Att2in+CIDER+DISC(1)}: \textcolor{green!50!black}{two people} riding a motorcycle on a city street\\
{\bf Att2in+CIDER+DISC(10)}: \textcolor{green!50!black}{two people} riding a motorcycle on a road\\
}
\begin{minipage}[t]{\linewidth}
\begin{minipage}[t]{.45\linewidth}
\includegraphics[height=4cm]{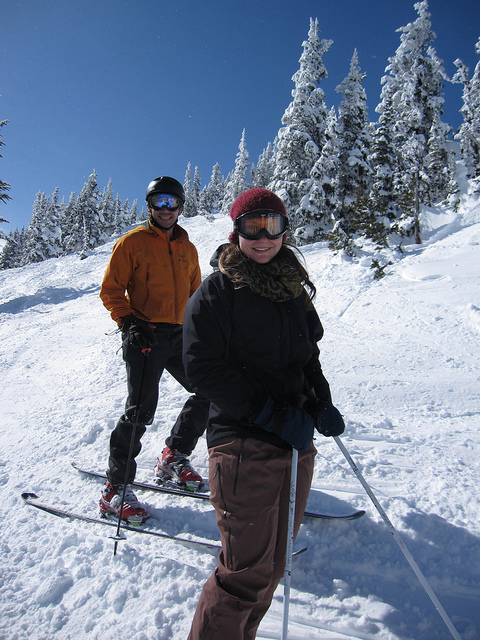}
\end{minipage}%
\hspace{1em}
\begin{minipage}[t]{.45\linewidth}
\includegraphics[height=4cm]{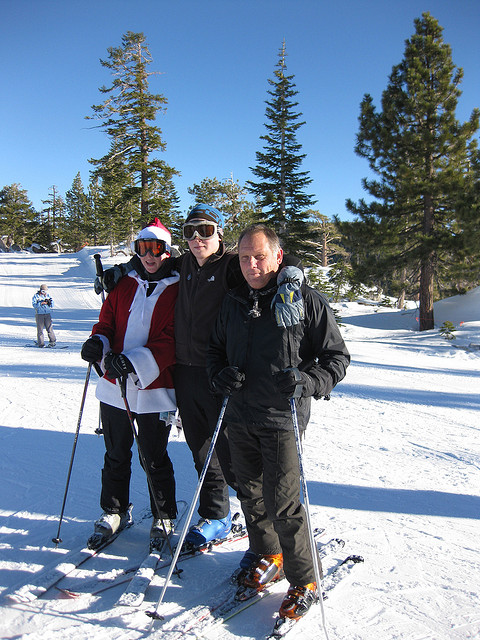}
\end{minipage}
\end{minipage}
{\footnotesize
{\bf Att2in+MLE}: a couple of people standing on top of a snow covered slope
\\
{\bf Att2in+CIDER~\cite{rennie2017self}}: a couple of people standing on skis in the snow\\
{\bf Att2in+CIDER+DISC(1)}: \textcolor{green!50!black}{two people} standing on skis in the snow\\
{\bf Att2in+CIDER+DISC(10)}: \textcolor{green!50!black}{two people} standing in the snow with a snow\\
}
\caption{Examples of cases where our method improves the cardinality accuracy of generated captions.}
\label{fig:disccap_cardinality}
\end{figure}

Following up the analysis of SPICE sub-scores in the paper, showing improvement due to our proposed objective, 
we show in Figures \ref{fig:disccap_color}, \ref{fig:disccap_attribute}, \ref{fig:disccap_cardinality} some examples demonstrating such improvement on color, attribute, and cardinality aspects of the scene. For each figure, the captions below are describing the target images(left) generated from different models (Att2in+MLE and Att2in+CIDER are baselines, and Att2in+CIDER+DISC(x) are our method with different $\lambda$ value). Right images are the corresponding distractors selected in val/test set; these pairs were included in AMT experiments.

\begin{figure}
\begin{minipage}[t]{\linewidth}
\begin{minipage}[t]{.45\linewidth}
\includegraphics[height=4cm]{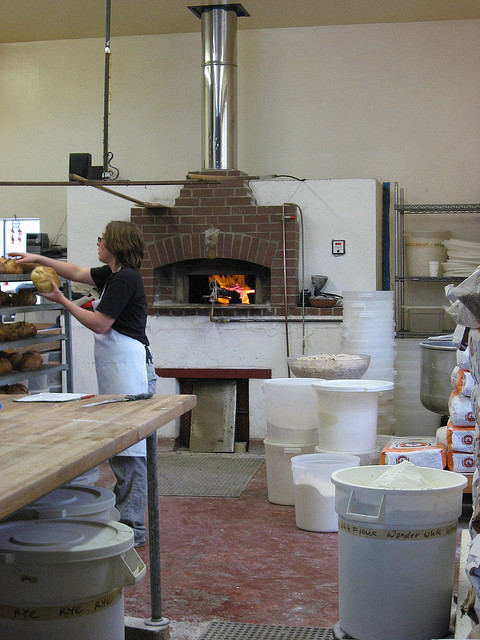}
\end{minipage}%
\hspace{1em}
\begin{minipage}[t]{.45\linewidth}
\includegraphics[height=4cm]{figures/disccap/COCO_val2014_000000505728.jpg}
\end{minipage}
\end{minipage}
{\footnotesize
{\bf Att2in+MLE}: a woman standing in a kitchen preparing food
\\
{\bf Att2in+CIDER~\cite{rennie2017self}}: a woman standing in a kitchen preparing food\\
{\bf Att2in+CIDER+DISC(1)}: a woman standing in a kitchen with \textcolor{green!50!black}{a fireplace}\\
{\bf Att2in+CIDER+DISC(10)}: a woman standing in a kitchen with \textcolor{green!50!black}{a brick oven}\\
}
%
\begin{minipage}[t]{\linewidth}
\begin{minipage}[t]{.45\linewidth}
\includegraphics[height=3cm]{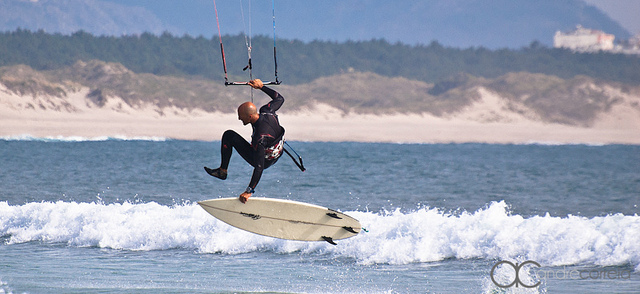}
\end{minipage}%
\hspace{1em}
\begin{minipage}[t]{.45\linewidth}
\includegraphics[height=3cm]{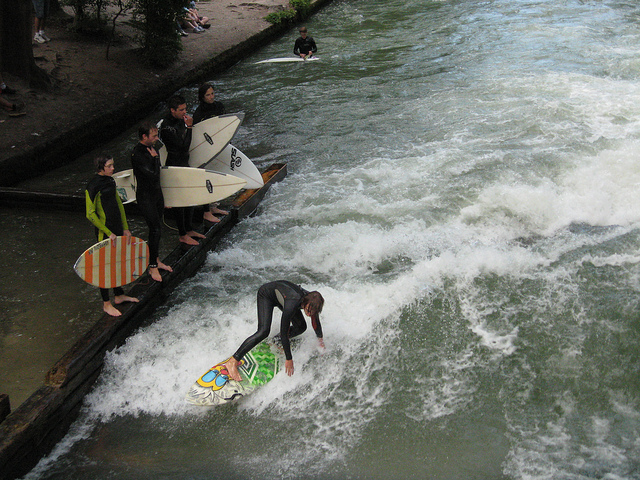}
\end{minipage}
\end{minipage}
{\footnotesize
{\bf Att2in+MLE}: a man on a surfboard in the water
\\
{\bf Att2in+CIDER~\cite{rennie2017self}}: a man riding a wave on a surfboard in the ocean\\
{\bf Att2in+CIDER+DISC(1)}: a man riding a \textcolor{green!50!black}{kiteboard} \textcolor{red}{on the ocean in the ocean}\\
{\bf Att2in+CIDER+DISC(10)}: a man \textcolor{green!50!black}{kiteboarding} \textcolor{red}{in the ocean on a ocean}\\
}
%
\begin{minipage}[t]{\linewidth}
\begin{minipage}[t]{.45\linewidth}
\includegraphics[height=3cm]{figures/disccap/COCO_train2014_000000493692.jpg}
\end{minipage}%
\hspace{1em}
\begin{minipage}[t]{.45\linewidth}
\includegraphics[height=3cm]{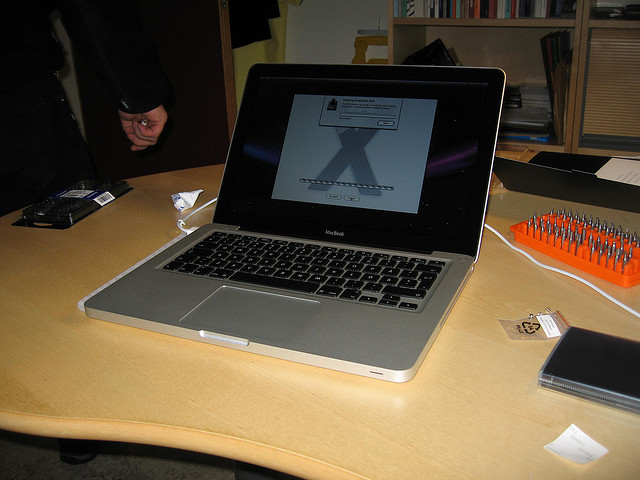}
\end{minipage}
\end{minipage}
{\footnotesize
{\bf Att2in+MLE}: a room with a laptop and a laptop
\\
{\bf Att2in+CIDER~\cite{rennie2017self}}: a laptop computer sitting on top of a table\\
{\bf Att2in+CIDER+DISC(1)}: a room with a laptop computer and \textcolor{green!50!black}{chairs} in it\\
{\bf Att2in+CIDER+DISC(10)}: a room with a laptops and \textcolor{green!50!black}{chairs} in a building\\
}
%
\begin{minipage}[t]{\linewidth}
\begin{minipage}[t]{.45\linewidth}
\includegraphics[height=3cm]{figures/disccap/COCO_train2014_000000388031.jpg}
\end{minipage}%
\hspace{1em}
\begin{minipage}[t]{.45\linewidth}
\includegraphics[height=3cm]{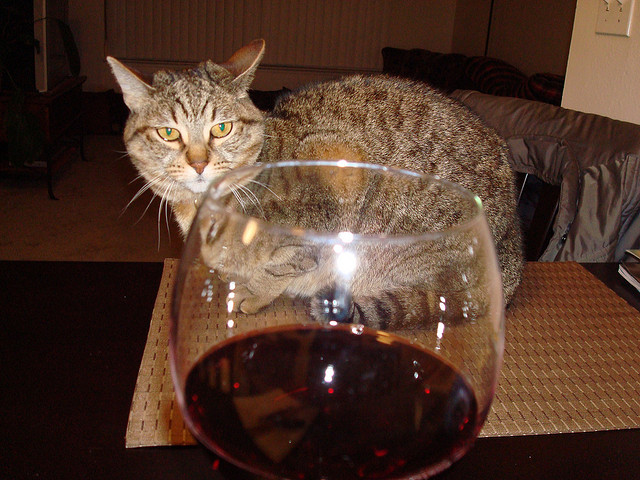}
\end{minipage}
\end{minipage}
{\footnotesize
{\bf Att2in+MLE}: a cat sitting next to a glass of wine
\\
{\bf Att2in+CIDER~\cite{rennie2017self}}: a cat sitting next to a glass of wine\\
{\bf Att2in+CIDER+DISC(1)}: a cat sitting next to a \textcolor{green!50!black}{bottles} of wine\\
{\bf Att2in+CIDER+DISC(10)}: a cat sitting next to a \textcolor{green!50!black}{bottles} of wine \textcolor{red}{bottles}\\
}
\caption{More examples.}
\label{fig:disccap_amt+}
\end{figure}

\begin{figure}
%
\begin{minipage}[t]{\linewidth}
\begin{minipage}[t]{.45\linewidth}
\includegraphics[height=3cm]{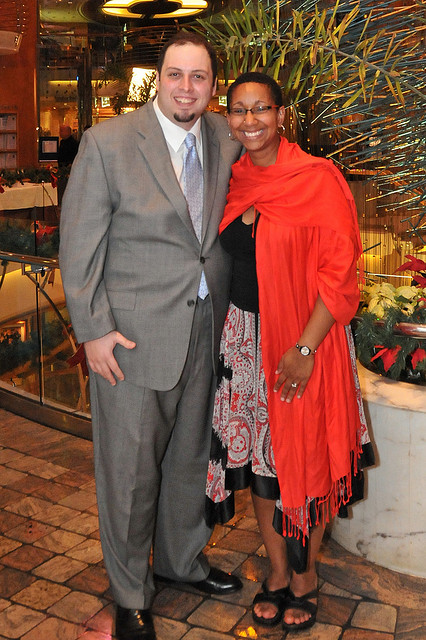}
\end{minipage}%
\hspace{1em}
\begin{minipage}[t]{.45\linewidth}
\includegraphics[height=3cm]{figures/disccap/COCO_train2014_000000428050.jpg}
\end{minipage}
\end{minipage}
{\footnotesize
{\bf Att2in+MLE}: a couple of people standing next to each other
\\
{\bf Att2in+CIDER~\cite{rennie2017self}}: a man and a woman standing in a room\\
{\bf Att2in+CIDER+DISC(1)}: a man and a woman standing in a room\\
{\bf Att2in+CIDER+DISC(10)}: \textcolor{red}{two men} standing in a room with a \textcolor{red}{red tie}\\
}
\begin{minipage}[t]{\linewidth}
\begin{minipage}[t]{.45\linewidth}
\includegraphics[height=3cm]{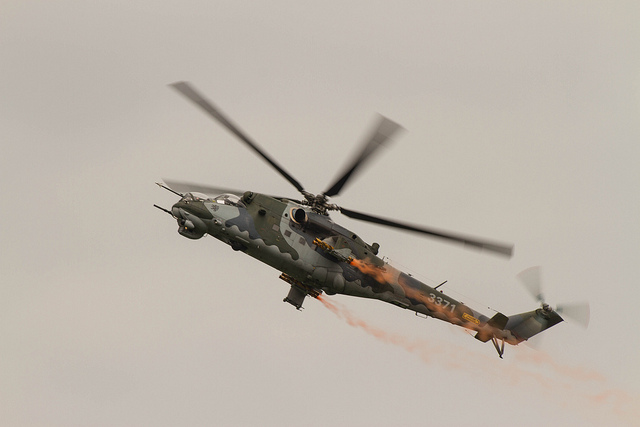}
\end{minipage}%
\hspace{1em}
\begin{minipage}[t]{.45\linewidth}
\includegraphics[height=3cm]{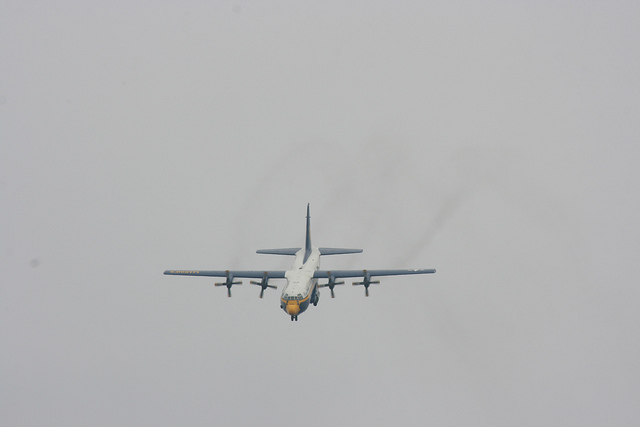}
\end{minipage}
\end{minipage}
{\footnotesize
{\bf Att2in+MLE}: a helicopter that is flying in the sky
\\
{\bf Att2in+CIDER~\cite{rennie2017self}}: a helicopter is flying in the sky\\
{\bf Att2in+CIDER+DISC(1)}: \textcolor{red}{a fighter jet} flying in the sky\\
{\bf Att2in+CIDER+DISC(10)}: \textcolor{red}{a fighter plane} flying in the sky with smoke\\
}
\begin{minipage}[t]{\linewidth}
\begin{minipage}[t]{.45\linewidth}
\includegraphics[height=3cm]{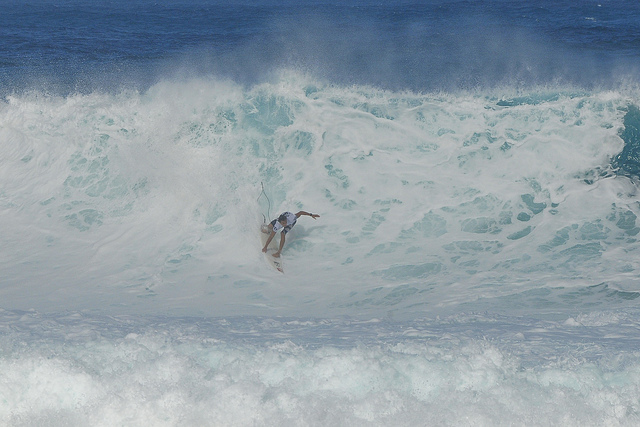}
\end{minipage}%
\hspace{1em}
\begin{minipage}[t]{.45\linewidth}
\includegraphics[height=3cm]{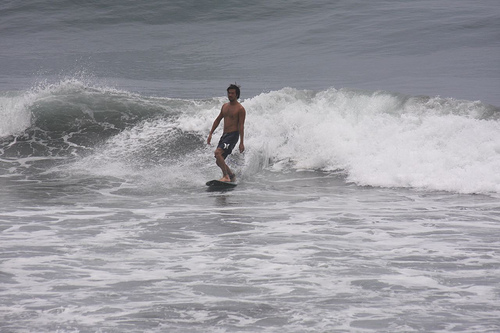}
\end{minipage}
\end{minipage}
{\footnotesize
{\bf Att2in+MLE}: a man riding a wave on top of a surfboard
\\
{\bf Att2in+CIDER~\cite{rennie2017self}}: a man riding a wave on a surfboard in the ocean\\
{\bf Att2in+CIDER+DISC(1)}: a person riding a wave on a surfboard in the ocean\\
{\bf Att2in+CIDER+DISC(10)}: a person  \textcolor{red}{kiteboarding} on a wave in the ocean\\
}
\caption{Some failure cases of our algorithm.}
\label{fig:disccap_failure}
\end{figure}

Figure \ref{fig:disccap_amt+} shows some more examples and Figure \ref{fig:disccap_failure} failure cases of our methods. We highlight in \textcolor{green!50!black}{green} caption elements that (subjectively) seem to aid discriminability, and in \textcolor{red}{red} the portions that seem incorrect or jarringly non-fluent.


\chapter{Appendix: Diversity analysis for image captioning models}\label{app:div}

\section{More Qualitative results}\label{app:div_more_qual}
In the following three pages, we show caption sets sampled by different decoding methods (DBS, BS, Top-K, Top-p, and SP) for three additional images. The model is Att2in model trained with XE.

  \begin{figure}
    \includegraphics[height=6cm]{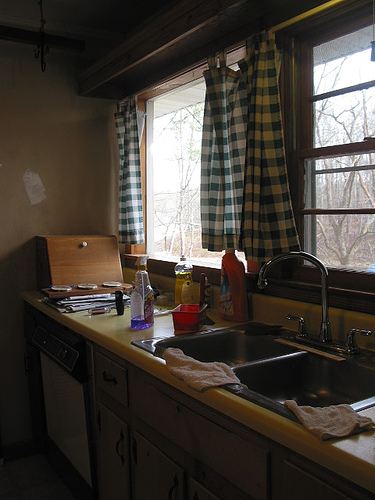}\\
    {\small
    {\bf DBS $\lambda$=3 $\MT$=1}:\\
    a kitchen with a sink and a window\\there is a sink and a window in the kitchen\\an empty kitchen with a sink and a window\\an image of a kitchen sink and window\\a sink in the middle of a kitchen
    
    {\bf BS $\MT$=0.75}:\\
    a kitchen with a sink and a window\\a kitchen with a sink a window and a window\\a kitchen sink with a window in it\\a kitchen with a sink and a sink\\a kitchen with a sink and a window in it
    
    {\bf Top-K K=3 $\MT$=0.75}:\\
    a kitchen with a sink and a mirror\\the kitchen sink has a sink and a window\\a sink and a window in a small kitchen\\a kitchen with a sink a window and a window\\a kitchen with a sink a window and a mirror
    
    {\bf Top-p p=0.8 $\MT$=0.75}:\\
    a kitchen with a sink a window and a window\\a kitchen with a sink a sink and a window\\a kitchen with a sink and a window\\a kitchen with a sink and a window\\a kitchen with a sink and a window
    
    {\bf SP $\MT$=0.5}:\\
    a kitchen with a sink a window and a window\\a sink sitting in a kitchen with a window\\a kitchen sink with a window on the side of the counter\\a kitchen with a sink and a window\\a kitchen with a sink and a window
    }
    \end{figure}
    \begin{figure}
    \includegraphics[height=6cm]{figures/diversity/captions/f3.jpeg}\\
    {\small
    {\bf DBS $\lambda$=3 $\MT$=1}:\\
    a person riding a bike down a street\\there is a man riding a bike down the street\\the woman is riding her bike on the street\\an image of a person riding a bike\\an older man riding a bike down a street
    
    {\bf BS $\MT$=0.75}:\\
    a man riding a bike down a street next to a train\\a person riding a bike down a street\\a person riding a bike on a street\\a man riding a bicycle down a street next to a train\\a man riding a bike down a street next to a red train
    
    {\bf Top-K K=3 $\MT$=0.75}:\\
    a person on a bicycle with a red train\\a man riding a bike down a street next to a red train\\a man riding a bike down a street next to a train\\a man riding a bike down a road next to a train\\a person on a bike is on a road
    
    {\bf Top-p p=0.8 $\MT$=0.75}:\\
    a woman riding a bicycle in a street\\a woman rides a bicycle with a train on it\\a man on a bicycle and a bike and a train\\a man rides a bike with a train on the back\\a woman is riding her bike through a city
    
    {\bf SP $\MT$=0.5}:\\
    a woman on a bike with a bicycle on the side of the road\\a person riding a bicycle down a street\\a man on a bicycle near a stop sign\\a man riding a bike next to a train\\a man riding a bike down the street with a bike
    }
    \end{figure}

    \begin{figure}
    \includegraphics[height=6cm]{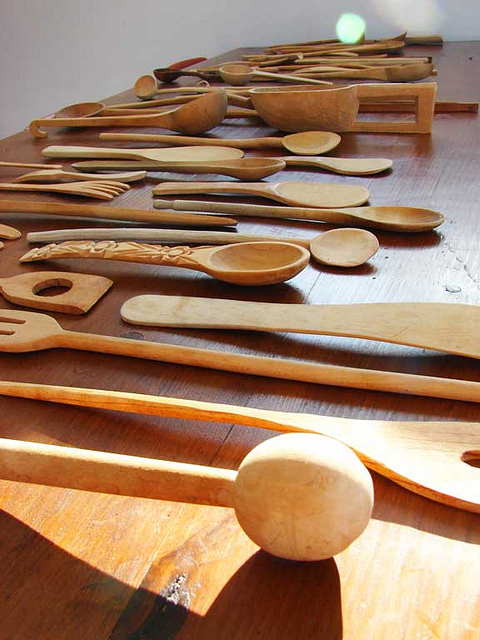}\\
    {\small
    {\bf DBS $\lambda$=3 $\MT$=1}:\\
    a wooden table topped with lots of wooden boards\\a bunch of different types of food on a cutting board\\there is a wooden cutting board on the table\\some wood boards on a wooden cutting board\\an assortment of vegetables on a wooden cutting board
    
    {\bf BS $\MT$=0.75}:\\
    a wooden table topped with lots of wooden boards\\a wooden cutting board topped with lots of wooden boards\\a wooden cutting board with a bunch of wooden boards\\a wooden cutting board with a wooden cutting board\\a wooden cutting board with a bunch of wooden boards on it
    
    {\bf Top-K K=3 $\MT$=0.75}:\\
    a wooden cutting board with a bunch of wooden boards\\a wooden table with several different items\\a wooden cutting board with some wooden boards\\a wooden cutting board with some wooden boards on it\\a bunch of different types of food on a cutting board
    
    {\bf Top-p p=0.8 $\MT$=0.75}:\\
    a bunch of wooden boards sitting on top of a wooden table\\a wooden cutting board with several pieces of bread\\a wooden cutting board with a bunch of food on it\\a bunch of different types of different colored UNK\\a wooden cutting board with a wooden board on top of it
    
    {\bf SP $\MT$=0.5}:\\
    a wooden cutting board with knife and cheese\\a wooden table topped with lots of wooden boards\\a wooden cutting board with chopped up and vegetables\\a wooden table topped with lots of wooden boards\\a wooden cutting board with some wooden boards on it
    }
    \captionof{figure}{More qualitative results.}\label{fig:div_qual_1}
    \end{figure}


\end{document}